\documentclass[11pt]{article}

\usepackage[textwidth=6.75in,textheight=9in,centering]{geometry}
\usepackage[T1]{fontenc}
\usepackage{times}
\usepackage{microtype}
\usepackage{graphicx}
\usepackage{subcaption}
\usepackage{tikz}
\usetikzlibrary{arrows.meta,positioning}
\usepackage{booktabs}
\usepackage[round]{natbib}
\usepackage{amsmath}
\usepackage{amssymb}
\usepackage{mathtools}
\usepackage{amsthm}
\usepackage{bm}
\usepackage{array}
\usepackage{tabularx}
\usepackage{xcolor}
\usepackage[colorlinks,linkcolor=blue!50!black,citecolor=blue!50!black,urlcolor=blue!50!black]{hyperref}
\usepackage[capitalize,noabbrev]{cleveref}

\theoremstyle{plain}
\newtheorem{theorem}{Theorem}[section]
\newtheorem{proposition}[theorem]{Proposition}
\newtheorem{lemma}[theorem]{Lemma}
\newtheorem{corollary}[theorem]{Corollary}
\theoremstyle{definition}
\newtheorem{definition}[theorem]{Definition}
\newtheorem{assumption}[theorem]{Assumption}
\newtheorem{example}[theorem]{Example}
\theoremstyle{remark}
\newtheorem{remark}[theorem]{Remark}

\usepackage[disable]{todonotes}

\newcommand{\R}{\mathbb R}
\newcommand{\E}{\mathbb E}
\newcommand{\Prob}{\mathbb P}
\newcommand{\Hh}{\mathcal H}
\newcommand{\Ss}{\mathcal S}
\newcommand{\Kk}{\mathcal K}
\newcommand{\Nn}{\mathcal N}
\newcommand{\Ff}{\mathbb F}
\newcommand{\Waff}{\mathcal W_{\mathrm{aff}}}
\newcommand{\WL}{\mathcal W_L}
\newcommand{\one}{\mathbf 1}
\newcommand{\ip}[2]{\langle #1,#2\rangle}
\newcommand{\norm}[1]{\lVert #1\rVert}
\newcommand{\abs}[1]{\lvert #1\rvert}
\newcommand{\Ran}{\operatorname{Ran}}
\newcommand{\Ker}{\operatorname{Ker}}
\newcommand{\Span}{\operatorname{span}}
\newcommand{\aff}{\operatorname{aff}}
\newcommand{\conv}{\operatorname{conv}}
\newcommand{\cone}{\operatorname{cone}}
\newcommand{\dist}{\operatorname{dist}}
\newcommand{\tr}{\operatorname{tr}}
\newcommand{\rank}{\operatorname{rank}}
\newcommand{\diag}{\operatorname{diag}}
\newcommand{\spec}{\operatorname{spec}}
\newcommand{\Cov}{\operatorname{Cov}}

\newcommand{\Reg}{\operatorname{Reg}}
\newcommand{\KL}{\operatorname{KL}}
\newcommand{\TV}{\operatorname{TV}}
\newcommand{\gw}{\operatorname{gw}}
\newcommand{\rad}{\operatorname{rad}}
\newcommand{\sd}{\operatorname{sd}}
\newcommand{\op}{\mathrm{op}}
\newcommand{\HS}{\mathrm{HS}}
\newcommand{\deff}{d_{\mathrm{eff}}}
\newcommand{\Gpar}{\Gamma_{\parallel}}
\newcommand{\Gperp}{\Gamma_{\perp}}
\newcommand{\Eop}{\mathcal E}
\DeclareMathOperator*{\argmin}{arg\,min}
\DeclareMathOperator*{\argmax}{arg\,max}
\DeclareMathOperator*{\esssup}{ess\,sup}

\newcommand{\Nfit}{N_{\mathrm{fit}}}
\newcommand{\dN}{d_{\Nn}}
\newcommand{\te}{\tilde e}            %
\newcommand{\formal}[1]{\hfill{\footnotesize\upshape[Formal: \cref{#1}.]}}
\newcommand{\apppart}[2]{\par\bigskip\noindent\rule{\textwidth}{0.6pt}\par\nopagebreak
  \noindent{\large\bfseries #1}\par\nopagebreak
  \ifx\relax#2\relax\else\noindent{\itshape #2}\par\nopagebreak\fi
  \noindent\rule{\textwidth}{0.6pt}\par\nopagebreak\medskip}

\hypersetup{pdftitle={Accuracy Is Not Value: Choosing Weak Supervisors by What the Student Inherits},pdfauthor={Romano Li, Xunfan Cai}}
\title{Accuracy Is Not Value:\\ Choosing Weak Supervisors by What the Student Inherits}
\author{Romano Li\\ Brown University\\ \texttt{romano.li@brown.edu} \and Xunfan Cai\\ \texttt{ffcaizy@gmail.com}}
\date{}

\begin{document}
\maketitle

\begin{abstract}
Weak supervisors are usually judged by their own errors, but what matters is
what a student learns from them. For squared loss and any linear learner
shared by all supervisors, we show that a supervisor's value is determined by
its errors \emph{after} they pass through the student's learning operator.
Consequently, the inner products of the supervisors' raw errors can be compatible with
every feasible value of the student's risk, and weighting supervisors by them
can be arbitrarily worse than the best weighting in relative excess risk. This
student-relative value
is not identifiable from unlabeled data, but what is missing is small: one
finite-dimensional projection of the target decides whether training on a
weak aggregate helps, another decides how to weight the supervisors. Known
teacher training rules, or a label-model condition stated in student-relative
form, shrink it further. Clean labels close the remaining gap at a cost
set by the geometry of the unresolved directions rather than by the dimension
of the student's representation, with matching
worst-case lower bounds. In Gaussian designs we also quantify when exact teacher
structure beats ordinary stacking and when weak supervisors, combined with
clean labels, beat every procedure that uses the clean labels alone. The guarantees extend to fitted ridge students and
finite unlabeled data.
\end{abstract}

\section{Introduction}\label{sec:intro}

Weak-to-strong (W2S) generalization studies a phenomenon that is both
counter-intuitive and common in modern machine learning: a more expressive
student trained only on labels from a weaker supervisor can end up predicting
better than the supervisor itself \citep{burns2024weak}. Existing theory
explains one key mechanism. The student does not copy all of the supervisor's
error; it inherits the error components it can represent and may filter out
the rest \citep{charikar2024quantifying,mulgund2025relating,xue2025representations}.

In practice, weak supervision rarely comes from a single source. Models of
different sizes or trained on different data, heuristic rules and labeling
functions, several LLM judges and several domain experts may all provide weak
labels at once \citep{ratner2016data,sang2024improving,agrawal2024ensemw2s}.
A W2S pipeline therefore faces a decision problem:
\begin{quote}
\emph{Which weak supervisors should a given student learn from, and how much
should it trust each of them?}
\end{quote}
Natural answers reuse tools built for the supervisors' own predictions:
picking the most accurate ones, forecast combination
\citep{bates1969combination}, or label models that infer reliability from
agreement \citep{dawid1979maximum,ratner2016data}. These tools solve their own
problems; applied to the choice of supervision for a downstream student, they
implicitly assume that the supervisors' own error geometry describes their
value to that student. Empirically the premise is already strained: W2S gains depend on
supervisor--student complementarity beyond accuracy \citep{goel2025great}.

We show that this premise fails in general. Because the student filters
error, what matters with several supervisors is not the raw error geometry
but the error geometry \emph{after} the student's learning operator, which we
call the \emph{student-relative geometry}. Filtering changes not only how
much a student gains from a supervisor, but also which supervisor is best,
which supervisors are redundant, and what the optimal weights are. A
two-dimensional example already shows the stakes: in two worlds where the weak
supervisors have the same average squared error ($5$) and the same error
spectrum, the student's risk is $3$ in one and $6$ in the other, depending
only on whether the large error direction lies outside the subspace the
student can represent (\cref{q1:thm:2d}).

\paragraph{Step 1: value is student-relative (\cref{sec:value}).}
Under a common linear learner the student's risk is an exact quadratic in the
weights. For a projection student only the transmitted part of the error Gram
enters, and every pairwise inner product of the supervisors' errors is
compatible with a whole interval of student risks; weighting by the raw Gram
can have arbitrarily large relative excess risk.

\paragraph{Step 2: the right object is not identified from unlabeled data
(\cref{sec:identification}).}
Every gain and every risk contrast is an observable quadratic minus one
unknown linear target moment, so gains and selection need two
finite-dimensional projections of the target. Two bounded, nondegenerate
worlds with identical unlabeled observations require opposite decisions. Known
teacher equations shrink the missing part through a \emph{leakage law}; a
label-model assumption, stated in student-relative form, reduces selection to
one unknown scalar.

\paragraph{Step 3: anchors close the gap (\cref{sec:anchors}).}
Clean labels, which we call \emph{anchors}, only estimate these moments.
Certifying gain signs costs labels set by the margin and the geometry of the
unresolved queries, uniformly over data-selected weights; selection costs
labels set by the \emph{unresolved trace} $\dN\le m-1$, not by the
dimension of the student's representation. Both rates have matching
worst-case lower bounds. When the hidden target lies in the unresolved
directions, regressing the residual label on them pays only the noise, so
exact teacher structure beats ordinary stacking when $\dN$ is below the
number $k$ of candidate contrasts; and weak
supervisors beat every rule that uses clean labels alone when anchors are few
relative to the target's effective dimension and the best aggregate captures
most of the signal.

\paragraph{Finite samples (\cref{sec:finite}).}
With a fitted ridge student and estimated Grams, the error splits into
student fitting, unlabeled integration and anchor selection.

\begin{figure*}[t]
  \centering
  \begin{subfigure}[t]{0.32\textwidth}
    \includegraphics[width=\linewidth]{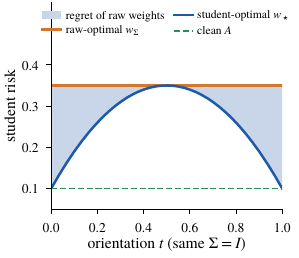}
    \caption{Same raw Gram, different students.}\label{fig:teaser-a}
  \end{subfigure}\hfill
  \begin{subfigure}[t]{0.32\textwidth}
    \includegraphics[width=\linewidth]{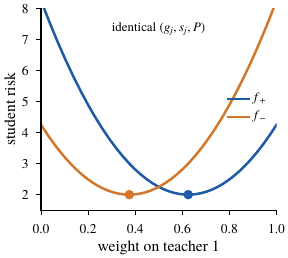}
    \caption{Same unlabeled observations, different optima.}\label{fig:teaser-b}
  \end{subfigure}\hfill
  \begin{subfigure}[t]{0.32\textwidth}
    \includegraphics[width=\linewidth]{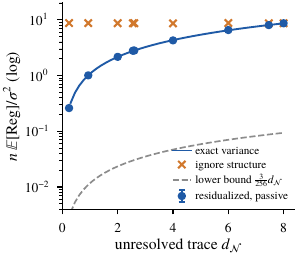}
    \caption{Anchor cost of selection tracks $\dN$.}\label{fig:teaser-c}
  \end{subfigure}
  \caption{\textbf{The three steps.} (a) With $\Sigma=I$ fixed, rotating the
  errors leaves raw-optimal weights unchanged while the student optimum moves;
  shaded: regret of raw weighting (\cref{thm:fixed-gram,thm:decisions}).
  (b) Identical unlabeled observations, opposite gains and optima
  (\cref{thm:counter}). (c) Normalized selection regret with passive anchors
  tracks $\dN$ (log scale; bounds of \cref{thm:selection}); ignoring teacher
  structure pays the full $k$.}
  \label{fig:teaser}
\end{figure*}

Together, these results separate three notions that are often conflated:
raw supervisor quality, identification of student-relative value, and the
clean-label complexity required to act on that value.

\section{Related Work}\label{sec:related}

\paragraph{Weak-to-strong theory.}
Existing theory explains why a \emph{single} weak supervisor can be improved
upon by a particular student. For convex student classes, the gain is
controlled by the student--supervisor misfit
\citep{charikar2024quantifying}, also beyond squared loss
\citep{mulgund2025relating,xu2026bregman}. High-dimensional and
representation analyses show how a teacher's error is inherited on the
subspace it shares with the student and attenuated elsewhere
\citep{dong2025discrepancies,ildiz2025high,xue2025representations}, and
regularization, early stopping and two-stage ridge models explain when the
student can win \citep{lang2024theoretical,medvedev2025weak,moniri2025mechanisms,
wu2026improved}; misfit already guides the choice among strong models trained
by different weak models \citep{charikar2024quantifying}, and label-free
metrics predict W2S trends under particular representation models
\citep{xue2025representations}. We study the joint aggregation decision for a
fixed student: its observational equivalence classes, when it is identified
from unlabeled data and, otherwise, how many clean labels it takes.

\paragraph{Multiple weak supervisors.}
Label models infer supervisor quality from agreement under conditional
independence \citep{dawid1979maximum,ratner2016data}; \cref{sec:identification}
states the student-relative version of that assumption. Empirical W2S methods combine, weight or route several weak models,
typically with hand-set or heuristic weights
\citep{sang2024improving,liu2024cosupervised,agrawal2024ensemw2s,
cui2025bayesian}. Supervisor--student complementarity matters empirically
beyond supervisor accuracy \citep{goel2025great}. We characterize which objective such methods should target, when it is
observable, and what it costs in clean labels.

\paragraph{Aggregation and few-label inference.}
Because $T(\sum_jw_jg_j)=\sum_jw_jTg_j$, training on a label mixture is
combining fixed candidate students, so given the right Gram and target
moments our selection is stacking or quadratic aggregation
\citep{wolpert1992stacked,breiman1996stacked,tsybakov2003optimal}, and
estimating the quadratic from unlabeled data is semi-supervised ERM
\citep{azriel2022semi}, which need not beat supervised ERM. Forecast
combination optimizes the raw combined error \citep{bates1969combination}.
Prediction-powered inference and its multi-proxy and ranking variants use a
few labels to estimate functionals or rankings
\citep{angelopoulos2023prediction,chatzi2024prediction,cowenbreen2026multiple};
with fixed candidate students, risk differences are such functionals. What is
specific here is the split into task spaces, the structural identification
boundary and its geometric label cost.
\Cref{app:related} gives a detailed comparison.

\section{Setup}\label{sec:setup}

Let $X\sim P_X$ and work in the real Hilbert space $\Hh=L^2(P_X)$ with
$\ip hk=\E[h(X)k(X)]$ and $\norm h^2=\E[h(X)^2]$, so that squared norms are
mean squared errors. The target regression function $f\in\Hh$ is unknown.
There are $m$ weak supervisors $g_j=f+e_j$, $j=1,\dots,m$, with errors $e_j$;
no mean-zero or independence assumption is made on the errors. For weights
$w\in\R^m$ put $g_w=\sum_jw_jg_j$ and $e_w=\sum_jw_je_j$. We use the affine
weights $\Waff=\{w:\one^\top w=1\}$, the simplex $\Delta_m$, and the bounded
domain $\WL=\{w\in\Waff:\norm{w}_1\le L\}$. The constraint $\one^\top w=1$ gives
$g_w=f+e_w$; otherwise the target itself enters the aggregate error.

\paragraph{Student.}
All aggregates are learned by one bounded linear learner $T:\Hh\to\Hh$. With
$s_j=Tg_j$ and $q_j=g_j-s_j$ (what the student does not absorb),
$s_w=Tg_w=\sum_jw_js_j$ and $q_w=\sum_jw_jq_j$. This includes linear and ridge
heads on a frozen representation, kernel ridge regression, early-stopped
gradient descent from zero initialization and spectral cut-off, all with
hyperparameters (step sizes, stopping time, regularization) that do not
depend on the labels. A nonzero initialization makes the learner affine,
with an offset that must be treated separately. Re-tuning hyperparameters per label mixture, or nonlinear
fine-tuning, breaks linearity in the labels. Linearity is also what
makes the problem tractable: training on a label mixture equals mixing fixed
candidate students, so selection becomes aggregation over a fixed dictionary
and nonlinear interactions between supervisors are excluded. The main case is the orthogonal
projection $T=P=P_\Ss$ onto a closed subspace $\Ss$, with $Q=I-P$; then the
clean-label student is $Pf$ and $A=A(f)=\norm{Qf}^2$ is the clean-label risk
floor of the representation.

\paragraph{Risks and gain.}
$R_W(w)=\norm{g_w-f}^2=\norm{e_w}^2$ and $R_T(w)=\norm{s_w-f}^2$, with
$R_S=R_P$. The W2S gain is always the same-aggregate comparison
$G_T(w)=R_W(w)-R_T(w)$ ($G=G_P$). Beating one's own aggregate, beating the
best single supervisor, and reaching the clean-label student are three
different benchmarks that do not imply one another (\cref{q1:thm:benchmarks}).

\paragraph{Error geometry.}
Let $E:\R^m\to\Hh$, $Ew=\sum_jw_je_j$, with adjoint $(E^*h)_j=\ip{e_j}h$.
The raw and transmitted error Grams are
\[
\Sigma=E^*E,\qquad \Gamma_T=E^*T^*TE,
\]
and in the projection case $\Gpar=E^*PE$, $\Gperp=E^*QE$, so that
$\Sigma=\Gpar+\Gperp$. All of these are \emph{uncentered} Gram matrices, not
covariances; ``centering'' only ever refers to the supervisor index.

\paragraph{Data and what is assumed known.}
Three independent samples are used: $\Nfit$ pseudo-labelled inputs to fit
the student, $N$ unlabeled inputs for prediction-only moments, and $n$
\emph{anchors} $(X_i,Y_i)$, $Y_i=f(X_i)+\xi_i$, $\E[\xi_i\mid X_i]=0$, for target
moments. \Cref{sec:value,sec:identification,sec:anchors} treat $T$, the
candidate students and every inner product of known functions (e.g.\
$\norm{g_w}^2-\norm{s_w}^2$, $\ip{f_0}{q_j}$) as exact; only target moments are
unknown, and the structured anchor estimator also evaluates the unresolved
directions $\Pi_\Nn h$ pointwise. \Cref{sec:finite} estimates the learner and
the prediction geometry; the structural input remains an assumption. Main-text
results are streamlined versions of the formal statements in
\cref{app:q1,app:q2,app:q3,app:q4}, to which each result points.

\section{The Value of a Supervisor Is Student-Relative}\label{sec:value}

Given the student's learning rule, what determines the value of a
weak-supervisor aggregate? Everything in this section is an exact population
identity.

\subsection{Exact risk}

Let $b_T=(I-T)f$, $B_T=\norm{b_T}^2$ (the algorithm's risk on clean labels)
and $c_T=E^*T^*b_T$ (the \emph{bias alignment}). Since
$s_w-f=T(f+e_w)-f=Te_w-b_T$, expanding the square gives the following.

\begin{proposition}[Exact risk]\label{prop:risk}
For every $w\in\Waff$,
\begin{align*}
R_T(w)&=B_T+w^\top\Gamma_Tw-2c_T^\top w,\\
G_T(w)&=w^\top(\Sigma-\Gamma_T)w-B_T+2c_T^\top w .
\end{align*}
The cross term vanishes for all targets and all error families if and only
if $T$ is an orthogonal projection. For $T=P$,
\[
R_S(w)=A+w^\top\Gpar w,\qquad G(w)=w^\top\Gperp w-A .
\]
\formal{q1:thm:multi,q1:prop:orth-only}
\end{proposition}

With one supervisor and $f\in\Ss$, $G=\norm{Qe}^2$ is the misfit--gain
identity of \citet{charikar2024quantifying}. So the supervisors' raw risk is
governed by $\Sigma$, and the student's risk by $\Gpar$. An error direction that is large in a supervisor's own risk is
filtered if it lies mostly in $\Ss^\perp$; an error that is small in raw risk
but aligned with $\Ss$ is inherited in full. For a general learner, $B_T$ is
not a floor: weak labels can partly cancel the algorithm's own shrinkage bias
through the $c_T$ term.

\subsection{A fixed raw Gram does not determine the student}

Is the gap between $\Sigma$ and $\Gpar$ an artifact of not using all of
$\Sigma$? It is not.

\begin{theorem}[Raw geometry does not determine student risk]\label{thm:fixed-gram}
Let $\Sigma\succeq0$, $d_\Ss=\dim\Ss$, $d_\perp=\dim\Ss^\perp$. A matrix
$\Gamma$ is the transmitted Gram $\Gpar$ of some error family with raw Gram
$\Sigma$ if and only if
\[
0\preceq\Gamma\preceq\Sigma,\qquad \rank\Gamma\le d_\Ss,\qquad
\rank(\Sigma-\Gamma)\le d_\perp .
\]
If $\rank\Sigma\le\min\{d_\Ss,d_\perp\}$, then for every fixed $w\in\Waff$
\[
R_S(w)\in[A,\,A+w^\top\Sigma w],\quad G(w)\in[-A,\,w^\top\Sigma w-A],
\]
and every point of these intervals is attained. The rank conditions are
necessary for the endpoint Grams $(\Sigma,0)$ and $(0,\Sigma)$, not for the scalar
range at a fixed $w$. In particular, when
$0<A<w^\top\Sigma w$, no function of $(\Sigma,w,A)$ determines the sign of
$G(w)$.
\formal{q1:thm:realize,q1:cor:fixed-gram-range,q1:cor:no-gram}
\end{theorem}

The raw geometry fixes only the sum of the two parts, not how much of it the
student transmits; a fixed error spectrum is a weaker version of the same
statement (\cref{q1:thm:orient}).

\subsection{Raw-optimal aggregation can fail the student}

Let $w_W^\star$ minimize the weak risk $w^\top\Sigma w$ and $w_S^\star$ the
student risk, both over $\Waff$. Raw-optimal weights are the forecast
combination weights $\Sigma^{-1}\one/\one^\top\Sigma^{-1}\one$ when $\Sigma\succ0$;
student-optimal weights replace $\Sigma$ by $\Gpar$.

\begin{corollary}[Raw weighting can be arbitrarily suboptimal]\label{thm:decisions}
Over affine weights, with $d=w_W^\star-w_S^\star$,
\[
R_S(w_W^\star)-R_S(w_S^\star)=d^\top\Gpar d;
\]
with simplex optimizers the right-hand side gains the nonnegative first-order
term $2d^\top\Gpar w_S^\star$. Moreover, for every $L>0$ there are two
supervisors with positive definite $\Gpar$ and $\Gperp$ (so that the excess
risk $R_S(w_S^\star)-A=1/\one^\top\Gpar^{-1}\one$ is strictly positive), whose
affine and simplex optimizers coincide within each objective, such that
\[
\frac{R_S(w_W^\star)-A}{R_S(w_S^\star)-A}>L .
\]
\formal{q1:cor:raw-regret,q1:thm:raw-arbitrary}
\end{corollary}

The ratio concerns excess risk over the floor $A$. In absolute terms the
regret over affine weights is $d^\top\Gpar d\le\norm d_1^2\max_j(\Gpar)_{jj}$, and on the
simplex it is at most $\max_j\norm{Pe_j}^2$ by convexity. In the construction it stays bounded while the
student-optimal excess risk vanishes, which requires a supervisor whose error is small in $\Ss$ but large
in $\Ss^\perp$. Raw weighting all but discards that supervisor, whereas the
student would filter the large part and should rely on it. The regret formula
looks like classical quadratic aggregation, but raw weighting optimizes
$\Sigma$ while the student needs $\Gpar$: the failure comes from optimizing
the wrong geometry, not from estimation noise. Maximizing the gain
is a third, distinct objective; on the simplex a single supervisor always
attains its maximum, and the result can be arbitrarily worse than
student-optimal weighting (\cref{q1:thm:gainmax}).

The same shift applies to the value of an additional supervisor. With
transmitted errors $\te_j=Pe_j$ and $\te_m^\star$ the current optimal
(minimum-norm) transmitted error over the simplex, supervisor $m+1$ strictly
helps if and only if $\ip{\te_m^\star}{\te_{m+1}}<\norm{\te_m^\star}^2$; raw and
student-relative redundancy can therefore disagree in both directions (\cref{q1:thm:marg-sim,q1:thm:marg-aff,q1:ex:redundancy}).

All of this is oracle geometry: $e_j=g_j-f$, $\Gpar$ and $A$ depend on the
unknown target. Can unlimited unlabeled data recover them?

\section{What Unlabeled Data Cannot Identify}\label{sec:identification}

\subsection{Observable quadratics and hidden target moments}

The population unlabeled observation is
\[
\mathcal O_T=(P_X,g_1,\dots,g_m,T),
\]
including all population inner products of known functions. A learner may
train arbitrarily many candidate students from $\mathcal O_T$, choose label
mixtures adaptively and use internal randomness, but observes no target
moment. Let $(K_g)_{jk}=\ip{g_j}{g_k}$, $(K_s)_{jk}=\ip{s_j}{s_k}$ and
$\mathsf D_T=K_g-K_s$. When risks are subtracted, $\norm f^2$ cancels.

\begin{theorem}[Task-specific identification]\label{thm:tasks}
For all $w,v\in\Waff$,
\begin{align*}
G_T(w)&=w^\top \mathsf D_Tw-2\ip f{q_w},\\
R_T(w)-R_T(v)&=\norm{s_w}^2-\norm{s_v}^2-2\ip f{s_w-s_v}.
\end{align*}
Let $U_T=\Span\{q_1,\dots,q_m\}$, $V_T=\Span\{s_j-s_m:j<m\}$ and
$k=\dim V_T\le m-1$. Two targets produce identical gains for all weights iff
their projections on $U_T$ coincide, and identical student-risk contrasts
for all pairs of weights iff their projections on $V_T$ coincide; identical
contrasts imply identical optimal-weight sets. For $T=P$, $q_j=Qg_j\in\Ss^\perp$
and $s_j-s_m\in\Ss$, hence $U\perp V$.
\formal{q2:thm:tasks}
\end{theorem}

The second equivalence concerns the full contrast information: two particular
targets with different projections on $V_T$ can still share an argmin by
accident. The exact condition for the \emph{entire} optimal-weight set over a
compatible class is \cref{thm:leakage}. Because $U\perp V$ for projection
students, gain certification and student selection are informationally
distinct problems. Even if it is known that $f\in\Ss$, so that $\ip f{q_w}=0$ and all
gains are identified, $\Pi_Vf$ can remain unknown and the optimal weights
unidentified (\cref{q2:cor:realizable}).

\subsection{Same unlabeled world, opposite decisions}

Non-identification is not an abstract dimension count. Let $u,z,v$ be
independent Rademacher coordinates, $\Ss=\Span\{u,z\}$,
\[
g_1=2u+z+v,\quad g_2=-2u+z-v,\quad
f_\pm=\pm\bigl(\tfrac12u+v\bigr).
\]
Both worlds share $P_X$, $g_1$, $g_2$, $P$ and the students $s_1=2u+z$,
$s_2=-2u+z$.

\begin{proposition}[Two indistinguishable worlds, opposite decisions]\label{thm:counter}
In both worlds $A=1$ and the clean student beats every supervisor. Supervisor
1 has gain $-1$ under $f_+$ and $3$ under $f_-$ (supervisor 2 the reverse),
and the unique affine and simplex student-optimal weights are $(5/8,3/8)$ and
$(3/8,5/8)$. Consequently, every procedure without anchors has worst-world
gain-sign error at least $1/2$, and every procedure returning affine weights
has worst-world expected selection regret at least $1/4$, even with unlimited
unlabeled data and candidate training.
\formal{q2:thm:counter,q2:cor:impossible}
\end{proposition}

That some target information is missing is elementary: the map
$f\mapsto f+h$, $e_j\mapsto e_j-h$ leaves every $g_j$ unchanged. The point of
\cref{thm:tasks,thm:counter} is which information, and that its absence
changes decisions in a benign world: the supervisors are distinct and
bounded, each world contains a positive gain, and the uniform aggregate beats
the best single supervisor. Disagreements reveal only centered errors
$J\Sigma J$, not the common error, which moves the optimal weights
(\cref{q2:thm:center,q2:prop:common}).

\subsection{Known teacher equations and the leakage law}

Sometimes the supervisors are produced by known population training rules,
\[
g_j=M_jf,
\]
with known bounded linear maps $M_j$ (e.g., least squares or ridge on the
supervisor's own features). Let $\Nn=\bigcap_j\Ker M_j$ be the common
nullspace and $\Kk=\Nn^\perp$. If the observation is compatible with at least
one target, the unlabeled observation determines the component $f_\Kk$, and
the compatible targets are exactly the affine fiber $f_\Kk+\Nn$. Define the
student \emph{leakages} $\ell_j=\Pi_\Nn Tg_j$.

\begin{theorem}[Leakage law]\label{thm:leakage}
Assume every $g_j\in\Kk$ (automatic when the $M_j$ are self-adjoint, as for
least-squares, ridge and spectral-filter supervisors; not automatic for
nonnormal maps). On the full compatible fiber:
\begin{enumerate}
\item all gains are identified iff $\ell_j=0$ for every $j$;
\item all student-risk contrasts, and the entire optimal-weight set over
  affine or simplex weights, are identified iff $\ell_1=\dots=\ell_m$;
\item hence gain identification implies selection identification, but not
  conversely.
\end{enumerate}
\formal{q2:thm:fiber,q2:thm:leakage,q2:thm:optset}
\end{theorem}

\emph{Proof idea.} A hidden shift $n\in\Nn$ moves the gain by
$-2\ip n{\Pi_\Nn q_w}$ and a contrast by $-2\ip n{\Pi_\Nn(s_w-s_v)}$, and
$\Pi_\Nn q_j=-\ell_j$.

To quantify how much of the selection problem remains open, write
$V=V_T=\Span\{Tg_j-Tg_m:j<m\}$ for the contrast space of the given learner
(not only of projections) and define the \emph{unresolved trace}
\[
\dN=\tr(\Pi_V\Pi_\Nn\Pi_V)=\sum_i\nu_i^2\in[0,k],
\]
where $\nu_i$ are the nonzero singular values of $\Pi_\Nn|_V$, i.e., the
cosines of the principal angles between $V$ and $\Nn$. It is an effective
unresolved dimension and need not be an integer. Without structure
$\Nn=\Hh$ and $\dN=k$; if $V\perp\Nn$, then $\dN=0$ and selection is recovered
without anchors. Even exactly trained teachers do not suffice in general
(\cref{q2:thm:struct-counter}). Exact equations are an idealization; with
equation error $\varepsilon$ and a bound $\rho$ on the hidden component, each
gain moves by at most $2\varepsilon C_w+2\rho\norm{\Pi_\Nn q_w}$, with $C_w$ a
stable-recovery constant (\cref{q2:prop:approx}).

\paragraph{Label-model assumptions.}
Between no assumption and known equations lie label models,
$g_j=a_jf+\varepsilon_j$. The version that matters is student-relative,
$\ip{\varepsilon_j}{s_k}=0$ for $j\ne k$, which the raw conditions
$\ip{\varepsilon_j}{\varepsilon_k}=\ip{\varepsilon_j}f=0$ do not imply. With $m\ge3$,
$\ip{g_j}{s_k}=a_j\ip f{s_k}$ then identifies all selection moments up to one
common scalar, which no unlabeled observation can fix; one anchor direction
gives $\E\Reg\le v_0^2/n$, independent of $m$ and $k$
(\cref{q2:thm:factor,q2:cor:factor-anchors}). Gains remain unidentified.

\paragraph{Sensitivity analysis.}
With a reference $f_0$ and the external bound $\norm{Q(f-f_0)}\le\rho_\perp$,
the sharp identified set of $G(w)$ is
$\norm{q_w}^2-2\ip{f_0}{q_w}\pm2\rho_\perp\norm{q_w}$ \citep{manski2003partial};
if it excludes zero, the sign is certified without anchors
(\cref{q2:thm:interval,q2:prop:joint}).

\section{How Many Anchors?}\label{sec:anchors}

\begin{table}[t]
\caption{\textbf{Decision--geometry map.} Each result answers a different
decision under its own observation model; they are not competing measures of
one difficulty. Constants and conditions are in the cited results.}
\label{tab:map}
\centering\footnotesize
\begin{tabularx}{\linewidth}{@{}>{\raggedright}p{0.25\linewidth}>{\raggedright}X>{\raggedright\arraybackslash}p{0.36\linewidth}@{}}
\toprule
Decision & Geometry & Anchor cost\tabularnewline
\midrule
Gain signs, one or all (\cref{thm:certificate}) & margin $\gamma$; width of queries $\mathbb A_\gamma$ & $\sigma^2\gamma^{-2}\log\tfrac1\delta$; $\sqrt n\gtrsim\gw(\mathbb A_\gamma)$\tabularnewline
Affine selection (\cref{thm:selection}) & unresolved trace $\dN$ & $\sigma^2\dN/n$\tabularnewline
Local selection (\cref{thm:selection}) & tangent cone $\mathcal T$ at the oracle & $\sigma^2\sd(\mathcal T)/n$\tabularnewline
Passive anchors (\cref{sec:coverage}) & cell coverage & $\mathfrak c_n+\mathfrak v_n$\tabularnewline
\bottomrule
\end{tabularx}
\end{table}

Anchors only estimate the decision moments of \cref{sec:identification}, at
a cost set by the geometry of the missing information (\cref{tab:map}). A
gain sign answers a deployment question (train the student on $w$, or use $w$
directly?), not how to weight, since maximizing the gain is the wrong
objective (\cref{q1:thm:gainmax}); selection chooses the weights.

Assume a known reference $f_0$ and residual subspace $\Nn$ with
$\bar f=f-f_0\in\Nn$; without structure take $f_0=0$ and $\Nn=\Hh$, and with
known teacher equations take $f_0=f_\Kk$. For any known query function $h$,
the anchors give the unbiased moment estimate
\[
\widehat\mu_h=\frac1n\sum_{i=1}^n\bigl(Y_i-f_0(X_i)\bigr)h(X_i),\qquad
\E\widehat\mu_h=\ip{\bar f}h,
\]
which requires neither the noise variance nor $\norm f^2$. Residualizing
against $f_0$ is a prediction-powered correction \citep{angelopoulos2023prediction}
aimed at the decision moments of \cref{sec:identification}.

\subsection{Gain certificates and sign complexity}

The gain depends only on the unresolved query directions
$h_j=\Pi_\Nn q_j$; replacing $\ip{\bar f}{h_j}$ by $\widehat\mu_{h_j}$ in
\cref{thm:tasks} defines $\widehat G(w)$. For simultaneous signs, write the
unresolved queries in coordinates as $v_i$, with margins $\gamma_i$ and noise
proxy $\Omega$, and let $\mathbb A_\gamma=\{0,\pm2\Omega^{1/2}v_i/\gamma_i\}$
be the margin-normalized query set, with Gaussian width
$\gw(\mathbb A)=\E\sup_{a\in\mathbb A}G^\top a$ and radius $\rad(\mathbb A)$.

\begin{theorem}[Anchor complexity for gain decisions]\label{thm:certificate}\label{thm:signs}
\leavevmode
\begin{enumerate}
\item[(i)] \emph{Uniform certification.} If the centered single-anchor
variables $(Y-f_0(X))h_j(X)-\ip{\bar f}{h_j}$ are sub-Gaussian with proxy
$\tau^2$, then with probability at least $1-\delta$,
\[
\sup_{w\in\WL}\abs{\widehat G(w)-G(w)}\le
2L\tau\sqrt{\frac{2\log(2m/\delta)}{n}}=:a_n ,
\]
so $\widehat G(w)\mp a_n$ certifies the sign of $G(w)$ simultaneously for all
$w\in\WL$, including weights selected with the same anchors.
\item[(ii)] \emph{Geometry-sensitive bound.} In the Gaussian moment submodel, all
$n_q$ signs are correct with probability at least $1-\delta$ whenever
\[
\sqrt n>\gw(\mathbb A_\gamma)+\rad(\mathbb A_\gamma)\sqrt{2\log(1/\delta)} .
\]
Consequently the cost is $\Theta(\sigma^2\gamma^{-2}\log(1/\delta))$,
independent of $m$, when all queries coincide, and grows as
$\sigma^2\gamma^{-2}\log m$ (for fixed $\delta$) for orthogonal queries; over the
scalar-product class the worst case is
$\Theta(\sigma^2\gamma^{-2}[\log m+\log(1/\delta)])$.
\end{enumerate}
\formal{q3:thm:uniform,q3:thm:sign-upper,q3:thm:extremes,q3:cor:sign-order}
\end{theorem}

Part (i) holds under general sub-Gaussian anchors; a pointwise interval for
a data-selected aggregate would lose coverage. Part (ii) is matched for
duplicated and orthogonal queries and in the worst case, but is not
instance-optimal for every dictionary. $\log m$ is a worst-case cost: what
matters are the unresolved, noise-whitened, margin-normalized queries, whose
clustering raw supervisor similarity does not imply (\cref{q3:prop:clusters}).

\subsection{Selection regret and the unresolved trace}

Selection depends only on the target's projection on $V$. Choose an isometry
$\Phi:\R^k\to V$ and put $t_j=\Phi^*s_j$, $\theta=\Phi^*f$ and
$\mathcal C=\{\sum_jw_jt_j:w\in\mathcal W\}$. For affine $w$,
$R_T(w)=\mathrm{const}+\norm{t(w)-\theta}^2$, so selection is the projection of
$\theta$ onto $\mathcal C$. Let $\varepsilon=\widehat\theta-\theta$ be the
error of the residualized estimate and $\mathcal T$ the tangent cone of
$\mathcal C$ at the oracle point.

\begin{theorem}[Selection regret is governed by $\dN$]\label{thm:selection}
For simplex, bounded affine or unrestricted affine weights, projecting
$\widehat\theta$ onto $\mathcal C$ gives
$0\le\Reg(\widehat w)\le\norm{\Pi_{\mathcal T}\varepsilon}^2\le\norm\varepsilon^2$.
If $\E[(Y-f_0(X))^2\mid X]\le v_0^2$, the residualized moment estimator
attains the first bound below; with Gaussian label noise, targets with
$\norm{f-f_0}\le\rho$, unresolved singular functions with $\sum_i\abs{z_i}\le M_z$,
and $n\ge\max\{1,M_z^2/32,\sigma^2r/(32\rho^2)\}$ ($r$ the number of nonzero
singular values), every affine-weight estimator obeys the second:
\[
\E\Reg(\widehat w)\le\frac{v_0^2\,\dN}{n},\qquad
\inf_{\widehat w}\sup_f\E_f\Reg(\widehat w)\gtrsim\frac{\sigma^2\dN}{n}.
\]
\formal{q3:thm:cone,q3:thm:expected,q3:thm:angle,q3:cor:k-lower}
\end{theorem}

Without structure $\dN=k\le m-1$. Within the lower-bound class
($v_0^2\le2\sigma^2$), affine selection thus has minimax rate
$\Theta(\sigma^2\dN/n)$ (lower-bound constant $3/256$). The $k/n$ rate is
classical; what is new is which moments must be estimated and that the
dimension is $\dN$. A poor reference puts the unexplained signal into
$v_0^2$; \cref{sec:compare} removes it.

\paragraph{Local geometry.}
When perturbations of size $\sigma/\sqrt n$ stay feasible, the regret on one
simplex is $O(\sigma^2\log m/n)$ at a vertex and $\Theta(\sigma^2(m-1)/n)$ in the
interior, the model-selection versus aggregation contrast
\citep{tsybakov2003optimal} in student-relative geometry
(\cref{q3:prop:vertex}).

\subsection{When are weak supervisors worth their anchors?}\label{sec:compare}

Write $\theta-\Phi^*f_0=\sum_ib_i\ip{z_i}{\bar f}$, where $z_i\in\Nn$ are the left
singular functions of $\Pi_\Nn|_V$ and $b_i\in\R^k$ its right singular vectors
scaled by $\nu_i$. \emph{Residual regression} fits $Y-f_0(X)$ on $z(X)$ by least
squares and plugs the coefficients into this formula; without structure
($\Nn=\Hh$) it is unregularized affine stacking. Let
$R_{\mathrm{agg}}=\min_{\Waff}\norm{s_w-\Pi_\Ss f}^2$, the best aggregate's excess over
$A_\Ss=\norm{f-\Pi_\Ss f}^2$.

\begin{proposition}[Structure, stacking, anchors alone]\label{prop:compare}
Let $f\in\Ss$, $n>k+1$, homoscedastic noise, and all functions linear in a
standard Gaussian vector.
\begin{enumerate}
\item[(i)] If $\bar f\in\Span\{z_i\}$, residual regression has
$\E\Reg=\sigma^2\dN/(n-r-1)$, and weak supervision with anchors has risk
$R_{\mathrm{agg}}+\sigma^2\dN/(n-r-1)$; unregularized stacking has $\E\Reg\ge\sigma^2k/(n-k-1)$.
\item[(ii)] Without structure, weak supervision with anchors has risk
$R_{\mathrm{agg}}+(\sigma^2+R_{\mathrm{agg}})k/(n-k-1)$.
\item[(iii)] If $f$ has coordinates $\vartheta\sim N(0,\Lambda)$ in $\Ss$ and the noise is
Gaussian, every procedure using only the anchors, tuned ridge included, has
Bayes risk at least $\sigma^2d_{\mathrm{eff}}(\sigma^2/n)/n$, where
$d_{\mathrm{eff}}(\gamma)=\tr\Lambda(\Lambda+\gamma I)^{-1}$.
\end{enumerate}
\formal{q3:thm:rr-gauss,q3:thm:stack,q3:thm:anchor-only,q3:prop:anchor-bayes,q3:cor:weak-bayes}
\end{proposition}

Weak supervision thus beats every anchor-only rule when $n$ is small relative
to the effective dimension of the target in the student's representation and
the aggregate captures most of the signal; for $R_{\mathrm{agg}}>0$ the advantage of
selection alone ends near $n\approx\sigma^2(d_{\mathrm{eff}}-\dN)/R_{\mathrm{agg}}$. The condition
$\bar f\in\Span\{z_i\}$ in (i) is strong: $\Nn$ is typically infinite-dimensional,
and any part of $\bar f$ outside the $r$ directions re-enters as variance, after
which structure need not beat stacking (\cref{q3:rem:rr-limits}). (i) then
measures the value of the side information in teacher equations, the noise
ratio $\dN/k$. Without structure, the setting of LLM supervisors, the
approximation error also inflates the variance, as in (ii). $R_{\mathrm{agg}}$ is not
observable, ridge stacking and hybrids that also fit the selected student's
residual on the anchors are not covered, and equation error adds a bias
(\cref{q3:cor:teacher-error}).

\subsection{Passive anchors and coverage}\label{sec:coverage}

Passive anchors may never cover rare directions. For a fixed cell
dictionary with $\Prob(X\in C_\ell)=p_\ell$ and target amplitude at most
$B_\ell$ on cell $\ell$, the noiseless minimax selection regret is exactly
$\mathfrak c_n=\sum_\ell p_\ell B_\ell^2(1-p_\ell)^n$; with Gaussian noise it is
of order $\mathfrak c_n+\mathfrak v_n$, $\mathfrak v_n=\sum_\ell\min\{p_\ell B_\ell^2,\sigma^2/n\}$
(\cref{q3:thm:cell-noiseless,q3:thm:cell-noisy}). With $B_\ell=p_\ell^{-1/2}$
all population Grams stay fixed while the regret moves from exponentially
small to the order of the number of cells (\cref{q3:prop:same-gram}): with an
unbounded envelope, the spectrum does not control passive coverage.

\section{Fitted Students and Finite Unlabeled Geometry}\label{sec:finite}

We now estimate the learner, the prediction moments and the target moments.
\Cref{sec:identification,sec:anchors} hold for any fixed common linear
learner; the projection-specific facts ($c_P=0$, $U\perp V$) are not used
here. Let $\psi(X)$ be a fixed
representation with synthesis operator $\Psi$ and feature operator
$C_\psi=\Psi^*\Psi$. For $\lambda>0$ the population ridge learner is
\[
T_\lambda=\Psi(C_\psi+\lambda I)^{-1}\Psi^*=\mathcal L(\mathcal L+\lambda I)^{-1},
\qquad \mathcal L=\Psi\Psi^* .
\]
Fitting on $\Nfit$ inputs with weak pseudo-labels gives the empirical learner
$\widehat T_\lambda$ and candidates $\widehat s_j=\widehat T_\lambda g_j$. A further
$N$ independent unlabeled inputs estimate the candidate Gram, and $n$
anchors estimate the target score. Let $R_\lambda$ be the risk of the
population-ridge dictionary and $R_D$ that of the fitted one.

\paragraph{Student fitting.} A \emph{residual normal equation} gives, for
each fixed $h$ and on a design event,
$\norm{(\widehat T_\lambda-T_\lambda)h}\lesssim
R_h[\sqrt{\deff(\lambda)/\Nfit}+\sqrt{\varkappa_\psi(\lambda)t/\Nfit}]$, with $\deff$
the effective dimension, $\varkappa_\psi$ the maximal regularized leverage and
$R_h$ a residual envelope, without assuming pseudo-labels independent of
inputs (\cref{q4:thm:learning}). It transfers to the risk and gain of every
bounded aggregate (\cref{q4:thm:geometry,q4:prop:risk-transfer}). For ridge
the cross term of \cref{prop:risk} must be kept: dropping it can flip the
gain sign.

\paragraph{Selection with an estimated Gram.}\label{sec:selection-gram}
Write $w=w_0+O\beta$ with a base point $w_0\in\Waff$ and $O$ spanning the
zero-sum directions, so $s_w=s_0+x^\top\beta$ with $s_0=s_{w_0}$ and contrast
features $x=O^\top(s_1,\dots,s_m)^\top$. Let $C_x$ be the candidate Gram,
$\widehat C_x$ its unlabeled estimate, $H_\alpha=C_x+\alpha I$,
$d_1(\alpha)=\tr[C_xH_\alpha^{-1}]$, $d_2(\alpha)=\tr[(C_xH_\alpha^{-1})^2]$ and
$\varkappa(\alpha)$ the maximal regularized leverage. Weights minimize the
$\alpha$-regularized empirical quadratic (Gram from $N$ unlabeled inputs,
linear term from $n$ anchors). A deterministic local conic oracle inequality
bounds the excess risk whenever
$\eta_\alpha=\norm{H_\alpha^{-1/2}(\widehat C_x-C_x)H_\alpha^{-1/2}}_\op<1/2$, without
an inverse smallest-eigenvalue factor (\cref{q4:thm:master}).

\paragraph{Finite-sample closure.}

Let $s_j^\circ=T_\lambda g_j$ be the population candidates and
$\widehat s_j=\widehat T_\lambda g_j$ the fitted ones, with
$\max_j\norm{\widehat s_j-s_j^\circ}\le\epsilon_D$ on the student-fitting
event of \cref{q4:thm:learning}. Fix a feasible comparator
$w^\circ=w_0+O\beta^\circ$ with $\norm{w^\circ}_1\le L$, independently of the
unlabeled and anchor samples. Assume the envelopes
$\abs{Y-s_0(X)}\le\Lambda_a$, $\E[(Y-s_0(X))^2\mid X]\le v_a^2$ and
$\abs{x(X)^\top\beta^\circ}\le\Lambda_\beta$, and put $t=\log(6m/\delta)$.

\begin{theorem}[Finite-sample closure]\label{thm:e2e}\label{thm:local}
With independent student-fitting, unlabeled and anchor samples and
$N\ge64\varkappa(\alpha)t$, with probability at least $1-\delta$ over the
unlabeled and anchor samples,
\begin{align*}
R_D(\widehat w)-R_\lambda(w^\circ)
\le{}&\underbrace{2L\sqrt{R_\lambda(w^\circ)}\,\epsilon_D+L^2\epsilon_D^2}_{\text{student fitting}}\\
&+\underbrace{24\,d_1(\alpha)\frac{\Lambda_\beta^2t}{N}}_{\text{unlabeled geometry}}
+\underbrace{24\,d_1(\alpha)\frac{v_a^2t}{n}}_{\text{anchors}}\\
&+\mathcal B_\alpha(w^\circ)+\mathcal R_{\mathrm{lev}}(n,N,\alpha,\delta),
\end{align*}
where $\mathcal B_\alpha(w^\circ)=6\alpha^2\norm{\beta^\circ}^2_{H_\alpha^{-1}}$ is the
regularization bias and
$\mathcal R_{\mathrm{lev}}=\tfrac{64}{3}\varkappa(\alpha)t^2(\Lambda_a^2/n^2+\Lambda_\beta^2/N^2)$
the leverage remainder. With known teacher structure, the residualized estimator
replaces the leading anchor variance $d_1(\alpha)v_a^2$ by $\dN(\alpha)v_r^2$,
where $v_r^2$ is the residual variance against $f_\Kk$ and
$\dN(\alpha)\to\dN$ as $\alpha\downarrow0$; its leverage and unlabeled
envelope change as well, and the structured leverage is not controlled by
$\varkappa(\alpha)$, so a smaller $\dN$ need not improve the whole bound in
proportion.
\formal{q4:thm:master,q4:thm:fast,q4:thm:affine,q4:thm:closure,q4:thm:structured,q4:prop:dN}
\end{theorem}

With $\alpha=64\kappa_x^2t/\min\{n,N\}$ the bias and remainder are of the
order of the variance terms (\cref{q4:cor:safe}); for unrestricted affine
weights in the homoscedastic anchor model the anchor term is exactly
$v_a^2d_2(\alpha)/n$ (\cref{q4:thm:affine}). The student-fitting term is first
order in $\epsilon_D$ unless an orthogonality condition holds
(\cref{q4:cor:orth}). The three bottlenecks use different samples: more
unlabeled data cannot replace anchors on uncovered directions, and more
anchors cannot repair the student's fitting error. Gain certificates
(\cref{q4:thm:gain4,q4:thm:integrated}), random features and linearized
fine-tuning (\cref{ext:scope}) are treated in the appendix.

\section{Synthetic Experiments}\label{sec:experiments}

\label{sec:synthetic}The synthetic experiments probe rates and constants the theory does not pin
down; identities are checked automatically (protocol in \cref{app:synthetic},
figures in \cref{app:x1,app:x2,app:x3,app:x4}).

\paragraph{Sign geometry.} With matched margins and noise, the passive anchors
needed for family-wise sign error $0.1$ stay flat (about $100$--$110$, within
resolution) for duplicated queries as $m$ grows from $2$ to $256$ and grow from $175$ to
$749$ for independent ones. A Gaussian-width prediction with no fitted
constant (the $n$ at which the $0.9$-quantile of the supremum of the
margin-normalized queries equals $\sqrt n$) lies within $6\%$ of every
empirical budget; the union bound of \cref{thm:certificate}(i) is $1.5$--$4$
times larger (\cref{fig:x3}(a), (h)).

\paragraph{Selection.} With passive anchors, $n\E[\Reg]/\sigma^2$ is $2.16$, $4.27$
and $8.65$ for $\dN=2,4,8$, against exact finite-sample values $2.16$, $4.31$
and $8.63$ (\cref{fig:teaser-c}); on one simplex it grows from $0.50$ to $7.11$
at a vertex and from $1$ to $511$ in the interior as $m$ goes from $2$ to $512$
(\cref{fig:x3}(c)). The exact risks of \cref{prop:compare}, including the
crossover with anchors alone, match simulation (\cref{fig:x3}(m)--(o)).

\paragraph{Finite samples.} Two designs with the same spectrum but regularized
leverage $3.4$ and $625$ have regret $0.063$ and $0.648$ at $N=64$; the two
regret components follow $d_1(\alpha)/N$ and $d_2(\alpha)/n$
(\cref{fig:x4}(a), (b)).

\paragraph{Accuracy versus value.}
In a constructed ensemble, supervisor accuracy and student value disagree
more when errors are structured (Spearman correlation $0.34$, against $0.70$
for isotropic errors; \cref{fig:x1}(b)); whether this happens in practice is an empirical question
we leave to future work.

\section{Discussion and Conclusion}\label{sec:conclusion}

A supervisor's value is a relation between its errors and the student's
learning operator. Unlabeled data cannot identify it, but what is missing is
finite-dimensional, reduced to the unresolved trace by teacher equations and
to one scalar by a student-relative label model. Anchors resolve the rest at
a cost set by its geometry. Without structure the resulting rule is
stacking; the theory says what it costs, when structure beats it, and when
weak supervisors are worth their anchors.

\paragraph{Limitations.}
Squared loss and one shared linear learner (frozen-representation heads,
kernel ridge, spectral filters; fine-tuning only under linearization); general
proper losses are open. Teacher equations and the label-model condition are
not verifiable from unlabeled data. The comparisons with stacking and with
anchors alone are exact only in Gaussian designs with homoscedastic noise.
All experiments are synthetic. Anchors are passive, and lower-bound constants are small.

\bibliographystyle{plainnat}
\bibliography{references}

\clearpage
\appendix
\raggedbottom
\section{Overview of the Appendix and Notation}\label{app:overview}

\begin{center}\small
\begin{tabular}{@{}lllr@{}}
\toprule
Main text & Appendix & Question answered & Page\\
\midrule
\cref{sec:value} & \cref{app:q1} & What determines a supervisor's value to a given student? & \pageref{app:q1}\\
\cref{sec:identification} & \cref{app:q2} & What can unlabeled data identify, and what is missing? & \pageref{app:q2}\\
\cref{sec:anchors} & \cref{app:q3} & How many clean labels does each decision need? & \pageref{app:q3}\\
\cref{sec:finite} & \cref{app:q4} & What survives with fitted students and finite unlabeled data? & \pageref{app:q4}\\
\cref{sec:experiments} & \cref{app:experiments} & Protocol, per-module details, figures & \pageref{app:experiments}\\
\cref{sec:related} & \cref{app:related} & Detailed comparison with related work & \pageref{app:related}\\
--- & \cref{app:ext} & Further results and their proofs (\cref{comp:ext:filters}--\ref{comp:short}) & \pageref{app:ext}\\
\bottomrule
\end{tabular}
\end{center}

\paragraph{Three ways to read the appendix.}
\begin{center}\small
\begin{tabularx}{\textwidth}{@{}>{\raggedright}p{0.24\textwidth}X@{}}
\toprule
Goal & Where to look\tabularnewline
\midrule
Check a main-text result & its ``[Formal: \dots]'' link, or the map at the start of each of \cref{app:q1,app:q2,app:q3,app:q4} (with page numbers)\tabularnewline
Understand the full theory & each appendix's summary and reading guide, then the banners that divide it into parts\tabularnewline
Check assumptions & the standing model opening each appendix; the checklist ending \cref{app:q1}; what is not established (\cref{q3:sec:summary3}); what is standard and what is new (\cref{app:boundary})\tabularnewline
\bottomrule
\end{tabularx}
\end{center}
Each appendix is self-contained up to the shared notation below and links to
any result it borrows from another.

\subsection{Notation translation}\label{app:translation}

Symbols that differ between the main text and the appendices:
\begin{center}\footnotesize
\begin{tabularx}{\textwidth}{@{}>{\raggedright}p{0.2\textwidth}>{\raggedright}p{0.22\textwidth}X@{}}
\toprule
Main text & Appendix & Meaning\tabularnewline
\midrule
$\te_j=Pe_j$ & $h_j$ (\cref{q1:sec:objectives,q1:sec:marginal}) & transmitted error of supervisor $j$; \cref{app:q1} states the marginal-value results for arbitrary vectors $h_j$ (for a general learner $h_j=Tg_j-f$)\tabularnewline
$\te_m^\star$ & $x_m^\star$ (\cref{q1:sec:marginal}) & minimum-norm point of the convex (or affine) hull of the current vectors\tabularnewline
$h_j=\Pi_\Nn q_j$ (\cref{sec:anchors}) & $h_j$ (\cref{app:q3}) & unresolved query direction (not the $h_j$ of \cref{app:q1})\tabularnewline
$M_j$, $\mathcal M$ & $M_j$, $\mathcal M$ & known teacher maps, $g_j=M_jf$; $A=\norm{Qf}^2$ is always the clean-label risk floor\tabularnewline
$R_\lambda$ (\cref{sec:finite}) & $R_\circ$ (\cref{app:q4}) & risk of the population-ridge candidates $s_j^\circ=T_\lambda g_j$\tabularnewline
$\Lambda_a$, $\Lambda_\beta$ (\cref{thm:e2e}) & $M_a$, $M_\beta$ (\cref{app:q4}) & envelopes of the anchor residual and of the comparator's contrast prediction $x^\top\beta^\circ$\tabularnewline
$\mathcal B_\alpha$, $\mathcal R_{\mathrm{lev}}$ & $6b_H(\beta,\alpha)$; leverage terms of \cref{q4:thm:fast} & regularization bias and leverage remainder\tabularnewline
$z_i$; $b_i$ (\cref{sec:compare}) & $z$; columns of $\mathsf B=U\diag(\nu)$ (\cref{q3:sec:compare}) & whitened unresolved features; map from regression coefficients to the target coordinate\tabularnewline
$R_{\mathrm{agg}}$ (\cref{sec:compare}) & $R_{\mathrm{agg}}(\Waff)$; $q_\parallel^{\Waff}$ for projection students & risk of the best affine aggregate in excess of $A_\Ss$\tabularnewline
\bottomrule
\end{tabularx}
\end{center}

\noindent\emph{Similar-looking symbols.} $N$ (unlabeled sample size) and
$\Nn$ (common nullspace); $T$ (learner) and $\mathcal T$ (tangent cone);
$\WL$ ($\ell_1$-bounded weights) and $\mathcal L=\Psi\Psi^*$; $\alpha$ (ridge
parameter) and $\alpha_{\mathcal M}$ (smallest singular value of the teacher
stack on $\Kk$); $\varkappa$
(leverage) and $\kappa_x$ (feature bound); $\mathsf D_T=K_g-K_s$ (prediction
Gram difference) and the fitting sample $D$; in $R_S$ the subscript means
``student''. In the cell model of \cref{q3:sec:cells} the number of cells equals
the contrast dimension $k$.

{\small
\subsection{Notation}\label{app:notation}
Appendices~B--E use the notation below. All matrices built from supervisor errors are \emph{uncentered Gram
matrices}, never covariances. ``Centering'' always means centering over the
supervisor index with $J$, never subtraction of an input mean.

\begin{center}\small
\begin{tabularx}{\textwidth}{@{}p{0.29\textwidth}X@{}}
\toprule
Symbol & Meaning\\
\midrule
$\Hh$, $\ip{\cdot}{\cdot}$, $\norm{\cdot}$ & real Hilbert space; $\Hh=L^2(P_X)$, $\ip hk=\E[h(X)k(X)]$ in statistical statements\\
$f$ & target (regression function); $\norm f^2$ never needs to be known\\
$g_j$, $e_j=g_j-f$, $j\le m$ & weak supervisors (teachers) and their errors\\
$\Waff,\ \Delta_m,\ \WL$ & $\{\one^\top w=1\}$; simplex; $\{\one^\top w=1,\ \norm w_1\le L\}$, $L\ge1$\\
$g_w,\ e_w,\ s_w,\ q_w$ & $\sum_jw_jg_j$, $\sum_jw_je_j$, $\sum_jw_js_j$, $g_w-s_w$\\
$\Ss$, $d_\Ss$, $P$, $Q=I-P$ & student class (closed subspace), its dimension, orthogonal projections\\
$A(f)=\norm{Qf}^2$ & clean-label approximation error (risk of $Pf$)\\
$T$; $s_j=Tg_j$, $q_j=g_j-s_j$ & common bounded linear learner (projection case $T=P$); students and residuals\\
$R_W(w)$, $R_T(w)$, $G_T(w)$ & $\norm{g_w-f}^2$, $\norm{s_w-f}^2$, $R_W-R_T$ (same-aggregate gain); $R_S=R_P$, $G=G_P$\\
$E$, $\Sigma=E^*E$ & error synthesis $Ew=\sum_jw_je_j$; raw error Gram\\
$\Gamma_T=E^*T^*TE$; $\Gpar=\Gamma_P$, $\Gperp=E^*QE$ & transmitted Gram; projection-case transmitted and filtered Grams\\
$b_T=(I-T)f$, $B_T=\norm{b_T}^2$ & clean-algorithm bias and risk (not a ceiling)\\
$c_T=E^*T^*b_T$, $\Lambda_T=\Sigma-\Gamma_T$, $\Eop_T=I-T^*T$ & bias alignment; energy-difference matrix; energy-removal operator\\
$K_g,K_s,K_q$; $\mathsf D_T=K_g-K_s$ & prediction Grams $(\ip{g_i}{g_j})$, $(\ip{s_i}{s_j})$, $(\ip{q_i}{q_j})$; observable gain quadratic\\
$\mu_q(f)_j=\ip f{q_j}$, $\mu_s(f)_j=\ip f{s_j}$ & hidden linear target moments\\
$J=I_m-\one\one^\top/m$; $\Upsilon_{jk}=\norm{g_j-g_k}^2$ & supervisor centering; squared disagreements\\
$U=\Span\{q_j\}$, $V=\Span\{s_j-s_m\}$, $k=\dim V$ & gain and selection task spaces\\
$\Phi:\R^k\to V$, $t_j=\Phi^*s_j$, $\theta=\Phi^*f$ & isometry onto $V$; candidate and target coordinates\\
$M_j$, $\mathcal M$, $\Nn=\Ker\mathcal M$, $\Kk=\Nn^\perp$ & known teacher maps ($g_j=M_jf$) and their stack; common nullspace; teacher-information space\\
$f=f_\Kk+f_\Nn$; $\ell_j=\Pi_\Nn Tg_j$; $d_\Nn$ & identified and hidden target parts; leakage; unresolved trace (an effective, possibly fractional dimension)\\
$f_0$; $\rho_\parallel,\rho_\perp,\rho$ & known reference; sensitivity radii\\
$n$; $(X_i,Y_i)$, $Y=f(X)+\xi$ & anchor budget and anchors, $\E[\xi\mid X]=0$\\
$\sigma$, $v_0$, $\tau$ & Gaussian noise level; residual second-moment bound; sub-Gaussian proxy\\
$\gamma$, $\delta$ & gain margin; failure probability\\
$N$, $N_{\mathrm{fit}}$ & unlabeled sample size for geometry; pseudo-labelled sample size for training\\
$\lambda$, $\alpha$ & student-training ridge; aggregation (selection) ridge\\
$\deff(\lambda)$, $d_1(\alpha)$, $d_2(\alpha)$, $\varkappa$ & effective dimensions; maximal regularized leverage\\
$O\in\R^{m\times(m-1)}$ & orthonormal basis of $\one^\perp$ (contrast coordinates)\\
$\mathcal T_{\mathcal C}(t)$, $\sd_\Omega$, $\gw$, $\rad$ & tangent cone; covariance-weighted statistical dimension; Gaussian width; radius\\
$\Reg_{\mathcal W}(\widehat w)$ & $R_T(\widehat w)-\inf_{w\in\mathcal W}R_T(w)$, selection regret\\
\bottomrule
\end{tabularx}
\end{center}

}

\clearpage
\section{Student-Relative Geometry: Formal Results and Proofs}\label{app:q1}

\paragraph{Summary.} \noindent
For squared loss and a student trained by a common linear learner $T$, we
give exact identities for the population risk of a student trained on any
affine aggregate of weak supervisors. The risk is
$R_T(w)=B_T+w^\top\Gamma_Tw-2c_T^\top w$: a clean-algorithm risk, a
\emph{transmitted} error Gram, and a bias-alignment term. For an orthogonal
projection the cross term vanishes, and only for a projection, giving
$R_S(w)=A+w^\top\Gpar w$ and gain $G(w)=w^\top\Gperp w-A$. We prove that
neither the error-energy spectrum nor the complete raw error Gram $\Sigma$
determines student risk: the realizable transmitted Grams for a fixed
$\Sigma$ are exactly the Loewner interval $0\preceq\Gamma\preceq\Sigma$ under
rank constraints, so student risk can take every value in
$[A,A+w^\top\Sigma w]$. We characterize weak-optimal, student-optimal and
gain-maximizing weights (including singular Grams and simplex constraints),
give exact oracle regret identities showing that raw weighting can be
arbitrarily suboptimal, and characterize exactly when an additional
supervisor has marginal value. Spectral-filter learners (ridge, gradient
descent, gradient flow, cutoff) and operator perturbations are treated
exactly. Everything here is \emph{oracle} geometry: it involves the unknown
target. Whether it is observable is the subject of \cref{app:q2}.

\paragraph{Where the main text uses this appendix.}
Formal versions of the main-text results; supporting results are cited in their proofs.

\begin{center}\small
\begin{tabularx}{\textwidth}{@{}>{\raggedright}p{0.24\textwidth}X@{}}
\toprule
Main-text result & Formal results\tabularnewline
\midrule
\cref{prop:risk} & \cref{q1:thm:single-general} (p.~\pageref{q1:thm:single-general}); \cref{q1:thm:multi} (p.~\pageref{q1:thm:multi}); \cref{q1:prop:orth-only} (p.~\pageref{q1:prop:orth-only})\tabularnewline
\cref{thm:fixed-gram} & \cref{q1:thm:realize} (p.~\pageref{q1:thm:realize}); \cref{q1:cor:fixed-gram-range} (p.~\pageref{q1:cor:fixed-gram-range}); \cref{q1:cor:no-gram} (p.~\pageref{q1:cor:no-gram})\tabularnewline
\cref{thm:decisions} & \cref{q1:thm:kkt} (p.~\pageref{q1:thm:kkt}); \cref{q1:cor:raw-regret} (p.~\pageref{q1:cor:raw-regret}); \cref{q1:thm:raw-arbitrary} (p.~\pageref{q1:thm:raw-arbitrary})\tabularnewline
Marginal value, \cref{sec:value} & \cref{q1:thm:marg-aff} (p.~\pageref{q1:thm:marg-aff}); \cref{q1:thm:marg-sim} (p.~\pageref{q1:thm:marg-sim})\tabularnewline
\bottomrule
\end{tabularx}
\end{center}

\paragraph{Notation.} As in \cref{app:notation,app:translation}.

\paragraph{Reading guide.} The results sit at four levels. The second
contains the central new results; the first is reused throughout; the third
extends the same multi-supervisor framework beyond projections using standard
linear learners and spectral filters; the fourth is an optional case study.
The first two are in this appendix, marked by banners; the last two are
summarized in \cref{app:ext}, with complete versions in
\cref{comp:ext:filters,comp:ext:perturb,comp:ext:moniri,comp:ext:rates,comp:ext:scope,comp:short}.
\begin{center}\small
\begin{tabularx}{\textwidth}{@{}>{\raggedright}p{0.2\textwidth}>{\raggedright}X>{\raggedright\arraybackslash}p{0.26\textwidth}@{}}
\toprule
Level & Content & Sections\tabularnewline
\midrule
Background and reusable identities & single-supervisor risk identity for any linear learner, projection split into transmitted and filtered error, aggregate risk, benchmarks & \cref{q1:sec:model,q1:sec:single,q1:sec:multi,q1:sec:benchmarks}\tabularnewline
Core contribution & a fixed spectrum or a fixed raw Gram does not determine the student; decision regret of raw weighting; marginal value of a supervisor; best single versus best aggregate & \cref{q1:sec:orientation,q1:sec:realizability,q1:sec:objectives,q1:sec:regret,q1:sec:marginal,q1:sec:best}\tabularnewline
Beyond projections & spectral views; ridge, early stopping, gradient flow and spectral cut-off; operator perturbation and projection limits & \cref{ext:filters,ext:perturb}\tabularnewline
Optional case study & proportional ridge-to-ridge model & \cref{ext:moniri}\tabularnewline
\bottomrule
\end{tabularx}
\end{center}
Symbols that differ from the main text are listed in \cref{app:translation};
in particular, the marginal-value results are stated for arbitrary vectors
$h_j$, which the main text instantiates as $\te_j=Pe_j$.

\apppart{Background and reusable identities}{Exact risk identities for one supervisor and for aggregates; used by every later appendix.}
\subsection{Scope, model and standing conventions}\label{q1:sec:model}

\paragraph{Role in the paper.} \cref{app:q1} answers the first of four questions:
\emph{which object determines whether a weak supervisor helps a given
student?} It contains deterministic population identities only. No
observability, estimation or finite-sample statement is made here.

\paragraph{Standing model.}
$\Hh$ is a real Hilbert space with inner product $\ip\cdot\cdot$ and norm
$\norm\cdot$; in statistical applications $\Hh=L^2(P_X)$. The target is a
fixed $f\in\Hh$. There are $m\ge1$ fixed weak supervisors $g_j=f+e_j$.
The student class $\Ss\subseteq\Hh$ is a closed linear subspace, with
orthogonal projections
\[
 P=P_\Ss,\qquad Q=I-P,\qquad P^2=P=P^*,\quad Q^2=Q=Q^*,\quad PQ=QP=0 .
\]
Every $h\in\Hh$ satisfies $h=Ph+Qh$ and $\norm h^2=\norm{Ph}^2+\norm{Qh}^2$.
Weights are restricted to
\[
 \Waff=\{w\in\R^m:\one^\top w=1\},\qquad
 \Delta_m=\{w\in\Waff:w\ge0\},\qquad
 \WL=\{w\in\Waff:\norm w_1\le L\}.
\]
For $w\in\Waff$, $g_w=\sum_jw_jg_j=f+e_w$ with $e_w=\sum_jw_je_j=Ew$, where
the \emph{error synthesis operator} $E:\R^m\to\Hh$ has adjoint
$(E^*h)_j=\ip{e_j}h$.

\begin{remark}[Why affine weights]\label{q1:rem:affine}
If $\one^\top w\ne1$ then $g_w-f=(\one^\top w-1)f+e_w$, and target-dependent
terms enter every risk. All identities below require $\one^\top w=1$, which
preserves the common target coefficient.
\end{remark}

\begin{assumption}[Frozen common linear learner]\label{q1:ass:T}
$T:\Hh\to\Hh$ is bounded and linear, identical for all supervisors and all
weights. Then $s_j=Tg_j$ and $s_w=Tg_w=\sum_jw_js_j$. Neither
self-adjointness, contractivity, idempotence nor $\Ran T\subseteq\Ss$ is
assumed unless stated. Hyperparameters (ridge level, stopping time) are held
fixed while the pseudo-labels or weights vary; retuning them as a nonlinear
function of each label mixture does not define one linear $T$.
\end{assumption}

The projection learner $T=P$ is the population least-squares fit over $\Ss$
(Proposition~\ref{q1:prop:clean-projection}). Risks and gain are
\[
 R_W(w)=\norm{g_w-f}^2,\qquad R_T(w)=\norm{Tg_w-f}^2,\qquad
 G_T(w)=R_W(w)-R_T(w),
\]
with $R_S=R_P$ and $G=G_P$. For a single error $e$ write
$R_W(e)=\norm e^2$, $R_T(e)=\norm{T(f+e)-f}^2$ and $G_T(e)=R_W(e)-R_T(e)$.
A positive gain compares a student with \emph{its own} weak aggregate; it
does not compare it with the best single supervisor
(Section~\ref{q1:sec:benchmarks}).

\subsection{Exact risk of a student trained on one weak supervisor}\label{q1:sec:single}

\subsubsection{Clean and weak-label projection students}

\begin{proposition}[Clean-label projection]\label{q1:prop:clean-projection}
The minimizer of $\norm{h-f}^2$ over $h\in\Ss$ is $Pf$, uniquely, with risk
$A(f)=\norm{Qf}^2$, and for every $h\in\Ss$,
$\norm{h-f}^2=\norm{h-Pf}^2+A(f)$. The same holds with $f$ replaced by any
$g\in\Hh$: the weak-label student is $Pg=Pf+Pe$, so
$Pg-Pf=Pe$.
\par{\upshape\small Immediate from Pythagoras: $h-f=(h-Pf)-Qf$ with $h-Pf\in\Ss$, $Qf\in\Ss^\perp$; and $Pg=Pf+Pe$.}
\end{proposition}

We call $A=A(f)$ the \emph{clean-label approximation error}. The class is
\emph{realizable} for $f$ if $f\in\Ss$ ($A=0$) and \emph{misspecified} if
$A>0$.

\subsubsection{General linear learners}

Define the clean-algorithm bias and its risk
\[
 b_T=(I-T)f,\qquad B_T=B_T(f)=\norm{b_T}^2 .
\]
$Tf$ is the output of the same algorithm on clean labels. In contrast $Pf$
minimizes risk over $\Ss$; $B_T$ is not a new approximation ceiling.

\begin{theorem}[Complete risk and gain identities]\label{q1:thm:single-general}
For every $f,e\in\Hh$, with $g=f+e$, $s=Tg$ and $q=g-s$,
\begin{align}
 R_T(e)&=B_T+\norm{Te}^2-2\ip{Te}{b_T},\label{q1:eq:single-risk}\\
 G_T(e)&=\ip e{(I-T^*T)e}+2\ip e{T^*(I-T)f}-B_T,\label{q1:eq:single-gain}\\
 G_T(e)&=\norm q^2+2\ip sq-2\ip fq=\norm g^2-\norm s^2-2\ip fq.\label{q1:eq:single-observable}
\end{align}
The term $2\ip sq$ cannot be discarded in general.
\end{theorem}
\begin{proof}
Linearity gives $T(f+e)-f=Te-b_T$; expanding its squared norm proves
\eqref{q1:eq:single-risk}. Subtracting from $\norm e^2$ and using
$\norm{Te}^2=\ip e{T^*Te}$ and $\ip{Te}{b_T}=\ip e{T^*(I-T)f}$ proves
\eqref{q1:eq:single-gain}. For \eqref{q1:eq:single-observable}, $g=s+q$ gives
$\norm{g-f}^2=\norm{s-f}^2+\norm q^2+2\ip{s-f}q$; subtract $\norm{s-f}^2$.
Finally $\norm g^2-\norm s^2=\norm q^2+2\ip sq$.
\end{proof}

\begin{proposition}[Range-constrained learners]\label{q1:prop:range}
If $\Ran T\subseteq\Ss$, then
\[
 B_T=A(f)+\norm{(P-T)f}^2,\qquad R_T(e)=A(f)+\norm{Te-(P-T)f}^2 .
\]
Thus $R_T\ge A(f)$, with equality iff $Te=(P-T)f$. There is no general
inequality $R_T\ge B_T$, and without the range restriction $A(f)$ need not
lower-bound $R_T$.
\end{proposition}
\begin{proof}
Since $Tf,Te,Pf\in\Ss$, $b_T=Qf+(P-T)f$ and $T(f+e)-f=Te-(P-T)f-Qf$; the
$\Ss$-terms are orthogonal to $Qf$. For the last claim take $\Ss=\{0\}$,
$T=I$, $e=0$, $f\ne0$: then $R_T=0<A(f)$. Example~\ref{q1:ex:clean-not-ceiling}
disproves $R_T\ge B_T$ even under the range restriction.
\end{proof}

\begin{proposition}[Universal disappearance of the cross term]\label{q1:prop:orth-only}
The following are equivalent: (i) $\ip{Te}{(I-T)f}=0$ for all $e,f\in\Hh$;
(ii) $T^*(I-T)=0$; (iii) $T$ is an orthogonal projection. For one fixed $f$
the condition is $T^*b_T=0$, which does not force $T$ to be a projection.
\end{proposition}
\begin{proof}
(i) states $\ip e{T^*(I-T)f}=0$ for all $e,f$, i.e.\ (ii). From (ii),
$T^*=T^*T$; taking adjoints, $T=T^*T$, so $T=T^*$ and $T=T^2$: a self-adjoint
idempotent is the orthogonal projection onto its closed range. Conversely a
projection has $T^*(I-T)=T-T^2=0$.
\end{proof}

\begin{example}[Clean-algorithm risk is not a ceiling]\label{q1:ex:clean-not-ceiling}
Let $\Hh=\Ss=\R$, $f=1$, $T=\tfrac12I$. Clean labels give $B_T=1/4$. The weak
supervisor $g=2$ ($e=1$) gives $\norm{Te}^2=1/4$, $2\ip{Te}{b_T}=1/2$ and
student risk $0$. With $g=f$ the weak risk is $0$ and the student risk is
$1/4$: realizability in $\Ss$ alone does not ensure nonnegative gain for a
shrinkage learner.
\end{example}

\subsubsection{The projection case: transmitted and filtered error}

\begin{theorem}[Exact weak-to-strong identity for a projection]\label{q1:thm:single-projection}
For every $e\in\Hh$,
\[
 \boxed{R_S(e)=A+\norm{Pe}^2,\qquad G(e)=\norm{Qe}^2-A,\qquad
 R_W(e)=\underbrace{\norm{Pe}^2}_{\text{transmitted}}+\underbrace{\norm{Qe}^2}_{\text{filtered}}.}
\]
Hence $R_S(e)<R_W(e)\iff\norm{Qe}^2>A$, with equality and reversal
characterized likewise. The student replaces the filtered error energy
$\norm{Qe}^2$ by the clean approximation cost $A$.
\end{theorem}
\begin{proof}
By Proposition~\ref{q1:prop:clean-projection}, $Pg-f=Pe-Qf$ with
$\ip{Pe}{Qf}=0$, so $R_S=\norm{Pe}^2+A$. Also $\norm e^2=\norm{Pe}^2+\norm{Qe}^2$.
This is Theorem~\ref{q1:thm:single-general} with $T=P$: $b_P=Qf$, $B_P=A$ and
$Pe\perp Qf$.
\end{proof}

\subsection{Multi-supervisor geometry}\label{q1:sec:multi}

\begin{definition}[Error Gram matrices]\label{q1:def:grams}
\[
 \Sigma=E^*E,\qquad \Gamma_T=E^*T^*TE,\qquad
 \Gpar=\Gamma_P=E^*PE,\qquad \Gperp=E^*QE,
\]
entrywise $\Sigma_{jk}=\ip{e_j}{e_k}$, $(\Gamma_T)_{jk}=\ip{Te_j}{Te_k}$,
$(\Gpar)_{jk}=\ip{Pe_j}{Pe_k}$, $(\Gperp)_{jk}=\ip{Qe_j}{Qe_k}$. Further
\[
 c_T=E^*T^*b_T,\quad(c_T)_j=\ip{Te_j}{b_T},\qquad
 \Lambda_T=\Sigma-\Gamma_T=E^*\Eop_TE,\quad \Eop_T=I-T^*T .
\]
$\Sigma$ is the \emph{raw}, $\Gamma_T$ the \emph{transmitted}, $\Gperp$ the
\emph{filtered} error Gram; $c_T$ is the \emph{bias alignment}. $\Lambda_T$
is an energy-difference matrix, not necessarily PSD.
\end{definition}

\begin{proposition}[PSD structure and exact decomposition]\label{q1:prop:psd}
$\Sigma,\Gamma_T,\Gpar,\Gperp$ are symmetric PSD, with
$w^\top\Sigma w=\norm{e_w}^2$, $w^\top\Gamma_Tw=\norm{Te_w}^2$,
$w^\top\Gpar w=\norm{Pe_w}^2$, $w^\top\Gperp w=\norm{Qe_w}^2$ for every
$w\in\R^m$, and $\boxed{\Sigma=\Gpar+\Gperp}$. Moreover
\[
 0\preceq\Gpar,\Gperp\preceq\Sigma,\qquad 0\preceq\Gamma_T\preceq\norm T_\op^2\Sigma,
\]
\[
 \rank\Gamma_T=\rank(TE)\le\min\{\rank\Sigma,\dim\overline{\Ran T}\},
\]
\[
 \rank\Gpar\le\min\{\rank\Sigma,d_\Ss\},\qquad
 \rank\Gperp\le\min\{\rank\Sigma,\dim\Ss^\perp\}.
\]
\end{proposition}
\begin{proof}
$w^\top\Gamma_Tw=\ip{TEw}{TEw}$ and similarly for the others; for $P$,
$\ip{Ew}{PEw}=\norm{PEw}^2$ because $P=P^*=P^2$. $\Gpar+\Gperp=E^*(P+Q)E=\Sigma$.
The Loewner bounds follow from $\norm{TEw}^2\le\norm T_\op^2\norm{Ew}^2$ and
$\Sigma-\Gpar=\Gperp\succeq0$. The kernel of $\Gamma_T$ equals that of $TE$,
since $w^\top\Gamma_Tw=\norm{TEw}^2$; rank--nullity and
$\rank E=\rank\Sigma$ give the rank bounds, together with
$\Ran(TE)\subseteq\overline{\Ran T}$.
\end{proof}

\begin{theorem}[Complete aggregate risk]\label{q1:thm:multi}
For every $w\in\Waff$,
\begin{align*}
 R_W(w)&=w^\top\Sigma w,\\
 R_T(w)&=B_T+w^\top\Gamma_Tw-2c_T^\top w,\\
 G_T(w)&=w^\top\Lambda_Tw+2c_T^\top w-B_T .
\end{align*}
At $T=P$: $B_P=A$, $c_P=0$, $\Gamma_P=\Gpar$, $\Lambda_P=\Gperp$, so
\[
 \boxed{R_S(w)=A+w^\top\Gpar w,\qquad G(w)=w^\top\Gperp w-A,\qquad
 R_W(w)=w^\top\Gpar w+w^\top\Gperp w .}
\]
\end{theorem}
\begin{proof}
Affinity gives $g_w=f+Ew$; insert $e=Ew$ in Theorem~\ref{q1:thm:single-general}.
For $T=P$, $b_P=Qf$ and $c_P=E^*PQf=0$.
\end{proof}

\begin{proposition}[Gram of the actual student errors]\label{q1:prop:HT}
Let $z_j=Tg_j-f=Te_j-b_T$ and $Z_Tw=\sum_jw_jz_j$. Then
\[
 H_T:=Z_T^*Z_T=\Gamma_T-c_T\one^\top-\one c_T^\top+B_T\one\one^\top\succeq0,
 \qquad R_T(w)=w^\top H_Tw\quad(w\in\Waff).
\]
Quadratic optimization over weights therefore applies to $H_T$, but not to
$\Gamma_T$ alone unless $c_T\in\Span\{\one\}$.
\par{\upshape\small Immediate from $\ip{Te_j-b_T}{Te_k-b_T}=(\Gamma_T)_{jk}-(c_T)_j-(c_T)_k+B_T$ and $w^\top\one=1$ on $\Waff$ (\cref{q1:thm:multi}).}
\end{proposition}

\subsection{Comparison benchmarks}\label{q1:sec:benchmarks}

For fixed errors, define the clean-label risk $R_{\mathrm{clean}}=A$, the best
individual weak risk $R_{\mathrm{best}}=\min_j\Sigma_{jj}$, the aggregate weak
risk $R_{\mathrm{agg}}(w)=w^\top\Sigma w$ and the student risk
$R_{\mathrm{student}}(w)=A+w^\top\Gpar w$ (projection learner).

\begin{theorem}[Exact benchmark comparisons]\label{q1:thm:benchmarks}
For every $w\in\Waff$,
\begin{align*}
 R_{\mathrm{student}}(w)-R_{\mathrm{clean}}&=w^\top\Gpar w=\norm{Pe_w}^2\ge0,\\
 R_{\mathrm{agg}}(w)-R_{\mathrm{student}}(w)&=w^\top\Gperp w-A,\\
 R_{\mathrm{best}}-R_{\mathrm{student}}(w)&=R_{\mathrm{best}}-A-w^\top\Gpar w .
\end{align*}
The student reaches the clean ceiling iff $Pe_w=0$; beats its own aggregate
iff $w^\top\Gperp w>A$; beats the best individual supervisor iff
$A+w^\top\Gpar w<R_{\mathrm{best}}$.
\end{theorem}
\begin{proof}
Theorem~\ref{q1:thm:multi} and the definitions.
\end{proof}

\apppart{Core contribution: raw geometry versus student-relative value}{Neither the error spectrum nor the full raw Gram determines the student; consequences for weights and marginal value.}
\subsection{Orientation: a fixed error spectrum does not determine the gain}\label{q1:sec:orientation}

A single fixed vector has no spectrum. To study spectra without a random-error
model, use a deterministic weighted error family $\{(\pi_a,e_a)\}_{a=1}^{M}$,
$\pi_a>0$, $\sum_a\pi_a=1$, with average risks
$\overline R_W=\sum_a\pi_aR_W(e_a)$, $\overline R_S=\sum_a\pi_aR_S(e_a)$,
$\overline G=\overline R_W-\overline R_S$, and the \emph{uncentered
error-energy operator} $C=\sum_a\pi_a\,e_a\otimes e_a$, where
$(u\otimes v)h=\ip vhu$. It is not a covariance and needs no mean-zero
assumption.

\begin{lemma}[Error-energy operator]\label{q1:lem:energy}
$C$ is finite-rank, self-adjoint and PSD, and $\tr(BC)=\sum_a\pi_a\ip{e_a}{Be_a}$
for bounded $B$. In particular $\tr C=\overline R_W$,
$\tr(PC)=\sum_a\pi_a\norm{Pe_a}^2$, $\tr(QC)=\sum_a\pi_a\norm{Qe_a}^2$, and
\[
 \overline R_W=\tr C,\qquad \overline R_S=A+\tr(PC),\qquad \overline G=\tr(QC)-A .
\]
\end{lemma}
\begin{proof}
Each $e_a\otimes e_a$ is self-adjoint with $\ip h{(e_a\otimes e_a)h}=\ip{e_a}h^2\ge0$;
its range is $\Span\{e_a\}$; $\tr(B(e_a\otimes e_a))=\ip{e_a}{Be_a}$ by
Lemma~\ref{q1:lem:rank-one}. Use $\ip{e_a}{Pe_a}=\norm{Pe_a}^2$ and average
Theorem~\ref{q1:thm:single-projection}.
\end{proof}

For this section $\Hh=\R^d$, $1\le d_\Ss=\rank P\le d-1$, $q=d-d_\Ss$, and a
prescribed spectrum $\lambda_1\ge\dots\ge\lambda_d\ge0$ with
$\Lambda=\diag(\lambda_i)$; $C_U=U\Lambda U^\top$ for $U\in O(d)$.

\begin{lemma}[Projection weights]\label{q1:lem:projweights}
For an orthonormal basis $(u_i)$ and $a_i=\norm{Qu_i}^2$: $0\le a_i\le1$,
$\sum_ia_i=q$, $a_i=1\iff u_i\in\Ran Q$, $a_i=0\iff u_i\in\Ran P$.
\par{\upshape\small Immediate from $\norm{Pu_i}^2+\norm{Qu_i}^2=1$ and $\sum_i\ip{u_i}{Qu_i}=\tr Q=q$.}
\end{lemma}

\begin{theorem}[Sharp fixed-spectrum orientation extrema]\label{q1:thm:orient}
\[
 \min_{U\in O(d)}\tr(QC_U)=\sum_{i=d_\Ss+1}^d\lambda_i,\qquad
 \max_{U\in O(d)}\tr(QC_U)=\sum_{i=1}^{q}\lambda_i .
\]
The maximum is attained iff
$E_{>\lambda_q}(C_U)\subseteq\Ran Q\subseteq E_{\ge\lambda_q}(C_U)$ and the
minimum iff $E_{<\lambda_{d_\Ss+1}}(C_U)\subseteq\Ran Q\subseteq E_{\le\lambda_{d_\Ss+1}}(C_U)$,
where $E_{>t}(C)$ etc.\ are spectral subspaces. For a simple spectrum the
extremal $\Ran Q$ are unique.
\end{theorem}
\begin{proof}
With $u_i$ the columns of $U$, $\tr(QC_U)=\sum_i\lambda_ia_i$
(Lemma~\ref{q1:lem:energy}). Put $M=\sum_{i\le q}(1-a_i)=\sum_{i>q}a_i$. Then
\[
 \sum_{i\le q}\lambda_i-\sum_i\lambda_ia_i
 =\sum_{i\le q}\lambda_i(1-a_i)-\sum_{i>q}\lambda_ia_i
 \ge\lambda_qM-\lambda_qM=0,
\]
since $\lambda_i\ge\lambda_q$ for $i\le q$ and $\lambda_i\le\lambda_q$ for
$i>q$. Equality requires $a_i=1$ whenever $\lambda_i>\lambda_q$ and $a_i=0$
whenever $\lambda_i<\lambda_q$; by Lemma~\ref{q1:lem:projweights} this is the
displayed inclusion, the remaining dimensions of $\Ran Q$ lying in
$E_{=\lambda_q}$. Conversely these inclusions make $\tr(QC_U)$ the sum of the
$q$ largest eigenvalues. For the minimum, $\tr(QC_U)=\tr C_U-\tr(PC_U)$ and
the upper bound applied to the rank-$d_\Ss$ projection $P$ gives
$\tr(PC_U)\le\sum_{i\le d_\Ss}\lambda_i$; the equality case is symmetric.
\end{proof}

\begin{proposition}[The orientation range is an interval]\label{q1:prop:orient-interval}
$\{\tr(QC_U):U\in O(d)\}=\bigl[\sum_{i>d_\Ss}\lambda_i,\ \sum_{i\le q}\lambda_i\bigr]$,
and the average gain ranges over this interval shifted by $-A$. Hence every
orientation gives strict improvement iff $\sum_{i>d_\Ss}\lambda_i>A$; some
orientation does iff $\sum_{i\le q}\lambda_i>A$; both strict signs occur iff
$\sum_{i>d_\Ss}\lambda_i<A<\sum_{i\le q}\lambda_i$.
\end{proposition}
\begin{proof}
If $\det U=-1$, replace $U$ by $UD$ with $D$ a diagonal sign matrix of
determinant $-1$; $D\Lambda D=\Lambda$. So the image of the continuous map
$U\mapsto\tr(QU\Lambda U^\top)$ on the connected group $SO(d)$ ($d\ge2$) is
an interval containing both endpoints of Theorem~\ref{q1:thm:orient} and nothing
outside them. Subtract $A$ (Lemma~\ref{q1:lem:energy}).
\end{proof}

\begin{theorem}[Two-dimensional sign reversal]\label{q1:thm:2d}
Let $\Hh=\R^2$, $\Ss=\Span\{v_1\}$, $f=s+\sqrt A\,v_2$ ($s\in\Ss$, $A>0$), and
$\lambda_1>\lambda_2\ge0$ with $\lambda_2<A<\lambda_1$. The worlds
$C_F=\lambda_2v_1\otimes v_1+\lambda_1v_2\otimes v_2$ (filterable) and
$C_I=\lambda_1v_1\otimes v_1+\lambda_2v_2\otimes v_2$ (imitable) have equal
spectra and $\overline R_W=\lambda_1+\lambda_2$, but
\[
 \overline R_S^F=A+\lambda_2,\ \overline G^F=\lambda_1-A>0;\qquad
 \overline R_S^I=A+\lambda_1,\ \overline G^I=\lambda_2-A<0,
\]
while $A<\lambda_1+\lambda_2$ in both worlds. They are realized by two
equally weighted errors, $e_1^F=\sqrt{2\lambda_1}v_2$,
$e_2^F=\sqrt{2\lambda_2}v_1$, and the same with $v_1,v_2$ exchanged.
For $(\lambda_1,\lambda_2,A)=(4,1,2)$: $\overline R_W=5$,
$\overline R_S\in\{3,6\}$, $\overline G\in\{2,-1\}$.
\end{theorem}
\begin{proof}
$\tr(PC_F)=\lambda_2$, $\tr(QC_F)=\lambda_1$ and conversely for $C_I$;
Lemma~\ref{q1:lem:energy}. $A<\lambda_1\le\lambda_1+\lambda_2$. The realization:
$\tfrac12(2\lambda_1)v_2\otimes v_2+\tfrac12(2\lambda_2)v_1\otimes v_1=C_F$.
\end{proof}

\subsection{A fixed raw error Gram does not determine the student}\label{q1:sec:realizability}

\begin{theorem}[Necessary and sufficient realizability of a student-relative split]\label{q1:thm:realize}
Let $\Sigma,G_\parallel,G_\perp\in\R^{m\times m}$ be symmetric and
$\Hh=\Ss\oplus\Ss^\perp$ with $d_\Ss=\dim\Ss$, $d_\perp=\dim\Ss^\perp$ (possibly
infinite). There exist $e_1,\dots,e_m\in\Hh$ with
$\ip{e_j}{e_k}=\Sigma_{jk}$, $\ip{Pe_j}{Pe_k}=(G_\parallel)_{jk}$ and
$\ip{Qe_j}{Qe_k}=(G_\perp)_{jk}$ iff
\[
 \boxed{G_\parallel\succeq0,\quad G_\perp\succeq0,\quad \Sigma=G_\parallel+G_\perp,\quad
 \rank G_\parallel\le d_\Ss,\quad \rank G_\perp\le d_\perp .}
\]
\end{theorem}
\begin{proof}
Necessity is Proposition~\ref{q1:prop:psd}. For sufficiency, with
$k_\parallel=\rank G_\parallel$ and $k_\perp=\rank G_\perp$, take Gram
factorizations $G_\parallel=B_\parallel^\top B_\parallel$,
$B_\parallel\in\R^{k_\parallel\times m}$, and
$G_\perp=B_\perp^\top B_\perp$ (Lemma~\ref{q1:lem:gram-factor}) with columns
$b_{\parallel,j},b_{\perp,j}$. The rank conditions give linear isometries
$J_\parallel:\R^{k_\parallel}\to\Ss$ and $J_\perp:\R^{k_\perp}\to\Ss^\perp$.
Put $e_j=J_\parallel b_{\parallel,j}+J_\perp b_{\perp,j}$. Then
$Pe_j=J_\parallel b_{\parallel,j}$, $Qe_j=J_\perp b_{\perp,j}$, so the
projected Grams are $G_\parallel,G_\perp$ and
$\ip{e_j}{e_k}=(G_\parallel+G_\perp)_{jk}=\Sigma_{jk}$.
\end{proof}

\begin{corollary}[All transmitted geometries for a fixed raw Gram]\label{q1:cor:feasible}
For fixed $\Sigma\succeq0$ the realizable transmitted Grams are exactly
\[
 \mathfrak G(\Sigma;d_\Ss,d_\perp)=\{G:0\preceq G\preceq\Sigma,\
 \rank G\le d_\Ss,\ \rank(\Sigma-G)\le d_\perp\}.
\]
If $d_\Ss,d_\perp\ge\rank\Sigma$, every Loewner split $0\preceq G\preceq\Sigma$
is realizable, and $G=\Sigma^{1/2}K\Sigma^{1/2}$ for a symmetric $K$ with
$0\preceq K\preceq\Pi_\Sigma$ (the projection onto $\Ran\Sigma$); one choice is
$K=\Sigma^{\dagger/2}G\Sigma^{\dagger/2}$.
\end{corollary}
\begin{proof}
Apply Theorem~\ref{q1:thm:realize} with $G_\parallel=G$, $G_\perp=\Sigma-G$. If
$0\preceq G\preceq\Sigma$ then $\Ker\Sigma\subseteq\Ker G\cap\Ker(\Sigma-G)$
(Lemma~\ref{q1:lem:kernel-inclusion}), so both ranks are at most $\rank\Sigma$.
For the parametrization: if $G=\Sigma^{1/2}K\Sigma^{1/2}$ with
$0\preceq K\preceq\Pi_\Sigma$ then $0\preceq G$ and
$\Sigma-G=\Sigma^{1/2}(\Pi_\Sigma-K)\Sigma^{1/2}\succeq0$. Conversely
$G=\Pi_\Sigma G\Pi_\Sigma$ by the kernel inclusion; with
$K=\Sigma^{\dagger/2}G\Sigma^{\dagger/2}$ we get $K\succeq0$,
$\Pi_\Sigma-K=\Sigma^{\dagger/2}(\Sigma-G)\Sigma^{\dagger/2}\succeq0$ and
$\Sigma^{1/2}K\Sigma^{1/2}=\Pi_\Sigma G\Pi_\Sigma=G$.
\end{proof}

\begin{corollary}[Sharp fixed-Gram range]\label{q1:cor:fixed-gram-range}
Suppose $\rank\Sigma\le\min\{d_\Ss,d_\perp\}$ and fix $w\in\Waff$. Across all
configurations with raw error Gram exactly $\Sigma$,
\[
 \boxed{R_S(w)\in[A,\ A+w^\top\Sigma w],\qquad G(w)\in[-A,\ w^\top\Sigma w-A],}
\]
and every point is attained. In particular the fully filterable split
$(\Gpar,\Gperp)=(0,\Sigma)$ and the fully imitable split $(\Sigma,0)$ are both
realizable: for every $w$, $R_S^F(w)=A$, $G^F(w)=w^\top\Sigma w-A$, while
$R_S^I(w)=A+w^\top\Sigma w$, $G^I(w)=-A$. The dimension conditions are
sharp for the endpoint \emph{matrices}: the split $(\Sigma,0)$ needs
$\rank\Sigma\le d_\Ss$ and $(0,\Sigma)$ needs $\rank\Sigma\le d_\perp$. They are
not necessary for the scalar range at a fixed $w$: with $\Sigma=I_2$,
$d_\Ss=d_\perp=1$ and $w=(1,0)$, placing $e_1$ in $\Ss$ or in $\Ss^\perp$ (and
$e_2$ in the other) attains both endpoints although $\rank\Sigma=2$.
\end{corollary}
\begin{proof}
For $t\in[0,1]$ the split $(t\Sigma,(1-t)\Sigma)$ is realizable
(Corollary~\ref{q1:cor:feasible}), with $R_S=A+t\,w^\top\Sigma w$ and
$G=(1-t)w^\top\Sigma w-A$. Conversely $0\le w^\top\Gpar w\le w^\top\Sigma w$.
At the endpoints $\Gpar=\Sigma$ or $\Gperp=\Sigma$, forcing the rank conditions.
\end{proof}

\begin{corollary}[No full-Gram-only diagnostic; individual reversal]\label{q1:cor:no-gram}
Under the conditions of Corollary~\ref{q1:cor:fixed-gram-range}, no function of
$(\Sigma,w,A)$ determines the sign of $G(w)$ whenever
$0<A<w^\top\Sigma w$. For every $j$ with $\Sigma_{jj}>A$, training on
supervisor $j$ alone gives $R_S^F(e_j)=A<\Sigma_{jj}$ but
$R_S^I(e_j)=A+\Sigma_{jj}>\Sigma_{jj}$. In world $F$ every aggregate reaches
the clean ceiling and beats the best supervisor when $A<\min_j\Sigma_{jj}$; in
world $I$ an aggregate is worse than the best supervisor exactly when
$A+w^\top\Sigma w>\min_j\Sigma_{jj}$.
\par{\upshape\small Immediate from the two worlds of \cref{q1:cor:fixed-gram-range}, which share $(\Sigma,w,A)$, with $w=\mathbf e_j$ and \cref{q1:thm:benchmarks}.}
\end{corollary}

\begin{remark}[What the fixed-Gram theorem does not establish]\label{q1:auto:what-the-fixed-gram-theorem-does-not-est}
$\Sigma$ depends on the unknown target through $e_j=g_j-f$, so it need not be
observable. Corollary~\ref{q1:cor:fixed-gram-range} shows that raw error
geometry is insufficient; it is not an observational non-identification
theorem. The same-observables construction is in \cref{app:q2}.
\end{remark}

\subsection{Three objectives and their exact optimization}\label{q1:sec:objectives}

Let $\mathcal C$ denote $\Waff$, $\Delta_m$ or another nonempty convex subset of
$\Waff$. The three natural objectives are
\[
 \mathcal W_W^\star=\argmin_{w\in\mathcal C}w^\top\Sigma w,\qquad
 \mathcal W_T^\star=\argmin_{w\in\mathcal C}R_T(w),\qquad
 \mathcal W_G^\star=\argmax_{w\in\mathcal C}G_T(w)
\]
(weak-optimal, student-optimal, gain-maximizing; the last only when finite and
attained). For $T=P$ they are the minimizers of $w^\top\Sigma w$ and
$w^\top\Gpar w$ and the maximizers of $w^\top\Gperp w$; $A$ does not affect
them. Since $\Sigma=\Gpar+\Gperp$, minimizing the sum need not minimize
either summand, and maximizing $w^\top\Gperp w$ rewards filterable error
while possibly increasing weak risk.

\subsubsection{Affine weights}

\begin{theorem}[PSD quadratic minimization over affine weights]\label{q1:thm:affine-psd}
Let $H\succeq0$ and $q_H^{\mathrm{aff}}=\min_{\one^\top w=1}w^\top Hw$.
\begin{enumerate}
\item If $\one\notin\Ran H$, there is $z\in\Ker H$ with $\one^\top z\ne0$, and
$q_H^{\mathrm{aff}}=0$, attained at $w=z/(\one^\top z)$.
\item If $\one\in\Ran H$, then $q_H^{\mathrm{aff}}=1/(\one^\top H^\dagger\one)$ and
the minimizers are exactly $H^\dagger\one/(\one^\top H^\dagger\one)+z$,
$z\in\Ker H$ (each such $z$ has $\one^\top z=0$).
\end{enumerate}
If $H\succ0$ the minimizer is unique: $w_H^\star=H^{-1}\one/(\one^\top H^{-1}\one)$.
\end{theorem}
\begin{proof}
$\Ran H=(\Ker H)^\perp$. If $\one\notin\Ran H$ its projection on $\Ker H$ is
nonzero, giving the $z$ of (i), and $w^\top Hw\ge0$. If $\one\in\Ran H$ then
$\one\perp\Ker H$, and for feasible $w$, since
$H^{\dagger/2}H^{1/2}=\Pi_{\Ran H}$,
$1=\one^\top w=\ip{H^{\dagger/2}\one}{H^{1/2}w}\le(\one^\top H^\dagger\one)^{1/2}(w^\top Hw)^{1/2}$.
The vector $w_0=H^\dagger\one/(\one^\top H^\dagger\one)$ attains equality;
adding $z\in\Ker H$ preserves feasibility and value. Equality in
Cauchy--Schwarz forces $H^{1/2}w=cH^{\dagger/2}\one$, and feasibility fixes
$c$, so $w-w_0\in\Ker H$.
\end{proof}

\begin{corollary}[Closed forms]\label{q1:cor:closed}
If $\Sigma\succ0$ and $\Gpar\succ0$, the unique affine weak-optimal and
student-optimal weights are $w_W^\star=\Sigma^{-1}\one/(\one^\top\Sigma^{-1}\one)$
and $w_S^\star=\Gpar^{-1}\one/(\one^\top\Gpar^{-1}\one)$, with optimal risks
$1/(\one^\top\Sigma^{-1}\one)$ and $A+1/(\one^\top\Gpar^{-1}\one)$.
They coincide iff $\Gpar\Sigma^{-1}\one\in\Span\{\one\}$; $\Gpar=c\Sigma$
($c>0$) is sufficient but not necessary.
\par{\upshape\small Immediate from \cref{q1:thm:affine-psd}: the affine first-order condition for $w^\top\Gpar w$ is $\Gpar w=\rho\one$; strict convexity gives uniqueness.}
\end{corollary}

For a general learner, write $\Gamma=\Gamma_T$, $c=c_T$, $B=B_T$ and
$R(w)=B+w^\top\Gamma w-2c^\top w$. Let $O\in\R^{m\times(m-1)}$ have
orthonormal columns spanning $\one^\perp$ and $w_0=\one/m$.

\begin{theorem}[General learner: affine optimum, including singular cases]\label{q1:thm:affine-general}
If $\Gamma\succ0$ the unique affine minimizer is
\[
 w_T^\star=\Gamma^{-1}c+\frac{1-\one^\top\Gamma^{-1}c}{\one^\top\Gamma^{-1}\one}\Gamma^{-1}\one,
 \qquad R(w_T^\star)=B-c^\top\Gamma^{-1}c+\frac{(1-\one^\top\Gamma^{-1}c)^2}{\one^\top\Gamma^{-1}\one};
\]
$c\in\Span\{\one\}$ recovers Corollary~\ref{q1:cor:closed}. In general, with
$H=O^\top\Gamma O\succeq0$ and $a=O^\top(c-\Gamma w_0)$, one has $a\in\Ran H$
and the affine minimizers are exactly $w_0+O(H^\dagger a+z)$, $z\in\Ker H$,
with value $R(w_0)-a^\top H^\dagger a$; the minimum is always finite and
attained, and unique iff $H\succ0$.
\end{theorem}
\begin{proof}
For $\Gamma\succ0$ complete the square:
$R(w)=B-c^\top\Gamma^{-1}c+(w-\Gamma^{-1}c)^\top\Gamma(w-\Gamma^{-1}c)$, and
minimize the last term subject to $\one^\top(w-\Gamma^{-1}c)=1-\one^\top\Gamma^{-1}c$
via $\Gamma v=\rho\one$. In general write $w=w_0+Ou$:
$R(w)=R(w_0)+u^\top Hu-2a^\top u$. With $A_0=TEO$ and $r_0=b_T-TEw_0$,
$H=A_0^*A_0$ and $a=A_0^*r_0$; if $Hz=0$ then $\norm{A_0z}^2=0$ and
$z^\top a=\ip{A_0z}{r_0}=0$, so $a\perp\Ker H$. Since $HH^\dagger a=a$,
$u^\top Hu-2a^\top u=(u-H^\dagger a)^\top H(u-H^\dagger a)-a^\top H^\dagger a$.
\end{proof}

\begin{remark}[Exact versus separately estimated geometry]\label{q1:auto:exact-versus-separately-estimated-geomet}
The range condition $a\in\Ran H$ is automatic for exact $(\Gamma,c)$ but can
fail when the two are estimated separately; a PSD estimated Hessian with an
incompatible linear term gives an unbounded affine objective. Compact weight
domains or explicit regularization repair the optimization but change the
problem (\cref{app:q4}).
\end{remark}

\subsubsection{Simplex and general convex domains}

\begin{theorem}[KKT conditions and oracle regret identity]\label{q1:thm:kkt}
A simplex weight $w^\star$ minimizes $R$ iff, with
$\rho=w^{\star\top}(\Gamma w^\star-c)$,
\[
 (\Gamma w^\star-c)_j\ge\rho\ \ \forall j,\qquad (\Gamma w^\star-c)_j=\rho\ \text{ if }w_j^\star>0 .
\]
For any convex $\mathcal C\subseteq\Waff$, its minimizer $w^\star$, and
$d=w-w^\star$, $w\in\mathcal C$:
\[
 \boxed{R(w)-R(w^\star)=d^\top\Gamma d+2d^\top(\Gamma w^\star-c)},\qquad d^\top(\Gamma w^\star-c)\ge0,
\]
and the linear term vanishes identically if $\mathcal C=\Waff$. In the
projection case ($c=0$, $\Gamma=\Gpar$), if $I=\{j:w_j^\star>0\}$ and
$\Gamma_{II}\succ0$, then $w_I^\star=\Gamma_{II}^{-1}\one_I/(\one_I^\top\Gamma_{II}^{-1}\one_I)$
and the inactive conditions are $\Gamma_{jI}w_I^\star\ge1/(\one_I^\top\Gamma_{II}^{-1}\one_I)$.
\end{theorem}
\begin{proof}
The Lagrangian $R(w)-2\rho(\one^\top w-1)-\nu^\top w$ ($\nu\ge0$) gives
$2(\Gamma w^\star-c)-2\rho\one-\nu=0$ and $\nu_jw_j^\star=0$; multiplying by
$w^\star$ identifies $\rho$; convexity makes KKT sufficient. Expanding
$R(w^\star+d)$ gives the identity; the directional derivative at a convex
minimizer gives $d^\top(\Gamma w^\star-c)\ge0$, and on $\Waff$ both $\pm d$ are
feasible. The active-set formula solves $\Gamma_{II}w_I=\rho\one_I$ with
$\one_I^\top w_I=1$.
\end{proof}

\begin{proposition}[Affine versus simplex versus single]\label{q1:prop:aff-sim}
For $H\succeq0$, $q_H^{\mathrm{aff}}\le q_H^\Delta\le\min_jH_{jj}$, where
$q_H^\Delta=\min_{\Delta_m}w^\top Hw$. If $h_1,\dots,h_m$ realize $H$ as a Gram
matrix, $q_H^{\mathrm{aff}}=\dist^2(0,\aff\{h_j\})$ and
$q_H^\Delta=\dist^2(0,\conv\{h_j\})$. Negative weights can strictly help:
$h_1=u$, $h_2=2u$ give $q_H^\Delta=1$ but $q_H^{\mathrm{aff}}=0$ at $(2,-1)$.
\par{\upshape\small Immediate from $w^\top Hw=\norm{\sum_jw_jh_j}^2$ and $\aff\supseteq\conv\supseteq\{h_j\}$.}
\end{proposition}

\subsubsection{Gain maximization is the wrong target}

\begin{theorem}[Gain maximization]\label{q1:thm:gainmax}
(i) For $H\succeq0$, $\sup_{\Waff}w^\top Hw<\infty$ iff $H=c\one\one^\top$,
$c\ge0$; then $w^\top Hw\equiv c$. So affine gain maximization with
$H=\Gperp$ is unbounded or uninformative. (ii) For $H\succeq0$,
$\max_{\Delta_m}w^\top Hw=\max_jH_{jj}$, attained at a vertex; hence
$\max_{\Delta_m}G=\max_j(\Gperp)_{jj}-A$ selects a single supervisor.
(iii) For a general learner, if $\Lambda_T\succeq0$ then
$\max_{\Delta_m}G_T=\max_j\{(\Lambda_T)_{jj}+2(c_T)_j-B_T\}$; for indefinite
$\Lambda_T$ a simplex maximum exists but can be interior. On $\Waff$, with
$H_G=O^\top\Lambda_TO$ and $\ell=O^\top(\Lambda_Tw_0+c_T)$, the supremum is
finite iff $H_G\preceq0$ and $\ell\in\Ran H_G$; then, with $N_G=-H_G$, the
maximizers are $w_0+O(N_G^\dagger\ell+z)$, $z\in\Ker N_G$, with value
$G_T(w_0)+\ell^\top N_G^\dagger\ell$.
\end{theorem}
\begin{proof}
(i) If $z\in\one^\perp$ has $z^\top Hz>0$, $w_0+tz$ makes the form diverge;
otherwise $\one^\perp\subseteq\Ker H$ and $H=c\one\one^\top$. (ii) Realize $H$
by vectors $h_j$; convexity of $\norm\cdot^2$ gives
$\norm{\sum w_jh_j}^2\le\sum w_jH_{jj}\le\max_jH_{jj}$. (iii) Convexity of
$G_T$ when $\Lambda_T\succeq0$ as in (ii); compactness for existence; on
$\Waff$, $G_T(w_0+Ou)=G_T(w_0)+u^\top H_Gu+2\ell^\top u$: a positive
direction, or a null direction not orthogonal to $\ell$, makes it unbounded;
otherwise complete the square.
\end{proof}

\begin{example}[Gain-maximizing weights can be arbitrarily poor]\label{q1:auto:gain-maximizing-weights-can-be-arbitrari}
With $\Gpar=\diag(L,\varepsilon)$, $\Gperp=\diag(L+1,1)$, $L>1>\varepsilon>0$,
simplex gain maximization selects supervisor $1$ with transmitted risk $L$,
whereas $w_S^\star=(\varepsilon,L)/(L+\varepsilon)$ has transmitted risk
$L\varepsilon/(L+\varepsilon)$; the ratio $(L+\varepsilon)/\varepsilon$ is
unbounded. For a non-contractive learner the gain maximizer can be interior:
$\Hh=\R$, $T=2I$, $f=0$, $e_1=1$, $e_2=-1$ gives $G_T(t,1-t)=-3(2t-1)^2$,
maximized at $t=\tfrac12$ while both vertices have negative gain.
\end{example}

\subsection{Oracle regret of raw weighting}\label{q1:sec:regret}

\begin{corollary}[Exact regret of raw-Gram weighting]\label{q1:cor:raw-regret}
Let $w_T^\star$ minimize $R_T$ and $w_W^\star$ minimize $w^\top\Sigma w$, both
over $\Waff$. Then
\[
 \boxed{R_T(w_W^\star)-R_T(w_T^\star)=(w_W^\star-w_T^\star)^\top\Gamma_T(w_W^\star-w_T^\star)},
\qquad
 R_W(w_T^\star)-R_W(w_W^\star)=(w_T^\star-w_W^\star)^\top\Sigma(w_T^\star-w_W^\star).
\]
In the projection case raw weighting is student-optimal iff
$w_W^\star-w_S^\star\in\Ker\Gpar$. On the simplex the full identity of
Theorem~\ref{q1:thm:kkt} applies, with a nonnegative first-order term that
vanishes when $w$ is supported on indices with $(\Gamma w^\star-c)_j=\rho$.
\par{\upshape\small Immediate from \cref{q1:thm:kkt} with $\mathcal C=\Waff$, applied to $R_T$ and to $w^\top\Sigma w$; $z^\top\Gpar z=0$ iff $z\in\Ker\Gpar$.}
\end{corollary}

\begin{theorem}[Raw weighting can be arbitrarily suboptimal]\label{q1:thm:raw-arbitrary}
For every $L>0$ there are $m=2$ and positive definite $\Gpar,\Gperp$, realizable
whenever $d_\Ss,d_\perp\ge2$, for which the affine and simplex optimizers
coincide within each objective, yet
$(R_S(w_W^\star)-A)/(R_S(w_S^\star)-A)>L$.
\end{theorem}
\begin{proof}
Take $\Gpar=\diag(\varepsilon,1)$, $\Gperp=\diag(M,\varepsilon)$,
$\Sigma=\diag(M+\varepsilon,1+\varepsilon)$. For $\diag(a,b)\succ0$ the affine
minimizer $(b,a)/(a+b)$ lies in the simplex, so
$w_S^\star=(1,\varepsilon)/(1+\varepsilon)$ and
$w_W^\star=(1+\varepsilon,M+\varepsilon)/(M+1+2\varepsilon)$. Then
$w_S^{\star\top}\Gpar w_S^\star=\varepsilon/(1+\varepsilon)$ while
$w_W^{\star\top}\Gpar w_W^\star=[\varepsilon(1+\varepsilon)^2+(M+\varepsilon)^2]/(M+1+2\varepsilon)^2\to1$
as $M\to\infty$, so the ratio tends to $(1+\varepsilon)/\varepsilon$. Choose
$\varepsilon$ with $(1+\varepsilon)/\varepsilon>L$, then $M$ large.
Realizability is Theorem~\ref{q1:thm:realize}.
\end{proof}

\subsection{Marginal value of an additional supervisor}\label{q1:sec:marginal}

The results below are stated for arbitrary vectors $h_j$ in a Hilbert space.
For the student risk of a general learner, apply them to $h_j=z_j=Tg_j-f$
(Proposition~\ref{q1:prop:HT}); for a projection, equivalently to $h_j=Pe_j$,
because $z_j=Pe_j-Qf$ with a common component $-Qf$ orthogonal to every
$Pe_k$, so $\norm{\sum_jw_jz_j}^2=\norm{\sum_jw_jPe_j}^2+A$ on $\Waff$. For
the weak risk use $h_j=e_j$.

\begin{theorem}[General affine marginal value]\label{q1:thm:marg-aff}
Let $\mathcal A_m=\aff\{h_1,\dots,h_m\}$, $x_m^\star$ its minimum-norm point,
$q_m=\norm{x_m^\star}^2$, $\mathcal L_m=\Span\{h_j-h_1:j\ge2\}$,
$r=(I-\Pi_{\mathcal L_m})(h_{m+1}-h_1)$ and
$q_{m+1}=\dist^2(0,\aff\{h_1,\dots,h_{m+1}\})$. If $r=0$ then $q_{m+1}=q_m$;
otherwise
\[
 \boxed{q_{m+1}=q_m-\frac{\ip{x_m^\star}r^2}{\norm r^2}}.
\]
The new vector strictly improves the affine optimum iff $r\ne0$ and
$\ip{x_m^\star}r\ne0$.
\end{theorem}
\begin{proof}
$x_m^\star$ is the projection of $0$ onto $h_1+\mathcal L_m$, so
$x_m^\star\perp\mathcal L_m$. The enlarged hull is
$x_m^\star+\mathcal L_m+\Span\{r\}$ because $x_m^\star-h_1\in\mathcal L_m$. If
$r\ne0$, since $r,x_m^\star\perp\mathcal L_m$,
$\norm{x_m^\star+\ell+tr}^2=\norm{x_m^\star+tr}^2+\norm\ell^2$; take $\ell=0$
and minimize the quadratic in $t$ at $t^\star=-\ip{x_m^\star}r/\norm r^2$.
\end{proof}

\begin{definition}[Geometric and risk redundancy]\label{q1:auto:geometric-and-risk-redundancy}
The vector $h_{m+1}$ is called affine-geometrically redundant if
$h_{m+1}\in\aff\{h_1,\dots,h_m\}$, i.e.\ $r=0$, and affine-risk redundant if $q_{m+1}=q_m$. Geometric redundancy
implies risk redundancy; the converse fails when $r\ne0$ and $r\perp x_m^\star$.
Applying the definitions to $h_j=Pe_j$ and to $h_j=e_j$ gives student-relative
and raw redundancy, which need not agree.
\end{definition}

\begin{theorem}[Schur-complement form]\label{q1:thm:schur}
Let $H\succ0$ and $\widetilde H=\left(\begin{smallmatrix}H&b\\b^\top&c\end{smallmatrix}\right)\succeq0$,
$\kappa=\one^\top H^{-1}\one$, $\beta=b^\top H^{-1}\one$, $\delta_0=1-\beta$,
$s=c-b^\top H^{-1}b\ge0$. If $s>0$, then
$\one^\top\widetilde H^{-1}\one=\kappa+\delta_0^2/s$,
$q_{\widetilde H}^{\mathrm{aff}}=1/(\kappa+\delta_0^2/s)\le1/\kappa$, strictly iff
$\delta_0\ne0$, and the new optimal weights are
$(\kappa+\delta_0^2/s)^{-1}\bigl(H^{-1}\one-H^{-1}b\,\delta_0/s,\ \delta_0/s\bigr)$.
If $s=0$ and $\alpha_0=H^{-1}b$: if $\one^\top\alpha_0=1$ the new vector is
affine-geometrically redundant; otherwise $q_{\widetilde H}^{\mathrm{aff}}=0$.
Applied to $\Gpar$ and to $\Sigma$, the student and raw criteria
$1-b_\parallel^\top\Gpar^{-1}\one\ne0$ and $1-b_{\mathrm{raw}}^\top\Sigma^{-1}\one\ne0$
imply neither each other.
\end{theorem}
\begin{proof}
The block inverse is
$\left(\begin{smallmatrix}H^{-1}+H^{-1}bs^{-1}b^\top H^{-1}&-H^{-1}bs^{-1}\\-s^{-1}b^\top H^{-1}&s^{-1}\end{smallmatrix}\right)$,
so $\one^\top\widetilde H^{-1}\one=\kappa+s^{-1}(\beta^2-2\beta+1)$; apply
Theorem~\ref{q1:thm:affine-psd}, and normalize $\widetilde H^{-1}\one$ for the
weights. If $s=0$, realize $\widetilde H$ by vectors; $\alpha_0$ is the
least-squares coefficient of $h_{m+1}$ on the independent $h_1,\dots,h_m$ and
$h_{m+1}=\sum_j\alpha_{0,j}h_j$. If $\one^\top\alpha_0\ne1$, the weights
$(-\alpha_0,1)/\delta_0$ sum to one and give the zero vector.
\end{proof}

\begin{theorem}[Simplex marginal value]\label{q1:thm:marg-sim}
Let $w_m^\star\in\Delta_m$ minimize $\norm{\sum_jw_jh_j}^2$, $x_m^\star=\sum_jw_{m,j}^\star h_j$,
$q_m=\norm{x_m^\star}^2$ and $b_j=\ip{h_j}{h_{m+1}}$. Adding $h_{m+1}$ strictly
improves the simplex optimum iff
\[
 \boxed{\ip{x_m^\star}{h_{m+1}}<\norm{x_m^\star}^2},\quad\text{equivalently}\quad b^\top w_m^\star<q_m .
\]
\end{theorem}
\begin{proof}
$x_m^\star$ is the projection of $0$ onto $C_m=\conv\{h_j\}$:
$\ip{x_m^\star}{y-x_m^\star}\ge0$ on $C_m$. The new set is
$\conv(C_m\cup\{h_{m+1}\})$, and the linear inequality holds there iff it holds
at the new extreme point. If it holds, $x_m^\star$ stays optimal; otherwise a
strictly closer point exists (the minimum-norm point is unique).
Finally $\ip{x_m^\star}{h_{m+1}}=b^\top w_m^\star$.
\end{proof}

\begin{example}[Raw and student-relative value can disagree]\label{q1:ex:redundancy}
Let $u\in\Ss$, $v\in\Ss^\perp$ be unit vectors and $t$ the weight of the new
supervisor. (a) $e_1=u+v$, $e_2=2v$: $\norm{e(t)}^2=2+2t^2\ge2$, so the new
supervisor is raw-redundant, but $\norm{Pe(t)}^2=(1-t)^2$ falls from $1$ to
$0$. (b) $e_1=u+v$, $e_2=u-v$: $\norm{e(t)}^2=1+(1-2t)^2$ falls from $2$ to $1$ at
$t=1/2$, so the new supervisor is raw-useful, but $Pe(t)=u$ for every $t$, so
it is student-redundant. Both statements hold over affine and simplex
weights. Raw redundancy
concerns the $e_j$, student redundancy the $Pe_j$; the components $Qe_j$ can
reverse the raw test without changing the transmitted vectors. High or low raw
correlation is neither necessary nor sufficient for student-relative value.
\end{example}

\subsection{Best single supervisor versus best aggregate}\label{q1:sec:best}

For $\mathcal C\in\{\Waff,\Delta_m\}$ let $q_\Sigma^{\mathcal C}=\min_{\mathcal C}w^\top\Sigma w$,
$q_\parallel^{\mathcal C}=\min_{\mathcal C}w^\top\Gpar w$.

\begin{theorem}[Exact hierarchy]\label{q1:thm:hierarchy}
$q_\Sigma^{\Waff}\le q_\Sigma^{\Delta_m}\le\min_j\Sigma_{jj}$ and
$A\le A+q_\parallel^{\Waff}\le A+q_\parallel^{\Delta_m}\le A+\min_j(\Gpar)_{jj}$.
The best aggregate student beats the best single supervisor iff
$A+q_\parallel^{\mathcal C}<\min_j\Sigma_{jj}$, beats the best weak aggregate
(possibly with different weights) iff $A+q_\parallel^{\mathcal C}<q_\Sigma^{\mathcal C}$,
and a student-optimal $w_S^\star$ beats its own aggregate iff
$w_S^{\star\top}\Gperp w_S^\star>A$.
\end{theorem}
\begin{proof}
Proposition~\ref{q1:prop:aff-sim} for $\Sigma$ and $\Gpar$, and
Theorem~\ref{q1:thm:multi}.
\end{proof}

\begin{proposition}[When mixing strictly beats every single vector]\label{q1:auto:when-mixing-strictly-beats-every-single-}
Let $H\succeq0$ be realized by $h_j$ and $J_{\min}=\argmin_j\norm{h_j}^2$.
Then $q_H^\Delta<\min_jH_{jj}$ iff no $j\in J_{\min}$ satisfies
$\ip{h_j}{h_k-h_j}\ge0$ for every $k$, i.e.\ no minimum-norm vertex is the
projection of $0$ onto the convex hull.
\end{proposition}
\begin{proof}
A vertex $h_j$ is the projection of $0$ onto the hull iff the variational
inequality holds at all vertices (linearity in $y$). If such a minimum-norm
vertex exists, $q_H^\Delta=\min_jH_{jj}$; otherwise the unique closest point
of the hull has strictly smaller norm.
\end{proof}

\paragraph{Against clean labels alone.} The hierarchy compares population
risks. When the aggregate must be selected with $n$ anchors, the competitor
is the student space fitted on those anchors alone. The comparison of
$A+q_\parallel^{\mathcal C}$ plus the selection cost with the anchor-only risk is made
in \cref{q3:sec:compare}:
every anchor-only procedure pays at least $\sigma^2d_{\mathrm{eff}}(\sigma^2/n)/n$, with
$d_{\mathrm{eff}}$ the effective dimension of the target in the student's
representation, so a small $q_\parallel^{\mathcal C}$ together with few anchors relative
to $d_{\mathrm{eff}}$ is what makes the weak supervisors worth their anchors.

\apppart{Supporting material}{}
\subsection{Relation to prior work}\label{q1:sec:related}

What in this appendix is standard and what is new, and detailed comparisons with prior work, are collected in \cref{app:boundary,app:related}.

\subsection{Linear-algebra facts}

\begin{lemma}[Trace of a rank-one operator]\label{q1:lem:rank-one}
For $u,v\in\R^d$ and $B\in\R^{d\times d}$, $\tr(B(u\otimes v))=\ip v{Bu}$; in
particular $\tr(u\otimes u)=\norm u^2$.
\par{\upshape\small Immediate from Parseval in an orthonormal basis.}
\end{lemma}

\begin{lemma}[Gram factorization]\label{q1:lem:gram-factor}
For symmetric $\Sigma\in\R^{m\times m}$ the following are equivalent:
$\Sigma\succeq0$; $\Sigma_{jk}=b_j^\top b_k$ for some vectors; $\Sigma=B^\top B$.
The minimal ambient dimension of such vectors is $\rank\Sigma$.
\end{lemma}
\begin{proof}
If $\Sigma=V\Lambda V^\top$ with $\Lambda\succeq0$, take $B=\Lambda^{1/2}V^\top$
and delete zero rows; conversely $w^\top\Sigma w=\norm{\sum_jw_jb_j}^2$;
$\rank(B^\top B)=\rank B$.
\end{proof}

\begin{lemma}[Zero quadratic forms and kernel inclusion]\label{q1:lem:kernel-inclusion}
If $H\succeq0$ then $z^\top Hz=0\iff Hz=0$. If $0\preceq G\preceq H$ then
$\Ker H\subseteq\Ker G$ and $\Ran G\subseteq\Ran H$.
\end{lemma}
\begin{proof}
$z^\top Hz=\sum_i\lambda_i(v_i^\top z)^2$ vanishes iff $z$ is orthogonal to the
eigenvectors with $\lambda_i>0$. If $Hz=0$ then $0\le z^\top Gz\le0$, so $Gz=0$;
take orthogonal complements.
\end{proof}

\subsection{Assumption checklist}

\begin{center}\small
\begin{tabularx}{\textwidth}{@{}p{0.31\textwidth}p{0.14\textwidth}X@{}}
\toprule
Result & Weights & Conditions\\
\midrule
Single and aggregate risk identities (Thms \ref{q1:thm:single-general}, \ref{q1:thm:multi}) & affine & bounded linear common $T$\\
Projection identities (Thm \ref{q1:thm:single-projection}) & affine & $T=P$, closed $\Ss$\\
Orientation range (Thm \ref{q1:thm:orient}) & --- & $\Hh=\R^d$, $1\le d_\Ss\le d-1$\\
Realizability (Thm \ref{q1:thm:realize}) & --- & PSD split, rank $\le$ dimensions\\
Affine optima (Thms \ref{q1:thm:affine-psd}, \ref{q1:thm:affine-general}) & affine & PSD; singular cases allowed\\
KKT and regret (Thm \ref{q1:thm:kkt}) & convex & PSD $\Gamma_T$\\
Marginal value (Thms \ref{q1:thm:marg-aff}--\ref{q1:thm:marg-sim}) & affine / simplex & none beyond Gram realization\\
Perturbation and limits (Section \ref{ext:perturb}) & bounded / compact & curvature for weight stability\\
\bottomrule
\end{tabularx}
\end{center}

\section{Identification: Formal Results and Proofs}\label{app:q2}

\paragraph{Summary.} \noindent
\cref{app:q1} showed that the value of a weak supervisor is determined by
student-relative error geometry. That geometry involves the unknown target.
For any common linear learner $T$, every same-aggregate gain equals an
observable quadratic minus twice an unknown linear target moment, and so does
every student-risk contrast. Consequently all gains are identified iff the
target's projection on $U_T=\Span\{q_j\}$ is, and all selection decisions iff
its projection on $V_T=\Span\{s_j-s_m\}$ is; for a projection learner
$U\perp V$, so gain certification and student selection need disjoint target
information. Even with unlimited unlabeled data and arbitrary candidate
training, two targets with identical observations can give opposite gains and
different optimal weights, in a bounded, nondegenerate example in which the
clean student beats every supervisor. Under a reference-radius assumption the
identified gain sets are sharp intervals. When the teachers' own training
rules are known exactly ($g_j=M_jf$), the compatible targets form an affine
fiber $f_\Kk+\Nn$, and a \emph{leakage law} decides identification: all gains
are identified iff every leakage $\ell_j=\Pi_\Nn Tg_j$ vanishes, all selection
decisions iff the leakages coincide. Hence gain identification implies
selection identification but not conversely. We give sharp bounded-fiber
intervals, an anchor-free robust-optimality test, and the distinction between
identification and stable recovery.

\paragraph{Where the main text uses this appendix.}
Formal versions of the main-text results; supporting results are cited in their proofs.

\begin{center}\small
\begin{tabularx}{\textwidth}{@{}>{\raggedright}p{0.24\textwidth}X@{}}
\toprule
Main-text result & Formal results\tabularnewline
\midrule
\cref{thm:tasks} & \cref{q2:thm:tasks} (p.~\pageref{q2:thm:tasks}); \cref{q2:cor:realizable} (p.~\pageref{q2:cor:realizable})\tabularnewline
\cref{thm:counter} & \cref{q2:thm:counter} (p.~\pageref{q2:thm:counter}); \cref{q2:cor:impossible} (p.~\pageref{q2:cor:impossible})\tabularnewline
\cref{thm:leakage} & \cref{q2:thm:fiber} (p.~\pageref{q2:thm:fiber}); \cref{q2:thm:struct-tasks} (p.~\pageref{q2:thm:struct-tasks}); \cref{q2:thm:leakage} (p.~\pageref{q2:thm:leakage}); \cref{q2:prop:approx} (p.~\pageref{q2:prop:approx})\tabularnewline
Label models, \cref{sec:identification} & \cref{q2:thm:factor} (p.~\pageref{q2:thm:factor}); \cref{q2:cor:factor-anchors} (p.~\pageref{q2:cor:factor-anchors})\tabularnewline
\bottomrule
\end{tabularx}
\end{center}

\paragraph{Notation.} As in \cref{app:notation,app:translation}.

\paragraph{Reading guide: what each observation layer adds.}
The appendix moves through nested observation layers (the label-model row is
an alternative to known teacher equations); each adds specific target
information and leaves a specific part unidentified.
\begin{center}\small
\begin{tabularx}{\textwidth}{@{}>{\raggedright}p{0.22\textwidth}>{\raggedright}X>{\raggedright}X>{\raggedright\arraybackslash}p{0.17\textwidth}@{}}
\toprule
Layer & Adds & Still unidentified & Sections\tabularnewline
\midrule
Unlabeled observables & weak labels, students and their Grams; centered supervisor errors $J\Sigma J$ & the moments $\Pi_Uf$ (gains) and $\Pi_Vf$ (selection); the common error & \cref{q2:sec:tasks,q2:sec:center,q2:sec:counter,q2:sec:partial}\tabularnewline
$\subset$ structured teacher observables & known training equations: $f$ is pinned down up to the common nullspace $\Nn$ & leakage $\Pi_\Nn$ of the decision spaces (leakage law) & \cref{q2:sec:structured,q2:sec:leakage,q2:sec:struct-counter,q2:sec:bounded,q2:sec:stability}\tabularnewline
or label-model assumption & supervisors are scaled targets plus errors orthogonal to the other supervisors' students & one scalar for selection; gains still need $\ip f{g_j}$ & \cref{q2:sec:factor}\tabularnewline
$\subset$ anchors & noisy values of $f$ at sampled inputs & nothing in the limit; finite-$n$ cost in \cref{app:q3} & \cref{q2:sec:labels}\tabularnewline
\bottomrule
\end{tabularx}
\end{center}

\paragraph{Two interval results.} \Cref{q2:sec:partial} and
\cref{q2:sec:bounded} both give sharp intervals but answer different
questions. The first is an external sensitivity analysis without teacher
structure: an analyst-chosen reference $f_0$ and radius $\rho_\perp$ for
$\norm{Q(f-f_0)}$. The second lives on the structured fiber: the teacher
equations fix $f$ up to $\Nn$, and a bound on the remaining component gives
intervals and an anchor-free robust-optimality test.

\paragraph{Identification map in brief.} Without structure, unlabeled data
identify neither gains nor selection (\cref{q2:thm:tasks,q2:thm:counter});
realizability identifies the gain but not the selection
(\cref{q2:cor:realizable}); with structured teachers, the leakage law decides
which of the two is identified (\cref{q2:thm:leakage}). The full map is
\cref{q2:sec:summary}.

\apppart{Unlabeled observations}{What weak labels and students reveal, and the two missing moments.}
\subsection{Model and observation experiments}\label{q2:sec:model}

\paragraph{Role in the paper.} \cref{app:q2} answers the second question: \emph{which
part of the student-relative geometry can be known without target labels, and
what exactly is missing?} The answer is a pair of finite-dimensional target
projections, reduced further by known teacher structure. \cref{app:q3} estimates them
from anchors.

\paragraph{Population geometry.} Let $(\mathcal X,\mathcal A_X,P_X)$ be a
probability space, $\Hh=L^2(P_X)$, $\ip hk=\E[h(X)k(X)]$; functions are
identified $P_X$-a.s. Fix a closed student subspace $\Ss$, $P=P_\Ss$, $Q=I-P$,
and for any other closed subspace $L$ let $\Pi_L$ be its orthogonal
projection. The target $f\in\Hh$ is unknown; there are $m\ge2$ weak
predictions $g_j$ with errors $e_j=g_j-f$, and a common bounded linear
learner $T$ (\cref{q1:ass:T}), with $s_j=Tg_j$, $q_j=g_j-s_j$; the
projection learner $T=P$ is the main case. For $w\in\Waff$,
$s_w=Tg_w=\sum_jw_js_j$ and $q_w=g_w-s_w$, and
\[
 R_W(w;f)=\norm{g_w-f}^2,\qquad R_T(w;f)=\norm{s_w-f}^2,\qquad G_T(w;f)=R_W-R_T .
\]
For $T=P$ write $R_S,G$; then (\cref{q1:thm:multi})
$R_S(w;f)=A(f)+w^\top\Gpar w$ and $G(w;f)=w^\top\Gperp w-A(f)$, with
$\Sigma=\Gpar+\Gperp$ the uncentered error Grams and $A(f)=\norm{Qf}^2$. All
of these depend on $f$.

\begin{definition}[Population unlabeled experiment]\label{q2:def:O}
The population unlabeled observation is
\[\mathcal O_T=(P_X,g_1,\dots,g_m,T):\]
the exact functions $g_j$, the learner, hence $s_j$ and $q_j$, and all
population inner products of known functions. It provides no target moment
and none of $\Sigma$, $\Gamma_T$, $\Gpar$, $\Gperp$, $A$ unless an extra assumption
implies it. Knowing a representation or architecture alone does not grant
knowledge of $T$ or of its population projection.
\end{definition}

An unlabeled learner may train arbitrarily many candidate students, choose
label mixtures adaptively, and use internal randomness, provided its
transcript is a measurable function of $\mathcal O_T$ and that randomness. A
nonlinear training rule need not satisfy $s_w=\sum_jw_js_j$ and is not covered
by the linear formulas.

\begin{definition}[Anchor experiment]\label{q2:def:anchors}
Additionally the learner may receive $n$ iid pairs $(X_i,Y_i)$, $X_i\sim P_X$,
$Y_i=f(X_i)+\xi_i$, $\E[\xi_i\mid X_i]=0$, independent of everything used to
build the dictionary, the learner and any reference function. Anchors are
passive draws, not actively chosen queries.
\end{definition}

\begin{definition}[Identification and sharpness]\label{q2:auto:identification-and-sharpness}
Given a class $\mathcal F$ of admissible targets, a functional $\mathsf T(f)$
is identified by an observation if it is constant over all $f\in\mathcal F$
compatible with that observation; a set of minimizers is identified when the
set is constant. An identified interval is sharp if it equals the set of
attainable values over the stated class, endpoints included.
\end{definition}

\subsection{Observable quadratics and hidden linear moments}\label{q2:sec:tasks}

Define the observable prediction Grams and the hidden target moments
\[
 (K_g)_{ij}=\ip{g_i}{g_j},\quad (K_s)_{ij}=\ip{s_i}{s_j},\quad (K_q)_{ij}=\ip{q_i}{q_j},\quad
 \mathsf D_T=K_g-K_s,
\]
\[
 \mu_q(f)_j=\ip f{q_j},\qquad \mu_s(f)_j=\ip f{s_j}.
\]

\begin{theorem}[Task-specific identification]\label{q2:thm:tasks}
For every $w,v\in\Waff$,
\begin{align}
 G_T(w;f)&=w^\top \mathsf D_Tw-2\mu_q(f)^\top w,\label{q2:eq:gain}\\
 R_T(w;f)-R_T(v;f)&=w^\top K_sw-v^\top K_sv-2\mu_s(f)^\top(w-v).\label{q2:eq:contrast}
\end{align}
Let $U_T=\Span\{q_1,\dots,q_m\}$ and $V_T=\Span\{s_j-s_m:j<m\}$, $k=\dim V_T$.
For two targets $f,\widetilde f$:
\begin{enumerate}
\item all gains on $\Delta_m$ (or on $\Waff$) coincide iff $\Pi_{U_T}f=\Pi_{U_T}\widetilde f$;
\item all student-risk contrasts on either domain coincide iff $\Pi_{V_T}f=\Pi_{V_T}\widetilde f$,
and then the student-optimal weight sets coincide.
\end{enumerate}
Moreover $k=\rank(JK_sJ)=\rank(J\Gamma_TJ)\le m-1$. For $T=P$:
$D_P=K_q$, so $G(w;f)=\norm{q_w}^2-2\ip f{q_w}$; $U=U_P\subseteq\Ss^\perp$,
$V=V_P\subseteq\Ss$, hence $U\perp V$; $\dim U=\rank K_q$ and
$k=\rank(J\Gpar J)$. For general $T$, $U_T$ and $V_T$ need not be orthogonal.
\end{theorem}
\begin{proof}
$\norm{g_w-f}^2-\norm{s_w-f}^2=\norm{g_w}^2-\norm{s_w}^2-2\ip f{g_w-s_w}$, and
$\norm{g_w}^2=w^\top K_gw$, $\norm{s_w}^2=w^\top K_sw$; the $\norm f^2$ terms
cancel. The contrast is the same computation. If $h=\widetilde f-f$, the gain
changes by $-2\ip h{q_w}$ and the contrast by $-2\ip h{s_w-s_v}$; these vanish
when $h\perp U_T$, respectively $h\perp V_T$. Conversely, equality at every
vertex $w=\mathbf e_j$ gives $\ip h{q_j}=0$, and at $(w,v)=(\mathbf e_j,\mathbf e_m)$
gives $\ip h{s_j-s_m}=0$. Equal contrasts give equal argmin sets. The
columns $s_j-\bar s$ ($\bar s=m^{-1}\sum_js_j$) span $V_T$ and have Gram
$JK_sJ$; since $g_j-\bar g=e_j-\bar e$, $s_j-\bar s=T(e_j-\bar e)$, whose
Gram is $J\Gamma_TJ$; a finite family and its Gram have equal rank. For
$T=P$, $\ip{s_i}{q_j}=0$ gives $K_g=K_s+K_q$. Nonorthogonality in general:
$\Hh=\R$, $T=\tfrac12I$, $g_1=1$, $g_2=2$ give $U_T=V_T=\R$.
\end{proof}

The second statement asserts sufficiency for equal optimal weight sets; it is
not claimed to be necessary merely for equality of those sets.

\begin{corollary}[Realizability identifies the gain, not the selection]\label{q2:cor:realizable}
For $T=P$, if $f\in\Ss$ is known, then $G(w;f)=\norm{q_w}^2\ge0$ is identified
without anchors, while $\Pi_Vf$ can remain unknown, so the optimal aggregate
need not be identified (Remark~\ref{q2:rem:realizable}). If all $q_j=0$, all gains
vanish. For a general learner, knowing $f\in\Ss$ does not identify the gain,
because $q_j$ can have a component in $\Ss$.
\par{\upshape\small Immediate from \cref{q2:thm:tasks}: $f\perp U\subseteq\Ss^\perp$ gives $\mu_q(f)=0$, while nothing fixes the projection of $f$ onto $V\subseteq\Ss$.}
\end{corollary}

\subsection{What disagreements and unlabeled predictions identify}\label{q2:sec:center}

Let $\bar g=m^{-1}\sum_jg_j$, $\bar e=\bar g-f$, $d_j=g_j-\bar g=e_j-\bar e$,
$J=I_m-\one\one^\top/m$ and $\Upsilon_{jk}=\norm{g_j-g_k}^2$.

\begin{theorem}[Centered Gram identification]\label{q2:thm:center}
$H_c:=J\Sigma J=-\tfrac12J\Upsilon J$, $(H_c)_{jk}=\ip{d_j}{d_k}$, and
$\Upsilon_{jk}=(H_c)_{jj}+(H_c)_{kk}-2(H_c)_{jk}$. With the full observation
$\mathcal O_P$, also $J\Gpar J=(\ip{Pd_j}{Pd_k})$ and $J\Gperp J=(\ip{Qd_j}{Qd_k})$
are observable. Disagreements $\Upsilon$ alone do not identify the projected
centered matrices.
\end{theorem}
\begin{proof}
$g_j-g_k=e_j-e_k$ gives $\Upsilon=b\one^\top+\one b^\top-2\Sigma$ with
$b=\diag\Sigma$; multiply by $J$ on both sides, using $J\one=0$. The synthesis
matrix of the $e_j$ times $J$ has columns $d_j$; its Gram is $J\Sigma J$. Apply
$P,Q$ to the known $d_j=g_j-\bar g$. For the last claim, in $\R^2$ with
$\Ss=\Span\{(1,0)\}$ the dictionaries $\{(1,0),(-1,0)\}$ and $\{(0,1),(0,-1)\}$
share $\Upsilon$ but have centered student Grams
$\left(\begin{smallmatrix}1&-1\\-1&1\end{smallmatrix}\right)$ and $0$.
\end{proof}

\begin{proposition}[The unidentified common component]\label{q2:prop:common}
$\Sigma=H_c+\alpha\one^\top+\one\alpha^\top+c\,\one\one^\top$ with
$\alpha_j=\ip{d_j}{\bar e}$, $c=\norm{\bar e}^2$, $\one^\top\alpha=0$, and the same
after applying $P$ or $Q$. For affine $w$, $e_w=\bar e+\sum_jw_jd_j$ and
$w^\top\Sigma w=w^\top H_cw+2\alpha^\top w+c$. The missing common error
affects the optimizing weights through $\alpha$, not only a constant offset.
\par{\upshape\small Immediate by expanding $\ip{d_j+\bar e}{d_k+\bar e}$ with $\sum_jd_j=0$.}
\end{proposition}

\begin{lemma}[Target-shift invariance]\label{q2:lem:gauge}
For every $h\in\Hh$ the targets $f$ and $f+h$ (with errors $e_j$ and $e_j-h$)
produce the same $\mathcal O_T$, and every unlabeled randomized transcript has
the same law under both.
\end{lemma}
\begin{proof}
$P_X$, $g_j$ and $T$ do not change; the target enters no unlabeled
observation; a common measurable map of the same observation and the same
internal randomness has the same law.
\end{proof}

\subsection{A bounded, nondegenerate impossibility}\label{q2:sec:counter}

\begin{theorem}[Same observations, opposite gains, different optima]\label{q2:thm:counter}
Let $X=(U_0,Z_0,V_0)$ have independent Rademacher coordinates with coordinate
functions $u,z,v$ (orthonormal in $\Hh$). Let $\Ss=\Span\{u,z\}$,
$g_1=2u+z+v$, $g_2=-2u+z-v$, $f_\pm=\pm(\tfrac12u+v)$. The two worlds have
identical $\mathcal O_P$, and:
\begin{enumerate}
\item the supervisors are distinct, $\norm{g_1-g_2}^2=20$;
\item $A(f_\pm)=1$, and the clean-label student beats each supervisor in each world;
\item supervisor $1$ has gain $-1$ in the plus world and $3$ in the minus world; supervisor $2$ the reverse;
\item the unique student-optimal affine and simplex weights are $(5/8,3/8)$ and $(3/8,5/8)$.
\end{enumerate}
The best single supervisor and the best single projected student switch
between worlds; in each world some supervisor has positive gain; and the
uniform aggregate student beats the best single supervisor in both worlds.
\end{theorem}
\begin{proof}
$s_1=2u+z$, $s_2=-2u+z$, $q_1=v$, $q_2=-v$. With $w=(t,1-t)$,
$s_w=(4t-2)u+z$ and $q_w=(2t-1)v$, so
\[
 R_S(w;f_\pm)=2+(4t-\tfrac52)^2,\ 2+(4t-\tfrac32)^2,
\]
\[
 R_W(w;f_+)=1+(4t-\tfrac52)^2+(2t-2)^2,\qquad R_W(w;f_-)=1+(4t-\tfrac32)^2+(2t)^2 .
\]
With $B=2t-1$ the gains are $B^2-2B$ and $B^2+2B$: at $t=1$, $-1$ and $3$; at
$t=0$, $3$ and $-1$. The student risks are strictly convex with minimizers
$5/8,3/8\in[0,1]$ and common minimum $2$. Single weak risks are $(13/4,45/4)$
and $(45/4,13/4)$; single student risks $(17/4,33/4)$ and $(33/4,17/4)$; clean
risks are $1$; at $t=\tfrac12$ the student is $z$ with risk $9/4<13/4$. Also
$\norm{f_\pm}^2=5/4$ and $\norm{g_j}^2=6$, so neither unbounded approximation
error, identical supervisors, nor a weak clean student explains the ambiguity.
\end{proof}

\begin{corollary}[Operational impossibility]\label{q2:cor:impossible}
With no anchors, any (possibly randomized) procedure has worst-world gain-sign
error at least $1/2$ at $w=\mathbf e_1$, and any procedure returning affine
weights has worst-world expected selection regret at least $1/4$, with
unlimited unlabeled data and candidate training. Uniform weights attain $1/4$.
\end{corollary}
\begin{proof}
The output law is common to both worlds (Lemma~\ref{q2:lem:gauge}); the two
error probabilities of a sign rule sum to one. If the output is $(T_0,1-T_0)$
and $B=4T_0-2$, the regrets are $(B-\tfrac12)^2$ and $(B+\tfrac12)^2$, whose
average expectation is $\E[B^2]+\tfrac14\ge\tfrac14$ (also when infinite).
\end{proof}

\begin{remark}[The realizable restriction]\label{q2:rem:realizable}
With the same dictionary and $f_\pm=\pm u/2$, $A=0$, the gain is $(2t-1)^2$ in
both worlds, but the optimal weights are still $5/8$ and $3/8$: selection
ambiguity persists under realizability. The opposite-gain claim deliberately
uses a misspecified regime; it would be false under the known constraint
$f\in\Ss$.
\end{remark}

\subsection{Sharp partial identification under reference radii}\label{q2:sec:partial}

\begin{assumption}[Known reference and sensitivity radii]\label{q2:ass:ball}
Fix an $\mathcal O_T$-measurable reference $f_0$ and known $\rho_\parallel,\rho_\perp\ge0$.
The admissible targets are exactly
$\mathcal F_0=\{f_0+h:\norm{Ph}\le\rho_\parallel,\ \norm{Qh}\le\rho_\perp\}$.
The radii are external assumptions, not quantities certified by
disagreements. With $f_0=\bar g$ they bound the two components of the common
error.
\end{assumption}

\begin{theorem}[Sharp gain interval]\label{q2:thm:interval}
For fixed $w\in\Waff$ let $c_0(w)=w^\top \mathsf D_Tw-2\ip{f_0}{q_w}$ and
$r(w)=2\rho_\parallel\norm{Pq_w}+2\rho_\perp\norm{Qq_w}$. The exact identified set
of $G_T(w;f)$ over $\mathcal F_0$ is $[c_0(w)-r(w),\,c_0(w)+r(w)]$. For $T=P$,
$Pq_w=0$ and $c_0(w)=\norm{q_w}^2-2\ip{f_0}{q_w}$, so the interval is
$c_0(w)\pm2\rho_\perp\norm{q_w}$. Strict positivity for every admissible target
holds iff $c_0>r$; strict negativity iff $c_0<-r$; both strict signs occur iff
$\abs{c_0}<r$; a zero endpoint permits zero gain and certifies no strict sign.
For $f_0=\bar g$ and $T=P$ the center is observable as
$\norm{Q(g_w-\bar g)}^2-\norm{Q\bar g}^2$.
\end{theorem}
\begin{proof}
By \eqref{q2:eq:gain}, $G_T(w;f_0+h)=c_0(w)-2\ip{Ph}{Pq_w}-2\ip{Qh}{Qq_w}$, and
Cauchy--Schwarz bounds the last two terms by $r(w)$. Choosing
$Ph=\pm\rho_\parallel Pq_w/\norm{Pq_w}$ and $Qh=\pm\rho_\perp Qq_w/\norm{Qq_w}$ with
a common sign (zero for a vanishing component) attains both endpoints;
scaling by $[-1,1]$ attains every intermediate value. The last identity is
$\norm{q_w-Q\bar g}^2-\norm{Q\bar g}^2=\norm{q_w}^2-2\ip{Q\bar g}{q_w}$.
\end{proof}

\begin{proposition}[Sharp joint set and contrasts]\label{q2:prop:joint}
Let $T=P$ and $b_0=(\ip{f_0}{q_j})_j$. The exact joint identified set for
$\mu_q(f)$ is $\{b_0+z:z\in\Ran K_q,\ z^\top K_q^\dagger z\le\rho_\perp^2\}$, so the
feasible gain surface is $\{w\mapsto w^\top K_qw-2(b_0+z)^\top w\}$ over this
ellipsoid; pointwise intervals must not be combined as an independent
product across weights. For a general learner and $w,v\in\Waff$, the
identified set of the contrast $R_T(w;f)-R_T(v;f)$ is
$d_0\pm\bigl(2\rho_\parallel\norm{P(s_w-s_v)}+2\rho_\perp\norm{Q(s_w-s_v)}\bigr)$
with $d_0=\norm{s_w-f_0}^2-\norm{s_v-f_0}^2$; for $T=P$ only $\rho_\parallel$
enters. Controlling only $\rho_\perp$ controls the projection gain but not the
student ordering; controlling only $\rho_\parallel$ has the reverse limitation.
\end{proposition}
\begin{proof}
Let $T_q:\R^m\to\Hh$, $T_qw=q_w$, with range $U$; then $K_q=T_q^*T_q$ and
$\mu_q(f)-b_0=T_q^*Qh$. Feasible $z$ lie in $\Ran T_q^*=\Ran K_q$. For such $z$,
$h_{\min}=T_qK_q^\dagger z\in U$ satisfies $T_q^*h_{\min}=z$ and
$\norm{h_{\min}}^2=z^\top K_q^\dagger z$, and every other preimage adds an element
of $\Ker T_q^*=U^\perp$, so a preimage of norm $\le\rho_\perp$ exists iff the
ellipsoid inequality holds; take $Ph=0$. One common $z$ fixes all weights. The
contrast equals $d_0-2\ip{h}{s_w-s_v}$; argue as in Theorem~\ref{q2:thm:interval}.
\end{proof}

\begin{remark}[Scope of sharpness]\label{q2:auto:scope-of-sharpness}
The sets are sharp for Assumption~\ref{q2:ass:ball} alone, without further
constraints such as a fixed $A(f)$, a target norm or a strict clean-student
advantage; intersecting with such constraints can shrink them. In the example
of Theorem~\ref{q2:thm:counter}, $\bar g=z$; with $f_0=z$, $\rho_\parallel=\sqrt5/2$
and $\rho_\perp=1$ both targets are admissible, the sharp interval at $\mathbf e_1$
is $[-1,3]$, and the two targets attain its endpoints.
\end{remark}

\apppart{Structured teacher observations}{Known training equations, the compatible fiber and the leakage law.}
\subsection{Structured teachers: known training equations}\label{q2:sec:structured}

Lemma~\ref{q2:lem:gauge} allows arbitrary target shifts. In practice the teachers
were themselves trained on the target, and knowing how restricts the shifts.

\begin{assumption}[Known exact teacher structure]\label{q2:ass:teachers}
For each $j$, $M_j:\Hh\to\Hh$ is a known bounded linear operator and the
observed prediction is exactly $g_j=M_jf$. The operators, hyperparameters,
representations and integration rules are fixed across compatible targets.
The observed tuple is compatible with at least one target. Unless added
explicitly, no target norm, realizability or teacher-risk bound restricts the
target.
\end{assumption}

Let $\mathcal Y=\Hh^m$ with $\ip yz_{\mathcal Y}=\sum_j\ip{y_j}{z_j}$,
$\mathcal Mf=(M_1f,\dots,M_mf)$, $y=(g_1,\dots,g_m)$; then
$\mathcal M^*z=\sum_jM_j^*z_j$. The student learner $T$ is common to all
teachers and weights; it may be chosen from the teacher data but is then fixed
on the whole compatible set.

\begin{definition}[Structured observation]\label{q2:auto:structured-observation}
$\mathcal O_{\mathcal M,T}=(P_X,M_1,\dots,M_m,g_1,\dots,g_m,T)$, with exact
functions and population inner products. The compatible set is
$\mathcal F_y=\{f\in\Hh:\mathcal Mf=y\}$.
\end{definition}

An architecture alone does not grant $M_j$ or $T$. An exact population
equation $g_j=M_jf$ is stronger than saying that a finite-sample teacher was
trained by ridge (Section~\ref{q2:sec:stability} treats approximate equations).

\begin{theorem}[Exact compatible fiber]\label{q2:thm:fiber}
Let $\Nn=\Ker\mathcal M=\bigcap_j\Ker M_j$ (common nullspace) and $\Kk=\Nn^\perp$
(teacher-information space). There is a unique $f_\Kk\in\Kk$ with
$\mathcal Mf_\Kk=y$, and $\boxed{\mathcal F_y=f_\Kk+\Nn}$. Moreover
$\Kk=\overline{\Ran\mathcal M^*}=\overline{\sum_j\Ran M_j^*}$. For $h\in\Hh$ the
moment $\ip fh$ is identified iff $h\in\Kk$; otherwise it takes every real
value on $\mathcal F_y$.
\end{theorem}
\begin{proof}
For a solution $f_*$, $f_\Kk:=\Pi_\Kk f_*$ is a solution since
$\mathcal M\Pi_\Nn f_*=0$; solutions differ by elements of $\Nn$, and two
solutions in $\Kk$ differ by an element of $\Kk\cap\Nn=\{0\}$. Also
$h\perp\Ran\mathcal M^*\iff\ip{\mathcal Mh}z_{\mathcal Y}=0\ \forall z\iff\mathcal Mh=0$,
so $\overline{\Ran\mathcal M^*}=\Nn^\perp$. For $f=f_\Kk+n$,
$\ip fh=\ip{f_\Kk}{\Pi_\Kk h}+\ip n{\Pi_\Nn h}$; if $\Pi_\Nn h\ne0$, $n=t\Pi_\Nn h$
sweeps $\R$.
\end{proof}

\subsubsection{Task identification on the fiber}

Retain $U_T,V_T$ from Theorem~\ref{q2:thm:tasks} and their structurally hidden
images $U_\Nn=\Pi_\Nn U_T$, $V_\Nn=\Pi_\Nn V_T$ (closed, finite-dimensional;
$\Pi_\Nn V_T$ need not equal $\Nn\cap V_T$).

\begin{theorem}[Identification boundaries for gains and contrasts]\label{q2:thm:struct-tasks}
On the full compatible fiber:
\begin{enumerate}
\item $G_T(w;\cdot)$ is identified iff $\Pi_\Nn q_w=0$; a contrast $R_T(w)-R_T(v)$ iff $\Pi_\Nn(s_w-s_v)=0$;
\item all gains on $\Delta_m$ or $\Waff$ are identified iff $U_T\subseteq\Kk$; all contrasts iff $V_T\subseteq\Kk$.
\end{enumerate}
A compatible shift $n\in\Nn$ changes the gain by $-2\ip n{\Pi_\Nn q_w}$ and the
contrast by $-2\ip n{\Pi_\Nn(s_w-s_v)}$; an unidentified fixed gain takes both
strict signs on the full fiber.
\end{theorem}
\begin{proof}
Equations \eqref{q2:eq:gain}--\eqref{q2:eq:contrast} and Theorem~\ref{q2:thm:fiber};
test at vertices for the ``all'' statements, as in Theorem~\ref{q2:thm:tasks}.
\end{proof}

\begin{theorem}[Identification of the entire optimal weight set]\label{q2:thm:optset}
For $\mathcal W=\Waff$ or $\Delta_m$, $\argmin_{\mathcal W}R_T(\cdot;f)$ is constant
over the full fiber $f_\Kk+\Nn$ iff $V_T\subseteq\Kk$.
\end{theorem}
\begin{proof}
\emph{Idea.} Sufficiency holds because every pairwise contrast is fixed on
the fiber. For necessity we exhibit a direction $n\in\Nn$ along which the
optimum moves: for affine weights the optimal coordinate is explicit; for
the simplex, optimality along a whole line of targets forces every contrast
to be orthogonal to $n$, and compactness gives a common minimizer.

Sufficiency: all contrasts are fixed. Necessity, affine: with an orthonormal
basis of $V_T$ and isometry $\Phi$, all $s_j=a_\circ+\Phi t_j$ with common
$a_\circ\perp V_T$, and
$R_T(w;f)=\norm{a_\circ-(I-\Pi_{V_T})f}^2+\norm{t(w)-\Phi^*f}^2$, where
$t(w)=\sum_jw_jt_j$ ranges over $\R^k$. The optimal coordinate is $\Phi^*f$; if
$V_T\not\subseteq\Kk$ there is $n\in\Nn$ with $\Phi^*n\ne0$, and $f_\Kk$,
$f_\Kk+n$ have different optima. Simplex: if $w_*$ is optimal for every
$f_\Kk+tn$, $t\in\R$, then against vertex $j$,
$\norm{s_*-f_\Kk}^2-\norm{s_j-f_\Kk}^2-2t\ip n{s_*-s_j}\le0$ for all $t$, forcing
$\ip n{s_*-s_j}=0$ for all $j$, hence $\ip n{s_i-s_j}=0$; a common minimizer
exists by compactness when the set is constant. Apply to every $n\in\Nn$.
\end{proof}

Failure of identification means that some additional target information is
necessary; anchors are one source, not the only one. On a bounded class a gain
can have an identified sign without an identified value, and optimal simplex
weights can stay fixed despite unidentified contrasts
(Section~\ref{q2:sec:bounded}).

\subsubsection{Teachers trained by population least squares or spectral filters}

\begin{proposition}[Exact population training equations]\label{q2:prop:training}
(i) A projection teacher onto a known closed $\mathcal W_j$ has $M_j=P_{\mathcal W_j}$.
(ii) A ridge teacher with feature synthesis $\Psi_j:\Ff_j\to\Hh$ and
$\lambda_j>0$ has $M_j=\Psi_j(\Psi_j^*\Psi_j+\lambda_jI)^{-1}\Psi_j^*=\mathcal L_j(\mathcal L_j+\lambda_jI)^{-1}$,
$\mathcal L_j=\Psi_j\Psi_j^*$, with $\Ker M_j=\Ker\mathcal L_j$. (iii) More
generally $M_j=a_j(\mathcal L_j)$ for a bounded Borel filter is self-adjoint,
and if $a_j(0)=0$ and $a_j>0$ on positive spectrum then $\Ker M_j=\Ker\mathcal L_j$;
zeros of the filter on further modes enlarge the kernel. Consequently
$\Kk=\overline{\mathcal W_1+\dots+\mathcal W_m}$ for projection teachers and
$\Kk=\overline{\sum_j(\Ker\mathcal L_j)^\perp}$ for positive filters, and the
teachers identify $f$ exactly iff $\bigcap_j\Ker M_j=\{0\}$.
\end{proposition}
\begin{proof}
(i) $\norm{h-f}^2=\norm{h-P_{\mathcal W_j}f}^2+\norm{(I-P_{\mathcal W_j})f}^2$ on
$\mathcal W_j$. (ii) \cref{ext:filters}; the multiplier $t/(t+\lambda_j)$ vanishes
only at $0$. (iii) $\norm{M_jh}^2=\int\abs{a_j}^2d\ip h{\mathsf E_j h}$ vanishes iff
the spectral measure of $h$ lives on the zero set of $a_j$. Then use
Theorem~\ref{q2:thm:fiber}.
\end{proof}

\subsection{The leakage law}\label{q2:sec:leakage}

\begin{assumption}[Predictions lie in the information space]\label{q2:ass:range}
Every $g_j\in\Kk$. This holds when all $M_j$ are self-adjoint (e.g.\
Proposition~\ref{q2:prop:training}): for $n\in\Nn$,
$\ip n{g_j}=\ip n{M_jf}=\ip{M_jn}f=0$. It is not automatic for nonnormal
teacher operators.
\end{assumption}

\begin{theorem}[Hidden information equals student leakage]\label{q2:thm:leakage}
Under Assumption~\ref{q2:ass:range}, define the leakages $\ell_j=\Pi_\Nn Tg_j$,
$\mathcal L_0=\Span\{\ell_j\}$ and $\mathcal L_c=\Span\{\ell_j-\ell_m:j<m\}$. Then
\[
 U_\Nn=\mathcal L_0,\qquad V_\Nn=\mathcal L_c\subseteq\mathcal L_0,\qquad
 \dim\mathcal L_c\le\dim\mathcal L_0\le\dim\mathcal L_c+1,
\]
and on the full compatible fiber:
\begin{enumerate}
\item all gains are identified iff $\ell_j=0$ for every $j$;
\item all student-risk contrasts, equivalently the entire optimal weight set
(affine or simplex), are identified iff $\ell_1=\dots=\ell_m$;
\item gain identification implies selection identification; the converse
fails (Example~\ref{q2:ex:sel-not-gain}).
\end{enumerate}
For a fixed weight, gain identification is equivalent to $\sum_jw_j\ell_j=0$.
\end{theorem}
\begin{proof}
$\Pi_\Nn q_j=\Pi_\Nn g_j-\Pi_\Nn Tg_j=-\ell_j$ and
$\Pi_\Nn(s_j-s_m)=\ell_j-\ell_m$; take spans. $\ell_j=(\ell_j-\ell_m)+\ell_m$
gives $\mathcal L_0=\mathcal L_c+\Span\{\ell_m\}$. Then apply
Theorems~\ref{q2:thm:struct-tasks} and~\ref{q2:thm:optset}; and
$\Pi_\Nn q_w=-\sum_jw_j\ell_j$.
\end{proof}

\begin{example}[Identified selection, unidentified gain]\label{q2:ex:sel-not-gain}
Let $u,v,z$ be orthonormal, $M_1=P_{\Span\{u,v\}}$, $M_2=P_{\Span\{u\}}$,
$g_1=u+v$, $g_2=u$, so $\Kk=\Span\{u,v\}$ and compatible targets include
$f=u+v+cz$, $c\in\R$. Let $t_0=(u+z)/\sqrt2$ and $P=P_{\Span\{t_0,v\}}$. Then
$s_1=(u+z)/2+v$, $s_2=(u+z)/2$, $\ell_1=\ell_2=z/2\ne0$, and for $w=(a,1-a)$,
$R_S(w;f)=\tfrac14+(\tfrac12-c)^2+(a-1)^2$: $a=1$ is the unique affine and
simplex optimum for every $c$. But $q_1=(u-z)/2$ and
$G((1,0);f)=\norm{q_1}^2-2\ip f{q_1}=c-\tfrac12$, unidentified and of either sign.
\end{example}

\subsection{Structured impossibility with exact, correctly trained teachers}\label{q2:sec:struct-counter}

\begin{theorem}[Same trained teachers, same risks, opposite decisions]\label{q2:thm:struct-counter}
Let $u,v,z$ be independent Rademacher coordinate functions, $M_1=P_{\Span\{u\}}$,
$M_2=P_{\Span\{v\}}$, $t_0=(u-v+2z)/\sqrt6$, $P=P_{\Span\{t_0\}}$, and
$f_\pm=u+v\pm2z$. Both worlds have the same operators and predictions
$g_1=u$, $g_2=v$; each teacher is the unique population least-squares fit in
its own feature space. In both worlds $\norm f^2=6$,
$R_W(\mathbf e_1)=R_W(\mathbf e_2)=5$ and $A(f)=10/3<5$. Teacher $1$ has gain
$1/6$ in the plus world and $-5/2$ in the minus world (teacher $2$ the reverse).
The unique simplex optima are $(1,0)$ and $(0,1)$, with risk $29/6<5$; the
affine optima are $(5/2,-3/2)$ and $(-3/2,5/2)$, with risk $10/3$.
\end{theorem}
\begin{proof}
$M_1f_\pm=u$, $M_2f_\pm=v$. Errors $-v\mp2z$ and $-u\mp2z$ have squared norm
$5$; $\ip{f_\pm}{t_0}=\pm4/\sqrt6$, $A=6-16/6$. With $Pu=t_0/\sqrt6$,
$Pv=-t_0/\sqrt6$, $w=(a,1-a)$, $b=2a-1$: $s_w=bt_0/\sqrt6$ and
$R_S(w;f_\pm)=6+(b^2\mp8b)/6=\tfrac{10}3+(b\mp4)^2/6$. At $b=\pm1$ the risks are
$29/6$ and $45/6$, gains $1/6$ and $-5/2$. On $[-1,1]$ the derivatives
$(b\mp4)/3$ have constant sign, giving the simplex optima; on $\R$ the minima
are at $b=\pm4$.
\end{proof}

\begin{corollary}[Zero-anchor lower bounds]\label{q2:cor:struct-zero}
Even if individual teacher risks, $\norm f^2$ and $A(f)$ are disclosed, every
unlabeled procedure has worst-world sign error $\ge1/2$ for teacher $1$ and
worst-world expected regret $\ge7/6$ (simplex) and $\ge8/3$ (affine); uniform
weights attain both.
\end{corollary}
\begin{proof}
All disclosed quantities agree across worlds. For $B=2\widehat w_1-1$ the simplex
regrets are $(B^2\mp8B+7)/6$, with average $(\E B^2+7)/6$; the affine regrets
are $(B\mp4)^2/6$, with average $(\E B^2+16)/6$.
\end{proof}

Here $\Kk=\Span\{u,v\}$, $f_+-f_-=4z\in\Nn$, and $\ell_1=z/3\ne\ell_2=-z/3$,
exactly as Theorem~\ref{q2:thm:leakage} requires for non-identified selection.

\subsection{Bounded fibers: sharp intervals and anchor-free decisions}\label{q2:sec:bounded}

Let $f_c\in\mathcal F_y$ be a known compatible reference (e.g.\ $f_\Kk$) and
$\rho\ge0$ an external radius; the bounded fiber is
$\mathcal F_{y,c,\rho}=\{f_c+n:n\in\Nn,\ \norm n\le\rho\}$.

\begin{theorem}[Sharp bounded-fiber intervals]\label{q2:thm:fiber-intervals}
With $u_w=\Pi_\Nn q_w$ and $d_{wv}=\Pi_\Nn(s_w-s_v)$,
\[
 G_T(w;f)\in[G_T(w;f_c)\mp2\rho\norm{u_w}],\qquad
 R_T(w;f)-R_T(v;f)\in[C_c(w,v)\mp2\rho\norm{d_{wv}}],
\]
$C_c(w,v)=R_T(w;f_c)-R_T(v;f_c)$, exactly. Strict positive gain for every
compatible target is equivalent to $G_T(w;f_c)>2\rho\norm{u_w}$; both strict
signs occur iff $\abs{G_T(w;f_c)}<2\rho\norm{u_w}$. For finitely many
$h_1,\dots,h_p$, the exact joint moment set is
$\{(\ip{f_c}{h_i})_i+z:z\in\Ran B_0,\ z^\top B_0^\dagger z\le\rho^2\}$,
$(B_0)_{ij}=\ip{\Pi_\Nn h_i}{\Pi_\Nn h_j}$.
\end{theorem}
\begin{proof}
The shift is $-2\ip n{u_w}$; Cauchy--Schwarz and $n=\pm\rho u_w/\norm{u_w}$
(scaled) give the interval; likewise for contrasts. For the joint claim let
$H_0a=\sum_ia_i\Pi_\Nn h_i$, so $B_0=H_0^*H_0$ and shifts are $H_0^*n$; for
$z\in\Ran B_0$, $n_0=H_0B_0^\dagger z$ has $H_0^*n_0=z$ and
$\norm{n_0}^2=z^\top B_0^\dagger z$, minimal among preimages.
\end{proof}

\begin{theorem}[Exact robust-optimality test without anchors]\label{q2:thm:robust}
A fixed simplex weight $w$ minimizes $R_T(\cdot;f)$ for every
$f\in\mathcal F_{y,c,\rho}$ iff, for every vertex $j$,
\[
 \boxed{\ip{s_w-f_c}{s_j-s_w}\ge\rho\norm{\Pi_\Nn(s_j-s_w)}}.
\]
Unidentified contrasts therefore need not prevent an anchor-free optimal
decision on a bounded class.
\end{theorem}
\begin{proof}
On $\conv\{s_j\}$, $s_w$ is optimal for $f$ iff $\ip{s_w-f}{s-s_w}\ge0$ for all
feasible $s$ (necessity from the one-sided derivative; sufficiency from
$\norm{s-f}^2-\norm{s_w-f}^2=2\ip{s_w-f}{s-s_w}+\norm{s-s_w}^2$); by linearity
it suffices to test vertices. For $f=f_c+n$ the vertex-$j$ quantity is
$\ip{s_w-f_c}{s_j-s_w}-\ip n{\Pi_\Nn(s_j-s_w)}$, whose minimum over the ball is
the first term minus $\rho\norm{\Pi_\Nn(s_j-s_w)}$. The worst target may differ
across vertices, which does not affect the equivalence.
\end{proof}

\begin{proposition}[What exact teacher-risk calibration adds]\label{q2:prop:calibration}
Under Assumption~\ref{q2:ass:range}, for $f=f_\Kk+n$,
$R_W(\mathbf e_j;f)=\norm{g_j-f_\Kk}^2+\norm n^2$; for projection teachers
$R_W(\mathbf e_j;f)=\norm f^2-\norm{g_j}^2$. So individual teacher risks reveal at
most $\norm n$. If $\norm n=\rho$ is fixed, the extremes of $\ip nh$ remain
$\pm\rho\norm{\Pi_\Nn h}$; all intermediate values are attained when
$\dim\Nn\ge2$, only the two endpoints when $\dim\Nn=1$ and $\rho>0$.
\end{proposition}
\begin{proof}
$g_j-f_\Kk\in\Kk\perp n$. For a projection teacher $f=g_j+(I-M_j)f$
orthogonally. On a sphere of $\Nn$, rotate in a plane containing
$\Pi_\Nn h$ to sweep the values; in one dimension the sphere has two points.
\end{proof}

\subsection{Exact identification versus stable recovery}\label{q2:sec:stability}

\begin{theorem}[Range criterion for stable recovery of a moment]\label{q2:thm:stable}
For $h\in\Hh$ the following are equivalent: (i) there is $C<\infty$ with
$\abs{\ip fh}\le C\norm{\mathcal Mf}_{\mathcal Y}$ for every $f\in\Hh$; (ii)
$h=\mathcal M^*z$ for some $z\in\mathcal Y$. Then $\ip fh=\ip yz_{\mathcal Y}$, and
the smallest constant is $C_h=\min\{\norm z:\mathcal M^*z=h\}$. Identification
needs $h\in\overline{\Ran\mathcal M^*}$; stable linear recovery in the
observation norm needs $h\in\Ran\mathcal M^*$.
\end{theorem}
\begin{proof}
If $h=\mathcal M^*z$, $\abs{\ip fh}=\abs{\ip{\mathcal Mf}z}\le\norm z\norm{\mathcal Mf}$.
Conversely $L(\mathcal Mf)=\ip fh$ is well defined on $\Ran\mathcal M$ by (i),
bounded with norm $\le C$, and extends to $\overline{\Ran\mathcal M}$; Riesz gives
$z_0\in\overline{\Ran\mathcal M}$ with $\ip fh=\ip{\mathcal Mf}{z_0}$ for all $f$,
so $h=\mathcal M^*z_0$. Other solutions add elements of
$\Ker\mathcal M^*=(\overline{\Ran\mathcal M})^\perp$, so $z_0$ has minimal norm,
equal to $\norm L$.
\end{proof}

\begin{example}[Injective teachers need not give stable identification]\label{q2:auto:injective-teachers-need-not-give-stable-}
On $\ell^2$ let $(Af)_j=f_j/j$ and $h_j=1/j$: $A$ is bounded, positive,
self-adjoint and injective, so $\Nn=\{0\}$ and every moment is identified. But
$h\notin\Ran A^*$ (a preimage would be constant), and $f^{(N)}_j=j/N$ for
$j\le N$ gives $\norm{Af^{(N)}}=N^{-1/2}\to0$ while $\ip{f^{(N)}}h=1$. The
targets have unbounded norms; regularized bounds on bounded classes are not
excluded.
\end{example}

\begin{proposition}[Robustness to imperfect teacher equations]\label{q2:prop:approx}
Keep the declared $\mathcal M$. If a known reference $f_c$ and the target satisfy
$\norm{\mathcal M(f-f_c)}\le\varepsilon$ and $\norm{\Pi_\Nn(f-f_c)}\le\rho$, and
$\Pi_\Kk h=\mathcal M^*z_h$, then
$\abs{\ip{f-f_c}h}\le\varepsilon\norm{z_h}+\rho\norm{\Pi_\Nn h}$. Hence a fixed gain
differs from its reference value by at most
$2\varepsilon C_{\Pi_\Kk q_w}+2\rho\norm{\Pi_\Nn q_w}$, and similarly for contrasts
with $h=s_w-s_v$. If $\norm{\mathcal Mf-y}\le\eta_0$ and $\norm{\mathcal Mf_c-y}\le\eta_c$
one may take $\varepsilon=\eta_0+\eta_c$. When teachers only approximately obey
their operators they need not lie in $\Kk$, so the leakage simplification must
be rechecked; the general $q_w$ and $s_w-s_v$ formulas remain valid.
\end{proposition}
\begin{proof}
Decompose $h$: $\ip dh=\ip{\mathcal Md}{z_h}+\ip{\Pi_\Nn d}{\Pi_\Nn h}$ with
$d=f-f_c$; Cauchy--Schwarz; minimize $\norm{z_h}$ by Theorem~\ref{q2:thm:stable}.
\end{proof}

For selection, an equation error becomes a squared bias plus an extra
residual variance, compared with ordinary stacking in \cref{q3:cor:teacher-error}; an ill-conditioned teacher stack
amplifies the error by $1/\alpha_{\mathcal M}$, and compact (e.g.\ kernel ridge) teachers
need a spectral truncation.

\subsection{Label-model assumptions: identification up to one scalar}\label{q2:sec:factor}

Between no assumption (\S\ref{q2:sec:counter}) and exact teacher equations
(\S\ref{q2:sec:leakage}) lie the statistical assumptions of weak-supervision
label models, which posit that supervisors are noisy versions of the target
with mutually unrelated errors. The following result shows what such an
assumption buys for selection, and which form of it is needed.

\begin{assumption}[Student-relative factor model]\label{q2:ass:factor}
There are scalars $a_j\ne0$ and functions $\varepsilon_j$ with
$g_j=a_jf+\varepsilon_j$, $j\le m$, such that
$\ip{\varepsilon_j}{s_k}=0$ for all $j\ne k$, where $s_k=Tg_k$.
\end{assumption}

\begin{theorem}[Factor-model identification of selection]\label{q2:thm:factor}
Under Assumption~\ref{q2:ass:factor} let $m\ge3$, $\mu_k=\ip f{s_k}$ and
$K_{jk}=\ip{g_j}{s_k}$, an unlabeled observable.
\begin{enumerate}
\item[(i)] $K_{jk}=a_j\mu_k$ for all $j\ne k$.
\item[(ii)] If every $\mu_k\ne0$, then $a_j/a_l=K_{jk}/K_{lk}$ for distinct $j,l$
and any $k\notin\{j,l\}$, so $\alpha=a/a_1$ is identified, and
$\mu=\upsilon\widetilde\mu$ with $\widetilde\mu_k=K_{jk}/\alpha_j$ (any $j\ne k$)
identified and the single scalar $\upsilon=1/a_1$ unknown.
\item[(iii)] $\upsilon$ is not identified: for every $t\ne0$ the map
$(f,a,\varepsilon)\mapsto(tf,a/t,\varepsilon)$ leaves every $g_j$, every
$s_k$ and Assumption~\ref{q2:ass:factor} unchanged and maps $\mu$ to $t\mu$.
\item[(iv)] Every risk contrast is
$R_T(w)-R_T(v)=\norm{s_w}^2-\norm{s_v}^2-2\upsilon\sum_j(w_j-v_j)\widetilde\mu_j$,
so all selection decisions are identified once one scalar moment
$\ip fh$, $h=\sum_jc_js_j$ with $\sum_jc_j\widetilde\mu_j\ne0$, is known.
Gains also need $\ip f{g_j}$, which (i)--(iii) do not identify.
\item[(v)] The raw label-model conditions $\ip{\varepsilon_j}{\varepsilon_k}=0$
($j\ne k$) and $\ip{\varepsilon_j}f=0$ do not imply the assumption, since
$\ip{\varepsilon_j}{s_k}=a_k\ip{\varepsilon_j}{Tf}+\ip{\varepsilon_j}{T\varepsilon_k}$; for
$T=P$ this is $a_k\ip{P\varepsilon_j}f+\ip{P\varepsilon_j}{P\varepsilon_k}$, a transmitted
orthogonality. For instance, $f=e_1$, $\varepsilon_1=e_2$ and $P$ the projection
onto $\Span\{e_1+e_2\}$ in $\R^2$ give $\ip{\varepsilon_1}f=0$ but
$\ip{P\varepsilon_1}f=\tfrac12$.
\end{enumerate}
\end{theorem}
\begin{proof}
(i) $\ip{g_j}{s_k}=a_j\ip f{s_k}+\ip{\varepsilon_j}{s_k}=a_j\mu_k$. (ii) For
$k\notin\{j,l\}$, $K_{jk}/K_{lk}=a_j/a_l$ since $\mu_k\ne0$; with $m\ge3$ such a
$k$ exists for every pair. Then $\mu_k=K_{jk}/a_j=\upsilon K_{jk}/\alpha_j$.
(iii) $(a_j/t)\,tf+\varepsilon_j=g_j$; $s_k=Tg_k$ is unchanged, so is the
orthogonality; $\ip{tf}{s_k}=t\mu_k$. (iv) Expand
$\norm{s_w-f}^2-\norm{s_v-f}^2$ and use $\ip f{s_w}=\sum_jw_j\mu_j$;
$\ip fh=\upsilon\sum_jc_j\widetilde\mu_j$ determines $\upsilon$. (v) Expand
$s_k=a_kTf+T\varepsilon_k$; in the example $Pe_2=\tfrac12(e_1+e_2)$.
\end{proof}

\begin{corollary}[One-dimensional anchor cost]\label{q2:cor:factor-anchors}
In the setting of Theorem~\ref{q2:thm:factor}(ii), let $v_\star\in V_T$ be the
identified element with $\ip{v_\star}{s_j-s_m}=\widetilde\mu_j-\widetilde\mu_m$, so that
$\Pi_{V_T}f=\upsilon v_\star$. If $v_\star=0$, selection is identified without
anchors. Otherwise, with a known reference $f_0$, put
$\widehat\upsilon=\bigl(\ip{f_0}{v_\star}+n^{-1}\sum_i(Y_i-f_0(X_i))v_\star(X_i)\bigr)/\norm{v_\star}^2$
and select by projecting $\widehat\upsilon v_\star$ as in
\cref{q3:thm:cone}. If
$\E[(Y-f_0(X))^2\mid X]\le v_0^2$, then for simplex, bounded affine or affine
weights $\E\Reg(\widehat w)\le v_0^2/n$, independent of $m$ and of the contrast
dimension $k$.
\end{corollary}
\begin{proof}
$\ip f{s_j-s_m}=\upsilon(\widetilde\mu_j-\widetilde\mu_m)=\ip{\upsilon v_\star}{s_j-s_m}$ for all
$j$, and both $\Pi_{V_T}f$ and $\upsilon v_\star$ lie in $V_T$, so they coincide.
The regret is at most the squared error of the estimated coordinate,
$(\widehat\upsilon-\upsilon)^2\norm{v_\star}^2$, by \cref{q3:thm:cone}. The
estimate of $\ip f{v_\star}$ is unbiased with variance at most
$v_0^2\norm{v_\star}^2/n$, so $\E(\widehat\upsilon-\upsilon)^2\norm{v_\star}^2\le v_0^2/n$.
\end{proof}

The theorem says what a label model must control: not the raw orthogonality
of supervisor errors, but their orthogonality to the \emph{other
supervisors' students}. When it holds, unlabeled data reduce the missing
selection information from the $k$ coordinates of $\Pi_{V_T}f$ to one scalar,
and one anchor direction suffices.

\apppart{What labels add and summary}{}
\subsection{What labels add}\label{q2:sec:labels}

If the entire joint law of $(X,Y)$ is known and $\E[Y\mid X]=f(X)$, the target
itself is identified in $L^2(P_X)$; nothing in \cref{app:q2} contradicts that. The
relevant question is quantitative: \emph{how many} anchors are needed to
estimate the finitely many moments that the decisions depend on --- the
projections of $f-f_c$ onto $U_\Nn$ (gains) and onto $V_\Nn$ (selection),
with $\Nn=\Hh$ and $f_c$ any reference in the unstructured case. This is \cref{app:q3}.
Two caveats belong here. First, if $\E[\xi\mid X]=b_\xi(X)$ is unknown, the
anchor law identifies only $f+b_\xi$: replacing $(f,b_\xi)$ by $(f+h,b_\xi-h)$
leaves it unchanged, so no unbiased-recovery statement covers arbitrary label
bias. Second, noisy residual products are biased for error Grams:
$\E[(g_j-Y)(g_k-Y)]=\Sigma_{jk}+\E\xi^2$, a multiple of $\one\one^\top$ that does
not change affine minimizers but is not the clean Gram (\cref{q3:prop:noise}).

\subsection{Identification map}\label{q2:sec:summary}

\begin{center}\small
\begin{tabularx}{\textwidth}{@{}p{0.30\textwidth}X@{}}
\toprule
Question & Exact answer\\
\midrule
What disagreements identify & $J\Sigma J=-\tfrac12J\Upsilon J$; with $\mathcal O_P$ also $J\Gpar J$, $J\Gperp J$; not the common error\\
What a gain needs & $\ip f{q_w}$; all gains $\iff\Pi_{U_T}f$\\
What selection needs & $\ip f{s_w-s_v}$; all contrasts $\iff\Pi_{V_T}f$; $U\perp V$ for $T=P$\\
Realizability & identifies the projection gain, not the selection\\
No labels at all & worst-world sign error $\ge\tfrac12$, regret $\ge\tfrac14$ (bounded, nondegenerate)\\
Radius assumption & sharp intervals $c_0\pm r(w)$; joint ellipsoid, not a product\\
Known teacher equations & fiber $f_\Kk+\Nn$; gains $\iff\ell_j=0$; selection $\iff\ell_j$ equal\\
Bounded fiber & sharp intervals; vertex test for anchor-free optimality\\
Stability & identification $h\in\overline{\Ran\mathcal M^*}$; stable recovery $h\in\Ran\mathcal M^*$\\
Factor (label-model) assumption & selection up to one scalar; one anchor direction, regret $\le v_0^2/n$\\
\bottomrule
\end{tabularx}
\end{center}

\paragraph{Relation to prior work.} What in this appendix is standard and what is new, and detailed comparisons with prior work, are collected in \cref{app:boundary,app:related}.

\section{Anchor Complexity: Formal Results and Proofs}\label{app:q3}

\paragraph{Summary.} \noindent
\cref{app:q2} reduced every gain and every selection decision to finitely many unknown
linear target moments, residualized against a known reference $f_0$ and, when
teacher structure is known, projected onto the common nullspace $\Nn$. We
estimate exactly these moments from $n$ passive anchors and quantify the
label cost. For gains, a single uniform deviation event certifies the sign of
every weight on $\WL$, including weights chosen from the same anchors; the
cost scales as $\gamma^{-2}$ in the margin and, for simultaneous signs, as the
Gaussian width of the margin-normalized unresolved query vectors: constant for
duplicated queries, $\log m$ in the worst case, both matched by lower bounds.
For selection, the regret of projecting the estimated target coordinate onto
the feasible prediction set is at most the squared projection of the
estimation error onto the local tangent cone; in expectation it is at most
$v_0^2d_\Nn/n$, where $d_\Nn=\tr(\Pi_V\Pi_\Nn\Pi_V)\le k\le m-1$, matched by a
$\sigma^2d_\Nn/n$ lower bound. When the hidden target lies in the $r$
unresolved directions, regressing the residual label on them removes the
signal from this variance: in Gaussian designs the selection risk is exactly
$\sigma^2d_\Nn/(n-r-1)$, against at least $\sigma^2k/(n-k-1)$ for ordinary
stacking without structure. Weak supervision with anchors beats every
anchor-only procedure, tuned ridge included, when the best aggregate's
excess risk plus this term is below $\sigma^2d_{\mathrm{eff}}(\sigma^2/n)/n$, with $d_{\mathrm{eff}}$
the effective dimension of the target; without structure the approximation
error re-enters as variance. Approximate structure adds a squared bias that
teacher-equation error controls, with a spectral truncation for compact
teachers. The same simplex dictionary costs order
$\log m$ at a vertex and $m-1$ in its interior. For passive anchors on a fixed
cell dictionary the noiseless minimax regret is exactly
$\sum_\ell p_\ell B_\ell^2(1-p_\ell)^n$, so dictionaries with identical
population Gram matrices can need exponentially different numbers of labels.

\paragraph{Where the main text uses this appendix.}
Formal versions of the main-text results; supporting results are cited in their proofs.

\begin{center}\small
\begin{tabularx}{\textwidth}{@{}>{\raggedright}p{0.24\textwidth}X@{}}
\toprule
Main-text result & Formal results\tabularnewline
\midrule
\cref{thm:certificate} & \cref{q3:thm:uniform} (p.~\pageref{q3:thm:uniform}); \cref{q3:thm:sign-upper} (p.~\pageref{q3:thm:sign-upper}); \cref{q3:thm:extremes} (p.~\pageref{q3:thm:extremes}); \cref{q3:cor:sign-order} (p.~\pageref{q3:cor:sign-order})\tabularnewline
\cref{thm:selection} & \cref{q3:thm:cone} (p.~\pageref{q3:thm:cone}); \cref{q3:thm:expected} (p.~\pageref{q3:thm:expected}); \cref{q3:thm:angle} (p.~\pageref{q3:thm:angle}); \cref{q3:prop:vertex} (p.~\pageref{q3:prop:vertex})\tabularnewline
\cref{prop:compare} & \cref{q3:thm:rr} (p.~\pageref{q3:thm:rr}); \cref{q3:thm:rr-gauss} (p.~\pageref{q3:thm:rr-gauss}); \cref{q3:thm:stack} (p.~\pageref{q3:thm:stack}); \cref{q3:thm:anchor-only} (p.~\pageref{q3:thm:anchor-only}); \cref{q3:prop:anchor-bayes} (p.~\pageref{q3:prop:anchor-bayes}); \cref{q3:cor:weak-bayes} (p.~\pageref{q3:cor:weak-bayes}); \cref{q3:cor:teacher-error} (p.~\pageref{q3:cor:teacher-error})\tabularnewline
\bottomrule
\end{tabularx}
\end{center}

\paragraph{Notation.} As in \cref{app:notation,app:translation}.

\paragraph{Reading guide: three branches.} After the common setup
(\cref{q3:sec:setup}) the appendix splits into two branches, each organized
as upper bound, lower bound, matched rate and limitation; a third compares
structured selection with explicit baselines, and passive coverage and the
bridge between observation models close it.
\begin{center}\small
\begin{tabularx}{\textwidth}{@{}>{\raggedright}p{0.17\textwidth}>{\raggedright}X>{\raggedright}X>{\raggedright\arraybackslash}p{0.22\textwidth}@{}}
\toprule
Branch & Upper bound & Lower bound and matched rate & Sections\tabularnewline
\midrule
I. Gain and sign certification & uniform certificate; Gaussian-width sign bound & two-point (margin) and packing ($\log m$) bounds; worst case $\sigma^2\gamma^{-2}[\log m+\log(1/\delta)]$ & \cref{q3:sec:gain,q3:sec:width,q3:sec:sign-lower}\tabularnewline
II. Selection and regret & local-cone regret; $v_0^2\dN/n$ & unresolved-angle bound; $\sigma^2\dN/n$; vertex $\log m$ versus interior $m-1$ & \cref{q3:sec:selection,q3:sec:select-lower}\tabularnewline
III. Comparisons & residual regression: exactly $\sigma^2\dN/(n-r-1)$ (Gaussian design) & stacking $\ge\sigma^2k/(n-k-1)$; every anchor-only rule $\ge\sigma^2d_{\mathrm{eff}}(\sigma^2/n)/n$ (Bayes) & \cref{q3:sec:compare}\tabularnewline
Passive coverage & exact noiseless formula $\mathfrak c_n$ & order $\mathfrak c_n+\mathfrak v_n$; equal Grams, different cost & \cref{q3:sec:cells,q3:sec:bridge}\tabularnewline
\bottomrule
\end{tabularx}
\end{center}
Constants are kept in the formal statements; the table and the summary give
orders. What is \emph{not} established is listed in \cref{q3:sec:summary3}.

\paragraph{Three observation experiments.} Rates proved in one model do not
transfer automatically to another.
\begin{center}\small
\begin{tabularx}{\textwidth}{@{}>{\raggedright}p{0.15\textwidth}>{\raggedright}p{0.17\textwidth}>{\raggedright}X>{\raggedright}X>{\raggedright\arraybackslash}X@{}}
\toprule
Experiment & Observed & Assumption & Used for & Does not imply\tabularnewline
\midrule
Passive anchors & pairs $(X_i,Y_i)$ & i.i.d.\ sampling; tails of $Y-f_0$; coverage of the relevant directions & actual label complexity; coverage results & the Gaussian sequence model\tabularnewline
Sub-Gaussian moments & $\widehat\vartheta=\vartheta+\epsilon$ & mgf proxy $\Omega/n$ & concentration upper bounds; certificates & passive realizability (needs envelopes)\tabularnewline
Gaussian moments & $\vartheta+n^{-1/2}\Omega^{1/2}G$ & known covariance & sharp local rates and lower bounds & empirical coverage of passive anchors\tabularnewline
\bottomrule
\end{tabularx}
\end{center}
\Cref{q3:sec:bridge} states when the Gaussian results apply to passive
anchors.

\apppart{Setup and observation experiments}{}
\subsection{Setup: residual moments and observation experiments}\label{q3:sec:setup}

\paragraph{Role in the paper.} \cref{app:q3} answers the third question: \emph{how
many clean labels are needed, and what determines that number?} Population
prediction geometry is treated as known here; finite unlabeled data and
fitted students are \cref{app:q4}.

\paragraph{Standing model.} $\Hh=L^2(P_X)$. Fixed teachers $g_j$ and students
$s_j$ ($s_j=Tg_j$ for a common bounded linear $T$, or any fixed functions with
$s_w=\sum_jw_js_j$), $q_j=g_j-s_j$. All results hold conditionally on an
independently fitted dictionary. Recall (\cref{q2:thm:tasks}) that for $w,v\in\Waff$
\begin{equation}\label{q3:eq:linear-forms}
 G_T(w;f)=\norm{g_w}^2-\norm{s_w}^2-2\ip f{q_w},\qquad
 R_T(w;f)-R_T(v;f)=\norm{s_w}^2-\norm{s_v}^2-2\ip f{s_w-s_v}.
\end{equation}

\begin{assumption}[Known reference and residual subspace]\label{q3:ass:residual}
A function $f_0$ and a closed subspace $\Nn\subseteq\Hh$ are known, fixed
independently of the anchors, with $\bar f:=f-f_0\in\Nn$. Unstructured case:
$\Nn=\Hh$ and $f_0$ is any known reference ($f_0=0$ allowed). Structured
case (\cref{q2:sec:structured}): $f_0=f_\Kk$ and $\Nn$ the common
nullspace of the teacher operators.
\end{assumption}

Under Assumption~\ref{q3:ass:residual}, $\ip{\bar f}h=\ip{\bar f}{\Pi_\Nn h}$ for
every $h$. Hence, with \emph{unresolved query directions}
$h_j=\Pi_\Nn q_j$ and the computable reference gain
$G_T(w;f_0)=w^\top \mathsf D_Tw-2\ip{f_0}{q_w}$,
\begin{equation}\label{q3:eq:residual-gain}
 G_T(w;f)=G_T(w;f_0)-2\sum_jw_j\ip{\bar f}{h_j}.
\end{equation}

\paragraph{Anchors.} $n$ iid pairs $(X_i,Y_i)$, $Y_i=f(X_i)+\xi_i$,
$\E[\xi_i\mid X_i]=0$, independent of the dictionary, $T$, $f_0$ and $\Nn$. For a
known feature $h$ put
\[
 \mu_h=\ip{\bar f}h,\qquad
 \widehat\mu_h=\frac1n\sum_{i=1}^n(Y_i-f_0(X_i))h(X_i),\qquad
 \E\widehat\mu_h=\E[\E[Y-f_0(X)\mid X]h(X)]=\mu_h ,
\]
assuming integrability. Neither the noise variance nor $\norm f^2$ is needed.

\begin{proposition}[Noise cancellation and the limits of noisy Grams]\label{q3:prop:noise}
For fixed $h,\widetilde h$,
$(h(X)-Y)^2-(\widetilde h(X)-Y)^2=h^2-\widetilde h^2-2Y(h-\widetilde h)$ has
expectation $\norm{h-f}^2-\norm{\widetilde h-f}^2$. With $\bar\sigma^2=\E\xi^2$,
\[
 \E[(g_j-Y)(g_k-Y)]=\Sigma_{jk}+\bar\sigma^2,\qquad
 \E[(Pg_j-Y)(Pg_k-Y)]=(\Gpar)_{jk}+A(f)+\bar\sigma^2 .
\]
The added matrices are multiples of $\one\one^\top$: they do not change affine
minimizers, but the expectations are not the error Grams.
\end{proposition}
\begin{proof}
$Y^2$ cancels in the paired difference; $\E[\xi\mid X]=0$ removes mixed terms.
Also $Pg_j-f=Pe_j-Qf$ with $\ip{Pe_j}{Qf}=0$; and $w^\top\one\one^\top w=1$ on $\Waff$.
\end{proof}

\paragraph{Three observation experiments.} Lower bounds fix the dictionary,
$P_X$ and all unlabeled information across alternatives.
\begin{enumerate}
\item \emph{Passive anchors}, as above; for Gaussian lower bounds,
$\xi\mid X\sim N(0,\sigma^2)$ independently.
\item \emph{Sub-Gaussian moment experiment}: a moment $\vartheta\in\R^d$ is
estimated by $\widehat\vartheta=\vartheta+\epsilon$ with
$\E e^{u^\top\epsilon}\le e^{u^\top\Omega u/(2n)}$ for all $u$; a covariance bound
alone does not imply this.
\item \emph{Gaussian moment experiment}: $Z=\vartheta+n^{-1/2}\Omega^{1/2}G$, $G\sim N(0,I_d)$.
\end{enumerate}
Passive empirical moments realize the second experiment when the
single-anchor vectors have proxy $\Omega$: e.g.\ if $\abs{Y-f_0(X)}\le B_0$ and
$\norm{\psi(X)}\le L_\psi$ for the feature vector $\psi$, then
$\Omega=B_0^2L_\psi^2I$ is valid, by Lemma~\ref{q3:lem:hoeffding} applied to
$u^\top(Y-f_0)\psi\in[-B_0L_\psi\norm u,B_0L_\psi\norm u]$ and independence.
Orthonormality in $L^2(P_X)$ gives no dimension-free pointwise bound on
$\psi(X)$: proxies, coverage and empirical geometry are part of the anchor
complexity. The passive realization of the Gaussian experiment is
Section~\ref{q3:sec:bridge}.

\apppart{Branch I: gain and sign certification}{Upper bounds, lower bounds and the matched rate for gain signs.}
\subsection{Gain recovery and adaptive certificates}\label{q3:sec:gain}

Let $\widehat\mu_j=\widehat\mu_{h_j}$ with $h_j=\Pi_\Nn q_j$ and
\[
 \widehat G(w)=G_T(w;f_0)-2\sum_jw_j\widehat\mu_j .
\]
In the unstructured case $h_j=q_j$ and this is the plug-in of
\eqref{q3:eq:linear-forms} with reference $f_0$. Any common learner $T$ is
allowed: no projection or contraction property is used.

\begin{assumption}[Scalar product concentration]\label{q3:ass:subg}
Each $Z_j=(Y-f_0(X))h_j(X)-\mu_{h_j}$ satisfies $\E e^{tZ_j}\le e^{t^2\tau^2/2}$
for all $t\in\R$. Independence across $j$ is not required. For example
$\abs{Y-f_0(X)}\le B_0$ and $\abs{h_j(X)}\le H_0$ give $\tau=B_0H_0$
(Lemma~\ref{q3:lem:hoeffding}); Gaussian label noise alone does not give a common
proxy without control of the multiplying features.
\end{assumption}

\begin{theorem}[Uniform deviation and adaptive sign certificates]\label{q3:thm:uniform}
Under Assumption~\ref{q3:ass:subg}, for $L\ge1$ and $0<\delta<1$, with
probability at least $1-\delta$,
\[
 \boxed{\sup_{w\in\WL}\abs{\widehat G(w)-G_T(w;f)}\le a_n:=2L\tau\sqrt{\frac{2\log(2m/\delta)}n}}
\]
(the simplex is $L=1$). On the same event, simultaneously for all
$w\in\WL$, $\widehat G(w)-a_n>0\Rightarrow G_T(w;f)>0$ and
$\widehat G(w)+a_n<0\Rightarrow G_T(w;f)<0$; this holds for anchor-dependent
weights. If every $h_j=0$, $\widehat G=G_T$ exactly and no anchors are needed.
\end{theorem}
\begin{proof}
By independence, $\E\exp(t(\widehat\mu_j-\mu_{h_j}))\le e^{t^2\tau^2/(2n)}$; the
exponential Markov bound at $t=nu/\tau^2$ and the same for $-Z_j$ give
$\Prob(\abs{\widehat\mu_j-\mu_{h_j}}>u)\le2e^{-nu^2/(2\tau^2)}$. A union bound over
$j$ with $u=\tau\sqrt{2\log(2m/\delta)/n}$ and, by \eqref{q3:eq:residual-gain},
$\abs{\widehat G(w)-G_T(w;f)}=2\abs{\sum_jw_j(\widehat\mu_j-\mu_{h_j})}\le2\norm w_1\max_j\abs{\widehat\mu_j-\mu_{h_j}}$
prove the bound. The certificates hold on this one event, which does not
depend on how $w$ is chosen.
\end{proof}

\begin{corollary}[Margin, sign consistency and label budget]\label{q3:cor:sign}
On a set $\mathcal W'\subseteq\WL$ with $\abs{G_T(w;f)}\ge\gamma>0$, with
probability $\ge1-\delta$ all plug-in signs are correct if
$n>8L^2\tau^2\gamma^{-2}\log(2m/\delta)$, and the confidence rule returns a
nonzero, correct certificate for every weight if
$n>32L^2\tau^2\gamma^{-2}\log(2m/\delta)$. Along a sequence of problems, sign
consistency holds whenever $n\gamma^2/(L^2\tau^2\log(2m/\delta_n))\to\infty$ with
$\delta_n\to0$.
\par{\upshape\small Immediate from \cref{q3:thm:uniform}: $a_n<\gamma$ prevents crossing zero, and $2a_n<\gamma$ gives $\widehat G-a_n\ge\gamma-2a_n>0$ for positive gains (symmetrically for negative ones).}
\end{corollary}

\begin{proposition}[Unrestricted affine uniformity needs care]\label{q3:prop:affine-unif}
With $\varepsilon=\widehat\mu-\mu$: if $J\varepsilon\ne0$ then
$\sup_{w\in\Waff}\abs{\widehat G(w)-G_T(w;f)}=\infty$; if $J\varepsilon=0$,
$\varepsilon=c\one$ and the supremum is $2\abs c$.
\end{proposition}
\begin{proof}
$w_t=\one/m+tJ\varepsilon$ is affine with
$\varepsilon^\top w_t=\varepsilon^\top\one/m+t\norm{J\varepsilon}^2$; if $J\varepsilon=0$,
$\varepsilon^\top w=c$ on $\Waff$.
\end{proof}

\begin{remark}[Selected weights and post-selection validity]\label{q3:auto:selected-weights-and-post-selection-vali}
The certificate is valid for a weight chosen with the same anchors only
because the event is uniform over $\WL$. A pointwise interval at the selected
weight is not: with $32$ zero gains, the naive pointwise rule loses coverage,
whereas the uniform (Bonferroni) rule and an independent confirmation split
keep it (post-selection experiment, \cref{app:experiments}). Positive certification of a selected weight does not
imply that its student beats the best single supervisor (\cref{q1:sec:benchmarks}).
\end{remark}

\subsection{Simultaneous signs: the geometry of the unresolved queries}\label{q3:sec:width}

Fix $n_q$ gain queries $w^{(1)},\dots,w^{(n_q)}$ (e.g.\ the vertices), with
unresolved directions $h_{(i)}=\Pi_\Nn q_{w^{(i)}}$, and an isometry
$\Phi_U:\R^d\to\Span\{h_{(i)}\}$ with coordinate functions $\psi$. Put
$v_i=\Phi_U^*h_{(i)}$, $\vartheta=\Phi_U^*\bar f$ and
$b_i=G_T(w^{(i)};f_0)$, so that $G_i=b_i-2v_i^\top\vartheta$ and
$\widehat G_i=b_i-2v_i^\top\widehat\vartheta$, with
$\widehat\vartheta=n^{-1}\sum_i(Y_i-f_0(X_i))\psi(X_i)$ in the passive case.

For compact $\mathbb A\subset\R^d$ containing $0$ let
$\gw(\mathbb A)=\E\sup_{a\in\mathbb A}G^\top a$ (Gaussian width) and
$\rad(\mathbb A)=\sup_{a\in\mathbb A}\norm a$, $G\sim N(0,I_d)$.

\begin{lemma}[Gaussian supremum]\label{q3:lem:gw}
For every $t>0$,
\[
 \Prob\Bigl\{\sup_{\mathbb A}G^\top a>\gw(\mathbb A)+\rad(\mathbb A)\sqrt{2t}\Bigr\}\le e^{-t},
\]
and for $\mathbb A=\{0,\pm a_1,\dots,\pm a_J\}$, $\gw(\mathbb A)\le\rad(\mathbb A)\sqrt{2\log(2J)}$.
\end{lemma}
\begin{proof}
$F(g)=\sup_{\mathbb A}g^\top a$ is $\rad(\mathbb A)$-Lipschitz. The Gaussian
logarithmic Sobolev inequality $\operatorname{Ent}(e^H)\le\tfrac12\E[e^H\norm{\nabla H}^2]$
with $H=uF$ gives, for $L_F(u)=\log\E e^{uF}$,
$uL_F'-L_F\le u^2\rad^2/2$, i.e.\ $(L_F/u)'\le\rad^2/2$; integrating from
$L_F(u)/u\to\E F$ as $u\downarrow0$ yields
$\log\E e^{u(F-\E F)}\le u^2\rad^2/2$, and Chernoff gives the tail (smooth
approximation covers the Lipschitz $F$; $\abs F\le\rad\norm G$ gives
integrability). For the finite set,
$e^{u\E\max_j\abs{G^\top a_j}}\le\sum_j\E(e^{uG^\top a_j}+e^{-uG^\top a_j})\le2Je^{u^2\rad^2/2}$;
optimize $\log(2J)/u+u\rad^2/2$.
\end{proof}

For the sub-Gaussian experiment, let $\varepsilon_\ell=2^{-\ell}\rad(\mathbb A)$,
choose $\varepsilon_\ell$-nets $\mathbb A_\ell\subseteq\mathbb A$ of sizes $N_\ell$
($\mathbb A_0=\{0\}$, $N_0=1$) and set
\[
 \mathfrak H_t(\mathbb A)=\sum_{\ell\ge1}3\varepsilon_\ell\sqrt{2[t+\log(2N_\ell N_{\ell-1})+2\log(\ell+1)]},
\]
finite for bounded finite-dimensional sets; $\mathfrak H_t(\{0\})=0$.

\begin{lemma}[Dyadic entropy bound]\label{q3:lem:entropy}
If $\epsilon$ has proxy $\Omega=I$, then
$\Prob\{\sup_{\mathbb A}a^\top\epsilon>\mathfrak H_t(\mathbb A)/\sqrt n\}\le e^{-t}$; for
general $\Omega$ replace $\mathbb A$ by $\Omega^{1/2}\mathbb A$.
\end{lemma}
\begin{proof}
For fixed $v$, $\Prob\{\abs{v^\top\epsilon}>\norm v\sqrt{2s/n}\}\le2e^{-s}$. Chain
$\pi_\ell(a)\in\mathbb A_\ell$ with $\norm{a-\pi_\ell(a)}\le\varepsilon_\ell$,
$\pi_0=0$; increments have norm $\le3\varepsilon_\ell$ (for $\ell=1$ because
$\norm{\pi_1(a)}\le\rad\le3\rad/2$). At most $N_\ell N_{\ell-1}$ increments
occur at level $\ell$; with $s_\ell=t+\log(2N_\ell N_{\ell-1})+2\log(\ell+1)$ the
level fails with probability $\le e^{-t}/(\ell+1)^2$, and
$\sum_{\ell\ge1}(\ell+1)^{-2}<1$. On the good event
$a^\top\epsilon=\sum_\ell(\pi_\ell(a)-\pi_{\ell-1}(a))^\top\epsilon\le n^{-1/2}\sum_\ell3\varepsilon_\ell\sqrt{2s_\ell}$.
For singular $\Omega$, $\epsilon$ has no component in $\Ker\Omega$ a.s.; write
$a^\top\epsilon=(\Omega^{1/2}a)^\top\Omega^{\dagger/2}\epsilon$.
\end{proof}

\begin{theorem}[Simultaneous sign upper bounds]\label{q3:thm:sign-upper}
Let $\Theta_\gamma=\{\vartheta:\abs{G_i}\ge\gamma_i\ \forall i\}$ and
$\mathbb A_\gamma=\{0,\pm2\Omega^{1/2}v_i/\gamma_i:i\le n_q\}$. In the Gaussian
moment experiment all signs are correct with probability $\ge1-\delta$ on
$\Theta_\gamma$ if
\[
 \boxed{\sqrt n>\gw(\mathbb A_\gamma)+\rad(\mathbb A_\gamma)\sqrt{2\log(1/\delta)}};
\]
in the sub-Gaussian experiment it suffices that $\sqrt n>\mathfrak H_{\log(1/\delta)}(\mathbb A_\gamma)$.
A sufficient worst-case Gaussian budget is
$\sqrt n>\rad(\mathbb A_\gamma)[\sqrt{2\log(2n_q)}+\sqrt{2\log(1/\delta)}]$: $\log n_q$
replaces the actual normalized geometry.
\end{theorem}
\begin{proof}
$\abs{\widehat G_i-G_i}/\gamma_i=2\abs{v_i^\top\epsilon}/\gamma_i$, so in the Gaussian
experiment $\max_i\abs{\widehat G_i-G_i}/\gamma_i=n^{-1/2}\sup_{\mathbb A_\gamma}G^\top a$;
Lemma~\ref{q3:lem:gw} makes it $<1$, and since $\abs{G_i}\ge\gamma_i$ no sign changes.
Lemma~\ref{q3:lem:entropy} gives the sub-Gaussian case; Lemma~\ref{q3:lem:gw} the
cardinality bound.
\end{proof}

\begin{proposition}[Duplicates and clusters]\label{q3:prop:clusters}
Duplicates in $\mathbb A_\gamma$ change neither width, radius nor covering
numbers. If $a_j=c_{\ell(j)}+\epsilon_j$ with $L_c$ centers, $\norm{c_\ell}\le R_0$
and $\norm{\epsilon_j}\le\varepsilon$, then
$\gw(\{0,\pm a_j\})\le R_0\sqrt{2\log(2L_c)}+\varepsilon\sqrt{2\log(2n_q)}$; for one
cluster the first term can be replaced by $\sqrt{2/\pi}\norm{c_1}$.
\end{proposition}
\begin{proof}
$\max_j\abs{G^\top a_j}\le\max_\ell\abs{G^\top c_\ell}+\max_j\abs{G^\top\epsilon_j}$;
Lemma~\ref{q3:lem:gw} and $\E\abs{G^\top c}=\sqrt{2/\pi}\norm c$.
\end{proof}

\subsection{Sign lower bounds}\label{q3:sec:sign-lower}

\begin{theorem}[Two-point and packing lower bounds]\label{q3:thm:sign-lower}
Fix a dictionary, design and admissible target class, with all unlabeled and
structural information identical across alternatives, in the passive Gaussian
experiment with noise $\sigma$. If two admissible targets $f,h$ have different
sign patterns and a procedure recovers the full pattern with error
$\le\delta<1/4$ at both, then $n\ge2\sigma^2\norm{f-h}^{-2}\log(1/(4\delta))$. If
$f_1,\dots,f_M$ have distinct patterns and $\norm{f_j-f_{\mathrm{ref}}}\le r_M$,
error $\le\delta$ at every alternative requires
$n\ge2\sigma^2r_M^{-2}[(1-\delta)\log M-\log2]$. The same holds in the isotropic
Gaussian moment experiment with Euclidean distances.
\end{theorem}
\begin{proof}
\emph{Strategy.} A procedure that recovers the sign pattern also identifies
which alternative generated the data, so its error is bounded below by a
testing error: Le Cam's two-point bound for one pair, Fano's inequality over
a packing of $M$ alternatives within radius $r_M$. Gaussian KL divergences
convert both into label counts.

Pattern recovery induces a test between any two alternatives; use
Lemmas~\ref{q3:lem:gausskl} and~\ref{q3:lem:testing}. For the family, decode the
index and use Lemma~\ref{q3:lem:fano} with $\KL(\Prob_j\Vert\Prob_{\mathrm{ref}})\le nr_M^2/(2\sigma^2)$.
\end{proof}

The distance from $f$ to the boundary $\{G_i=0\}$ is
$\abs{G_i(f)}/(2\norm{h_{(i)}})$ (attained by
$\Delta=G_i(f)h_{(i)}/(2\norm{h_{(i)}}^2)$), but a structured class can exclude
that boundary point: a large width alone does not give a lower bound;
admissible alternatives must exist.

Both ends of the sign complexity are attained by one family of
dictionaries, in which every residual query is one of $r$ orthogonal
Rademacher coordinates.

\begin{definition}[Rademacher query dictionaries]\label{q3:def:rad-dict}
Let $u,v_1,\dots,v_r$ be independent Rademacher functions, $P$ the projection
onto $\Span\{u\}$, and $\pi:\{1,\dots,m\}\to\{1,\dots,r\}$ onto. Put
$g_j=c_ju+v_{\pi(j)}$ with $\abs{c_j}\ge1$ and the pairs $(c_j,\pi(j))$
distinct, so $s_j=c_ju$ and $q_j=v_{\pi(j)}$. For a target $f$ in
$\Span\{v_1,\dots,v_r\}$ write $\theta_\ell=\ip f{v_\ell}$.
\end{definition}

\begin{lemma}[Reduction to $r$ coordinates]\label{q3:lem:rad-reduction}
In a Rademacher query dictionary,
$G(\mathbf e_j;f)=1-2\theta_{\pi(j)}$ and
$R_W(\mathbf e_j)-A(f)=c_j^2+G(\mathbf e_j;f)$. Hence the full sign pattern of
the $m$ supervisors is a function of the $r$ signs of $1-2\theta_\ell$;
whenever every gain exceeds $-1$, the clean student beats every
supervisor; and with reference $f_0$, the plug-in
$\widehat G_j=G(\mathbf e_j;f_0)-\tfrac2n\sum_i(Y_i-f_0(X_i))v_{\pi(j)}(X_i)$ is
unbiased.
\end{lemma}
\begin{proof}
$\norm{g_j}^2-\norm{s_j}^2=\norm{v_{\pi(j)}}^2=1$ and $\ip f{q_j}=\theta_{\pi(j)}$
give the gain; $R_W(\mathbf e_j)=c_j^2+\norm{f-v_{\pi(j)}}^2$ and $A(f)=\norm f^2$
give the second identity; $c_j^2\ge1>-G$ gives the third claim; and
$G(\mathbf e_j;f)-G(\mathbf e_j;f_0)=-2\ip{f-f_0}{v_{\pi(j)}}$ gives unbiasedness.
\end{proof}

\begin{theorem}[The two extremes of sign complexity]\label{q3:thm:extremes}\label{q3:thm:easy}\label{q3:thm:hard}
Consider a Rademacher query dictionary with Gaussian anchor noise $\sigma$ and
margin $0<\gamma\le\tfrac12$.
\begin{enumerate}
\item[(i)] \emph{Duplicated queries} ($r=1$, any $m$). With $f_\pm=\tfrac{1\mp\gamma}2v_1$
all gains equal $\pm\gamma$. For $f_0=v_1/2$, $\widehat G_j-G\sim N(0,4\sigma^2/n)$,
so the sign error is $\Phi(-\gamma\sqrt n/(2\sigma))\le e^{-n\gamma^2/(8\sigma^2)}$;
conversely every sign estimator has worst-world error
$\ge\tfrac14e^{-n\gamma^2/(2\sigma^2)}$, so error $\delta<\tfrac14$ needs
$n\ge2\sigma^2\gamma^{-2}\log(1/(4\delta))$. The cost is
$\sigma^2\gamma^{-2}\log(1/\delta)$, independent of $m$.
\item[(ii)] \emph{Independent queries} ($r=m$, $\pi=\mathrm{id}$, $c_j=1$). With
$f_0=\tfrac{1-\gamma}2\sum_jv_j$ and $f_\ell=f_0+\gamma v_\ell$,
$G(\mathbf e_j;f_\ell)=-\gamma$ if $j=\ell$ and $+\gamma$ otherwise, and
$A(f_\ell)=[(1+\gamma)^2+(m-1)(1-\gamma)^2]/4$ is the same for all $\ell$.
Recovering all signs with worst-world error $\le\delta$ needs
$n\ge2\sigma^2\gamma^{-2}[(1-\delta)\log m-\log2]$, while
$n>8(\sigma^2+\gamma^2)\gamma^{-2}\log(2m/\delta)$ suffices for the plug-in. For
$\gamma\le\sigma$ and fixed $\delta$ both are of order $\sigma^2\gamma^{-2}\log m$.
\end{enumerate}
\end{theorem}
\begin{proof}
\emph{Shared reduction.} By Lemma~\ref{q3:lem:rad-reduction} the gains are
$1-2\theta_\ell$, the alternatives differ only in $\theta$, and recovering the
sign pattern identifies the alternative; Theorem~\ref{q3:thm:sign-lower} then
gives the lower bounds, with $\norm{f_+-f_-}=\gamma$ (two points) in (i) and
$\norm{f_\ell-f_0}=\gamma$ with $M=m$ distinct patterns (packing) in (ii).
\emph{Upper bounds.} $(Y-f_0)v_{\pi(j)}=\ip{f-f_0}{v_{\pi(j)}}+\rho_j+\xi v_{\pi(j)}$,
where $\rho_j$ is the centered part of $(f-f_0)v_{\pi(j)}$. In
(i) $f-f_0=\mp\tfrac\gamma2v_1$, so $\rho_j=0$ and the estimate is exactly
Gaussian with variance $4\sigma^2/n$. In (ii) $f-f_0=\gamma v_\ell$, so
$\rho_j$ is $0$ for $j=\ell$ and a centered $\pm\gamma$ variable otherwise
($\cosh(t\gamma)\le e^{t^2\gamma^2/2}$); with the Gaussian term the proxy is
$\tau^2=\sigma^2+\gamma^2$, and the scalar tail of Theorem~\ref{q3:thm:uniform} at
$\gamma/2$ with a union bound over $j$ gives the sufficient budget.
\end{proof}

\begin{corollary}[Order of simultaneous sign complexity]\label{q3:cor:sign-order}
For $L=1$, $m\ge16$, $0<\gamma\le\min\{\tfrac12,\sigma\}$ and $0<\delta\le1/16$, the
worst-case noisy label complexity of simultaneous supervisor-gain signs over
scalar-product classes with $\tau^2\le2\sigma^2$ is of order
$\sigma^2\gamma^{-2}(\log m+\log(1/\delta))$: the upper bound is
Corollary~\ref{q3:cor:sign}, the two necessary terms are
Theorem~\ref{q3:thm:extremes}(i) and (ii); the larger of two
necessary terms is at least half their sum. This is a worst case over
dictionaries: duplicated queries cost no $\log m$, and
Theorem~\ref{q3:thm:sign-upper} interpolates through the width. Optimal
dependence on $L>1$ is not established.
\end{corollary}

\apppart{Branch II: selection and regret}{Local-cone regret, the unresolved trace and matching lower bounds.}
\subsection{Selection: residualized coordinates and local-cone regret}\label{q3:sec:selection}

\paragraph{Reduced oracle.} If $k=\dim V=0$ all candidate students coincide
and every regret below is zero; assume $k\ge1$. Let $\phi_\ell=\Phi\mathbf
e_\ell$ ($\ell\le k$) be a known orthonormal basis of $V=\Span\{s_j-s_m\}$,
$t_j=\Phi^*s_j$, $\theta=\Phi^*f$, $\theta_0=\Phi^*f_0$, and
$a_\circ=(I-\Pi_V)s_m$ (common to all $s_j$). For a weight domain
$\mathcal W\in\{\Waff,\Delta_m\}$ or any nonempty compact convex
$\mathcal W\subseteq\WL$ put $\mathcal C=\{t(w)=\sum_jw_jt_j:w\in\mathcal W\}$
($\mathcal C=\R^k$ for $\Waff$, $\conv\{t_j\}$ for $\Delta_m$).

\begin{proposition}[Reduced oracle]\label{q3:prop:reduced}
For every affine $w$, $R_T(w;f)=\norm{a_\circ-(I-\Pi_V)f}^2+\norm{t(w)-\theta}^2$.
The optimal prediction coordinate is $t^\star=\Pi_{\mathcal C}\theta$, and for
$w\in\mathcal W$, $\Reg_{\mathcal W}(w)=\norm{t(w)-t^\star}^2+2\ip{t(w)-t^\star}{t^\star-\theta}$
with nonnegative second term. For affine weights and $T=P$,
$\Reg_{\Waff}(w)=\norm{t(w)-\theta}^2=(w-w^\star)^\top\Gpar(w-w^\star)$ for any
optimal $w^\star$.
\end{proposition}
\begin{proof}
$s_w-f=\Phi(t(w)-\theta)+[a_\circ-(I-\Pi_V)f]$ is an orthogonal decomposition
and $\Phi$ is an isometry. The projection variational inequality
$\ip{t^\star-\theta}{t-t^\star}\ge0$ ($t\in\mathcal C$) gives the regret formula.
For $T=P$ and $d=w-w^\star$, $\one^\top d=0$, so
$\sum_jd_jPe_j=s_w-s_{w^\star}=\Phi(t(w)-\theta)$.
\end{proof}

\paragraph{Residualized estimator.} Put $\chi_\ell=\Pi_\Nn\phi_\ell$,
$\chi=(\chi_1,\dots,\chi_k)^\top$, $H_\Nn=\Phi^*\Pi_\Nn\Phi=(\ip{\chi_\ell}{\chi_b})$ and
\[
 d_\Nn=\tr H_\Nn=\sum_{\ell}\norm{\chi_\ell}^2=\tr(\Pi_V\Pi_\Nn\Pi_V)\in[0,\dim\Pi_\Nn V],
\]
the sum of squared singular values of $\Pi_\Nn|_V$ (it may be fractional;
$d_\Nn=k$ when $\Nn=\Hh$). Estimate
\[
 \widehat\theta=\theta_0+\frac1n\sum_{i=1}^n(Y_i-f_0(X_i))\chi(X_i),\qquad
 \widehat t=\Pi_{\mathcal C}\widehat\theta,
\]
and choose any measurable $\widehat w\in\mathcal W$ with $t(\widehat w)=\widehat t$
(e.g.\ minimum Euclidean norm in the fiber); the risk does not depend on the
choice. Under Assumption~\ref{q3:ass:residual},
$\ip f{\phi_\ell}=\ip{f_0}{\phi_\ell}+\ip{\bar f}{\chi_\ell}$, so $\E\widehat\theta=\theta$.
With $\Nn=\Hh$ this is the plain moment estimator; with structure it samples
only the unresolved components.

\begin{lemma}[Projection facts]\label{q3:lem:proj}
For nonempty closed convex $C$: $p=\Pi_Cz$ iff $\ip{z-p}{v-p}\le0$ for all
$v\in C$; $\Pi_C$ and $I-\Pi_C$ are nonexpansive. For a closed convex cone
$K$: $\sup_{v\in K}\{2\ip zv-\norm v^2\}=\norm{\Pi_Kz}^2$ and
$\sup_{v\in K,\norm v\le1}\ip zv=\norm{\Pi_Kz}$.
\end{lemma}
\begin{proof}
The characterization follows by differentiating $\norm{z-p-t(v-p)}^2$ at
$t=0^+$ and, conversely, from
$\norm{z-v}^2=\norm{z-p}^2+\norm{v-p}^2-2\ip{z-p}{v-p}$. Adding the
inequalities for $p=\Pi_Cz$, $q=\Pi_Cy$ gives $\norm{p-q}^2\le\ip{p-q}{z-y}$,
hence both nonexpansiveness claims (expand
$\norm{(z-p)-(y-q)}^2$). For a cone, $v=0$ and $v=2p$ give $\ip{z-p}p=0$, and
then $\ip{z-p}v\le0$, so $\ip zv\le\ip pv\le\norm p\norm v$ with equality at
$v=p/\norm p$; and $2\ip zv-\norm v^2=\norm z^2-\norm{z-v}^2$ is maximized at
$v=p$ with value $\norm p^2$.
\end{proof}

Let $\mathcal T=\mathcal T_{\mathcal C}(t^\star)=\overline{\cone(\mathcal C-t^\star)}$ be
the tangent cone at the oracle and, for $\Omega\succeq0$,
$\sd_\Omega(\mathcal T)=\E\norm{\Pi_{\mathcal T}(\Omega^{1/2}G)}^2$ ($\sd_I$ is the
statistical dimension \citep{amelunxen2014living}).

\begin{theorem}[Deterministic local-cone regret]\label{q3:thm:cone}
With $\varepsilon=\widehat\theta-\theta$,
\[
 \boxed{0\le\Reg_{\mathcal W}(\widehat w)\le\norm{\Pi_{\mathcal T}\varepsilon}^2\le\norm\varepsilon^2,}
\]
with $\Reg_{\Waff}(\widehat w)=\norm\varepsilon^2$ for affine weights.
\end{theorem}
\begin{proof}
\emph{Strategy.} Compare the projected estimate with the oracle point through
the variational inequality of the projection; the regret is then a concave
quadratic in the displacement $d$, which lies in the tangent cone, and its
supremum over the cone is the squared projection of the error onto the cone.

Let $d=\widehat t-t^\star\in\mathcal T$. The variational inequality for
$\widehat t$ at $t^\star$ is $\ip{\widehat t-\theta-\varepsilon}d\le0$. Hence
$\Reg=\norm{\widehat t-\theta}^2-\norm{t^\star-\theta}^2=2\ip{\widehat t-\theta}d-\norm d^2
\le2\ip\varepsilon d-\norm d^2\le\sup_{v\in\mathcal T}\{2\ip\varepsilon v-\norm v^2\}=\norm{\Pi_{\mathcal T}\varepsilon}^2$
by Lemma~\ref{q3:lem:proj}; the last inequality is nonexpansiveness with $\Pi_{\mathcal T}0=0$.
For $\mathcal C=\R^k$, $\widehat t=\widehat\theta$ and $t^\star=\theta$.
\end{proof}

\begin{theorem}[Expected selection regret]\label{q3:thm:expected}
If $\E[(Y-f_0(X))^2\mid X]\le v_0^2$ a.s., then for every domain above
\[
 \E\Reg_{\mathcal W}(\widehat w)\le\frac1n\Big(\E\big[(Y-f_0(X))^2\norm{\chi(X)}^2\big]-\norm{\theta-\theta_0}^2\Big)
 \le\boxed{\frac{v_0^2d_\Nn}n}.
\]
If $d_\Nn=0$ the exact optimal decision is obtained without anchors; $n\ge v_0^2d_\Nn/\epsilon$
gives expected regret $\le\epsilon$, and $n\ge v_0^2d_\Nn/(\epsilon\delta)$ gives regret
$\le\epsilon$ with probability $\ge1-\delta$ (Markov).
\end{theorem}
\begin{proof}
$Z=(Y-f_0(X))\chi(X)$ has mean $\theta-\theta_0$, and independent anchors give
$\E\norm\varepsilon^2=n^{-1}(\E\norm Z^2-\norm{\E Z}^2)$, with
$\E\norm Z^2\le v_0^2\E\norm{\chi(X)}^2=v_0^2d_\Nn$. Apply Theorem~\ref{q3:thm:cone}.
\end{proof}

\begin{remark}[What $v_0$ contains; relation to ERM and stacking]\label{q3:rem:v0}
$v_0^2$ bounds the conditional second moment of the \emph{residual label}
$Y-f_0(X)$, i.e.\ noise variance plus $(f-f_0)^2$: the signal not explained
by the reference is paid as variance. The moment estimator is unbiased but is
not the least-squares (stacking) estimator, which regresses $Y$ on the
candidate predictions and pays only the noise once the design is well
conditioned. This explains why, in the selection experiments of \cref{app:experiments}, ERM/stacking beats the
moment estimator when $\norm{f-f_0}$ is large, while the moment estimator's
$d_\Nn$ advantage appears when structure removes most of $V$. A good reference
($f_0=f_\Kk$, or a pilot fit on independent anchors) reduces $v_0$; the
paired-anchor and regularized versions of \cref{app:q4} address the same issue, and
residual regression (Section~\ref{q3:sec:compare}) removes the explained-signal
term altogether when the residual is linear in the unresolved features. The
$k/n$ rate itself is classical for quadratic aggregation; the contribution is
that the relevant dimension is the \emph{unresolved} trace $d_\Nn$, not $m$
or $\rank\Gpar$.
\end{remark}

\begin{assumption}[Directional sub-Gaussianity]\label{q3:ass:vector}
$\E\exp(u^\top(Z-\E Z))\le e^{\nu^2\norm u^2/2}$ for all $u\in\R^k$, where
$Z=(Y-f_0(X))\chi(X)$. For instance $\abs{Y-f_0(X)}\le B_0$ and
$\norm{\chi(X)}\le L_\chi$ give $\nu=B_0L_\chi$, which may grow with $k$; it does
not follow from second moments.
\end{assumption}

\begin{theorem}[High-probability selection regret]\label{q3:thm:hp-select}
Under Assumption~\ref{q3:ass:vector}, with probability $\ge1-\delta$,
$\norm\varepsilon\le2\nu\sqrt{2(k\log5+\log(2/\delta))/n}$ and
$\Reg_{\mathcal W}(\widehat w)\le8\nu^2(k\log5+\log(2/\delta))/n$. Moreover, with the
same probability, $\Reg_{\mathcal W}(\widehat w)\le n^{-1}\mathfrak H_{\log(1/\delta)}(\nu(\mathcal T\cap B_2))^2$.
In the Gaussian moment experiment $\widehat\theta=\theta+n^{-1/2}\Omega^{1/2}G$,
\[
 \E\Reg\le\frac{\sd_\Omega(\mathcal T)}n,\qquad
 \Reg\le\frac1n\Big[\sqrt{\sd_\Omega(\mathcal T)}+\sqrt{2\norm\Omega_\op\log(1/\delta)}\Big]^2
 \ \text{w.p.}\ \ge1-\delta .
\]
For affine weights $\sd_\Omega(\R^k)=\tr\Omega$.
\end{theorem}
\begin{proof}
Lemma~\ref{q3:lem:net} gives a $\tfrac12$-net $\mathcal M$ of the sphere with
$\abs{\mathcal M}\le5^k$ and $\norm x\le2\max_{a\in\mathcal M}\abs{a^\top x}$; each
$a^\top\varepsilon$ has proxy $\nu^2/n$; a union bound and
Theorem~\ref{q3:thm:cone} give the first claim. By Lemma~\ref{q3:lem:proj},
$\norm{\Pi_{\mathcal T}\varepsilon}=\sup_{v\in\mathcal T\cap B_2}v^\top\varepsilon$; apply
Lemma~\ref{q3:lem:entropy}. In the Gaussian experiment positive homogeneity
gives $\norm{\Pi_{\mathcal T}\varepsilon}^2=n^{-1}\norm{\Pi_{\mathcal T}\Omega^{1/2}G}^2$; the map
$G\mapsto\norm{\Pi_{\mathcal T}\Omega^{1/2}G}$ is $\norm\Omega_\op^{1/2}$-Lipschitz with
mean $\le\sd_\Omega^{1/2}$, so the Gaussian concentration of
Lemma~\ref{q3:lem:gw}'s proof applies.
\end{proof}

\begin{proposition}[Exact vertex selection]\label{q3:prop:vertex-exact}
Let $\mathcal C=\conv\{t_1,\dots,t_m\}$, $t^\star=t_1$; discard vertices equal to
$t_1$ and set $d_j=t_j-t_1$, $c_j=\ip{t_1-\theta}{d_j}$. If every $c_j>0$, then
$\Pi_{\mathcal C}(\theta+\varepsilon)=t_1$ iff $\varepsilon^\top d_j\le c_j$ for all $j$. In the
Gaussian experiment, with $\mathbb A_\star=\{0,\Omega^{1/2}d_j/c_j\}$, the regret
is exactly zero with probability $\ge1-\delta$ whenever
$\sqrt n>\gw(\mathbb A_\star)+\rad(\mathbb A_\star)\sqrt{2\log(1/\delta)}$. The relevant
margins are the normal margins $c_j$; the discrete gaps
$\norm{t_j-\theta}^2-\norm{t_1-\theta}^2=\norm{d_j}^2+2c_j$ can be positive with
$c_j\le0$.
\end{proposition}
\begin{proof}
The variational inequality at $t_1$ is linear in $v$, so it suffices to check
vertices; a violation requires $\sup_{\mathbb A_\star}G^\top a>\sqrt n$;
Lemma~\ref{q3:lem:gw}.
\end{proof}

Duplicated candidates change neither $\mathcal C$ nor $\mathcal T$; the cone
depends on the dictionary \emph{and} on the oracle location, so the same
dictionary can be easy at one target and hard at another
(Section~\ref{q3:sec:select-lower}).

\subsection{Selection lower bounds}\label{q3:sec:select-lower}

\begin{lemma}[A simplex containing a cube]\label{q3:lem:cube}
For $c\in\R^k_{>0}$ let $t_0=-c$ and $t_j=-c+2kc_j\mathbf e_j$ ($j\le k$). These
$k+1$ points are affinely independent and
$\conv\{t_0,\dots,t_k\}\supseteq\prod_\ell[-c_\ell,c_\ell]$.
\end{lemma}
\begin{proof}
For $x$ in the box, $w_j=(x_j+c_j)/(2kc_j)\in[0,1/k]$ and $w_0=1-\sum_jw_j\ge0$
give $\sum_jw_jt_j=x$.
\end{proof}

\begin{theorem}[Feasible packing lower bound]\label{q3:thm:packing}
In the isotropic Gaussian moment experiment with noise $\sigma$, if
$\theta_1,\dots,\theta_M\in\mathcal C$ are $2s$-separated and within $r$ of a
reference, every estimator satisfies
$\max_j\E_{\theta_j}\Reg\ge s^2[1-(nr^2/(2\sigma^2)+\log2)/\log M]$. The same holds
for passive Gaussian anchors when admissible targets have these coordinates and
squared $L^2$ distance $\le r^2$ to a common reference.
\end{theorem}
\begin{proof}
Feasibility makes the oracle risk zero, so the regret is
$\norm{\widehat t-\theta_j}^2$. Decoding the nearest packing point errs only if
the loss is $\ge s^2$; Lemmas~\ref{q3:lem:gausskl} and~\ref{q3:lem:fano}.
\end{proof}

\begin{theorem}[Unresolved-angle lower bound]\label{q3:thm:angle}
Let $\nu_1,\dots,\nu_r>0$ be the nonzero singular values of $\Pi_\Nn|_V$ with
singular pairs $v_i\in V$, $z_i\in\Nn$ orthonormal,
$\Pi_\Nn v_i=\nu_iz_i$, $\Pi_Vz_i=\nu_iv_i$, so $d_\Nn=\sum_i\nu_i^2$. Assume
$M_z=\operatorname*{ess\,sup}_x\sum_i\abs{z_i(x)}<\infty$, fix $\rho,\sigma>0$ and
$n\ge\max\{1,M_z^2/32,\sigma^2r/(32\rho^2)\}$. In the Gaussian-noise anchor
experiment there are targets with $f-f_0\in\Nn$, $\norm{f-f_0}\le\rho$ and
$\E[(Y-f_0(X))^2\mid X]\le2\sigma^2$ such that every affine-weight estimator has
\[
 \boxed{\sup_f\E_f\Reg_{\Waff}(\widehat w;f)\ge\frac{3\sigma^2d_\Nn}{256n}.}
\]
The same holds for the simplex when all target coordinates used below are
feasible.
\end{theorem}
\begin{proof}
\emph{Strategy.} An Assouad-type hypercube argument: perturb the target along
the unresolved singular directions $z_i$ with signs $\omega\in\{\pm1\}^r$,
so that each wrong sign costs regret proportional to $\nu_i^2$ while
neighboring hypotheses stay statistically close; summing the per-coordinate
testing errors gives $d_\Nn=\sum_i\nu_i^2$.

($v_i,\nu_i^2$ diagonalize $\Pi_V\Pi_\Nn|_V$ and $z_i=\Pi_\Nn v_i/\nu_i$.) Let
$a=\sigma/\sqrt{32n}$ and $f_\omega=f_0+a\sum_i\omega_iz_i$, $\omega\in\{\pm1\}^r$.
Then $f_\omega-f_0\in\Nn$, $\norm{f_\omega-f_0}^2=a^2r\le\rho^2$ and
$\abs{f_\omega-f_0}\le aM_z\le\sigma$, giving the second-moment bound.
$\Pi_Vf_\omega=\Pi_Vf_0+a\sum_i\nu_i\omega_iv_i$, and the affine regret is the
squared error of the estimated $V$-coordinate. Decoding $\omega_i$ by the sign
of the estimated coefficient on $v_i$ (after subtracting $\Pi_Vf_0$), a sign error
costs $\ge a^2\nu_i^2$. Neighbors differ by $2az_i$, KL $=1/16$, so the sum of
the two errors is $\ge3/4$ (Lemma~\ref{q3:lem:testing}) and the average error per
coordinate is $\ge3/8$. Summing, the Bayes risk is
$\ge\tfrac38a^2\sum_i\nu_i^2=3\sigma^2d_\Nn/(256n)$.
\end{proof}

\begin{corollary}[Unstructured matching bound]\label{q3:cor:k-lower}
For every $k\ge1$, $b>0$, $\sigma>0$ and $n\ge\max\{k,\sigma^2/(8b^2)\}$ there
is a fixed dictionary with $m=k+1$ distinct supervisors, $\Nn=\Hh$, $f_0=0$,
for which every estimator has $\sup_f\E\Reg_{\mathcal W}\ge3\sigma^2k/(256n)$ for
both affine and simplex weights, the clean student has risk $0$, every
supervisor has risk $\ge1$, and $v_0^2\le\tfrac{33}{32}\sigma^2$. So
Theorem~\ref{q3:thm:expected} is minimax sharp up to constants.
\end{corollary}
\begin{proof}
Let $X=(I,Z)$ with $I$ uniform on $\{1,\dots,k\}$, $Z$ an independent
Rademacher; $\phi_\ell=\sqrt k\one\{I=\ell\}$, $v=Z\perp\Span\{\phi_\ell\}$. With
Lemma~\ref{q3:lem:cube} ($c=b\one$) let $g_j=\Phi t_j+v$ and $P$ the projection
onto $\Span\{\phi_\ell\}$. Take $\psi_\ell=\phi_\ell$, $\nu_\ell=1$ in the proof
of Theorem~\ref{q3:thm:angle}: $a=\sigma/\sqrt{32n}\le b/2$ keeps every
$a\omega$ in the simplex; $R(g_j)=\norm{t_j-a\omega}^2+1$; and
$\E[Y^2\mid X]=ka^2+\sigma^2\le\tfrac{33}{32}\sigma^2$ since $n\ge k$.
\end{proof}

\paragraph{The same simplex: a vertex versus its interior.} Let
$\mathcal C=\conv\{\mathbf e_1,\dots,\mathbf e_m\}\subset\R^m$ in the Gaussian moment
experiment with $\widehat\theta=\theta+\varsigma G$ ($\varsigma^2=\sigma^2/n$); its
contrast dimension is $m-1$.

\begin{proposition}[Logarithmic vertex complexity]\label{q3:prop:vertex}
At $\theta=\mathbf e_1$ the tangent cone
$\mathcal T_1=\{d:\sum_jd_j=0,\ d_j\ge0\ (j\ge2)\}$ has
$\sd_I(\mathcal T_1)\le4\log(2m)+6$, so
$\E\norm{\Pi_{\mathcal C}(\mathbf e_1+\varsigma G)-\mathbf e_1}^2\le\varsigma^2[4\log(2m)+6]$.
For $M=m-1\ge4$ and $a=\varsigma\sqrt{\log M}/4\le1$, every estimator has
$\sup_{\theta\in\mathcal C\cap B(\mathbf e_1,\sqrt2a)}\E\norm{\widehat t-\theta}^2\ge\varsigma^2\log M/128$,
while projection has worst-case risk $\le2\varsigma^2\sd_I(\mathcal T_1)+4a^2=O(\varsigma^2\log m)$
on this neighborhood. At a relative-interior point the cone is the whole
$(m-1)$-dimensional contrast space and, in the local regime where a cube of
side of order $\varsigma$ around the point stays in the simplex, the matching
order is $(m-1)\varsigma^2$; since the simplex has bounded diameter, this
requires $\varsigma$ small relative to the distance to the boundary.
\end{proposition}
\begin{proof}
For $d\in\mathcal T_1$, $\alpha=\sum_{j\ge2}d_j=-d_1\le\norm d$ and
$G^\top d=\sum_{j\ge2}(G_j-G_1)d_j\le\norm d\max_j(G_j-G_1)_+$, so
$\norm{\Pi_{\mathcal T_1}G}\le\max_j\abs{G_j}+\abs{G_1}$ and
$\sd_I\le2\E\max_jG_j^2+2$; $\Prob(\max_jG_j^2>u)\le2me^{-u/2}$ integrates to
$\E\max_jG_j^2\le2\log(2m)+2$. Apply Theorem~\ref{q3:thm:cone}. For the lower
bound take $\theta_j=(1-a)\mathbf e_1+a\mathbf e_j$, $j=2,\dots,m$: pairwise distance
and distance to $\mathbf e_1$ are $\sqrt2a$, so $s=a/\sqrt2$, $r=\sqrt2a$, and the
divergence to $\mathbf e_1$ is $a^2/\varsigma^2=\log M/16$; Theorem~\ref{q3:thm:packing} with
$1-\tfrac1{16}-\log2/\log M\ge\tfrac7{16}$ gives $\tfrac7{32}a^2\ge\varsigma^2\log M/128$.
For the upper bound, couple through $G$: by nonexpansiveness of $I-\Pi_{\mathcal C}$
the error at $\theta$ differs from that at $\mathbf e_1$ by at most
$\norm{\theta-\mathbf e_1}\le\sqrt2a$. For an interior $\theta_c$ use
Theorem~\ref{q3:thm:angle}'s cube in an orthonormal basis of the contrast space,
feasible when $a\sqrt{m-1}\le\min_j(\theta_c)_j$.
\end{proof}

\begin{remark}
The vertex lower bound is local minimax over a shrinking neighborhood, not a
pointwise bound at $\mathbf e_1$ (the constant estimator $\mathbf e_1$ has zero risk
there). It holds in the Gaussian moment experiment; passive anchors need the
bridge of Section~\ref{q3:sec:bridge}.
\end{remark}
\apppart{Branch III: what structure and weak supervision buy}{Residual regression; structured selection against ordinary stacking; weak supervision against anchors alone; approximate structure.}
\subsection{Residual regression and when anchors are worth spending}\label{q3:sec:compare}

The moment estimator of Section~\ref{q3:sec:selection} pays the unexplained
signal through $v_0^2$ (Remark~\ref{q3:rem:v0}). Regressing the residual label on
the unresolved features, and reconstructing the target coordinate with the
\emph{population} Gram, removes that cost when the residual is linear in those
features. This section computes the resulting risk and uses it for three
comparisons, each with an explicitly specified competitor: structured
selection against ordinary stacking, weak supervision with anchors against
anchors alone, and approximate against exact structure. Least squares,
residualization, inverse-Wishart moments \citep{muirhead1982aspects},
ridgeless risk formulas \citep{hastie2022surprises,belkin2020two} and Gaussian
conjugacy are standard; what is new is which comparisons they settle in the
decision geometry of this paper, and under which assumptions.

\paragraph{Setting.} As in Section~\ref{q3:sec:selection}, with
Assumption~\ref{q3:ass:residual}, $\bar f=f-f_0\in\Nn$ and
$H_\Nn=\Phi^*\Pi_\Nn\Phi$. In this section the label noise is homoscedastic,
$\E[\xi^2\mid X]=\sigma^2$, and anchors are independent; heteroscedastic noise
replaces $\sigma^2(Z_A^\top Z_A)^{-1}$ below by the sandwich
$(Z_A^\top Z_A)^{-1}Z_A^\top\diag(\sigma^2(X_i))Z_A(Z_A^\top Z_A)^{-1}$. Let $r=\rank H_\Nn$
and write $H_\Nn=U\diag(\nu_1^2,\dots,\nu_r^2)U^\top$ with $U\in\R^{k\times r}$
orthonormal; the $\nu_i>0$ are the nonzero singular values of $\Pi_\Nn|_V$, so
$d_\Nn=\sum_i\nu_i^2$. Put
\[
 z(X)=\diag(\nu)^{-1}U^\top\chi(X)\in\R^r,\qquad \mathsf B=U\diag(\nu).
\]
Then $\E[zz^\top]=I_r$, $\chi=\mathsf Bz$ almost surely (a direction $a\perp\Ran H_\Nn$
has $\E(a^\top\chi)^2=a^\top H_\Nn a=0$), $\mathsf B\mathsf B^\top=H_\Nn$ and
$\tr(\mathsf B^\top\mathsf B)=d_\Nn$. The coordinates $z_i$ are the left singular
functions of $\Pi_\Nn|_V$ (the $z_i$ of Theorem~\ref{q3:thm:angle}). Decompose the
hidden target into its part in their span and the rest,
\[
 \bar f=z^\top\beta+\bar f_\perp,\qquad \beta_i=\ip{z_i}{\bar f},\quad \bar f_\perp\perp\Span\{z_i\}.
\]
Since $\bar f\in\Nn$, $\ip{\phi_\ell}{\bar f}=\ip{\chi_\ell}{\bar f}$, hence
$\theta=\theta_0+\E[\chi\bar f]=\theta_0+\mathsf B\beta$ whatever $\bar f_\perp$ is.

\begin{definition}[Residual regression]\label{q3:def:rr}
With $Z_A\in\R^{n\times r}$ the matrix of rows $z(X_i)^\top$ and $\widehat C_z=Z_A^\top Z_A/n\succ0$, put
$\widehat\beta=(Z_A^\top Z_A)^{-1}Z_A^\top(Y-f_0(X))$ and $\widehat\theta_{\mathrm{rr}}=\theta_0+\mathsf B\widehat\beta$,
and select by projecting $\widehat\theta_{\mathrm{rr}}$ onto $\mathcal C$ as in
Section~\ref{q3:sec:selection}. The regression coefficient is not the target
moment; the moment is reconstructed with the population $\mathsf B$.
\end{definition}

\begin{theorem}[Conditional residual-regression risk]\label{q3:thm:rr}
Let $\bar f_{\perp,A}=(\bar f_\perp(X_i))_{i\le n}$. Conditionally on the anchor inputs,
\[
 \E_\xi\norm{\widehat\theta_{\mathrm{rr}}-\theta}^2
 =\norm{\mathsf B(Z_A^\top Z_A)^{-1}Z_A^\top \bar f_{\perp,A}}^2+\frac{\sigma^2}n\tr(\mathsf B\widehat C_z^{-1}\mathsf B^\top).
\]
This is the conditional expected affine selection regret and bounds the
regret for every closed convex $\mathcal C$ (Theorem~\ref{q3:thm:cone}). If
$\bar f_\perp=0$ it contains no term in the signal amplitude $\norm\beta$. On
$\mathcal E_\eta=\{(1-\eta)I_r\preceq \widehat C_z\preceq(1+\eta)I_r\}$, $0<\eta<1$, the noise
term lies between $\sigma^2d_\Nn/\{n(1+\eta)\}$ and $\sigma^2d_\Nn/\{n(1-\eta)\}$.
\end{theorem}
\begin{proof}
Substituting $Y-f_0(X)=Z_A\beta+\bar f_{\perp,A}+\xi$ gives
$\widehat\beta-\beta=(Z_A^\top Z_A)^{-1}Z_A^\top \bar f_{\perp,A}+(Z_A^\top Z_A)^{-1}Z_A^\top\xi$; the second term is
conditionally centered with covariance $\sigma^2(Z_A^\top Z_A)^{-1}$, so the cross term
vanishes after multiplying by $\mathsf B$. On $\mathcal E_\eta$,
$\widehat C_z^{-1}$ lies between $I_r/(1+\eta)$ and $I_r/(1-\eta)$, and
$\tr(\mathsf BM\mathsf B^\top)$ is monotone in $M\succeq0$ with $\tr(\mathsf B\mathsf B^\top)=d_\Nn$.
\end{proof}

\begin{remark}[The well-specified case $\bar f_\perp=0$ is a strong assumption]\label{q3:rem:rr-limits}
$\bar f_\perp=0$ says that the part of the target that no teacher sees lies
entirely in the $r$ directions $z_i$ along which the students' contrasts meet
$\Nn$. Teacher equations do not imply this: $\Nn$ is typically
infinite-dimensional and nothing places $\bar f$ inside $\Span\{z_i\}$. When
$\bar f_\perp\ne0$, it is orthogonal to $z$ in population but not in a sample,
so it re-enters as variance (Theorem~\ref{q3:thm:rr-gauss}): exact teacher
equations alone do not give a noise-only passive-anchor rate. All $r$
residual directions must also be covered for the regression to be stable; a
small fractional $d_\Nn$ does not by itself make $n$ large enough.
\end{remark}

\begin{lemma}[Inverse Gaussian Gram]\label{q3:lem:wishart}
If the rows of $G\in\R^{n\times r}$ are independent $N(0,I_r)$ and $n>r+1$, then
$\E[(G^\top G)^{-1}]=I_r/(n-r-1)$ \citep{muirhead1982aspects}.
\end{lemma}
\begin{proof}
By the Schur complement, $[(G^\top G)^{-1}]_{11}=\{g_1^\top(I-P_{G_{-1}})g_1\}^{-1}$, where $g_1$ is the
first column and $P_{G_{-1}}$ projects onto the span of the others; given $G_{-1}$
the denominator is $\chi^2_{n-r+1}$, and $\E W^{-1}=1/(\nu-2)$ for
$W\sim\chi^2_\nu$. Column symmetry equalizes the diagonal, and reversing the sign
of one column shows the off-diagonal means vanish (they are integrable by
Cauchy--Schwarz).
\end{proof}

\begin{theorem}[Noise-only unresolved-trace rate]\label{q3:thm:rr-gauss}
If $\bar f_\perp=0$ and $z(X)\sim N(0,I_r)$, then for $n>r+1$
\[
 \boxed{\E\Reg_{\Waff}(\widehat w_{\mathrm{rr}})=\frac{\sigma^2d_\Nn}{n-r-1},}
\]
and this bounds the expected regret for every closed convex $\mathcal C$. If
instead $\bar f_\perp(X)$ is independent of $z(X)$, with $\tau_\perp^2=\norm{\bar f_\perp}^2$, replace
$\sigma^2$ by $\sigma^2+\tau_\perp^2$. Under the same Gaussian, well-specified
model the moment estimator of Section~\ref{q3:sec:selection} has
\[
 \E\Reg_{\Waff}(\widehat w_{\mathrm{mom}})=\frac{(\sigma^2+\norm\beta^2)d_\Nn+\norm{\mathsf B\beta}^2}n .
\]
\end{theorem}
\begin{proof}
Average Theorem~\ref{q3:thm:rr} with Lemma~\ref{q3:lem:wishart}. With independent
$\bar f_\perp$, condition only on $Z_A$: the entries of $\bar f_{\perp,A}+\xi$ are independent with
second moment $\sigma^2+\tau_\perp^2$, and reversing the sign of one row of $Z_A$
leaves $Z_A^\top Z_A$ and the law of $Z_A$ unchanged, so only the diagonal of their
conditional second-moment matrix contributes after averaging. For the moment estimator, the single-anchor vector
$\mathsf Bz(z^\top\beta+\xi)$ has covariance $\mathsf B[(\norm\beta^2+\sigma^2)I_r+\beta\beta^\top]\mathsf B^\top$,
since $\E[z_az_bz_cz_d]=\delta_{ab}\delta_{cd}+\delta_{ac}\delta_{bd}+\delta_{ad}\delta_{bc}$
gives $\E[zz^\top(z^\top\beta)^2]=\norm\beta^2I_r+2\beta\beta^\top$; take the trace and divide by
$n$.
\end{proof}

Residual regression removes exactly the explained-signal terms of the moment
estimator, at the price of estimating and inverting an $r$-dimensional design
Gram; near $n=r$ or at very low signal it need not help.

\paragraph{Gaussian coordinates.} Several statements below assume that all
functions involved (students, target, $\phi$, $z$, $\psi$) are linear in a
standard Gaussian vector, so that functions orthogonal in $L^2(P_X)$ are
independent random variables. This makes the comparisons exact; it is an
idealization of real representations.

\paragraph{Structure versus ordinary stacking.} Ordinary affine stacking
regresses $Y-s_m(X)$ on the contrasts $s_j(X)-s_m(X)$ by unregularized least
squares; since $s_m=a_\circ+\Phi t_m$ and the contrasts span $V$, its
coordinate estimate is the least-squares fit of $Y-a_\circ(X)$ on $\phi(X)$. Write
$f=a_\circ+\Phi\theta+v$ with $v=(I-\Pi_V)f-a_\circ\perp V$, so that the best
affine aggregate has risk $\norm v^2$ (Proposition~\ref{q3:prop:reduced}). With $F_A$
the matrix of rows $\phi(X_i)^\top$, its coordinate error is
$(F_A^\top F_A)^{-1}F_A^\top(v_A+\xi)$. Ordinary stacking is residual regression with
\emph{trivial} structure: $f_0=a_\circ$ (so $\theta_0=0$), $\Nn=\Hh$, $z=\phi$,
$\mathsf B=I_k$, $\beta=\theta$ and $\bar f_\perp=v$.

\begin{theorem}[Structure versus ordinary stacking]\label{q3:thm:stack}
Assume exact structure with $\bar f_\perp=0$ and $\sigma^2>0$. Conditionally on the same
anchor inputs, structured residual regression has strictly smaller expected
affine regret than ordinary stacking whenever
\[
 \tr\{\mathsf B(Z_A^\top Z_A)^{-1}\mathsf B^\top\}<\tr\{(F_A^\top F_A)^{-1}\},
\]
with no condition on $v$; if both normalized Grams are within $\eta$ of the
identity, $d_\Nn/(1-\eta)<k/(1+\eta)$ suffices. If moreover $z(X)$ and $\phi(X)$
are standard Gaussian and $n>k+1$, then
\[
 \E\Reg(\widehat w_{\mathrm{rr}})=\frac{\sigma^2d_\Nn}{n-r-1}\quad\text{while}\quad
 \E\Reg(\widehat w_{\mathrm{stack}})\ge\frac{\sigma^2k}{n-k-1},
\]
with equality $(\sigma^2+\norm v^2)k/(n-k-1)$ in Gaussian coordinates. Under
$\bar f_\perp=0$, structure strictly helps whenever $d_\Nn<k$, for every signal
amplitude. If $\bar f_\perp\ne0$ is independent of $z$, the comparison is
$(\sigma^2+\tau_\perp^2)d_\Nn/(n-r-1)$ against $(\sigma^2+\norm v^2)k/(n-k-1)$, and
$\tau_\perp^2$ and $\norm v^2$ are not ordered in general.
\end{theorem}
\begin{proof}
The structured risk is $\sigma^2\tr\{\mathsf B(Z_A^\top Z_A)^{-1}\mathsf B^\top\}$ by
Theorem~\ref{q3:thm:rr}; by the same computation with trivial structure,
stacking's is
\[
 \norm{(F_A^\top F_A)^{-1}F_A^\top v_A}^2+\sigma^2\tr\{(F_A^\top F_A)^{-1}\},
\]
and the first term is nonnegative. Gram stability gives
the second claim; Lemma~\ref{q3:lem:wishart}, Theorem~\ref{q3:thm:rr-gauss} with
$\tau_\perp^2=\norm v^2$ and $r\le k$ give the Gaussian ones. No independence
between $z$ and $\phi$ is needed.
\end{proof}

\begin{example}[Principal angles]\label{q3:ex:angles}
Let $z_\ell,\zeta_\ell$ ($\ell\le k$) be independent standard Gaussian coordinate
functions, $\Nn\supseteq\Span\{z_\ell\}$ with $\zeta_\ell\perp\Nn$, and
$\phi_\ell=\nu_\ell z_\ell+\sqrt{1-\nu_\ell^2}\,\zeta_\ell$ with $0<\nu_\ell\le1$. Then
$\chi_\ell=\nu_\ell z_\ell$, $H_\Nn=\diag(\nu_\ell^2)$, $r=k$ and $d_\Nn=\sum_\ell\nu_\ell^2$. For
any target $f=f_0+\sum_\ell\beta_\ell z_\ell$ the ratio of the structured risk to
the stacking noise floor is exactly $d_\Nn/k$, independently of $\beta$. (These
are frozen dictionaries; the example does not claim that every such
dictionary arises from a prescribed teacher-training rule.)
\end{example}

The comparison should be read for what it is. The competitor is
unregularized stacking that does not use the structural reference and
nullspace; stacking allowed to use them can implement Definition~\ref{q3:def:rr}.
Theorem~\ref{q3:thm:stack} therefore quantifies the value of the side
information carried by teacher equations (the noise ratio $d_\Nn/k$), rather
than exhibiting a better method for the same information. Ridge-regularized
stacking trades part of its $k$-dimensional variance for a bias toward its
prior weight and is not covered. For simplex weights the structured bound
still holds, but affine stacking's variance does not lower-bound simplex
stacking.

\paragraph{Weak supervision with anchors versus anchors alone.} Let $\Ss$ be a
student space of finite dimension $d_\Ss$ containing all candidates, with
orthonormal basis $\psi$, $A_\Ss=\norm{(I-\Pi_\Ss)f}^2$ and
\[
 R_{\mathrm{agg}}(\mathcal W)=\min_{w\in\mathcal W}\norm{s_w-\Pi_\Ss f}^2 ,
\]
the excess of the best aggregate over $A_\Ss$, so that the best aggregate has
risk $A_\Ss+R_{\mathrm{agg}}(\mathcal W)$. For projection students ($P=\Pi_\Ss$)
$s_w-\Pi_\Ss f=\sum_jw_jPe_j$, so $A_\Ss=A$ and $R_{\mathrm{agg}}(\mathcal W)=q_\parallel^{\mathcal W}$,
the minimum transmitted-error energy of \cref{q1:sec:best}. Write $R_{\mathrm{agg}}=R_{\mathrm{agg}}(\Waff)$. \emph{Anchor-only least
squares} fits $Y$ on $\psi(X)$ with the $n$ anchors alone:
$\widehat\vartheta=\Psi_A^+Y$ for the design $\Psi_A$ with rows $\psi(X_i)^\top$, ordinary
least squares when $n\ge d_\Ss$ and the minimum-norm interpolant when $n<d_\Ss$.
It uses the same $n$ labels but no supervisor information. Since it is
unregularized, it is a weak competitor when the target's energy is
concentrated on few directions; Proposition~\ref{q3:prop:anchor-bayes} replaces it
by the best anchor-only procedure.

\begin{theorem}[Weak supervision against anchor-only least squares]\label{q3:thm:anchor-only}
Let $\sigma^2>0$.
\begin{enumerate}
\item[(i)] Assume exact structure with $\bar f_\perp=0$. If $\Psi_A$ has full column rank, weak supervision with residual
regression has smaller conditional expected risk than anchor-only least
squares whenever
$R_{\mathrm{agg}}(\mathcal W)+\sigma^2\tr\{\mathsf B(Z_A^\top Z_A)^{-1}\mathsf B^\top\}<\sigma^2\tr\{(\Psi_A^\top \Psi_A)^{-1}\}$.
\item[(ii)] Let $f\in\Ss$ and $\psi(X)\sim N(0,I_{d_\Ss})$. Anchor-only least squares has
\[
 \E\norm{\psi^\top\widehat\vartheta-f}^2=
 \begin{cases}
 \dfrac{\sigma^2d_\Ss}{n-d_\Ss-1}, & n>d_\Ss+1,\\[2ex]
 \norm f^2\Bigl(1-\dfrac n{d_\Ss}\Bigr)+\dfrac{\sigma^2n}{d_\Ss-n-1}, & n<d_\Ss-1,
 \end{cases}
\]
the classical OLS and isotropic ridgeless formulas \citep{hastie2022surprises,belkin2020two}.
\item[(iii)] Let $f\in\Ss$, affine weights and Gaussian coordinates. With exact structure,
$\bar f_\perp=0$ and $n>r+1$, weak supervision with anchors has expected risk
\[
 \mathcal R_{\mathrm{weak}}=R_{\mathrm{agg}}+\frac{\sigma^2d_\Nn}{n-r-1};
\]
without structure ($f_0=a_\circ$, $\Nn=\Hh$) and $n>k+1$,
\[
 \mathcal R_{\mathrm{weak}}^{\mathrm{unstr}}=R_{\mathrm{agg}}+\frac{(\sigma^2+R_{\mathrm{agg}})k}{n-k-1}.
\]
Either beats anchor-only least squares if and only if it is below the risk in (ii);
for convex $\mathcal W$ the conditions with $R_{\mathrm{agg}}(\mathcal W)$ are sufficient.
\end{enumerate}
\end{theorem}
\begin{proof}
The weak procedure has risk at most $A_\Ss+R_{\mathrm{agg}}(\mathcal W)$ plus its coordinate error
(Theorem~\ref{q3:thm:cone}), with equality for affine weights. With
$f_{\Ss^\perp}=(I-\Pi_\Ss)f$ and $\vartheta=\E[\psi f]$, OLS has
$\widehat\vartheta-\vartheta=(\Psi_A^\top \Psi_A)^{-1}\Psi_A^\top(f_{\Ss^\perp,A}+\xi)$ and conditional risk
$A_\Ss+\norm{(\Psi_A^\top \Psi_A)^{-1}\Psi_A^\top f_{\Ss^\perp,A}}^2+\sigma^2\tr\{(\Psi_A^\top \Psi_A)^{-1}\}$; drop the middle
term for (i). For (ii) with $f\in\Ss$ and $n>d_\Ss+1$ use Lemma~\ref{q3:lem:wishart}.
For $n<d_\Ss-1$, $\Psi_A\Psi_A^\top$ is invertible almost surely and
$\widehat\vartheta-\vartheta=-(I-P_A)\vartheta+\Psi_A^\top(\Psi_A\Psi_A^\top)^{-1}\xi$, with $P_A$ the projection onto the row
space of $\Psi_A$; the two parts are orthogonal. The row space of a Gaussian
matrix is uniformly distributed, so $\E\norm{(I-P_A)\vartheta}^2=\norm\vartheta^2(1-n/d_\Ss)$,
and $\Psi_A\Psi_A^\top$ is Wishart with $d_\Ss$ degrees of freedom, so
$\sigma^2\E\tr\{(\Psi_A\Psi_A^\top)^{-1}\}=\sigma^2n/(d_\Ss-n-1)$. For (iii), the structured
case is Theorem~\ref{q3:thm:rr-gauss}; without structure Theorem~\ref{q3:thm:stack} gives
$(\sigma^2+\norm v^2)k/(n-k-1)$ with $\norm v^2=A_\Ss+R_{\mathrm{agg}}=R_{\mathrm{agg}}$ for $f\in\Ss$.
\end{proof}

Without structure the approximation error re-enters as variance: the factor
$\sigma^2+R_{\mathrm{agg}}$ in $\mathcal R^{\mathrm{unstr}}_{\mathrm{weak}}$. This is the relevant formula when
the supervisors are language models or heuristics with no known training
equations.

\begin{proposition}[Anchors alone on the sphere]\label{q3:prop:anchor-lower}
Let $n<d_\Ss$. For any design, any noise law not depending on $f$, and any
(randomized) procedure using only the $n$ anchors and the knowledge that
$f\in\Ss$ with $\norm f=R$, the risk averaged over $f$ uniform on that
sphere, hence also the worst case over the sphere, is at least $R^2(1-n/d_\Ss)$.
\end{proposition}
\begin{proof}
Write $f=\psi^\top\vartheta$. Given the inputs, the labels depend on $\vartheta$ only through
$\Psi_A\vartheta$. The reflection $\vartheta\mapsto P_A\vartheta-(I-P_A)\vartheta$ preserves the sphere, the
uniform law and $\Psi_A\vartheta$, so given the data the component $(I-P_A)\vartheta$ has
conditional mean zero, and any estimate errs by at least its conditional
second moment in that component. Its expectation is
$R^2(d_\Ss-\rank \Psi_A)/d_\Ss\ge R^2(1-n/d_\Ss)$. The risk of a prediction $h$ is at least
$\norm{\Pi_\Ss h-f}^2$.
\end{proof}

Proposition~\ref{q3:prop:anchor-lower} is a statement about isotropic targets
(no free lunch with $n<d_\Ss$ labels); it is not pointwise. The advantage of
weak supervision is pointwise and holds on the set $\{f:R_{\mathrm{agg}}(f)\text{ small}\}$ of
targets close to the affine hull of the candidates; for a typical $f$ on the
sphere, $R_{\mathrm{agg}}(f)$ is itself of order $\norm f^2(1-k/d_\Ss)$. Real
representations have decaying spectra and targets aligned with their leading
directions, so the relevant anchor-only cost is set by an effective dimension.

\begin{proposition}[Effective-dimension lower bound for anchors alone]\label{q3:prop:anchor-bayes}
Let $f=\psi^\top\vartheta\in\Ss$ with prior $\vartheta\sim N(0,\Lambda)$, $\Lambda=\diag(\lambda_j)\succ0$, independent
of the inputs, Gaussian noise $N(0,\sigma^2)$, and any passive design with
$\E[\psi\psi^\top]=I$. Every procedure using only the $n$ anchors (and knowing
$\Lambda$ and $\sigma^2$) has Bayes risk
\[
 \E\norm{h-f}^2\ \ge\ \E\tr\Bigl\{\Bigl(\Lambda^{-1}+\frac{\Psi_A^\top\Psi_A}{\sigma^2}\Bigr)^{-1}\Bigr\}
 \ \ge\ \frac{\sigma^2}n\,d_{\mathrm{eff}}\Bigl(\frac{\sigma^2}n\Bigr),\qquad
 d_{\mathrm{eff}}(\gamma)=\sum_j\frac{\lambda_j}{\lambda_j+\gamma},
\]
with equality in the first inequality for ridge regression on the anchors
with penalty $\sigma^2\vartheta^\top\Lambda^{-1}\vartheta$, the posterior mean.
\end{proposition}
\begin{proof}
Given the data, $\vartheta$ is Gaussian with covariance $(\Lambda^{-1}+\Psi_A^\top\Psi_A/\sigma^2)^{-1}$ and
mean the stated ridge estimate; the Bayes risk of any estimate of $\vartheta$ is at
least the expected trace of the posterior covariance, with equality at the
posterior mean, and $\norm{h-f}^2\ge\norm{\Pi_\Ss h-f}^2$. The map
$M\mapsto\tr\{(\Lambda^{-1}+M)^{-1}\}$ is convex on positive semidefinite matrices
(operator convexity of the inverse) and $\E\Psi_A^\top\Psi_A=nI$, so Jensen's
inequality gives $\tr\{(\Lambda^{-1}+nI/\sigma^2)^{-1}\}=\sum_j\lambda_j\sigma^2/(\sigma^2+n\lambda_j)$.
\end{proof}

\begin{corollary}[When weak supervision is worth its anchors]\label{q3:cor:weak-bayes}
Under the prior of Proposition~\ref{q3:prop:anchor-bayes} and the assumptions of
Theorem~\ref{q3:thm:anchor-only}(iii) for prior-almost every $f$, weak supervision with anchors has smaller Bayes
risk than \emph{every} procedure using only the anchors, including tuned ridge,
whenever
\[
 \boxed{\E\,\mathcal R_{\mathrm{weak}}<\frac{\sigma^2}n\,d_{\mathrm{eff}}\Bigl(\frac{\sigma^2}n\Bigr)}
\]
(with $\mathcal R^{\mathrm{unstr}}_{\mathrm{weak}}$ in place of $\mathcal R_{\mathrm{weak}}$ without structure), the
expectation being over the prior.
\end{corollary}
\begin{proof}
Average the exact conditional risks of Theorem~\ref{q3:thm:anchor-only}(iii) over the
prior and compare with Proposition~\ref{q3:prop:anchor-bayes}.
\end{proof}

The right side is near $\tr\Lambda=\E\norm f^2$ while $n\lambda_j\lesssim\sigma^2$ for the
leading directions, and near $\sigma^2d_{\mathrm{eff}}/n$ afterwards. Weak supervision is
therefore worth its anchors when $n$ is small relative to the effective
dimension of the target in the student's representation and the best
aggregate captures most of the signal. With structure, the condition fails
roughly once $n\gtrsim\sigma^2\{d_{\mathrm{eff}}(\sigma^2/n)-d_\Nn\}/R_{\mathrm{agg}}$; in the isotropic
case $d_{\mathrm{eff}}\approx d_\Ss$ for large $n$. The condition is sufficient only:
it compares an exact weak risk with a lower bound. Three limits remain.
$R_{\mathrm{agg}}$ is not observable, so this is a population comparison, not a
selection rule. The weak procedure here uses anchors only to select weights;
hybrid estimators that also fit the residual $Y-s_{\widehat w}$ on the anchors (or use
the selected student as a ridge prior) are not capped by $R_{\mathrm{agg}}$ as $n$
grows and are not analyzed. And ridge stacking is not covered.

\paragraph{Approximate structure.} Let a reference $\widetilde f_0$ and a
nullspace $\widetilde\Nn$ be proposed independently of the anchors, and repeat the
construction with $\widetilde\chi=\Pi_{\widetilde\Nn}\phi=\widetilde{\mathsf B}\widetilde z$,
$\E\widetilde z\widetilde z^\top=I_{\widetilde r}$, $\widetilde H=\widetilde{\mathsf B}\widetilde{\mathsf B}^\top$ and
$f-\widetilde f_0=\widetilde z^\top\widetilde\beta+\widetilde f_\perp$, $\widetilde f_\perp\perp\Span\{\widetilde z_i\}$. The population
target of the reconstructed coordinate is $\Phi^*\widetilde f_0+\widetilde{\mathsf B}\widetilde\beta$, whose bias is
\[
 b=-\Phi^*(I-\Pi_{\widetilde\Nn})(f-\widetilde f_0),
\]
because $\widetilde{\mathsf B}\widetilde\beta=\E[\widetilde\chi(f-\widetilde f_0)]=\Phi^*\Pi_{\widetilde\Nn}(f-\widetilde f_0)$.

\begin{theorem}[Approximate structure: a bias--variance budget]\label{q3:thm:approx-structure}
On an invertible design $\widetilde Z_A$, with
$c_A=\widetilde{\mathsf B}(\widetilde Z_A^\top\widetilde Z_A)^{-1}\widetilde Z_A^\top\widetilde f_{\perp,A}$,
\[
 \E_\xi\norm{\widehat\theta_{\mathrm{str}}-\theta}^2=\norm{b+c_A}^2+\sigma^2\tr\{\widetilde{\mathsf B}(\widetilde Z_A^\top\widetilde Z_A)^{-1}\widetilde{\mathsf B}^\top\},
\]
the conditional expected affine regret (an upper bound for convex $\mathcal C$).
If $\widetilde z(X)$ is standard Gaussian, $\widetilde f_\perp(X)$ is independent of $\widetilde z(X)$
with $\norm{\widetilde f_\perp}^2=\tau^2$, and $n>\widetilde r+1$, then
$\E\norm{\widehat\theta_{\mathrm{str}}-\theta}^2=\norm b^2+(\sigma^2+\tau^2)\tr\widetilde H/(n-\widetilde r-1)$, and
with standard Gaussian $\phi(X)$ and $n>k+1$, structure beats ordinary affine
stacking whenever
\[
 \boxed{\norm b^2<\frac{\sigma^2k}{n-k-1}-\frac{(\sigma^2+\tau^2)\tr\widetilde H}{n-\widetilde r-1}.}
\]
\end{theorem}
\begin{proof}
The coordinate error is
$b+c_A+\widetilde{\mathsf B}(\widetilde Z_A^\top\widetilde Z_A)^{-1}\widetilde Z_A^\top\xi$; conditional centering removes the
noise cross term. In the Gaussian case, the conditional mean of $c_A$ given
$\widetilde Z_A$ is odd under $\widetilde Z_A\mapsto-\widetilde Z_A$, so its cross term with the
deterministic $b$ averages to zero, and $\E\norm{c_A}^2$ plus the noise term
equals $(\sigma^2+\tau^2)\E\tr\{\widetilde{\mathsf B}(\widetilde Z_A^\top\widetilde Z_A)^{-1}\widetilde{\mathsf B}^\top\}$ as in the
proof of Theorem~\ref{q3:thm:rr-gauss}; apply Lemma~\ref{q3:lem:wishart} and compare
with Theorem~\ref{q3:thm:stack}.
\end{proof}

\begin{corollary}[From teacher-equation error to decision bias]\label{q3:cor:teacher-error}
Let the observed teacher outputs be $y=\mathcal Mf+\delta$ with $\norm\delta\le\epsilon$.
\begin{enumerate}
\item[(i)] \emph{Closed range} (finite-dimensional teacher features, e.g.\
projection teachers or ridge on finitely many features). If
$\norm{\mathcal Mv}\ge\alpha_{\mathcal M}\norm v$ on $\Kk=\Nn^\perp$ with $\alpha_{\mathcal M}>0$, then with
$\widetilde f_0=\mathcal M^\dagger y$ and $\widetilde\Nn=\Nn$, the reference error $e_0=\widetilde f_0-f_\Kk$
lies in $\Kk$ with $\norm{e_0}\le\epsilon/\alpha_{\mathcal M}$, the bias is $b=\Phi^*e_0$ and
\[
 \norm b^2\le\lambda_{\max}(I_k-H_\Nn)\,\epsilon^2/\alpha_{\mathcal M}^2 .
\]
If, in addition, $\bar f_\perp=0$, $(z(X),e_0(X))$ is jointly Gaussian, $\phi(X)$ is
standard Gaussian and $n>k+1$, structure beats ordinary affine stacking
whenever
\[
 \boxed{\lambda_{\max}(I_k-H_\Nn)\frac{\epsilon^2}{\alpha_{\mathcal M}^2}+\Bigl(\sigma^2+\frac{\epsilon^2}{\alpha_{\mathcal M}^2}\Bigr)\frac{d_\Nn}{n-r-1}<\frac{\sigma^2k}{n-k-1}.}
\]
\item[(ii)] \emph{Compact teachers} (e.g.\ kernel ridge teachers
$M_j=\mathcal L_j(\mathcal L_j+\lambda_jI)^{-1}$ in infinite dimensions, where $\alpha_{\mathcal M}=0$ and (i)
is void). Let $\mathcal M=\sum_i\varsigma_iu_i\otimes v_i$ be a singular value decomposition, fix
$\tau>0$, and put $\Kk_\tau=\overline{\Span}\{v_i:\varsigma_i>\tau\}$, $\Nn_\tau=\Kk_\tau^\perp\supseteq\Nn$ and
$\mathcal M_\tau^\dagger=\sum_{\varsigma_i>\tau}\varsigma_i^{-1}v_i\otimes u_i$. With $\widetilde f_0=\mathcal M_\tau^\dagger y$ and
$\widetilde\Nn=\Nn_\tau$,
\[
 b=\Phi^*\mathcal M_\tau^\dagger\delta,\qquad \norm b^2\le\lambda_{\max}(\Phi^*\Pi_{\Kk_\tau}\Phi)\,\epsilon^2/\tau^2,
\]
and the variance is governed by $d_{\Nn_\tau}=\tr(\Phi^*\Pi_{\Nn_\tau}\Phi)\ge d_\Nn$, which
decreases to $d_\Nn$ as $\tau\downarrow0$. The threshold $\tau$ trades decision bias
against unresolved trace.
\end{enumerate}
For a general proposed nullspace and $\norm{\widetilde f_0-f_\Kk}\le\epsilon_0$,
$\norm b\le\epsilon_0+\norm{(\Pi_{\widetilde\Nn}-\Pi_\Nn)(f-f_\Kk)}$.
\end{corollary}
\begin{proof}
(i) The lower bound on $\Kk$ makes the range closed and $\norm{\mathcal M^\dagger}\le1/\alpha_{\mathcal M}$ (as in
\cref{ext:short}), and
$\mathcal M^\dagger\mathcal M=\Pi_\Kk$, so $e_0=\mathcal M^\dagger\delta\in\Kk$ with
$\norm{e_0}\le\epsilon/\alpha_{\mathcal M}$. With $\widetilde\Nn=\Nn$, $b=-\Phi^*\Pi_\Kk(f-\widetilde f_0)=\Phi^*e_0$, and
$\norm{\Phi^*\Pi_\Kk}_\op^2=\lambda_{\max}(\Phi^*\Pi_\Kk\Phi)=\lambda_{\max}(I_k-H_\Nn)$. In the Gaussian case
$e_0\in\Kk$ is uncorrelated with, hence independent of, $z$ (whose coordinates lie
in $\Nn$), so $\widetilde f_\perp=-e_0$ with $\tau^2=\norm{e_0}^2\le\epsilon^2/\alpha_{\mathcal M}^2$ and $\widetilde H=H_\Nn$; apply
Theorem~\ref{q3:thm:approx-structure}. (ii) $\mathcal M_\tau^\dagger\mathcal M=\Pi_{\Kk_\tau}$ and
$\norm{\mathcal M_\tau^\dagger}\le1/\tau$, so $\Pi_{\Kk_\tau}(f-\widetilde f_0)=-\mathcal M_\tau^\dagger\delta$ and
$b=-\Phi^*\Pi_{\Kk_\tau}(f-\widetilde f_0)=\Phi^*\mathcal M_\tau^\dagger\delta$, with
$\norm{\Phi^*\Pi_{\Kk_\tau}}_\op^2=\lambda_{\max}(\Phi^*\Pi_{\Kk_\tau}\Phi)$. As $\tau\downarrow0$,
$\Pi_{\Nn_\tau}\to\Pi_\Nn$ strongly, and the trace over the finite-dimensional $V$
converges. The general bound: $(I-\Pi_{\widetilde\Nn})(f-f_\Kk)=(\Pi_\Nn-\Pi_{\widetilde\Nn})(f-f_\Kk)$
because $f-f_\Kk\in\Nn$, and $\Phi^*$ and $I-\Pi_{\widetilde\Nn}$ are contractions.
\end{proof}

Equation error thus consumes a squared-bias budget \emph{and} a
residual-variance budget, and an ill-conditioned teacher stack (small $\alpha_{\mathcal M}$,
or small $\tau$ for compact teachers) can turn a small equation error into a
large decision bias. This converts \cref{q2:prop:approx}, which bounds one fixed moment, into a selection-risk
comparison.

\begin{remark}[Implementation and accounting]\label{q3:rem:rr-implement}
(i) The identities use the population $\mathsf B$ and $\theta_0$. An implementable
estimate differing from the ideal one by $\Delta$ satisfies
$\E\norm{\widehat\theta-\theta}^2\le(1+a)\E\norm{\widehat\theta_{\mathrm{ideal}}-\theta}^2+(1+a^{-1})\E\norm\Delta^2$ for
every $a>0$; the extra term must come from the finite-unlabeled bounds of \cref{app:q4},
and a claimed improvement must compare such an upper bound with an exact risk
or a lower bound of the competitor. (ii) Conditional bounds on $\mathcal E_\eta$ are
not unconditional; with a bounded fallback of regret at most $\bar L$ on the
complement, add $\bar L\,\Prob(\mathcal E_\eta^c)$. (iii) Without noise, covered and
well-specified residual regression and anchor-only least squares are both
exact, so no strict advantage follows. (iv) Labels used to train teachers,
references or pilots are side information and count toward the budget.
\end{remark}

\apppart{Passive anchors and coverage}{Exact coverage complexity and when Gaussian results apply to passive anchors.}
\subsection{Passive anchors on a fixed dictionary: exact coverage complexity}\label{q3:sec:cells}

This section stays in the passive experiment, with no Gaussian moment model.
Let $X=(I,Z)$ with $I\in\{0,1,\dots,k\}$, $\Prob(I=\ell)=p_\ell>0$ for
$\ell\le k$, $\sum_{\ell\ge1}p_\ell\le1$, and $Z$ an independent Rademacher. The
target class is $f_b=\sum_{\ell=1}^kb_\ell\one\{I=\ell\}$ with $\abs{b_\ell}\le B_\ell$
(known $p_\ell,B_\ell$). Put $\phi_\ell=\one\{I=\ell\}/\sqrt{p_\ell}$,
$\theta_\ell=b_\ell\sqrt{p_\ell}$, $c_\ell=B_\ell\sqrt{p_\ell}$, take the vertices of
Lemma~\ref{q3:lem:cube}, $s_j=\Phi t_j$, $g_j=s_j+Z$, and $P$ the projection onto
$\Span\{\phi_\ell\}$, so $Pg_j=s_j$. Every target coordinate lies in the simplex
hull, so the best aggregate and the clean student $Pf_b=f_b$ have risk $0$,
while $R(g_j)=\norm{s_j-f_b}^2+1\ge1$. Let $\mathcal R_n$ be the minimax expected
selection regret (affine or simplex) and
\[
 \mathfrak c_n=\sum_{\ell=1}^kp_\ell B_\ell^2(1-p_\ell)^n,\qquad
 \mathfrak v_n=\sum_{\ell=1}^k\min\Big\{p_\ell B_\ell^2,\frac{\sigma^2}n\Big\}.
\]

\begin{theorem}[Exact noiseless coverage formula]\label{q3:thm:cell-noiseless}
For noiseless anchors $Y=f_b(X)$, $\mathcal R_n=\mathfrak c_n$ for every $n\ge0$.
\end{theorem}
\begin{proof}
Let $K_\ell=\#\{i:I_i=\ell\}\sim\mathrm{Bin}(n,p_\ell)$. Estimate $b_\ell$ by its
observed value if $K_\ell>0$ and by $0$ otherwise; the estimate is in the box,
hence a feasible aggregate, with risk $\sum_\ell p_\ell b_\ell^2(1-p_\ell)^n\le\mathfrak c_n$.
Conversely, under independent uniform priors on $b_\ell\in\{\pm B_\ell\}$, on
$\{K_\ell=0\}$ the data are independent of $b_\ell$, so any $\widehat b_\ell$ has
conditional risk $\widehat b_\ell^2+B_\ell^2\ge B_\ell^2$; multiplying by $p_\ell$ and
summing gives Bayes risk $\ge\mathfrak c_n$.
\end{proof}

\begin{theorem}[Coverage plus estimation noise]\label{q3:thm:cell-noisy}
With $Y=f_b(X)+\xi$, $\xi\sim N(0,\sigma^2)$ independent, $\sigma>0$, $n\ge1$:
\[
 \max\Big\{\mathfrak c_n,\tfrac3{256}\mathfrak v_n\Big\}\le\mathcal R_n\le\mathfrak c_n+2\mathfrak v_n,
 \qquad\text{hence}\qquad\tfrac3{512}(\mathfrak c_n+\mathfrak v_n)\le\mathcal R_n\le2(\mathfrak c_n+\mathfrak v_n).
\]
\end{theorem}
\begin{proof}
\emph{Upper.} Use the clipped cell average on observed cells and $0$ on
unobserved cells; clipping to $[-B_\ell,B_\ell]$ cannot increase error. Since
$1/K\le2/(K+1)$ and $\E(K+1)^{-1}=\int_0^1(1-p+pt)^ndt\le1/((n+1)p)$ for
$K\sim\mathrm{Bin}(n,p)$, the coordinate risk is
$\le B_\ell^2(1-p_\ell)^n+2\sigma^2/((n+1)p_\ell)$; the zero estimator has risk
$\le B_\ell^2$. Choosing the better one from known parameters and summing with
weights $p_\ell$ gives $\le\mathfrak c_n+\sum_\ell\min\{p_\ell B_\ell^2,2\sigma^2/(n+1)\}\le\mathfrak c_n+2\mathfrak v_n$.
\emph{Coverage lower.} The noiseless prior argument is unchanged by noise.
\emph{Noise lower.} With $a_\ell=\min\{B_\ell,\sigma/\sqrt{32np_\ell}\}$ and uniform
signs, neighbors differ in $L^2$ by $4p_\ell a_\ell^2$, KL
$2np_\ell a_\ell^2/\sigma^2\le1/16$, so average sign error $\ge3/8$, each costing
$\ge p_\ell a_\ell^2$; thus $\mathcal R_n\ge\tfrac38\sum_\ell\min\{p_\ell B_\ell^2,\sigma^2/(32n)\}\ge\tfrac3{256}\mathfrak v_n$.
Finally $\max\{x,cy\}\ge\tfrac c2(x+y)$ for $c\le1$.
\end{proof}

\paragraph{Interpretation.} (i) Fixed design: $\mathfrak c_n\le e^{-np_{\min}}\sum_\ell p_\ell B_\ell^2$,
exponentially faster than $k/n$. (ii)
\[
 \mathfrak v_n=\sum_{\ell:\,np_\ell B_\ell^2<\sigma^2}p_\ell B_\ell^2+\frac{\sigma^2}n\,\#\{\ell:np_\ell B_\ell^2\ge\sigma^2\}
\]
counts statistically resolved directions and caps unresolved ones by their
signal energy. (iii) Worst case: with $p_\ell=1/(2n)$, $B_\ell=B$, $n\ge k$,
Bernoulli's inequality gives $\mathfrak c_n\ge B^2k/(4n)$ --- the noiseless
$B^2k/n$ rate of Theorem~\ref{q3:thm:expected} (with $v_0=B$, $f_0=0$) is minimax
sharp over bounded targets, by a design that varies with $n$.

\begin{proposition}[Population Grams do not determine anchor complexity]\label{q3:prop:same-gram}
With $B_\ell=p_\ell^{-1/2}$, all student, teacher and residual Gram matrices and
the target coordinate set $[-1,1]^k$ are independent of $(p_\ell)$, while
$\mathcal R_n^{\mathrm{noiseless}}=\sum_\ell(1-p_\ell)^n$. Uniform cells
$p_\ell=1/k$ give $\le ke^{-n/k}$; rare cells $p_\ell=\varepsilon/k$ give $\ge k-n\varepsilon$.
\end{proposition}
\begin{proof}
$c_\ell=1$, so the $t_j$ do not depend on $p$; $\ip{s_i}{s_j}=t_i^\top t_j$,
$\ip{g_i}{g_j}=t_i^\top t_j+1$, $\ip{q_i}{q_j}=1$. Apply
Theorem~\ref{q3:thm:cell-noiseless} with $p_\ell B_\ell^2=1$.
\end{proof}

The pointwise envelopes (leverage) differ between these designs; the example
shows precisely that population $L^2$ geometry alone cannot determine passive
anchor complexity, not hardness under a common bounded-leverage assumption.

\subsection{When Gaussian moment results apply to passive anchors}\label{q3:sec:bridge}

Suppose $f=f_0+\psi^\top\vartheta$ with $\E[\psi\psi^\top]=I_d$, the model
covering the whole unknown regression component, and Gaussian noise. With
design matrix $\Psi_A$ (rows $\psi(X_i)^\top$) of full column rank, the OLS
estimate $\widehat\vartheta=(\Psi_A^\top\Psi_A)^{-1}\Psi_A^\top(Y-f_0)$ satisfies, conditionally on
the design, $\widehat\vartheta-\vartheta\sim N(0,\sigma^2(\Psi_A^\top\Psi_A)^{-1})$: every Gaussian
moment result applies conditionally with $\Omega_A=n\sigma^2(\Psi_A^\top\Psi_A)^{-1}$
($L\Omega_AL^\top$ for $L\vartheta$). Representing only the gain-query functions
does not remove omitted regression components, and plain empirical moments
need not be jointly Gaussian.

\begin{proposition}[Empirical-geometry event]\label{q3:prop:bridge}
Assume that the features are bounded, $\norm{\psi(X)}^2\le\Lambda$ a.s.\ (necessarily $\Lambda\ge d$), and that
$n\ge64\Lambda\log(2d/\delta_D)$. Then, with probability at least $1-\delta_D$,
\[
 \tfrac34I\preceq n^{-1}\Psi_A^\top\Psi_A\preceq\tfrac54I,\qquad
 \tfrac45\sigma^2I\preceq\Omega_A\preceq\tfrac43\sigma^2I;
\]
a conditional failure
probability $\delta$ becomes $\delta+\delta_D$.
\end{proposition}
\begin{proof}
$Z_i=\psi_i\psi_i^\top-I$ are centered, $\norm{Z_i}_\op\le\Lambda$ and
$\E Z_i^2\preceq\E[\norm\psi^2\psi\psi^\top]\preceq\Lambda I$. Matrix Bernstein
\citep{tropp2012user} gives $\norm{n^{-1}\sum Z_i}_\op\le\sqrt{2\Lambda t/n}+2\Lambda t/(3n)$
with probability $\ge1-2de^{-t}$; at $t=\log(2d/\delta_D)$ and
$n\ge64\Lambda t$ this is $\le0.19<\tfrac14$.
\end{proof}

The empirical-geometry requirement can dominate a localized Gaussian
complexity and cannot be omitted from a passive claim; the cell results need
no bridge and account for unobserved directions explicitly.

\apppart{Summary and supporting tools}{}
\subsection{Frozen candidates, anchor reuse, and what is matched}\label{q3:sec:summary3}

\paragraph{What is matched, and in which model.}
\begin{center}\small
\begin{tabularx}{\textwidth}{@{}>{\raggedright}p{0.15\textwidth}>{\raggedright}X>{\raggedright}X>{\raggedright\arraybackslash}p{0.22\textwidth}@{}}
\toprule
Claim & Upper bound & Lower bound & Observation model\tabularnewline
\midrule
One gain sign & $\tau^2\gamma^{-2}\log(1/\delta)$ & $\sigma^2\gamma^{-2}\log(1/\delta)$ (Thm~\ref{q3:thm:extremes}(i)) & sub-Gaussian (upper), passive Gaussian (lower)\tabularnewline
All signs on $\WL$ & $L^2\tau^2\gamma^{-2}\log(m/\delta)$; Gaussian width & $\log m$ worst case, none for duplicates (Thm~\ref{q3:thm:extremes}) & Gaussian moments; passive via \S\ref{q3:sec:bridge}\tabularnewline
Selection & $v_0^2d_\Nn/n$; $\sd_\Omega(\mathcal T)/n$ (in expectation) & $\sigma^2d_\Nn/n$ (Thm~\ref{q3:thm:angle}) & passive anchors (lower bound: Gaussian noise)\tabularnewline
Vertex vs.\ interior & $\log m$ vs.\ $m-1$ & both (Prop.~\ref{q3:prop:vertex}) & Gaussian moments\tabularnewline
Selection by residual regression & exactly $\sigma^2d_\Nn/(n-r-1)$ (Thm~\ref{q3:thm:rr-gauss}) & $\sigma^2d_\Nn/n$ in a different design class (Thm~\ref{q3:thm:angle}) & Gaussian design, homoscedastic noise\tabularnewline
Structure vs.\ stacking & ratio $d_\Nn/k$ of noise terms (Thm~\ref{q3:thm:stack}) & stacking $\ge\sigma^2k/(n-k-1)$ & Gaussian design; $\bar f_\perp=0$; OLS stacking\tabularnewline
Anchors alone & exact OLS/min-norm risk (Thm~\ref{q3:thm:anchor-only}); Bayes risk of tuned ridge (Prop.~\ref{q3:prop:anchor-bayes}) & $\sigma^2d_{\mathrm{eff}}(\sigma^2/n)/n$ for any procedure (Prop.~\ref{q3:prop:anchor-bayes}); $R^2(1-n/d_\Ss)$ worst case over the sphere (Prop.~\ref{q3:prop:anchor-lower}) & Gaussian prior and noise; any passive design\tabularnewline
Passive cells & $\mathfrak c_n+2\mathfrak v_n$ & $\max\{\mathfrak c_n,\tfrac3{256}\mathfrak v_n\}$; exact when noiseless & passive cells\tabularnewline
\bottomrule
\end{tabularx}
\end{center}
Not established: optimal $L>1$ dependence for signs; a matching
high-probability selection lower bound under Assumption~\ref{q3:ass:vector};
unconditional simplex matching in Theorem~\ref{q3:thm:angle}; exact comparisons
of Section~\ref{q3:sec:compare} beyond Gaussian designs and homoscedastic noise,
without the well-specification $\bar f_\perp=0$, against ridge-regularized
stacking, and for hybrid estimators that also fit the residual of the selected
student on the anchors; an observable rule for whether weak supervision is
worth its anchors.

\paragraph{Frozen candidates and anchor reuse.} The paired statistic
$n^{-1}\sum_i[g(X_i)^2-\widetilde s(X_i)^2-2Y_i(g-\widetilde s)(X_i)]$ is unbiased
for $\norm{g-f}^2-\norm{\widetilde s-f}^2$ for any frozen pair
(Proposition~\ref{q3:prop:noise}); replacing $\norm g^2-\norm{\widetilde s}^2$ by
$\norm{Qg}^2$ is invalid unless $\widetilde s=Pg$. Candidates built with the
same anchors need fresh anchors or a uniform bound over the enlarged class,
and one anchor sample estimating both $U$- and $V$-moments needs a union
bound, since $U\perp V$ does not make the errors independent.

\paragraph{Relation to prior work.} What in this appendix is standard and what is new, and detailed comparisons with prior work, are collected in \cref{app:boundary,app:related}.

\subsection{Standard tools}\label{q3:app:conc}

\begin{lemma}[Bounded-variable sub-Gaussian inequality]\label{q3:lem:hoeffding}
If $Z\in[a,b]$ a.s., then $\E e^{t(Z-\E Z)}\le e^{t^2(b-a)^2/8}$ for all $t\in\R$.
\end{lemma}
\begin{proof}Standard \citep{boucheron2013concentration}.\end{proof}

\begin{lemma}[A finite norming net]\label{q3:lem:net}
The unit sphere of $\R^k$ has a $\tfrac12$-net $\mathcal M$ with $\abs{\mathcal M}\le5^k$,
and $\norm x\le2\max_{a\in\mathcal M}\abs{a^\top x}$ for all $x$.
\end{lemma}
\begin{proof}Standard volumetric argument \citep{boucheron2013concentration}.\end{proof}

\label{q3:app:info}

\begin{lemma}[Gaussian-regression KL]\label{q3:lem:gausskl}
With a common design law and labels $N(f(X),\sigma^2)$ vs.\ $N(h(X),\sigma^2)$,
the $n$-anchor laws satisfy $\KL=n\norm{f-h}^2/(2\sigma^2)$.
\end{lemma}
\begin{proof}Direct computation; the design laws cancel.\end{proof}

\begin{lemma}[Binary testing]\label{q3:lem:testing}
For any (randomized) test between $\mathbb P,\mathbb Q$ with
$D=\KL(\mathbb P\Vert\mathbb Q)<\infty$: the error sum is $\ge1-\TV\ge1-\sqrt D$, and
the maximal error is $\ge\tfrac14e^{-D}$.
\end{lemma}
\begin{proof}Standard \citep{tsybakov2009introduction}.\end{proof}

\begin{lemma}[Finite-family testing]\label{q3:lem:fano}
Let $J_0$ be uniform on $\{1,\dots,M\}$, $M\ge2$, with conditional laws
$\mathbb P_j$. If $\KL(\mathbb P_j\Vert\mathbb Q)\le\beta$ for all $j$ and some
reference $\mathbb Q$ (in particular if all pairwise KL are $\le\beta$), every
estimator has $\Prob(\widehat J_0\ne J_0)\ge1-(\beta+\log2)/\log M$.
\end{lemma}
\begin{proof}Fano's inequality \citep{tsybakov2009introduction}.\end{proof}

\section{Finite-Sample Learning: Formal Results and Proofs}\label{app:q4}

\paragraph{Summary.} \noindent
Appendices~\ref{app:q1}--\ref{app:q3} treat the student learner and all prediction-only moments as known
population objects. \cref{app:q4} removes these oracles. The student is ridge regression
(finite-dimensional, kernel, random-feature or frozen-representation) fitted on
$N_{\mathrm{fit}}$ pseudo-labelled inputs; prediction Grams are estimated from
$N$ unlabeled inputs; target moments from $n$ anchors. We prove: (i) an exact
conditional risk identity for the fitted student, with the in-sample transmitted
Gram shown to be a different quadratic; (ii) functionwise learning deviations
$\norm{(\widehat T_\lambda-T_\lambda)h}$ via a residual normal equation, giving
entrywise transfer of the transmitted-error geometry and of risks and gains,
and spectral rates under source and effective-dimension conditions; (iii) for
downstream aggregation with an estimated Gram, a deterministic local oracle
inequality $R(\widehat\beta)-R(\beta)\le\norm{\Pi_{\mathcal T_\beta}H_\alpha^{-1/2}\zeta_\beta}^2/(1-2\eta)$
whose stochastic terms scale as $d_1(\alpha)(v^2/n+M^2/N)$, with no inverse
smallest-eigenvalue factor, an exact affine variance governed by $d_2(\alpha)$,
and a structured version whose anchor term is the regularized unresolved trace
$d_\Nn(\alpha)\to d_\Nn$; (iv) uniform finite-sample gain certificates; and
(v) an end-to-end comparator inequality whose learning contribution is
first-order in general and quadratic under an orthogonality condition, with a
counterexample showing the first-order term can occur.

\paragraph{Where the main text uses this appendix.}
Formal versions of the main-text results; supporting results are cited in their proofs.

\begin{center}\small
\begin{tabularx}{\textwidth}{@{}>{\raggedright}p{0.24\textwidth}X@{}}
\toprule
Main-text result & Formal results\tabularnewline
\midrule
\cref{thm:e2e} & \cref{q4:thm:master} (p.~\pageref{q4:thm:master}); \cref{q4:thm:fast} (p.~\pageref{q4:thm:fast}); \cref{q4:thm:affine} (p.~\pageref{q4:thm:affine}); \cref{q4:thm:structured} (p.~\pageref{q4:thm:structured}); \cref{q4:thm:closure} (p.~\pageref{q4:thm:closure}); \cref{q4:thm:integrated} (p.~\pageref{q4:thm:integrated})\tabularnewline
\bottomrule
\end{tabularx}
\end{center}

\paragraph{Notation.} As in \cref{app:notation,app:translation}.

\begin{figure}[h]
\centering
\begin{tikzpicture}[
  font=\footnotesize, node distance=2mm and 6mm, inner sep=2pt,
  smp/.style={draw, rounded corners=2pt, fill=black!6, align=center, text width=27mm},
  obj/.style={draw, align=center, text width=34mm},
  outb/.style={draw, thick, align=center, text width=40mm},
  ->, >=Stealth]
\node[smp] (D) {$D$: $\Nfit$ pseudo-labelled inputs};
\node[smp, below=of D] (U) {$U$: $N$ unlabeled inputs};
\node[smp, below=of U] (A) {$A$: $n$ clean anchors};
\node[obj, right=of D] (T) {fitted $\widehat T_\lambda$, candidates $\widehat s_j$};
\node[obj, right=of U] (K) {candidate Gram $\widehat C_x$ ($\widehat\Gamma_U$)};
\node[obj, right=of A] (M) {target score $\widehat a$ (moments $\widehat\mu_h$)};
\node[outb, right=of K] (W) {weights $\widehat w$ from $\beta^\top\widehat C_x\beta-2\widehat a^\top\beta+\alpha\norm\beta^2$};
\node[outb, right=of W] (R) {$R_D(\widehat w)$ vs.\ $R_\lambda(w^\circ)$; gain certificates};
\draw (D) -- (T); \draw (U) -- (K); \draw (A) -- (M);
\draw (T.east) -- (W); \draw (K) -- (W); \draw (M.east) -- (W); \draw (W) -- (R);
\draw (T) -- (K);
\end{tikzpicture}

\medskip
\begin{tikzpicture}[
  font=\footnotesize, node distance=2.5mm and 6mm, inner sep=2pt,
  res/.style={draw, align=center, text width=40mm},
  key/.style={draw, thick, align=center, text width=40mm},
  ->, >=Stealth]
\node[res] (L) {residual normal equation \cref{q4:lem:residual,q4:thm:learning}};
\node[res, right=of L] (Gate) {stability gate \cref{q4:thm:gate}};
\node[res, right=of Gate] (Aff) {affine identity, $d_2$ \cref{q4:thm:affine}};
\node[res, below=of L] (G) {geometry, risk, gain transfer \cref{q4:thm:geometry,q4:prop:risk-transfer}};
\node[key, below=of Gate] (Mst) {local conic oracle inequality \cref{q4:thm:master}};
\node[res, below=of Aff] (Str) {structured rate, $\dN(\alpha)$ \cref{q4:thm:structured,q4:prop:dN}};
\node[res, below=of G] (Cert) {gain certificate \cref{q4:thm:gain4}};
\node[res, below=of Mst] (Fast) {fast rate; safe $\alpha$ \cref{q4:thm:fast,q4:cor:safe}};
\node[key, below=of Fast] (Cl) {end-to-end closure \cref{q4:thm:closure,q4:thm:integrated}};
\draw (L) -- (G); \draw (G) -- (Cert);
\draw (Gate) -- (Mst); \draw (Gate) -- (Aff); \draw (Mst) -- (Fast); \draw (Mst) -- (Str);
\draw (Fast) -- (Cl); \draw (Cert) |- ([yshift=-1mm]Cl.west); \draw (Str) |- (Cl);
\draw (G.west) -- ++(-3mm,0) |- ([yshift=1mm]Cl.west);
\end{tikzpicture}
\caption{\emph{Top:} sample flow; the three samples are independent, and
the fitted candidates enter both the Gram (integrated over $U$) and the
score (over $A$). \emph{Bottom:} dependencies among the main results. Left:
the fitted learner (stage~1); middle: selection with an estimated Gram
(stage~2); right: refinements (stage~3). Thick boxes are the local oracle
inequality and the closure behind \cref{thm:e2e}. Alongside the local fast
bound, \cref{q4:thm:slow} is a robust $N^{-1/2}$-type bound under finite
moments; the safe regime is a checkable sufficient condition; the affine
identity is an exact refinement; the structured rate replaces the ambient
anchor complexity by $\dN(\alpha)$.}
\label{q4:fig:guide}
\end{figure}
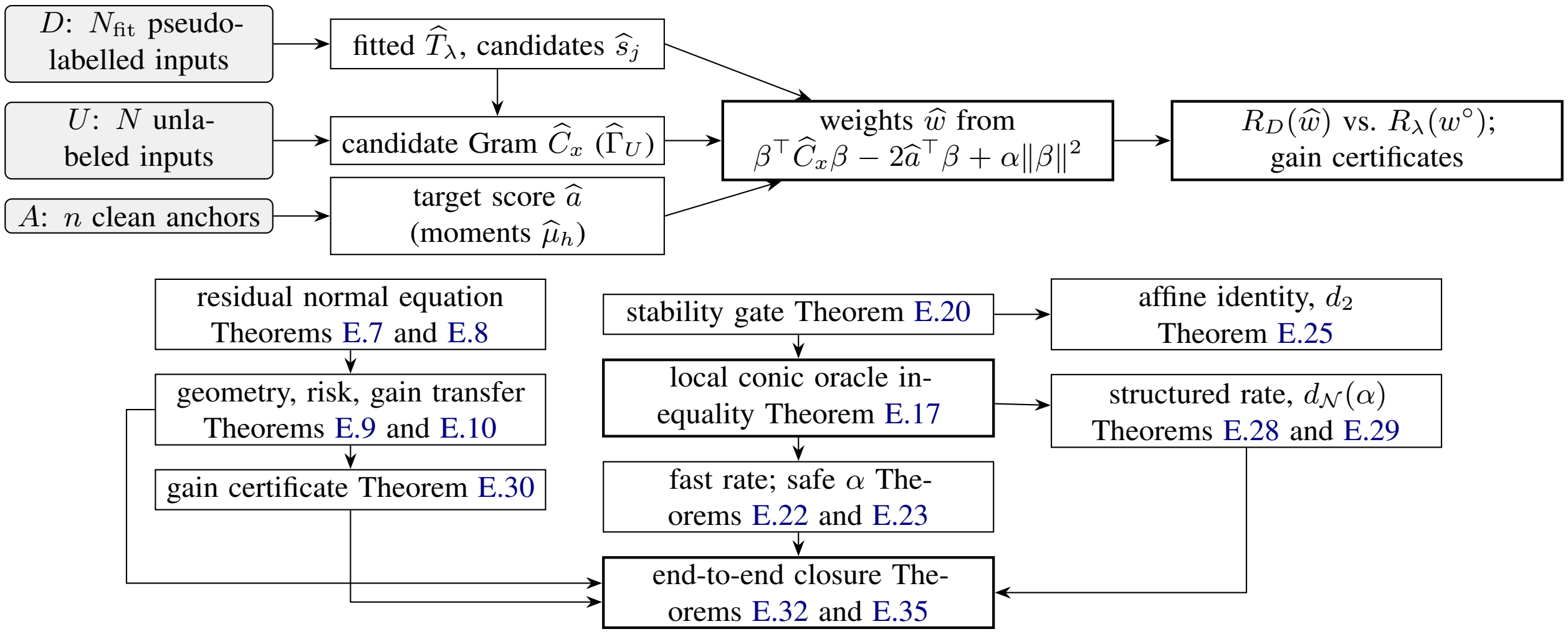

\paragraph{Reading guide.}
This appendix follows three independent samples through the pipeline
(\cref{q4:fig:guide}, top) and proves one end-to-end inequality from a short
chain of results (bottom), in four stages marked by banners. Spectral
learning rates, Gram diagnostics, same-sample dependence, kernels, random
features and linearization are summarized in \cref{app:ext}.

\paragraph{Notation by stage.}
\begin{center}\footnotesize
\begin{tabularx}{\textwidth}{@{}>{\raggedright}p{0.12\textwidth}>{\raggedright}p{0.3\textwidth}X@{}}
\toprule
Stage & Symbols & Meaning\tabularnewline
\midrule
1. Fitting & $\psi,\Psi$; $C_\psi$, $C_D$; $\lambda$; $T_\lambda$, $\widehat T_\lambda$; $\deff$, $\varkappa_\psi$, $\eta_D$ & representation; population and empirical feature operators; student ridge; population and fitted learners; effective dimension, leverage, relative design error\tabularnewline
& $\Gamma_\lambda$, $\Gamma_D$, $\Gamma_D^{\mathrm{in}}$, $\widehat\Gamma_U$; $\varepsilon_{\mathrm{op}}$ & transmitted Grams of $T_\lambda$ over $P_X$, of $\widehat T_\lambda$ over $P_X$ (deployed risk), of $\widehat T_\lambda$ over the training inputs (in-sample; \emph{not} the deployed risk, \cref{q4:prop:gap}); estimator of $\Gamma_D$ from $U$; risk-transfer error\tabularnewline
2. Selection & $w=w_0+O\beta$, $s_0$, $x$; $C_x$, $\widehat C_x$, $a$, $\widehat a$; $\alpha$, $H_\alpha$, $d_1$, $d_2$, $\varkappa$, $\eta_\alpha$ & contrast coordinates; population and estimated Gram and score; aggregation ridge, regularized Gram, effective dimensions, leverage, relative Gram error\tabularnewline
3. Structure & $f_\Kk$, $\Nn$, $\dN(\alpha)$; $\epsilon_G$ & structured reference, residual space, regularized unresolved trace; certificate radius\tabularnewline
4. Closure & $s_j^\circ$, $R_\circ=R_\lambda$, $\Delta_{w^\circ}$, $\epsilon_D$ & population candidates and their risk; fitting perturbation and its bound\tabularnewline
\bottomrule
\end{tabularx}
\end{center}

\apppart{Stage 1: fitting the student (sample $D$)}{The fitted learner, its exact conditional identities and its deviation from the population learner.}
\subsection{Model, learner and sampling protocol}\label{q4:sec:model}

\paragraph{Role in the paper.} \cref{app:q4} answers the fourth question: \emph{do the
population conclusions survive finite samples, and at what cost?} The
contribution boundary is explicit: ridge normal equations, effective-dimension
rates, quadratic aggregation, matrix concentration and kernel resolvent
perturbation are standard \citep{hsu2012random,tropp2012user}. What is specific
here is the decomposition of a weak-to-strong pipeline into student fitting,
unlabeled integration and anchor selection, with the exact cross terms that
connect them.

\paragraph{Population objects.} $\Hh=L^2(P_X)$, supervisors $g_j=f+e_j$,
$E w=\sum_jw_je_j$, $\Sigma=E^*E$, $\WL$ and $\Delta_m=\mathcal W_1$. Except in
explicitly affine statements, $\mathcal W\subseteq\WL$ is nonempty compact convex.
A separable Hilbert space $\Ff$ and a measurable feature map $\psi:\mathcal X\to\Ff$
with $\norm{\psi(X)}_\Ff\le\kappa$ a.s.\ define the synthesis
$\Psi:\Ff\to\Hh$, $(\Psi\vartheta)(x)=\ip{\psi(x)}\vartheta_\Ff$, bounded by $\kappa$, with
\[
 \Psi^*h=\E[\psi(X)h(X)],\qquad C_\psi=\Psi^*\Psi=\E[\psi\otimes\psi],\qquad\mathcal L=\Psi\Psi^*,
\]
$\Ss=\overline{\Ran\Psi}$, $P=P_\Ss$, $Q=I-P$, $A(f)=\norm{Qf}^2$. ($C_\psi$ is the
feature second-moment operator, not the error Gram $\Sigma$.) For $\lambda>0$,
$C_{\psi,\lambda}=C_\psi+\lambda I$, $\vartheta_\lambda(h)=C_{\psi,\lambda}^{-1}\Psi^*h$, and the
population ridge learner
\[
 T_\lambda h=\Psi\vartheta_\lambda(h)=\Psi(C_\psi+\lambda I)^{-1}\Psi^*h=\mathcal L(\mathcal L+\lambda I)^{-1}h,
 \qquad0\preceq T_\lambda\preceq I,\ \Ran T_\lambda\subseteq\Ss .
\]
With $T=T_\lambda$ in \cref{app:q1}'s notation: $b_\lambda=(I-T_\lambda)f$, $B_\lambda=\norm{b_\lambda}^2$,
$(\Gamma_\lambda)_{jk}=\ip{T_\lambda e_j}{T_\lambda e_k}$, $(c_\lambda)_j=\ip{T_\lambda e_j}{b_\lambda}$, and
$R_\lambda(w)=B_\lambda+w^\top\Gamma_\lambda w-2c_\lambda^\top w$,
$G_\lambda(w)=w^\top\Sigma w-R_\lambda(w)$ (\cref{q1:thm:single-general}).

\paragraph{Sampling protocol.} Three independent samples:
$D=(X_i)_{i\le N_{\mathrm{fit}}}$ trains the student on pseudo-labels;
$U=(X^U_i)_{i\le N}$ estimates prediction-only moments; anchors
$A=((X^A_i,Y_i))_{i\le n}$, $Y=f(X^A)+\xi$, $\E[\xi\mid X^A]=0$, estimate target
moments. Bounded-label results assume $\abs Y\le M_y$. Feature maps,
supervisors and hyperparameters may be random but are built independently of
$D,U,A$ (a sigma-field $\mathcal F_0$); all statements are conditional on
$\mathcal F_0$. Inputs $\psi(X_i)$ and $g_j(X_i)$ are generally dependent;
independence is only across sampled inputs once functions are frozen.
Dependence beyond this protocol is \cref{ext:scope}.

\paragraph{Regularization symbols.} $\lambda$ is the student-training ridge;
$\alpha$ is the downstream aggregation ridge. They play different roles and
are never identified.

\subsection{The empirical learner and exact conditional identities}\label{q4:sec:empirical}

Let $S_D:\Ff\to\R^{N_{\mathrm{fit}}}$,
$S_D\vartheta=N_{\mathrm{fit}}^{-1/2}(\ip{\psi(X_i)}\vartheta)_i$, and for a specified
pointwise representative $h_D=N_{\mathrm{fit}}^{-1/2}(h(X_i))_i$,
$\norm h_D=\norm{h_D}_2$; $C_D=S_D^*S_D$. The fitted coefficient and its
population prediction are
\[
 \widehat\vartheta_\lambda(h)=(C_D+\lambda I)^{-1}S_D^*h_D,\qquad\widehat T_\lambda h=\Psi\widehat\vartheta_\lambda(h)
\]
(random-design ridge for $\Ff=\R^d$, kernel ridge for $\psi(x)=k(x,\cdot)$).

\begin{proposition}[Optimization characterization]\label{q4:prop:opt}
$\widehat\vartheta_\lambda(h)$ uniquely minimizes $\norm{S_D\vartheta-h_D}_2^2+\lambda\norm\vartheta^2$,
$\norm{\widehat\vartheta_\lambda(h)}\le\lambda^{-1/2}\norm h_D$ and
$\norm{\widehat T_\lambda h}\le\kappa\lambda^{-1/2}\norm h_D$. The in-sample smoother
$H_D=S_D(C_D+\lambda I)^{-1}S_D^*$ satisfies $0\preceq H_D\preceq I$.
\end{proposition}
\begin{proof}
The normal equation is $(C_D+\lambda I)\vartheta=S_D^*h_D$, and $C_D+\lambda I\succeq\lambda I$.
Comparing the objective with its value at $0$ gives
$\lambda\norm{\widehat\vartheta}^2\le\norm h_D^2$; $\norm\Psi_\op\le\kappa$. The smoother eigenvalues
are $\varsigma^2/(\varsigma^2+\lambda)\in[0,1)$ for singular values $\varsigma$ of $S_D$.
\end{proof}

\begin{remark}[No unrestricted empirical operator]\label{q4:auto:no-unrestricted-empirical-operator}
On a nonatomic $P_X$ point evaluation is not defined on $L^2$ classes, so
$\widehat T_\lambda$ is a sampling-and-fitting map on \emph{specified functions}, not a
bounded operator on $\Hh$; all empirical-to-population bounds concern
specified functions or a specified finite-dimensional space.
\end{remark}

Given $D$, put $\widehat s_j=\widehat T_\lambda g_j$, $\widehat s_w=\sum_jw_j\widehat s_j$,
$\widehat b=f-\widehat T_\lambda f$, $\widehat B=\norm{\widehat b}^2$,
$(\Gamma_D)_{jk}=\ip{\widehat T_\lambda e_j}{\widehat T_\lambda e_k}$,
$(c_D)_j=\ip{\widehat T_\lambda e_j}{\widehat b}$ (population inner products).

\begin{theorem}[Conditional population-risk identity]\label{q4:thm:cond-risk}
For every realized design and affine $w$,
\[
 \boxed{R_D(w):=\norm{\widehat s_w-f}^2=\widehat B+w^\top\Gamma_Dw-2c_D^\top w,\qquad
 G_D(w)=w^\top\Sigma w-R_D(w),}
\]
and $R_D(w)=A(f)+\norm{\widehat s_w-Pf}^2$: $A(f)$ lower-bounds the fitted
student's risk, while $\widehat B$ need not.
\end{theorem}
\begin{proof}
$\widehat s_w-f=\widehat T_\lambda Ew-\widehat b$ by linearity and $\one^\top w=1$; expand.
$\widehat s_w\in\Ss$ gives the orthogonal split $(\widehat s_w-Pf)-Qf$.
\end{proof}

\paragraph{Training risk is a different quadratic.} The in-sample risk
$R_D^{\mathrm{in}}(w)=\norm{H_D(g_w)_D-f_D}_2^2$ expands with
$b_D^{\mathrm{in}}=(I-H_D)f_D$, $(\Gamma^{\mathrm{in}}_D)_{jk}=\ip{H_D(e_j)_D}{H_D(e_k)_D}$ and
$(c^{\mathrm{in}}_D)_j=\ip{H_D(e_j)_D}{b^{\mathrm{in}}_D}$; with
$\vartheta_j=\widehat\vartheta_\lambda(e_j)$,
$(\Gamma_D-\Gamma^{\mathrm{in}}_D)_{jk}=\ip{\vartheta_j}{(C_\psi-C_D)\vartheta_k}$. More generally:

\begin{proposition}[Exact generalization gap]\label{q4:prop:gap}
For every $\vartheta\in\Ff$, possibly depending on $D$,
$\norm{\Psi\vartheta-f}^2-\norm{S_D\vartheta-f_D}^2=\ip\vartheta{(C_\psi-C_D)\vartheta}-2\ip\vartheta{\Psi^*f-S_D^*f_D}+\norm f^2-\norm f_D^2$.
\par{\upshape\small Immediate by expanding both squares; the identity is pointwise, so no independence is used.}
\end{proposition}

\subsection{Learning deviation of the fitted student}\label{q4:sec:learning}

\begin{lemma}[Hilbert-valued sample mean]\label{q4:lem:hilbert}
If $Z_1,\dots,Z_q$ are iid in a separable Hilbert space, $\norm Z\le M$ and
$\E\norm Z^2\le v$, then with probability $\ge1-e^{-t}$,
$\norm{q^{-1}\sum_iZ_i-\E Z}\le\sqrt{v/q}+M\sqrt{2t/q}$. For scalars in an interval
of length $\ell$, $\Prob(\abs{\bar Z-\E Z}>\ell\sqrt{t/(2q)})\le2e^{-t}$.
\end{lemma}
\begin{proof}
$W=\norm{q^{-1}\sum(Z_i-\E Z)}$ has $\E W\le(\E W^2)^{1/2}\le\sqrt{v/q}$. Replacing one
observation changes $W$ by at most $2M/q$; the Doob martingale with
conditional Hoeffding increments (\cref{q3:lem:hoeffding}, restated: an increment in an interval of length $c$ has
$\E e^{s\Delta}\le e^{s^2c^2/8}$) gives $\Prob(W-\E W\ge u)\le e^{-2u^2/(4M^2/q)}$; take
$u=M\sqrt{2t/q}$. The scalar case is Hoeffding's inequality.
\end{proof}

\paragraph{Leverage and effective dimension.} With $z_\lambda(x)=C_{\psi,\lambda}^{-1/2}\psi(x)$,
\[
 \varkappa_\psi(\lambda)=\esssup_x\norm{z_\lambda(x)}^2,\qquad\deff(\lambda)=\E\norm{z_\lambda(X)}^2=\tr(C_\psi C_{\psi,\lambda}^{-1}),
 \qquad\deff(\lambda)\le\varkappa_\psi(\lambda)\le\kappa^2/\lambda,
\]
and $\Delta_D=C_{\psi,\lambda}^{-1/2}(C_D-C_\psi)C_{\psi,\lambda}^{-1/2}$, $\eta_D=\norm{\Delta_D}_\op$.

\begin{proposition}[Design event]\label{q4:prop:design}
With probability $\ge1-e^{-t}$,
$\eta_D\le\eta_{N_{\mathrm{fit}}}(\lambda,t):=\sqrt{\varkappa_\psi\deff/N_{\mathrm{fit}}}+\varkappa_\psi\sqrt{2t/N_{\mathrm{fit}}}$.
\end{proposition}
\begin{proof}
Lemma~\ref{q4:lem:hilbert} in the Hilbert--Schmidt space with $Z=z_\lambda\otimes z_\lambda$,
$\norm Z_\HS\le\varkappa_\psi$, $\E\norm Z_\HS^2=\E\norm{z_\lambda}^4\le\varkappa_\psi\deff$, and
$\norm\cdot_\op\le\norm\cdot_\HS$.
\end{proof}

This elementary gate needs roughly $N_{\mathrm{fit}}\gtrsim\deff^2\log(1/\delta)$ when
$\varkappa_\psi\lesssim\deff$; the matrix-Bernstein gate of Theorem~\ref{q4:thm:gate} is
sharper when $\Ff$ is finite-dimensional.

\begin{lemma}[Residual normal equation]\label{q4:lem:residual}
With $r_\lambda(h)=h-T_\lambda h$ and
$\xi_D(h)=N_{\mathrm{fit}}^{-1}\sum_i\psi(X_i)r_\lambda(h)(X_i)-\E[\psi r_\lambda(h)]$,
$\widehat\vartheta_\lambda(h)-\vartheta_\lambda(h)=(C_D+\lambda I)^{-1}\xi_D(h)$ for every design.
\end{lemma}
\begin{proof}
$\E[\psi r_\lambda(h)]=\Psi^*h-C_\psi\vartheta_\lambda(h)=\lambda\vartheta_\lambda(h)$, and
$(C_D+\lambda I)(\widehat\vartheta-\vartheta_\lambda)=N_{\mathrm{fit}}^{-1}\sum\psi(X_i)h(X_i)-C_D\vartheta_\lambda-\lambda\vartheta_\lambda=\xi_D(h)$.
\end{proof}

\begin{theorem}[Functionwise learning deviation]\label{q4:thm:learning}
If $\eta_D<1$, then
$\norm{(\widehat T_\lambda-T_\lambda)h}\le\norm{C_{\psi,\lambda}^{-1/2}\xi_D(h)}/(1-\eta_D)$. If $h$ is fixed
independently of $D$ with $\norm{r_\lambda(h)}_\infty\le R_h$, then except with
probability $2e^{-t}$, provided $\eta_{N_{\mathrm{fit}}}(\lambda,t)<1$,
\[
 \boxed{\norm{(\widehat T_\lambda-T_\lambda)h}\le\varepsilon_h:=\frac{R_h}{1-\eta_{N_{\mathrm{fit}}}(\lambda,t)}
 \Big[\sqrt{\frac{\deff(\lambda)}{N_{\mathrm{fit}}}}+\sqrt{\frac{2\varkappa_\psi(\lambda)t}{N_{\mathrm{fit}}}}\Big].}
\]
\end{theorem}
\begin{proof}
$C_D+\lambda I=C_{\psi,\lambda}^{1/2}(I+\Delta_D)C_{\psi,\lambda}^{1/2}$, $\norm{(I+\Delta_D)^{-1}}_\op\le(1-\eta_D)^{-1}$
and $\norm{\Psi C_{\psi,\lambda}^{-1/2}}_\op\le1$; apply Lemma~\ref{q4:lem:residual}. For
concentration use Lemma~\ref{q4:lem:hilbert} with $Z=z_\lambda r_\lambda(h)$:
$\norm Z\le R_h\sqrt{\varkappa_\psi}$, $\E\norm Z^2\le R_h^2\deff$; union with
Proposition~\ref{q4:prop:design} (no independence between the events is needed).
\end{proof}

For fixed $h_1,\dots,h_q$, one design event and $q$ residual events with
$t=\log((q+1)/\delta)$ hold jointly with probability $\ge1-\delta$; then
$\norm{(\widehat T_\lambda-T_\lambda)\sum_\ell a_\ell h_\ell}\le\sum_\ell\abs{a_\ell}\varepsilon_{h_\ell}$, and for an
orthonormal basis $(v_\ell)$ of a fixed finite-dimensional $V'$,
$\norm{(\widehat T_\lambda-T_\lambda)|_{V'}}_\op\le(\sum_\ell\varepsilon_{v_\ell}^2)^{1/2}$ (Cauchy--Schwarz).

\subsection{Transfer of transmitted-error geometry, risks and gains}\label{q4:sec:transfer}

Let $\delta_j=(\widehat T_\lambda-T_\lambda)e_j$, $\delta_f=(\widehat T_\lambda-T_\lambda)f$ with
$\norm{\delta_j}\le\varepsilon_j$, $\norm{\delta_f}\le\varepsilon_f$, $a_j=\norm{T_\lambda e_j}$, $b=\norm{b_\lambda}$.

\begin{theorem}[Entrywise geometry transfer]\label{q4:thm:geometry}
$\abs{(\Gamma_D-\Gamma_\lambda)_{jk}}\le a_j\varepsilon_k+a_k\varepsilon_j+\varepsilon_j\varepsilon_k$,
$\abs{(c_D-c_\lambda)_j}\le b\varepsilon_j+a_j\varepsilon_f+\varepsilon_j\varepsilon_f$ and
$\abs{\widehat B-B_\lambda}\le2b\varepsilon_f+\varepsilon_f^2$. With $D_Ew=\sum_jw_j\delta_j$ and
$d_E=\norm{D_E}_\op\le(\sum_j\varepsilon_j^2)^{1/2}$,
$\norm{\Gamma_D-\Gamma_\lambda}_\op\le2\norm{T_\lambda E}_\op d_E+d_E^2$.
\end{theorem}
\begin{proof}
Expand $\ip{T_\lambda e_j+\delta_j}{T_\lambda e_k+\delta_k}$, use $\widehat b=b_\lambda-\delta_f$, and
Cauchy--Schwarz; for the operator bound expand
$(T_\lambda E+D_E)^*(T_\lambda E+D_E)-(T_\lambda E)^*T_\lambda E$.
\end{proof}

\begin{proposition}[Direct risk and gain transfer]\label{q4:prop:risk-transfer}
If $\max_j\norm{(\widehat T_\lambda-T_\lambda)g_j}\le\varepsilon_g$ and
$K_\lambda=\sup_{w\in\mathcal W}\norm{T_\lambda g_w-f}$, then
\[
 \boxed{\sup_{w\in\mathcal W}\abs{R_D(w)-R_\lambda(w)}=\sup_{w\in\mathcal W}\abs{G_D(w)-G_\lambda(w)}\le\varepsilon_{\mathrm{op}}:=2K_\lambda L\varepsilon_g+L^2\varepsilon_g^2.}
\]
\end{proposition}
\begin{proof}
With $u=T_\lambda g_w-f$ and $d=\widehat s_w-T_\lambda g_w$, $\norm d\le L\varepsilon_g$ and
$\norm{u+d}^2-\norm u^2=2\ip ud+\norm d^2$; the weak risk is common to both gains.
\end{proof}

\paragraph{What an empirical $\widehat\Gamma$ estimates.} Three objects must not be
interchanged: $\Gamma_\lambda$ (population learner, population integration),
$\Gamma_D$ (fitted learner, population integration) and $\Gamma_D^{\mathrm{in}}$ (fitted
learner, training integration). Even $\Gamma_D$ is unobserved, because
$\widehat T_\lambda e_j=\widehat s_j-\widehat T_\lambda f$ contains the unknown clean fitted student. If
the functions $\widehat T_\lambda e_j$ were available, frozen and bounded by $M_h$, the
empirical Gram on $U$ would satisfy
$\norm{\widehat\Gamma_U-\Gamma_D}_{\max}\le M_h^2\sqrt{2\log(2m^2/\delta)/N}$ (Hoeffding and a union
bound) --- an oracle diagnostic, not a label-free algorithm.

\begin{proposition}[Proxy bias in the transmitted Gram]\label{q4:prop:proxy}
Let $\tilde t$ be a proxy for $\widehat T_\lambda f$ built independently of $U$, $\varpi=\tilde t-\widehat T_\lambda f$
with $\norm\varpi\le\tau_p$, $h_j=\widehat T_\lambda e_j$ and $\tilde h_j=\widehat s_j-\tilde t=h_j-\varpi$. Then
$\abs{\ip{\tilde h_j}{\tilde h_k}-(\Gamma_D)_{jk}}\le\tau_p(\norm{h_j}+\norm{h_k})+\tau_p^2$; if
$\norm{\tilde h_j}_\infty\le M_{\tilde h}$, the empirical Gram of $\tilde h_j$ on $U$ is within
$a_j\varepsilon_k+a_k\varepsilon_j+\varepsilon_j\varepsilon_k+\tau_p(\norm{h_j}+\norm{h_k})+\tau_p^2+M_{\tilde h}^2\sqrt{2\log(2m^2/\delta)/N}$
of $(\Gamma_\lambda)_{jk}$ on the learning and evaluation events. A small proxy error
is an additional requirement; selection (Section~\ref{q4:sec:agg}) avoids
estimating $\widehat T_\lambda f$ altogether.
\end{proposition}
\begin{proof}
$\ip{h_j-\varpi}{h_k-\varpi}-\ip{h_j}{h_k}=-\ip{h_j}\varpi-\ip\varpi{h_k}+\norm\varpi^2$;
combine with Theorem~\ref{q4:thm:geometry} and entrywise Hoeffding.
\end{proof}

\apppart{Stage 2: selection with an estimated Gram (samples $U$ and $A$)}{Slow and local fast bounds, the safe regime and the affine refinement.}
\subsection{Downstream aggregation: coordinates and population-geometry baselines}\label{q4:sec:agg}

From here on the student predictions $s_1,\dots,s_m$ ($m\ge2$) are frozen given
$\mathcal F_0$: $s_j=Pg_j$, $T g_j$, or $\widehat s_j=\widehat T_\lambda g_j$. No projection or
contraction property is used; the only operational identity is
$s_w=\sum_jw_js_j$, valid for a common linear learner with fixed design and
hyperparameters (not for a nonlinear learner retrained per $w$).

\paragraph{Coordinates.} Fix $w_0\in\mathcal W$ independently of $U,A$ (e.g.\ a
pilot on independent data) and $O\in\R^{m\times(m-1)}$ with $O^\top O=I$, $O^\top\one=0$.
Put $s_0=s_{w_0}$, $x(X)=O^\top(s_1(X),\dots,s_m(X))^\top$,
$\mathcal B=\{\beta:w_0+O\beta\in\mathcal W\}$ ($\R^{m-1}$ or compact convex), so
$s_{w_0+O\beta}=s_0+x^\top\beta$, and
\[
 C_x=\E[xx^\top],\qquad a=\E[x(Y-s_0(X))]=\E[x(f-s_0)],\qquad
 R(\beta)=\norm{s_0-f}^2+\beta^\top C_x\beta-2a^\top\beta .
\]
Neither $\norm f^2$ nor the noise variance is needed.

\begin{lemma}[Singular coordinates]\label{q4:lem:quotient}
$\rank C_x=k=\dim V$. If $v\in\Ker C_x$ then $x^\top v=0$ a.s.\ and $v^\top a=0$. For
affine weights $\beta_*=C_x^\dagger a$ is optimal and
$R(\beta)-\inf R=\norm{C_x^{1/2}(\beta-\beta_*)}^2$; on a compact convex domain the optimal
prediction is unique (its weight need not be). Replacing $O$ by $OO'$ with $O'$
orthogonal rotates $\beta$ and leaves spectra, leverages and Euclidean penalties
unchanged.
\end{lemma}
\begin{proof}
$v\mapsto x^\top v$ has Gram $C_x$ and range $V$; $v^\top C_xv=0$ forces $x^\top v=0$ a.s.,
so $a\in\Ran C_x$ and $C_x\beta_*=a$; substitute $\beta=\beta_*+d$. Strict convexity of
the squared norm in the prediction gives uniqueness.
\end{proof}

\begin{lemma}[Uniform-objective regret]\label{q4:lem:slow}
For $q(w)=w^\top Kw-2\mu^\top w$ with $K\succeq0$ and an estimate $\widehat q$ built from
$\widehat K,\widehat\mu$ with $\norm{\widehat K-K}_{\max}\le\alpha_K$, $\norm{\widehat\mu-\mu}_\infty\le\alpha_\mu$, any
minimizer $\widehat w$ of $\widehat q$ on $\mathcal W\subseteq\WL$ has
$q(\widehat w)-\min_{\mathcal W}q\le2L^2\alpha_K+4L\alpha_\mu$.
\end{lemma}
\begin{proof}
$\abs{\widehat q-q}\le\norm w_1^2\alpha_K+2\norm w_1\alpha_\mu=:\varepsilon$ on $\WL$, and
$q(\widehat w)\le\widehat q(\widehat w)+\varepsilon\le\widehat q(w^\star)+\varepsilon\le q(w^\star)+2\varepsilon$.
\end{proof}

\begin{theorem}[Finite-moment slow bound]\label{q4:thm:slow}
Let $\widehat K_{jk}=N^{-1}\sum_is_j(X^U_i)s_k(X^U_i)$ and $\widehat\mu_j=n^{-1}\sum_iY_is_j(X^A_i)$, with
$\abs{s_j}\le M_s$ and $\abs Y\le M_y$. With probability $\ge1-\delta$,
$\norm{\widehat K-K_s}_{\max}\le\alpha_K=M_s^2\sqrt{2\log(4m^2/\delta)/N}$,
$\norm{\widehat\mu-\mu_s}_\infty\le\alpha_\mu=M_yM_s\sqrt{2\log(4m/\delta)/n}$, and the plug-in minimizer
satisfies $R_T(\widehat w)-\min_{\mathcal W}R_T\le2L^2\alpha_K+4L\alpha_\mu$. The same bound with
$h=q$ ($K_g-K_s$, $\mu_q$) gives simultaneous gain intervals (Section~\ref{q4:sec:gain4}).
\end{theorem}
\begin{proof}
$R_T(w)=w^\top K_sw-2\mu_s^\top w+\norm f^2$; Hoeffding entrywise
(Lemma~\ref{q4:lem:hilbert}), $\E[Ys_j(X)]=\ip f{s_j}$, union bounds with $\delta/2$
each, and Lemma~\ref{q4:lem:slow}.
\end{proof}

This bound is transparent but slow ($N^{-1/2}$ and $n^{-1/2}$), and $M_s$ can
scale as $\kappa M_g\lambda^{-1/2}$ for ridge students (Proposition~\ref{q4:prop:opt}).

\begin{proposition}[Prediction-space bound with population geometry]\label{q4:prop:pred-space}
If the prediction-only inner products are known exactly and only the target
coordinate $\theta=\Phi^*f$ on an orthonormal basis of $V$ is estimated by
$\widehat\theta$, then $\Reg_{\mathcal W}\le\norm{\widehat\theta-\theta}^2$ (\cref{q3:thm:cone}). If the prediction-only part of the objective is known only up to a
uniform error $\epsilon_u$, then $\Reg_{\mathcal W}\le4\norm{\widehat\theta-\theta}^2+4\epsilon_u$.
\end{proposition}
\begin{proof}
For the second claim let $x=\Reg$, $d=t(\widehat w)-t^\star$, $\zeta=\norm{\widehat\theta-\theta}$. The projection inequality
gives $x\ge\norm d^2$; estimated optimality gives
$x\le2\norm{\widehat\theta-\theta}\norm d+2\epsilon_u\le2\zeta\sqrt x+2\epsilon_u\le x/2+2\zeta^2+2\epsilon_u$.
\end{proof}

\begin{example}[Equal spectrum, different coverage]\label{q4:ex:coverage}
In one dimension compare $x=\varepsilon Z$ ($Z$ Rademacher) with $x=\one_E$,
$\Prob(E)=\varepsilon^2$: both have $C_x=\varepsilon^2$ and identical effective dimensions. In
the first $\widehat C_x=C_x$ always; in the second $\Prob(\widehat C_x=0)=(1-\varepsilon^2)^N$ and the
leverage is $1/(\varepsilon^2+\alpha)$. With $s_0=0$, $Y=x$ and even the exact score
$a=C_x$, on $\{\widehat C_x=0\}$ the unregularized criterion is unbounded below, a
pseudoinverse returns $0$ (regret $\varepsilon^2$) and ridge returns $C_x/\alpha$ (regret
$C_x(C_x/\alpha-1)^2$). Neither the population spectrum nor empirical
invertibility controls finite-sample coverage; regularized leverage does.
\end{example}

\subsection{A local oracle inequality with an estimated Gram}\label{q4:sec:local}

For $\alpha>0$ let $H_\alpha=C_x+\alpha I$,
$\widehat C_x=N^{-1}\sum_ix(X^U_i)x(X^U_i)^\top$, $\widehat H_\alpha=\widehat C_x+\alpha I$,
$\widehat a=n^{-1}\sum_ix(X^A_i)(Y_i-s_0(X^A_i))$, and
\[
 \widehat\beta_\alpha=\argmin_{\beta\in\mathcal B}\{\beta^\top\widehat H_\alpha\beta-2\widehat a^\top\beta\},\qquad\widehat w=w_0+O\widehat\beta_\alpha
\]
(unique; $\widehat H_\alpha^{-1}\widehat a$ in the affine case). With positive eigenvalues
$\mu_1,\dots,\mu_k$ of $C_x$,
\[
 d_1(\alpha)=\tr(C_xH_\alpha^{-1})=\sum_j\frac{\mu_j}{\mu_j+\alpha},\qquad
 d_2(\alpha)=\tr[(C_xH_\alpha^{-1})^2],\qquad
 \varkappa(\alpha)=\esssup_xx^\top H_\alpha^{-1}x,
\]
\[
 \eta_\alpha=\norm{H_\alpha^{-1/2}\Delta_UH_\alpha^{-1/2}}_\op,
\]
$\Delta_U=\widehat C_x-C_x$; $0\le d_2\le d_1\le k$, $d_1\le\varkappa$, and $\eta_\alpha\le\eta<1$ gives
$(1-\eta)H_\alpha\preceq\widehat H_\alpha\preceq(1+\eta)H_\alpha$. No lower bound on $\mu_k$ is assumed;
$\norm v_M^2=v^\top Mv$.

\begin{theorem}[Local conic oracle inequality]\label{q4:thm:master}
For any PSD $\widehat C_x$, any score $\widehat a$ and any comparator $\beta\in\mathcal B$, let
$\zeta_\beta=(\widehat a-a)-\Delta_U\beta-\alpha\beta$ and
$\mathcal T_\beta=\overline{\cone\{H_\alpha^{1/2}(\beta'-\beta):\beta'\in\mathcal B\}}$. If $\eta_\alpha\le\eta<\tfrac12$,
\[
 \boxed{R(\widehat\beta_\alpha)-R(\beta)\le\frac{\norm{\Pi_{\mathcal T_\beta}(H_\alpha^{-1/2}\zeta_\beta)}^2}{1-2\eta}
 \le\frac{\norm{\zeta_\beta}^2_{H_\alpha^{-1}}}{1-2\eta}
 \le\frac3{1-2\eta}\Big(\norm{\widehat a-a}^2_{H_\alpha^{-1}}+\norm{\Delta_U\beta}^2_{H_\alpha^{-1}}+b_H(\beta,\alpha)\Big),}
\]
with $b_H(\beta,\alpha)=\alpha^2\norm\beta^2_{H_\alpha^{-1}}$. It holds deterministically and simultaneously
for all comparators, which need not be population minimizers.
\end{theorem}
\begin{proof}
With $d=\widehat\beta_\alpha-\beta$ and $g=C_x\beta-a$, the empirical variational inequality
and $\widehat H_\alpha\beta-\widehat a=g-\zeta_\beta$ give $g^\top d+d^\top\widehat H_\alpha d\le\zeta_\beta^\top d$. Hence
$R(\widehat\beta_\alpha)-R(\beta)=2g^\top d+d^\top C_xd\le2\zeta_\beta^\top d-d^\top(C_x+2\Delta_U+2\alpha I)d$, and
$C_x+2\Delta_U+2\alpha I=H_\alpha+2\Delta_U+\alpha I\succeq(1-2\eta)H_\alpha$. With $v=H_\alpha^{1/2}d\in\mathcal T_\beta$ and
$u=H_\alpha^{-1/2}\zeta_\beta$ the bound is $2u^\top v-(1-2\eta)\norm v^2$, whose supremum over the cone
is $\norm{\Pi_{\mathcal T_\beta}u}^2/(1-2\eta)$ (rescale \cref{q3:lem:proj}). Then
$\norm{\Pi_{\mathcal T_\beta}u}\le\norm u$ and $\norm{x+y+z}^2\le3(\norm x^2+\norm y^2+\norm z^2)$.
\end{proof}

A uniform objective bound pays $\sup_\beta\abs{\beta^\top\Delta_U\beta}\asymp N^{-1/2}$; the local
inequality pays the dual norm of the \emph{vector} $\Delta_U\beta$, squared, after
exploiting curvature. The only global requirement is the relative event
$\eta_\alpha<\tfrac12$. It is a prediction-risk statement; Euclidean weights can stay
unstable. The cone form is the finite-unlabeled analogue of \cref{app:q3}'s local-cone
regret: if $\widehat a-a$ has proxy $\Omega/n$ in the sense of \cref{app:q3}, then with the comparator,
cone and proxy fixed independently of the noise,
$\norm{\Pi_{\mathcal T_\beta}H_\alpha^{-1/2}(\widehat a-a)}$ obeys \cref{app:q3}'s entropy and statistical-dimension bounds
with $H_\alpha^{-1/2}\Omega H_\alpha^{-1/2}$.

\begin{lemma}[Matrix and vector Bernstein]\label{q4:lem:bernstein}
For independent centered self-adjoint $d\times d$ matrices with $\norm{Z_i}_\op\le\mathsf L$
and $\norm{\sum_i\E Z_i^2}_\op\le v$, except with probability $2de^{-t}$,
$\norm{\sum_iZ_i}_\op\le\sqrt{2vt}+2\mathsf Lt/3$. For centered vectors in $\R^{d}$ with
$\norm{Z_i}\le\mathsf L$ and $\sum_i\E\norm{Z_i}^2\le v$ the same holds for $\norm{\sum Z_i}$ with
failure $\le2(d+1)e^{-t}$.
\end{lemma}
\begin{proof}
For centered $Z$ with $\norm Z\le\mathsf L$, $Z^j\preceq\mathsf L^{j-2}Z^2$ and $j!\ge2\cdot3^{j-2}$ give
$\E e^{\theta Z}\preceq\exp(\theta^2\E Z^2/(2(1-\theta\mathsf L/3)))$ for $\theta\mathsf L<3$. Lieb's concavity
theorem gives $\E\tr\exp(\theta\sum Z_i)\le\tr\exp(\sum\log\E e^{\theta Z_i})$ \citep{tropp2012user},
hence $\Prob(\lambda_{\max}\ge u)\le d\exp(-u^2/(2(v+\mathsf Lu/3)))$ at $\theta=u/(v+\mathsf Lu/3)$; the
choice $u=\sqrt{2vt}+2\mathsf Lt/3$ satisfies $u^2\ge2t(v+\mathsf Lu/3)$. Apply to $-Z_i$. For
vectors use the dilation $\begin{psmallmatrix}0&z^\top\\z&0\end{psmallmatrix}$, of norm $\norm z$
and squared-moment norm $\le\sum\E\norm{Z_i}^2$.
\end{proof}

\begin{lemma}[Whitened product averages]\label{q4:lem:products}
If $\abs Z\le M$ and $\E[Z^2\mid X]\le v_Z^2$, then for $q$ iid copies,
$e_q=q^{-1}\sum_ix_iZ_i-\E[xZ]$ satisfies, with failure $\le2me^{-t}$,
\[
 \norm{e_q}_{H_\alpha^{-1}}\le\mathfrak r_q(v_Z,M;t):=\sqrt{\frac{2v_Z^2d_1(\alpha)t}q}+\frac{4M\sqrt{\varkappa(\alpha)}\,t}{3q},
\]
\[
 \E\norm{e_q}^2_{H_\alpha^{-1}}=\frac{\E[Z^2x^\top H_\alpha^{-1}x]-\norm{\E xZ}^2_{H_\alpha^{-1}}}q\le\frac{v_Z^2d_1}q .
\]
\end{lemma}
\begin{proof}
$u=H_\alpha^{-1/2}xZ$ has $\norm{u-\E u}\le2M\sqrt\varkappa$ and $\E\norm{u-\E u}^2\le v_Z^2d_1$;
Lemma~\ref{q4:lem:bernstein} in dimension $m-1$; independence kills cross terms.
\end{proof}

\begin{theorem}[Finite-unlabeled stability gate]\label{q4:thm:gate}
With failure $\le2(m-1)e^{-t}$, $\eta_\alpha\le\sqrt{2\varkappa t/N}+2\varkappa t/(3N)$; in
particular $N\ge64\varkappa(\alpha)t$ gives $\eta_\alpha\le\tfrac14$.
\end{theorem}
\begin{proof}
$z=H_\alpha^{-1/2}x$, $A_\alpha=\E zz^\top\preceq I$, $Z=zz^\top-A_\alpha$ has $\norm Z\le\varkappa$ and
$\E Z^2\preceq\E[\norm z^2zz^\top]\preceq\varkappa I$; Lemma~\ref{q4:lem:bernstein}; $1/\sqrt{32}+1/96<\tfrac14$.
\end{proof}

\begin{assumption}[Residual and comparator envelopes]\label{q4:ass:env}
$\abs{Y-s_0(X)}\le M_a$, $\E[(Y-s_0(X))^2\mid X]\le v_a^2$; the comparator $\beta\in\mathcal B$ is
fixed independently of $U,A$ with $\abs{x^\top\beta}\le M_\beta$.
\end{assumption}

\begin{theorem}[Fast regret with an estimated Gram]\label{q4:thm:fast}
Under Assumption~\ref{q4:ass:env}, with $t=\log(6m/\delta)$ and $N\ge64\varkappa(\alpha)t$,
with probability $\ge1-\delta$,
\[
\begin{aligned}
 R(\widehat\beta_\alpha)-R(\beta)&\le2\{\mathfrak r_n(v_a,M_a;t)+\mathfrak r_N(M_\beta,M_\beta;t)+\sqrt{b_H(\beta,\alpha)}\}^2\\
 &\le\boxed{24d_1(\alpha)t\Big(\frac{v_a^2}n+\frac{M_\beta^2}N\Big)+\frac{64}3\varkappa(\alpha)t^2\Big(\frac{M_a^2}{n^2}+\frac{M_\beta^2}{N^2}\Big)+6b_H(\beta,\alpha)}.
\end{aligned}
\]
Taking $\beta$ a population minimizer bounds selection regret. There is no
$1/\mu_k$ factor.
\end{theorem}
\begin{proof}
$\widehat a-a$ is the product average with $Z=Y-s_0$; $\Delta_U\beta=N^{-1}\sum x_i(x_i^\top\beta)-\E[x(x^\top\beta)]$
is the product average with $Z=x^\top\beta$. The gate and the two vector events fail
with probability $\le[2(m-1)+4m]e^{-t}\le\delta$; on their intersection
$1/(1-2\eta)\le2$. Theorem~\ref{q4:thm:master} and the triangle inequality give the
first bound; $\mathfrak r_q^2\le4v^2d_1t/q+32M^2\varkappa t^2/(9q^2)$ gives the second.
\end{proof}

\begin{corollary}[An explicit implementable regime]\label{q4:cor:safe}
Let $\norm{x(X)}\le\kappa_x$ a.s.\ and $\norm\beta\le R_0$, and take
\[
 \alpha=\frac{64\kappa_x^2t}{\min\{n,N\}}.
\]
Then $\varkappa\le\kappa_x^2/\alpha=\min\{n,N\}/(64t)$, so the gate holds, $b_H\le\alpha R_0^2$, $M_\beta\le\kappa_xR_0$, and the bound is
\[
 \lesssim\frac{t\,[d_1(\alpha)(v_a^2+\kappa_x^2R_0^2)+M_a^2+\kappa_x^2R_0^2]}{\min\{n,N\}}
\]
with numerical constants, requiring no whitening and no smallest-eigenvalue estimate.
\end{corollary}
\begin{proof}
$H_\alpha\succeq\alpha I$. Substituting $\varkappa\le\min\{n,N\}/(64t)$ makes each $n^{-2}$ and $N^{-2}$
term at most $\tfrac13tM^2/\min\{n,N\}$, and $6b_H\le384\kappa_x^2tR_0^2/\min\{n,N\}$. (The choice $\alpha\propto1/N$ leaves a term $\propto Nt/n^2$ that is not small when $N\gg n$;
the $\min\{n,N\}$ choice removes it.)
\end{proof}

\begin{remark}[Scope]\label{q4:auto:scope}
$b_H$ is an explicit regularization cost. For an unrestricted affine oracle
$\norm{C_x^\dagger a}$ may diverge as eigenvalues shrink, so a uniform fast rate cannot
be inferred from $d_1\le k$; one must keep the spectral bias, bound the
comparator, or assume a source condition. A compact simplex bounds tangent
coordinates but not the envelope and coverage requirements.
\end{remark}

\subsection{Affine refinement: exact variance and two effective dimensions}\label{q4:sec:affine}

Let $\mathcal B=\R^{m-1}$, $\beta_*=C_x^\dagger a$, $\beta_\alpha=H_\alpha^{-1}a$, $\tilde\beta_U=\widehat H_\alpha^{-1}a$,
$\Omega_a=\Cov(x(Y-s_0))$, and
$b_P(\alpha)=\norm{C_x^{1/2}(\beta_\alpha-\beta_*)}^2=\sum_j\alpha^2\mu_j\beta_{*,j}^2/(\mu_j+\alpha)^2$.

\begin{theorem}[Conditional affine risk identity]\label{q4:thm:affine}
For every realized $U$,
$\E_A[\Reg_{\Waff}\mid U]=\norm{C_x^{1/2}(\tilde\beta_U-\beta_*)}^2+n^{-1}\tr(C_x\widehat H_\alpha^{-1}\Omega_a\widehat H_\alpha^{-1})$.
If $\E[(Y-s_0)^2\mid X]\le v_a^2$ and $\eta_\alpha\le\eta<1$,
\[
 \E_A[\Reg_{\Waff}\mid U]\le2b_P(\alpha)+\frac{2\norm{\Delta_U\beta_\alpha}^2_{H_\alpha^{-1}}}{(1-\eta)^2}+\frac{v_a^2d_2(\alpha)}{n(1-\eta)^2}:
\]
$d_2$ governs the anchor variance; the unlabeled geometry term involves $d_1$.
\end{theorem}
\begin{proof}
$\widehat\beta_\alpha-\tilde\beta_U=\widehat H_\alpha^{-1}(\widehat a-a)$ with conditional mean $0$ and covariance $\Omega_a/n$;
expand around $\tilde\beta_U$ (Lemma~\ref{q4:lem:quotient}). $v^\top\Omega_av\le v_a^2v^\top C_xv$ gives
$\Omega_a\preceq v_a^2C_x$. With $M=C_x^{1/2}\widehat H_\alpha^{-1}C_x^{1/2}\preceq(1-\eta)^{-1}M_0$,
$M_0=C_x^{1/2}H_\alpha^{-1}C_x^{1/2}$, eigenvalue monotonicity gives
$\tr M^2\le(1-\eta)^{-2}\tr M_0^2=(1-\eta)^{-2}d_2$ without commutation. The resolvent
identity $\tilde\beta_U-\beta_\alpha=-\widehat H_\alpha^{-1}\Delta_U\beta_\alpha$ and
$\norm{C_x^{1/2}H_\alpha^{-1/2}}\le1$ give $\norm{C_x^{1/2}(\tilde\beta_U-\beta_\alpha)}\le(1-\eta)^{-1}\norm{\Delta_U\beta_\alpha}_{H_\alpha^{-1}}$;
insert $\beta_\alpha$ and use $(x+y)^2\le2x^2+2y^2$. Finally
$\beta_{\alpha,j}=\mu_j\beta_{*,j}/(\mu_j+\alpha)$.
\end{proof}

On $\mathcal E=\{\eta_\alpha\le\eta\}$ with $\abs{x^\top\beta_\alpha}\le M_\alpha$, averaging over $U$ gives the
truncated statement
$\E[\one_{\mathcal E}\Reg_{\Waff}]\le2b_P+2M_\alpha^2d_1/(N(1-\eta)^2)+v_a^2d_2/(n(1-\eta)^2)$; the
complement needs a separate envelope or fallback.

\begin{lemma}[Norm bound from a matrix proxy]\label{q4:lem:proxy}
If $\E e^{u^\top Z}\le e^{u^\top\Sigma_Zu/2}$ for all $u$, then with failure $\le e^{-t}$,
$\norm Z^2\le\tr\Sigma_Z+2\sqrt{\tr(\Sigma_Z^2)t}+2\norm{\Sigma_Z}_\op t$.
\end{lemma}
\begin{proof}
For $0<q<(2\norm{\Sigma_Z})^{-1}$ and an independent Gaussian $G$,
$\E e^{q\norm Z^2}=\E e^{\sqrt{2q}G^\top Z}\le\E e^{qG^\top\Sigma_ZG}=\prod_j(1-2q\lambda_j)^{-1/2}$; with
$-\log(1-u)-u\le u^2/(2(1-u))$,
$\log\E e^{q(\norm Z^2-\tr\Sigma_Z)}\le q^2\tr(\Sigma_Z^2)/(1-2q\norm{\Sigma_Z})$; choose
$q=\sqrt t/(\sqrt{\tr\Sigma_Z^2}+2\norm{\Sigma_Z}\sqrt t)$ in Chernoff's bound.
\end{proof}

\begin{proposition}[Affine high-probability refinement]\label{q4:prop:affinehp}
If $\E e^{u^\top(x(Y-s_0)-a)}\le e^{\sigma_a^2u^\top C_xu/2}$ (stronger than a second-moment bound;
Gaussian label noise alone does not give it), then conditionally on $U$ with
$\eta_\alpha\le\eta<1$, with probability $\ge1-e^{-t}$,
$\Reg_{\Waff}\le4b_P+4\norm{\Delta_U\beta_\alpha}^2_{H_\alpha^{-1}}/(1-\eta)^2+2\sigma_a^2(d_2+2\sqrt{d_2t}+2t)/(n(1-\eta)^2)$.
\end{proposition}
\begin{proof}
$C_x^{1/2}\widehat H_\alpha^{-1}(\widehat a-a)$ has proxy $(\sigma_a^2/n)M^2$ with trace
$\le\sigma_a^2d_2/(n(1-\eta)^2)$, operator norm $\le\sigma_a^2/(n(1-\eta)^2)$ and squared trace
$\le\sigma_a^4d_2/(n^2(1-\eta)^4)$ ($\tr M_0^4\le\tr M_0^2$); Lemma~\ref{q4:lem:proxy} and the bias part
of Theorem~\ref{q4:thm:affine}.
\end{proof}

\subsection{Paired anchors and reference residualization}\label{q4:sec:paired}

The anchor inputs can also estimate geometry: with $C_A=n^{-1}\sum x_i^Ax_i^{A\top}$ and
$a_A=n^{-1}\sum x^A_i(Y_i-s_0(X^A_i))$, the error in Theorem~\ref{q4:thm:master} at a fixed
comparator is exactly
\[
 \zeta_\beta=\frac1n\sum_ix^A_i\big(Y_i-s_0(X^A_i)-x_i^{A\top}\beta\big)-\E[x(Y-s_0-x^\top\beta)]-\alpha\beta,
\]
and the gate uses $n$ in place of $N$. One vector radius for the comparator
residual $Y-s_0-x^\top\beta$ suffices. In the noiseless well-specified case
$Y=s_0+x^\top\beta_*$ the stochastic term vanishes identically and, with $\alpha=0$ on a
covered quotient, the fit is exact --- so the $1/N$ term of the split
estimator is not a lower bound for all methods. This is the finite-sample
counterpart of \cref{app:q3}'s remark that the moment estimator pays the signal
$(f-f_0)^2$ as variance: a pilot reference $w_0$ (fitted on independent data)
reduces both $Y-s_0$ and $x^\top\beta_*$, and the penalty then shrinks towards the pilot.
Fitting the pilot on the same anchors needs a uniform or stability argument.
The corresponding regression on the unresolved features, with the
population Gram, is analyzed in \cref{q3:sec:compare}; with finite unlabeled data its reconstruction
map must be estimated, and the resulting difference enters through part (i)
of \cref{q3:rem:rr-implement}. We do not derive that bound
here.

\apppart{Stage 3: teacher structure and certificates}{The anchor term under known teacher structure and finite-sample gain certificates.}
\subsection{Structured teacher information with finite moments}\label{q4:sec:structured}

Suppose \cref{app:q2}'s structured model holds, $f=f_\Kk+\bar f$ with $\bar f\in\Nn$, and the
functions $f_\Kk$ and $h_j=\Pi_\Nn x_j$ ($j\le m-1$) are available pointwise,
independently of $U,A$ (an extra structural input; their population inner
products are not granted). Let $h=(h_j)$, $C_\Nn=\E[hh^\top]\preceq C_x$,
$d_\Nn(\alpha)=\tr(H_\alpha^{-1}C_\Nn)\le d_1(\alpha)$ and $\varkappa_\Nn(\alpha)=\esssup h^\top H_\alpha^{-1}h$ (not ordered with
$\varkappa$ in general). Use $\widehat C_x$ and the structured score
\[
 \widehat a_{\mathrm{str}}=\frac1N\sum_ix^U_i\big(f_\Kk(X^U_i)-s_0(X^U_i)\big)+\frac1n\sum_ih^A_i\big(Y_i-f_\Kk(X^A_i)\big),
\]
unbiased because $\E[h(f-f_\Kk)]=\E[x(f-f_\Kk)]$.

\begin{theorem}[Structured fast rate]\label{q4:thm:structured}
Fix $\beta\in\mathcal B$ and suppose $\abs{Y-f_\Kk}\le M_r$, $\E[(Y-f_\Kk)^2\mid X]\le v_r^2$ and
$\abs{f_\Kk-s_0-x^\top\beta}\le M_g$. With $t=\log(6m/\delta)$, $N\ge64\varkappa t$,
$r_{A,\Nn}=\sqrt{2v_r^2d_\Nn(\alpha)t/n}+4M_r\sqrt{\varkappa_\Nn}t/(3n)$ and $r_{U,g}=\mathfrak r_N(M_g,M_g;t)$, with
probability $\ge1-\delta$,
\[
 \boxed{R(\widehat\beta_{\mathrm{str}})-R(\beta)\le2\big(r_{A,\Nn}+r_{U,g}+\sqrt{b_H(\beta,\alpha)}\big)^2 .}
\]
The anchor term depends on $d_\Nn(\alpha)$; unlabeled integration and the gate
still see the full contrast geometry.
\end{theorem}
\begin{proof}
The local error is $(\mathbb P_A-\mathbb P)[h(Y-f_\Kk)]+(\mathbb P_U-\mathbb P)[x(f_\Kk-s_0-x^\top\beta)]-\alpha\beta$.
Lemma~\ref{q4:lem:products} with $H_\alpha^{-1/2}h$ (second moment $v_r^2d_\Nn$, envelope
$2M_r\sqrt{\varkappa_\Nn}$) and with $Z=f_\Kk-s_0-x^\top\beta$; intersect with the gate;
Theorem~\ref{q4:thm:master}.
\end{proof}

\begin{proposition}[Recovery of the unresolved trace]\label{q4:prop:dN}
On the quotient, $d_\Nn(\alpha)\to\tr(C_x^\dagger C_\Nn)=\tr(\Pi_V\Pi_\Nn\Pi_V)=d_\Nn$ as $\alpha\downarrow0$,
the dimension of \cref{app:q3}'s Theorem ``Expected selection regret''.
\end{proposition}
\begin{proof}
With $C_x=U_k\diag(\mu_j)U_k^\top$, $\phi_j=x^\top U_k\mathbf e_j/\sqrt{\mu_j}$ is an orthonormal basis of
$V$ and $\sum_j\norm{\Pi_\Nn\phi_j}^2=\sum_j\mu_j^{-1}\mathbf e_j^\top U_k^\top C_\Nn U_k\mathbf e_j=\tr(C_x^\dagger C_\Nn)$;
null directions contribute nothing since $x^\top v=0$ a.s.
\end{proof}

If $C_\Nn=0$ the anchor term is absent ($n=0$ allowed), yet finite unlabeled
integration error remains: anchor-free identification in \cref{app:q2} does not mean
exact finite-sample integrals.

\paragraph{Estimated structural inputs.} If $\tilde f_\Kk$, $\tilde h$ are fixed from
independent pilot data, $e_f=\tilde f_\Kk-f_\Kk$, $e_h=\tilde h-h$, the population score
bias is exactly $b_{\mathrm{str}}=\E[(x-h)e_f]+\E[e_h(\bar f-e_f)]$ (expand
$\E[(x-\tilde h)(\tilde f_\Kk-f)]$ and use $\E[(x-h)\bar f]=0$); add $\norm{b_{\mathrm{str}}}_{H_\alpha^{-1}}$
inside the parentheses of Theorem~\ref{q4:thm:structured}. Since $x-h\in\Kk$ and
$h\in\Nn$ coordinatewise, $\E\norm{H_\alpha^{-1/2}(x-h)}^2=d_1-d_\Nn$, so with
$\epsilon_h^2=\E\norm{H_\alpha^{-1/2}e_h}^2$,
$\norm{b_{\mathrm{str}}}_{H_\alpha^{-1}}\le\sqrt{d_1-d_\Nn}\norm{e_f}+\epsilon_h(\norm{\bar f}+\norm{e_f})$ (Cauchy--Schwarz).
Cross-fitting supplies independence but does not remove either first-order
term, and small unweighted nuisance error is insufficient near small
eigenvalues.

\subsection{Finite-sample gain certification}\label{q4:sec:gain4}

For frozen students $s_j$ (fitted or not), $q_j=g_j-s_j$, $D_s=K_g-K_s$ and
$\mu_q=(\ip f{q_j})_j$, the gain is $G(w)=w^\top D_sw-2\mu_q^\top w$ for affine $w$ (\cref{q2:thm:tasks}; $\norm f^2$ cancels). Estimate
$\widehat D_{jk}=N^{-1}\sum_i[g_jg_k-s_js_k](X^U_i)$, $\widehat\mu_{q,j}=n^{-1}\sum_iY_iq_j(X^A_i)$ and
$\widehat G(w)=w^\top\widehat Dw-2\widehat\mu_q^\top w$.

\begin{theorem}[Uniform finite-sample gain certificate]\label{q4:thm:gain4}
If $\abs{g_j}\le M_g$, $\abs{s_j}\le M_s$, $\abs Y\le M_y$, put
$\alpha_D=(M_g^2+M_s^2)\sqrt{2\log(4m^2/\delta)/N}$, $\alpha_q=M_y(M_g+M_s)\sqrt{2\log(4m/\delta)/n}$ and
$\epsilon_G=L^2\alpha_D+2L\alpha_q$. With probability $\ge1-\delta$,
$\sup_{\WL}\abs{\widehat G-G}\le\epsilon_G$; hence $\widehat G(w)>\epsilon_G\Rightarrow G(w)>0$,
$\widehat G(w)<-\epsilon_G\Rightarrow G(w)<0$ and $\abs{G(w)}>\epsilon_G\Rightarrow$ correct sign, for all
weights including those chosen with the same $U,A$. If moreover
$\sup_{\WL}\abs{G_D-G_\lambda}\le\varepsilon_{\mathrm{op}}$ (Proposition~\ref{q4:prop:risk-transfer}), then
$\abs{G_\lambda(w)}>\epsilon_G+\varepsilon_{\mathrm{op}}$ implies $\operatorname{sign}\widehat G(w)=\operatorname{sign}G_\lambda(w)$ ---
a statement about the population learner, distinct from certifying the fitted
student.
\end{theorem}
\begin{proof}
Summands are bounded by $M_g^2+M_s^2$ and $M_y(M_g+M_s)$; $\E[Yq_j]=\ip f{q_j}$;
entrywise Hoeffding with union bounds; $\abs{\widehat G-G}\le\norm w_1^2\alpha_D+2\norm w_1\alpha_q$. The
event is uniform over $\WL$.
\end{proof}

For a margin $\gamma$ it suffices that $L^2\alpha_D<\gamma/2$ and $2L\alpha_q<\gamma/2$, i.e.\
$N\gtrsim L^4(M_g^2+M_s^2)^2\gamma^{-2}\log(m/\delta)$ and
$n\gtrsim L^2M_y^2(M_g+M_s)^2\gamma^{-2}\log(m/\delta)$; with known prediction geometry and a
reference $f_0$, \cref{app:q3}'s residualized certificate replaces $M_y(M_g+M_s)$ by the proxy
$\tau$ of $(Y-f_0)\Pi_\Nn q_j$. No noise-variance subtraction is needed. If
$\E[\xi\mid X]=b_\xi(X)\ne0$, the estimator has bias $-2\ip{b_\xi}{q_w}$, which a valid
certificate must bound. A fast selection rate gives no $n^{-1}$ gain-value or
gain-sign rate: certificates control the quadratic and linear moments of $G$
separately and need a margin.

\apppart{Stage 4: end-to-end closure}{}
\subsection{End-to-end guarantees}\label{q4:sec:closure}

Distinguish the population dictionary $s^\circ_j=T_\lambda g_j$ from the fitted
$\widehat s_j=\widehat T_\lambda g_j$; Sections~\ref{q4:sec:agg}--\ref{q4:sec:structured} apply
conditionally to the fitted dictionary with independent $U,A$. Let
$R_D(w)=\norm{\widehat s_w-f}^2$, $R_\circ(w)=\norm{s^\circ_w-f}^2$, $\Delta_w=\widehat s_w-s^\circ_w$.

\begin{proposition}[Population-oracle comparison]\label{q4:prop:oracle-comp}
On an event where $\sup_{\mathcal W}\abs{R_D-R_\circ}\le\varepsilon_{\mathrm{op}}$ and
$R_D(\widehat w)-\inf_{\mathcal W}R_D\le\varepsilon_{\mathrm{sel}}$,
$R_D(\widehat w)\le\inf_{\mathcal W}R_\circ+\varepsilon_{\mathrm{op}}+\varepsilon_{\mathrm{sel}}$ and
$R_\circ(\widehat w)-\inf_{\mathcal W}R_\circ\le2\varepsilon_{\mathrm{op}}+\varepsilon_{\mathrm{sel}}$.
\end{proposition}
\begin{proof}
$R_D(\widehat w)\le R_D(w^\circ_\star)+\varepsilon_{\mathrm{sel}}\le R_\circ(w^\circ_\star)+\varepsilon_{\mathrm{op}}+\varepsilon_{\mathrm{sel}}$; for the
second replace $R_\circ(\widehat w)$ by $R_D(\widehat w)+\varepsilon_{\mathrm{op}}$.
\end{proof}

\begin{theorem}[Comparator-local end-to-end inequality]\label{q4:thm:closure}
For a feasible comparator $w^\circ=w_0+O\beta^\circ$ fixed given $\mathcal F_0$ (not
depending on $U,A$) and any valid bound $\mathcal E_{\mathrm{sel}}(\beta^\circ)$ on
$R_D(\widehat w)-R_D(w^\circ)$ from Theorems~\ref{q4:thm:master}, \ref{q4:thm:fast}
or~\ref{q4:thm:structured},
\[
 \boxed{R_D(\widehat w)-R_\circ(w^\circ)\le\mathcal E_{\mathrm{sel}}(\beta^\circ)+2\ip{s^\circ_{w^\circ}-f}{\Delta_{w^\circ}}+\norm{\Delta_{w^\circ}}^2 .}
\]
If $\max_j\norm{\widehat s_j-s^\circ_j}\le\epsilon_D$ and $\norm{w^\circ}_1\le L_\circ$, the learning terms are at
most $2\sqrt{R_\circ(w^\circ)}L_\circ\epsilon_D+L_\circ^2\epsilon_D^2$; if
$\ip{s^\circ_{w^\circ}-f}{\Delta_{w^\circ}}=0$ they equal $\norm{\Delta_{w^\circ}}^2$ exactly.
\end{theorem}
\begin{proof}
Add and subtract $R_D(w^\circ)$ and expand $\norm{s^\circ_{w^\circ}+\Delta_{w^\circ}-f}^2$;
$\norm{\Delta_{w^\circ}}\le L_\circ\epsilon_D$ and Cauchy--Schwarz.
\end{proof}

\begin{corollary}[Two orthogonality regimes]\label{q4:cor:orth}
The learning term is quadratic if (i) $s^\circ_{w^\circ}=f$, or (ii) $w^\circ$ is an
unconstrained affine population oracle and $\Delta_{w^\circ}\in V^\circ=\Span\{s^\circ_j-s^\circ_m\}$.
\end{corollary}
\begin{proof}
(ii) At an affine oracle the derivative along every zero-sum direction
vanishes, so $s^\circ_{w^\circ}-f\perp V^\circ$.
\end{proof}

Student realizability does not supply orthogonality: even if $f\in\Ss$ and
$s^\circ_j=Pg_j$, the aggregate need not equal $f$, and a simplex oracle gives only
a one-sided variational inequality.

\begin{example}[The first-order term occurs]\label{q4:auto:the-first-order-term-occurs}
Let every population candidate be a unit function $u$, $f=0$, and every fitted
candidate $(1+\varepsilon)u$. Selection regret is zero in both dictionaries, the
squared fitting discrepancy is $\varepsilon^2$, but $R_D-R_\circ=2\varepsilon+\varepsilon^2$. So no
$\epsilon_D^2$-only bound follows from fitting closeness.
\end{example}

\begin{theorem}[Learning, selection and certification]\label{q4:thm:integrated}
Under the split protocol, let $\mathcal E_{\mathrm{learn}}$ (probability
$\ge1-\delta_{\mathrm{learn}}$) give $\max_j\norm{(\widehat T_\lambda-T_\lambda)g_j}\le\varepsilon_g$, hence
$\varepsilon_{\mathrm{op}}$ of Proposition~\ref{q4:prop:risk-transfer}, and let the independent
moment events (conditional failure $\le\delta_{\mathrm{mom}}$) give selection error
$\varepsilon_{\mathrm{sel}}$ (Theorem~\ref{q4:thm:slow}, Proposition~\ref{q4:prop:pred-space},
Theorem~\ref{q4:thm:fast} or~\ref{q4:thm:structured}) and gain error $\epsilon_G$
(Theorem~\ref{q4:thm:gain4}). With probability $\ge1-\delta_{\mathrm{learn}}-\delta_{\mathrm{mom}}$,
simultaneously: the two inequalities of Proposition~\ref{q4:prop:oracle-comp},
$\sup_{\mathcal W}\abs{\widehat G-G_D}\le\epsilon_G$ and $\sup_{\mathcal W}\abs{\widehat G-G_\lambda}\le\epsilon_G+\varepsilon_{\mathrm{op}}$. A
uniform random-feature or network discrepancy (\cref{ext:scope}) adds to
the comparison budget by one more triangle inequality.
\end{theorem}
\begin{proof}
$\Prob(\text{failure})\le\Prob(\mathcal E_{\mathrm{learn}}^c)+\E[\one_{\mathcal E_{\mathrm{learn}}}\Prob(\text{moment failure}\mid\mathcal F_0,D)]$;
combine Propositions~\ref{q4:prop:risk-transfer} and~\ref{q4:prop:oracle-comp} with the
triangle inequality for gains.
\end{proof}

\paragraph{Sample-budget interface.} If $\epsilon_D^2\le e_D$ with probability
$\ge1-\delta_D$ (e.g.\ from \cref{ext:rates},
$e_D\lesssim((1+\log(1/\delta_D))/N_{\mathrm{fit}})^{2r/(2r+p)}$) and the selection event fails with probability
$\le\delta_S$, the learning contribution is $2\sqrt{R_\circ(w^\circ)}L_\circ\sqrt{e_D}+L_\circ^2e_D$ in
general and $L_\circ^2e_D$ under Corollary~\ref{q4:cor:orth}, plus the explicit
$n,N,\alpha$ terms, with joint failure $\le\delta_D+\delta_S$. This separates student
fitting, unlabeled integration and anchor selection. For a predetermined
finite set of feature maps or $\lambda$ values, add a union bound; a continuum
needs a uniform argument, and choosing $\lambda$ as a nonlinear function of each
label mixture does not define the common linear learner of the identities.

\apppart{Supporting material}{}
\subsection{Relation to prior work}\label{q4:sec:related4}

What in this appendix is standard and what is new, and detailed comparisons with prior work, are collected in \cref{app:boundary,app:related}.

\section{Experimental Details}\label{app:experiments}

\subsection{Synthetic experiments: protocol}\label{app:synthetic}

\paragraph{Code and environment.} All synthetic experiments are in
\texttt{experiments/}: one module per theory appendix (X1--X4 for
\cref{app:q1,app:q2,app:q3,app:q4}), shared helpers in \texttt{w2s/}, and
\texttt{run\_all.py}, which reproduces every table and figure (about two
minutes on two cores; \texttt{--quick} gives a 30-second smoke test).
Dependencies are pinned in \texttt{requirements.txt} (Python 3.13, NumPy
2.5.3, SciPy 1.18.1, Matplotlib 3.11.2); each module writes
\texttt{results/xK/run.json} with its seed, arguments, package versions,
platform and runtime.

\paragraph{Randomness.} Module $K$ has one master seed (default
$2026110K$). Every sub-experiment draws from its own stream, built as a NumPy
\texttt{SeedSequence} from the master seed and a hash of the sub-experiment
name, so streams do not depend on execution order and changing one
sub-experiment leaves the others' random numbers unchanged. Within a panel,
methods are evaluated on the same simulated anchor and unlabeled samples
(common random numbers), except the sample-splitting certificate in
\cref{fig:x3}(f), which needs its own draws.

\paragraph{Exact computations versus Monte Carlo.} X1 and X2 are exact
population computations (in coefficient space, or on finite designs such as
the eight-point cube of \cref{thm:counter}); randomness there only draws
configurations, and shaded bands are exact identified sets or ranges. X3 and
X4 simulate anchors, unlabeled inputs and pseudo-labelled fitting samples.
X3 uses passive anchors with Rademacher features throughout, except the
local-cone panel, whose theorem concerns the Gaussian moment experiment
(\cref{q3:sec:bridge} gives the bridge); X4 uses finite designs with exact
population risks, so only the samples are random.

\paragraph{Error bars, benchmarks and budgets.} Error bars are $\pm1.96$
Monte Carlo standard errors of the plotted mean. Lines labelled
``bound'', ``prediction'' or ``exact'' are theoretical benchmarks evaluated
at the simulated configuration, not fits; reference slopes are guides. Label
budgets (``anchors needed'') are where the Monte Carlo family-wise error
first falls to the target level $0.1$ on a fixed logarithmic grid (adjacent
budgets differ by a factor of about $1.15$), interpolated log-linearly between
the two neighbouring grid points. With $2{,}000$ replicates the standard error
of the error rate near $0.1$ is about $0.007$, so budget differences below
about $10\%$ are within resolution and are not interpreted.

\paragraph{Automatic checks.} Every module writes \texttt{checks.csv} and
exits with an error if a check fails. Exact identities (risk identities,
realizability, KKT conditions, leakage law, coverage formula, paired-anchor
exactness) are checked to relative tolerance $10^{-9}$ (some limits to
$10^{-3}$); Monte Carlo statements (an exact variance or law, the validity of
a certificate, a bound that must hold on average) are checked against
$3$--$4.5$ standard errors; deterministic inequalities (the local oracle
inequality per replicate) are checked to $10^{-10}$. The current run passes
all $30$, $30$, $25$ and $12$ check groups of X1--X4.

\begin{center}\small
\begin{tabularx}{\textwidth}{@{}>{\raggedright}p{0.09\textwidth}>{\raggedright}p{0.22\textwidth}>{\raggedright}X>{\raggedright}p{0.17\textwidth}>{\raggedright\arraybackslash}p{0.14\textwidth}@{}}
\toprule
Module & Theory & Data-generating process & Scale & Output\tabularnewline
\midrule
X1 & \cref{app:q1}: value is student-relative & exact projection and ridge geometries in coefficient space; random Gaussian errors & 300 identity configurations; $2\times20{,}000$ orientations; $2\times2{,}000$ dictionaries; $60\times50$ candidates & \cref{fig:x1}; \texttt{results/x1}\tabularnewline
X2 & \cref{app:q2}: identification & eight-point cube world; random structured teachers & 20,000 targets per radius; 500 structured instances & \cref{fig:x2}; \texttt{results/x2}\tabularnewline
X3 & \cref{app:q3}: anchor complexity & passive anchors, Rademacher features, Gaussian label noise; cell dictionaries & 2,000--50,000 replicates per point & \cref{fig:x3}; \texttt{results/x3}\tabularnewline
X4 & \cref{app:q4}: finite samples & finite designs ($64$ points) with exact population risks; fitted ridge students & 400--20,000 replicates per point & \cref{fig:x4}; \texttt{results/x4}\tabularnewline
\bottomrule
\end{tabularx}
\end{center}

\newcommand{\xrow}[5]{#1 & #2 & #3 & #4 & #5\tabularnewline}
\newcommand{\xtable}[1]{%
\begin{center}\small
\begin{tabularx}{\textwidth}{@{}>{\raggedright}p{0.175\textwidth}>{\raggedright}p{0.2\textwidth}>{\raggedright}X>{\raggedright}p{0.17\textwidth}>{\raggedright\arraybackslash}p{0.06\textwidth}@{}}
\toprule
Sub-experiment & Result checked & Design & Sweep, replicates & Panels\tabularnewline
\midrule
#1
\bottomrule
\end{tabularx}
\end{center}}

\subsection{X1: student-relative value}\label{app:x1}

\xtable{
\xrow{\texttt{identities}}{risk identity, Gram split, KKT and marginal-value identities (\cref{q1:sec:multi,q1:sec:objectives,q1:sec:marginal})}{$d=12$, $d_\Ss=5$, $m=6$; Gaussian target and errors; projection and general learners}{300 configurations}{checks}
\xrow{\texttt{fixed\_\allowbreak gram}}{\cref{thm:fixed-gram}}{four-dimensional family with $\Sigma=I$ fixed, rotated relative to $\Ss$}{401 rotations}{(a)}
\xrow{\texttt{teacher\_\allowbreak ranking}}{accuracy $\ne$ value}{$d=60$, $d_\Ss=15$, $m=10$, $A=0.2$; error energies $\mathrm{U}(0.5,2)$; transmitted fraction isotropic (Gaussian) or structured ($\mathrm{Beta}(0.6,0.6)$)}{2,000 dictionaries per regime}{(b), (i)}
\xrow{\texttt{raw\_\allowbreak weighting}}{\cref{thm:decisions}; gain maximization (\cref{q1:thm:gainmax})}{two supervisors, $\Gpar=\mathrm{diag}(\varepsilon,1)$, $\Gperp=\mathrm{diag}(M,\varepsilon)$}{$M\in[1,10^4]$ (81 values), $\varepsilon\in\{0.1,0.03,0.01\}$}{(c), (g)}
\xrow{\texttt{marginal\_\allowbreak value}}{simplex value criterion (\cref{q1:thm:marg-sim})}{$d=30$, $d_\Ss=10$, $m=6$; near-duplicate candidates mixing existing errors}{60 dictionaries $\times$ 50 candidates}{(d), (h)}
\xrow{\texttt{orientation}}{fixed spectrum (\cref{q1:thm:orient})}{2-d example with spectrum $(8,2)$; Haar rotations with $d=10$, $d_\Ss=5$, $m=10$}{361 angles; $2\times20{,}000$ rotations}{(e), (f)}
\xrow{\texttt{general\_\allowbreak learner}}{cross term $c_T$ (\cref{prop:risk})}{ridge filter on a 6-d example}{$\lambda\in[10^{-3},10]$ (121 values)}{(j)}
}

\par\noindent\begin{minipage}{\textwidth}\centering

\includegraphics[width=0.8\textwidth]{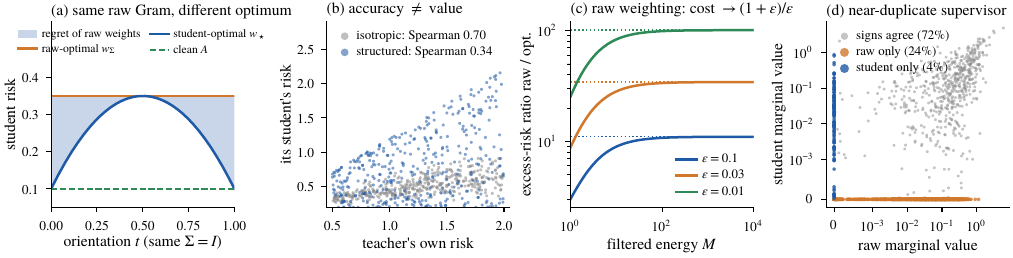}\\[2pt]
\includegraphics[width=0.8\textwidth]{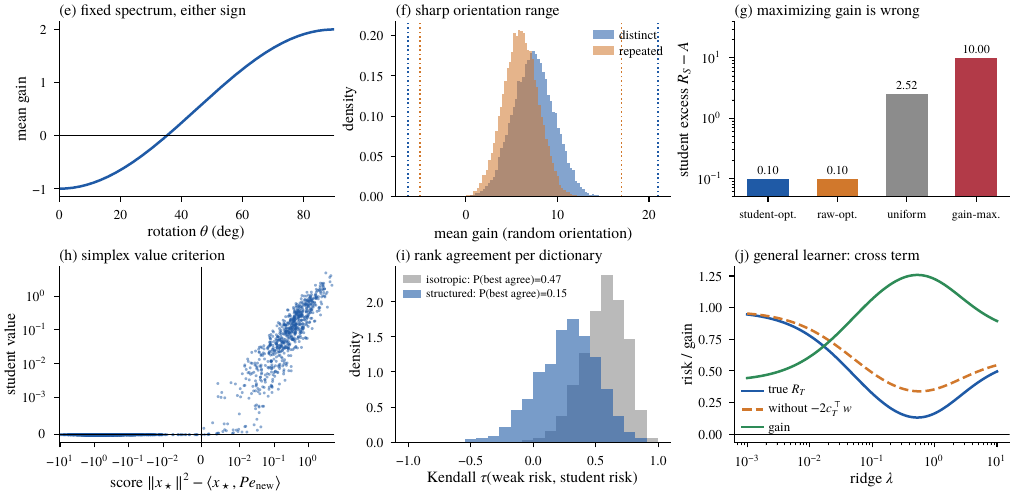}
\captionof{figure}{\textbf{X1: value is student-relative} (exact computations). Fixed raw
geometry leaves student value undetermined (a); raw and student-relative
rankings, optima and marginal values can disagree (b), (d), (h), (i); raw
weighting can be arbitrarily worse (c), and maximizing the gain is a different
objective (g). Panels (e)--(f) verify the sharp orientation range; (j) shows the
bias-alignment term required for a ridge learner.}
\label{fig:x1}
\end{minipage}\par\medskip

\subsection{X2: identification}\label{app:x2}

\xtable{
\xrow{\texttt{same\_\allowbreak observables}}{\cref{thm:tasks,thm:counter}}{eight-point cube world of \cref{thm:counter}, misspecified and realizable targets; hash check that both worlds produce identical observables}{401 weights}{(a), (d), (e), (f)}
\xrow{\texttt{partial\_\allowbreak identification}}{sharp gain interval (\cref{q2:thm:interval})}{same world, reference $f_0=z$, random compatible targets on the sphere of each radius}{161 radii, 20,000 targets each}{(b), (g), (h)}
\xrow{\texttt{leakage\_\allowbreak regimes}}{\cref{thm:leakage}}{three-dimensional structured example: zero, common and different leakage; hidden coordinate swept over $c\in[-2,2]$ (on the full fiber the unidentified gain is unbounded)}{401 hidden coordinates}{(c)}
\xrow{\texttt{leakage\_\allowbreak ensemble}}{\cref{thm:leakage}; bounded-fiber interval}{$D=14$, $m=4$ rank-2 projection teachers, rank-5 student, Gaussian targets}{500 instances}{(i)}
\xrow{\texttt{structured\_\allowbreak counterexample}}{exact trained teachers, either sign (\cref{q2:sec:struct-counter})}{three-dimensional teachers and a rank-1 student}{151 targets}{(j)}
}

\par\noindent\begin{minipage}{\textwidth}\centering

\includegraphics[width=0.8\textwidth]{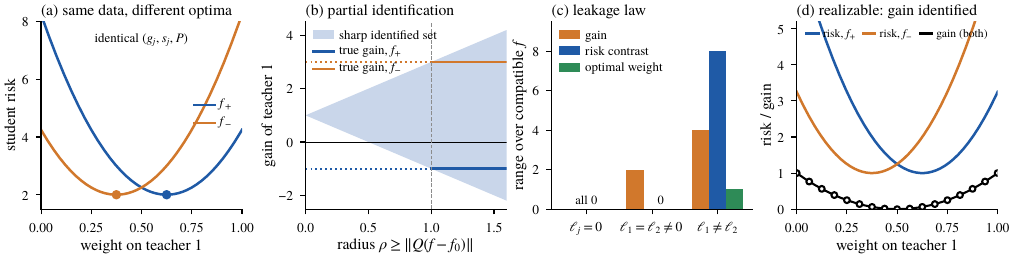}\\[2pt]
\includegraphics[width=0.8\textwidth]{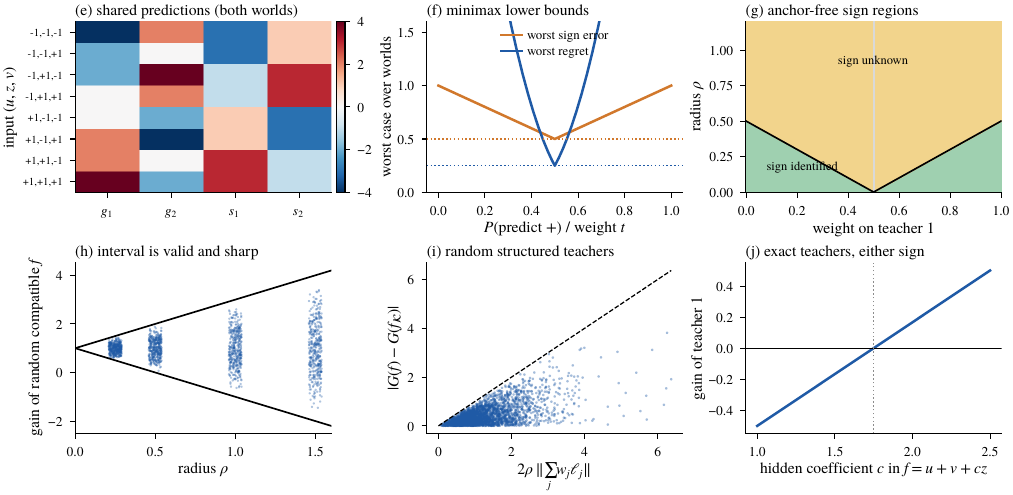}
\captionof{figure}{\textbf{X2: identification} (exact computations). Identical observations
admit opposite gains and optima (a), (e), (f); the sharp interval contains every
compatible target and is attained (b), (g), (h); the leakage law separates the
three regimes (c; ranges of gain, contrast and weight share one axis but not units, over a bounded sweep), (i); realizability identifies the gain but not the selection
(d); exact trained teachers still allow either gain sign (j).}
\label{fig:x2}
\end{minipage}\par\medskip

\subsection{X3: anchor complexity}\label{app:x3}

\xtable{
\xrow{\texttt{single\_\allowbreak sign}}{margin cost (\cref{q3:cor:sign})}{one query, Gaussian estimate, $\sigma=1$}{$\gamma\in\{0.05,0.1,0.2,0.4\}$, 30 budgets; 40,000}{(e)}
\xrow{\texttt{post\_\allowbreak selection}}{uniform certificate after selection (\cref{thm:certificate})}{$m=32$ null queries, select the largest estimated gain, $\delta=0.1$; naive, uniform and split intervals}{$n\in\{32,\dots,2048\}$; 4,000}{(f)}
\xrow{\texttt{sign\_\allowbreak geometry}}{\cref{thm:certificate}(ii)}{duplicated, clustered and independent queries, $\gamma=0.25$; prediction: the $n$ at which the Monte Carlo $(1-\delta)$-quantile (50,000 draws) of the one-sided Gaussian supremum of the margin-normalized queries, with the exact per-query covariance, equals $\sqrt n$; no fitted constant}{$m\in\{2,\dots,256\}$, 41 budgets in $[16,4096]$; 2,000}{(a), (g), (h)}
\xrow{\texttt{unstructured\_\allowbreak selection}}{$k/n$ rate; moment, cross-fitted residualized moment and ERM}{$k$ uniform cells, $\sigma=0.5$}{$k\in\{2,4,8,16\}$, $n\in[16,4096]$; 20,000}{(i)}
\xrow{\texttt{structured\_\allowbreak selection}}{\cref{thm:selection}}{$k=8$ exact structured teachers; ten unresolved-angle settings}{$n\in[32,2048]$; 4,000}{(b), (j)}
\xrow{\texttt{local\_\allowbreak cones}}{vertex versus interior (\cref{q3:prop:vertex})}{Gaussian moment experiment on the simplex}{$m\in\{2,\dots,512\}$; 200,000 draws, 50,000 replicates}{(c)}
\xrow{\texttt{cells}}{coverage (\cref{sec:coverage})}{$k=16$ cells, uniform or rare, $B_\ell^2=1/p_\ell$; $\sigma\in\{0,0.5\}$}{$n\in\{0,10,\dots,400\}$; 40,000}{(d), (l)}
\xrow{\texttt{redundancy}}{duplicates do not change prediction-space selection}{$k=4$, $n=128$; dictionary duplicated 1--16 times}{4,000}{(k)}
\xrow{\texttt{comparisons}}{\cref{prop:compare}; exact risks of \cref{q3:sec:compare}}{Gaussian coordinates, $k=8$, $\nu_i^2=\tfrac14$ ($\dN=2$), $\norm\beta^2=4.5$, $R_{\mathrm{agg}}=0.1$, $d_\Ss=50$, $\sigma=1$; anchors alone by OLS and by Bayes ridge under isotropic and $j^{-2}$ priors with $\tr\Lambda=\norm f^2=5$; reference error $e_0\in\Kk$ along one $V$-direction}{$n\in[12,2048]$; reference error $0$--$0.6$ at $n=64$; 3,000}{(m)--(o)}
}

\par\noindent\begin{minipage}{\textwidth}\centering

\includegraphics[width=0.8\textwidth]{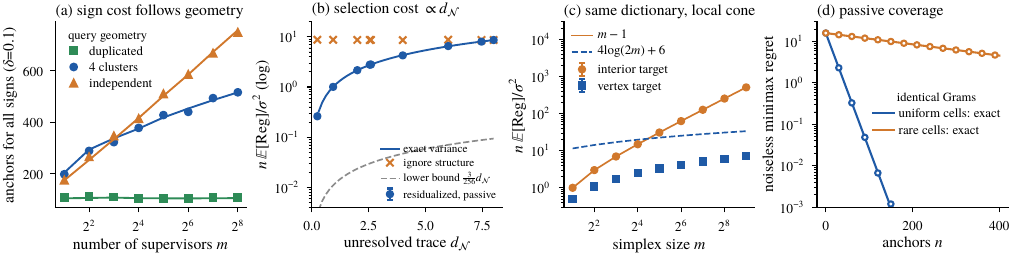}\\[2pt]
\includegraphics[width=0.8\textwidth]{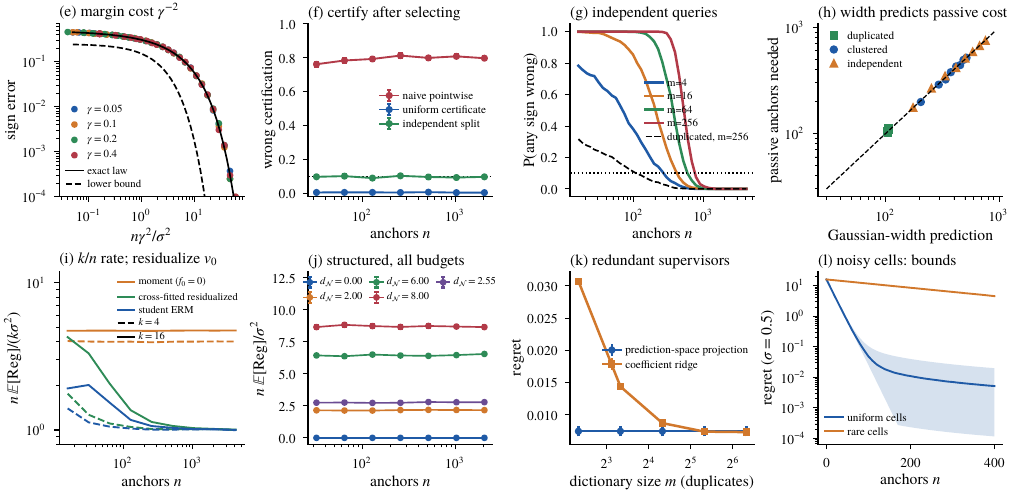}\\[2pt]
\includegraphics[width=0.8\textwidth]{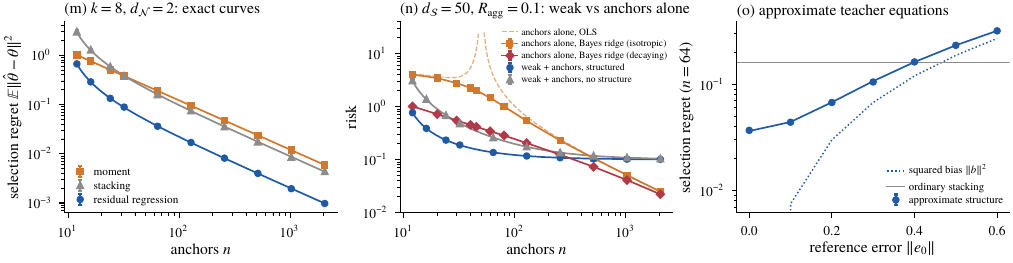}
\captionof{figure}{\textbf{X3: anchor complexity} (passive anchors; (c) uses the Gaussian moment
experiment). Sign cost follows the query geometry and the Gaussian-width
prediction (a), (g), (h), with the exact margin law (e); uniform certificates
stay valid after selection (f). Selection regret tracks $\dN$ (b), (j), with the
$k/n$ rate and residualization (i); vertex and interior differ (c); duplicates do
not affect prediction-space selection (k); coverage, not the Gram, sets the
passive cost (d), (l). Residual regression pays only the noise, while the
moment estimator and ordinary stacking pay the signal or $k$ (m). Against
anchors alone (n), the comparison depends on the competitor: OLS has the
ridgeless peak at $n=d_\Ss$, while tuned (Bayes) ridge has no peak and, with a
decaying spectrum (small $d_{\mathrm{eff}}$), overtakes weak supervision without
structure already at a few dozen anchors and structured weak supervision at a
few hundred; a reference
error adds the bias of \cref{q3:thm:approx-structure} and eventually loses to
stacking (o). Lines in (m)--(o) are exact risks, not fits.}
\label{fig:x3}
\end{minipage}\par\medskip

\subsection{X4: finite samples}\label{app:x4}

\xtable{
\xrow{\texttt{two\_\allowbreak bottlenecks}}{local oracle inequality (\cref{q4:thm:master,q4:thm:fast})}{$k=6$, 64-point sign design, $\alpha=10^{-3}$; one of $n,N$ varied, the other $8192$}{$n,N\in[64,8192]$; 5,000}{(a), (e)}
\xrow{\texttt{coverage}}{equal spectrum, different coverage (\cref{q4:ex:coverage})}{spectrum $(1,10^{-1},10^{-2},10^{-3})$; dense versus rare-cell design}{$N\in[8,16384]$; 5,000}{(b)}
\xrow{\texttt{fitted\_\allowbreak student}}{learning deviation, risk and gain transfer (\cref{q4:thm:learning,q4:prop:risk-transfer})}{six Rademacher coordinates, the student sees four with spectrum $(1,0.3,0.08,0.02)$; $m=3$; $n=N=2048$}{$\Nfit\in[8,4096]$; 3,000}{(c)}
\xrow{\texttt{learning\_\allowbreak term}}{\cref{thm:e2e}, two orthogonality regimes}{same world; generic target versus a comparator that fits $f$}{$\Nfit\in[8,65536]$; 400}{(d)}
\xrow{\texttt{affine\_\allowbreak d2}}{conditional affine identity (\cref{q4:thm:affine})}{$k=6$, spectrum $10^0$--$10^{-3}$, $n=N=256$}{$\alpha\in\{10^{-4},\dots,1\}$; $12\times20{,}000$}{(f)}
\xrow{\texttt{paired}}{paired anchors (\cref{q4:sec:paired})}{$k=6$, 64-point design, $\sigma\in\{0,0.5\}$}{$n\in[8,1024]$; 5,000}{(g)}
\xrow{\texttt{certificates}}{finite-sample gain certificate (\cref{q4:thm:gain4})}{fitted student, $\delta=0.05$, three gain shifts}{4,000}{(h)}
}

\par\noindent\begin{minipage}{\textwidth}\centering

\includegraphics[width=0.8\textwidth]{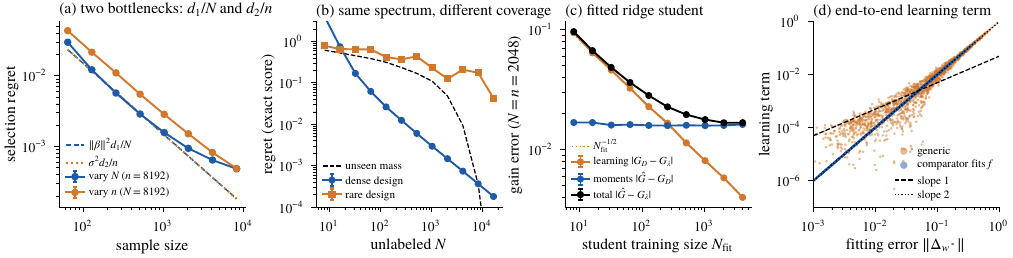}\\[2pt]
\includegraphics[width=0.8\textwidth]{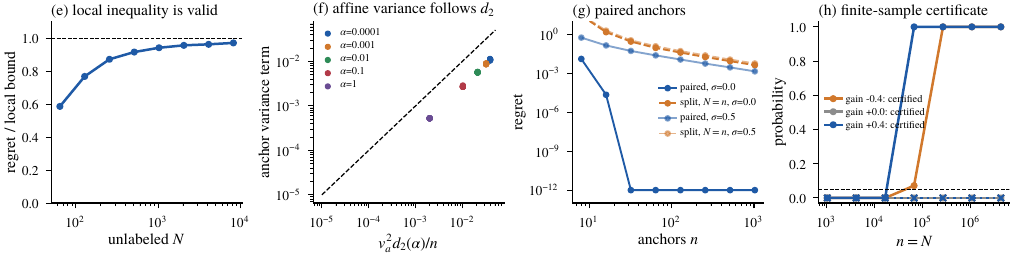}
\captionof{figure}{\textbf{X4: finite samples.} Selection regret has separate $d_1(\alpha)/N$
and $d_2(\alpha)/n$ terms (a), (f), and the local oracle inequality holds in every
replicate (e); leverage, not the spectrum, governs coverage (b). The fitted
student's gain error splits into learning and moments (c); the end-to-end
learning term is first order in general and second order when the comparator
fits $f$ (d). Paired anchors cancel the noise-free stochastic term (g); the
finite-sample certificate is valid but conservative (h).}
\label{fig:x4}
\end{minipage}\par\medskip

\subsection{Vector-valued outputs}
Multi-class experiments use class-probability (or one-hot) vectors. The
extension is formal.

\begin{proposition}[Vector-valued outputs]\label{prop:vector}
Let $\Hh_C=L^2(P_X;\R^C)$ with $\ip hk=\E[h(X)^\top k(X)]$. When $f$ is the
vector of class probabilities and $Y$ the one-hot label,
$\E\norm{h(X)-Y}_2^2=\E\norm{f(X)-Y}_2^2+\norm{h-f}^2$, so
$\norm{h-f}^2$ is the \emph{excess} Brier risk of $h$ (equivalently, its
population squared probability error).
\begin{enumerate}\renewcommand{\labelenumi}{(\roman{enumi})}
\item $\Hh_C$ is a real Hilbert space. Every result of
  \cref{sec:value,sec:identification} uses only the Hilbert structure of
  $\Hh$ and a bounded linear learner $T$, and therefore holds verbatim in
  $\Hh_C$ for unconstrained vector regression. If $f$ is additionally
  required to be a probability vector, the compatible set becomes
  $(f_\Kk+\Nn)\cap\{f:f(x)\in\Delta_C\text{ a.s.}\}$, a bounded subset of the
  fiber: the identities and sufficient conditions still hold, but the
  necessity parts of the full-fiber characterizations (which move along whole
  lines $f_\Kk+tn$) and the sharp ranges must be re-derived on the bounded
  set, as in \cref{q2:sec:bounded}. A common head fitted by multi-output least squares or ridge on a
  shared representation acts coordinatewise, $T_C=T\otimes I_C$, and is a
  projection whenever $T$ is.
\item The anchor estimates of \cref{sec:anchors} are averages of the scalar
  products $(Y-f_0(X))^\top h(X)$, with
  $\E[(Y-f_0(X))^\top h(X)]=\ip{f-f_0}h$ when $\E[Y\mid X]=f(X)$. The
  sub-Gaussian assumptions are imposed on these scalars; e.g.\ if
  $\norm{Y-f_0(X)}_2\le B_0$ and $\norm{h(X)}_2\le H_0$, the proxy is
  $\tau=B_0H_0$, with no factor of $C$.
\item For the fitted learner of \cref{sec:finite},
  $\norm{(\widehat T_C-T_C)h}^2=\sum_{c\le C}\norm{(\widehat T-T)h_c}^2$.
  Applying \cref{q4:thm:learning} to each coordinate $h_c$, with its own
  scalar residual envelope $R_{h_c}$ and confidence parameter $t+\log C$,
  gives, with probability at least $1-2e^{-t}$ simultaneously for all
  coordinates,
  $\norm{(\widehat T_C-T_C)h}\le\bigl(\sum_{c\le C}\varepsilon_{h_c}(t+\log C)^2\bigr)^{1/2}$,
  where $\varepsilon_{h_c}(\cdot)$ is the scalar bound of \cref{q4:thm:learning}.
\end{enumerate}
\end{proposition}

\begin{proof}
(i) Completeness of $L^2(P_X;\R^C)$ is standard, and the proofs of
\cref{sec:value,sec:identification} use only inner products, orthogonal
projections and bounded linear maps. Coordinatewise fitting gives
$T_C=T\otimes I_C$; if $T=T^*=T^2$, the same holds for $T_C$.
(ii) Linearity of expectation and $\E[Y-f(X)\mid X]=0$; Hoeffding's lemma for
a scalar in $[-B_0H_0,B_0H_0]$.
(iii) The squared norm is a sum over coordinates, and $\widehat T_C$ acts on
each coordinate as the scalar $\widehat T$; apply the scalar bound to each
$h_c$ at level $t+\log C$, take a union bound over the $C$ coordinates
($2Ce^{-t-\log C}=2e^{-t}$) and sum the squares.
\end{proof}

\begin{remark}[Two routes for vector outputs]
Part (iii) is the coordinatewise route: it reuses the scalar theorem, pays
$\log C$ in the confidence level, and measures each coordinate with its own
scalar envelope. A direct route would rerun the residual-normal-equation
argument of \cref{q4:thm:learning} once in $\Hh_C$, with a Hilbert-valued
concentration inequality and the vector envelope
$\sup_x\norm{r_\lambda(h)(x)}_2$; this can avoid the $\log C$ term, but we do
not state or use it. Parts (i) and (ii) need neither route.
\end{remark}

\section{Extended Related Work}\label{app:related}

This appendix expands \cref{sec:related}. For each nearby line of work it
states how the target object, the observables and the object of the
complexity bounds differ from ours.
\Cref{tab:closest} summarizes the closest works; the paragraphs below give
details. This appendix is the only place where the theory appendices'
relation to prior work is discussed in detail.

\begin{table}[h]
\caption{\textbf{Closest work.} ``Learner'' is the downstream model that the
combined supervision trains; ``Ident.'' marks whether the work asks what is
identified from the available observations.}
\label{tab:closest}
\centering\footnotesize
\begin{tabularx}{\textwidth}{@{}>{\raggedright}p{0.15\textwidth}>{\raggedright}p{0.15\textwidth}>{\raggedright}p{0.13\textwidth}>{\raggedright}p{0.1\textwidth}>{\raggedright}p{0.1\textwidth}>{\raggedright}p{0.07\textwidth}>{\raggedright\arraybackslash}X@{}}
\toprule
Work & Target object & Observables & Clean labels & Learner & Ident. & What this paper adds\tabularnewline
\midrule
Misfit--gain \citep{charikar2024quantifying,mulgund2025relating} & gain of one supervisor; choice among weak models by misfit & student and supervisor predictions & none & convex class / projection & no & joint aggregation; identification boundaries; label cost\tabularnewline
Representations \citep{xue2025representations} & prediction gap of weakly vs.\ cleanly trained students & representations & none & ridge head & no & absolute risk and decisions, which the gap does not determine; non-identification\tabularnewline
Label models \citep{dawid1979maximum,ratner2016data} & supervisor accuracies & agreement patterns & none & none or any & yes (under independence) & what independence must control; post-learning value\tabularnewline
Ensemble W2S \citep{sang2024improving,agrawal2024ensemw2s,liu2024cosupervised} & a combined pseudo-label & weak predictions & none or few & fine-tuned model & no & the objective to optimize and when it is observable\tabularnewline
Stacking, aggregation \citep{wolpert1992stacked,breiman1996stacked,tsybakov2003optimal} & best combination of fixed predictors & labeled data & all & none & no & which moments are needed; $\dN$ instead of $k$; unlabeled Gram\tabularnewline
Forecast combination \citep{bates1969combination} & combined forecast error & raw error covariance & historical outcomes & none & no & the transmitted, not the raw, error is optimized\tabularnewline
PPI, MultiPPI \citep{angelopoulos2023prediction,cowenbreen2026multiple} & a population functional & many predictions, few labels & few & none & no & the functional is a post-learning risk; decision complexity\tabularnewline
Partial identification \citep{manski2003partial} & bounds on a parameter & observed distribution & none & --- & yes & sharp sets for student-relative gains\tabularnewline
\bottomrule
\end{tabularx}
\end{table}

\subsection{What is standard and what is new}\label{app:boundary}

\begin{itemize}
\item \emph{\Cref{app:q1}.} Representation-dependent filtering of a single
supervisor's error, the misfit--gain identity for projections and the
insufficiency of a spectrum are known
\citep{charikar2024quantifying,xue2025representations,mulgund2025relating}
and serve as the base case; the proofs are elementary quadratic and
projection arguments. New are the complete realizability of student-relative
splits of a fixed raw Gram, the regret of raw weighting and the
student-relative marginal value, with general-learner versions.
\item \emph{\Cref{app:q2}.} That unlabeled predictions cannot reveal an error
common to all supervisors is elementary. New are the task-specific
identification boundaries, the bounded nondegenerate counterexample, the
leakage law with the anchor-free robust test for structured teachers, and the
student-relative factor-model result.
\item \emph{\Cref{app:q3}.} Sub-Gaussian concentration, Gaussian width,
statistical dimension, constrained least squares and testing lower bounds are
standard, as are least squares, residualization and inverse-Wishart moments.
New are their reduction to the two decision problems, the residualization
that makes selection depend on $\dN$, the exact comparisons of structured
selection with ordinary stacking and of weak supervision with anchors alone
(\cref{q3:sec:compare}), and the fixed-dictionary passive constructions with
matching lower bounds.
\item \emph{\Cref{app:q4}.} Ridge, stacking and effective dimensions are
standard. New is a conditional decision guarantee for multiple supervisors:
operator deviation, target-moment estimation, weight regret and gain-sign
stability; it is not a minimax theorem for every $(n,N,\alpha)$ regime.
\end{itemize}

\subsection{Comparison by line of work}

\paragraph{Weak-to-strong theory.}
We see three layers. First, misfit and gain: for convex student classes the
gain is controlled by the student--supervisor misfit
\citep{charikar2024quantifying}, also beyond squared loss
\citep{mulgund2025relating,xu2026bregman}. \citet{charikar2024quantifying}
already use this relation to choose among strong models trained by different
weak models (the one with the smallest weak error minus misfit) and raise
ensembling as future work; what we add is the joint aggregation decision, its
observational equivalence classes and identification boundaries, and the
label cost of recovering it. For one supervisor and a
realizable target, \cref{prop:risk} with a projection student reduces to this
identity. Second, high-dimensional and representation analyses. In
ridgeless regression, a weak teacher's variance is inherited on the subspace
it shares with the student and attenuated on their discrepancy subspace
\citep{dong2025discrepancies}. Distillation from a weak surrogate can beat
strong labels at a fixed budget but cannot improve the data scaling law
\citep{ildiz2025high}. Label-free representation metrics predict W2S trends
under particular representation models with ridge-regularized linear heads
\citep{xue2025representations}. Their object is the prediction gap between
weakly and cleanly trained students, which they relate to the risk through
additional conditions; we study the absolute risk, the same-aggregate gain
and selection, which the prediction gap does not determine in general
(\cref{prop:risk} adds the bias alignment). Their predictions hold under
particular representation models; our non-identification is uniform over
targets compatible with the same unlabeled observations. Our transmitted/filtered decomposition extends
the student-relative viewpoint of these works to any fixed linear learner and
to decisions among several supervisors. Third, regularization, early stopping
and scaling. Early-stopped random-feature students provably beat their
teacher \citep{medvedev2025weak}; an early-stopped linear head is a spectral
filter and hence a common linear learner in our sense. The same work proves
a population limitation, $R_S\ge\tfrac34R_W^2$, under shrinking optimality of
teacher and student; an empirical ridge optimizer or an anchor-selected
mixture need not satisfy that condition, and nothing here claims to defeat
it. Two-stage random-features ridge regression can improve the scaling law
\citep{wu2026improved}; their unnormalized penalty corresponds to
$\Nfit\lambda$ in \cref{sec:finite}, and we do not reproduce their
deterministic equivalents. \citet{moniri2025mechanisms} identify regularization
compensation in a proportional ridge-to-ridge model. Our decomposition offers
an interpretation of their result rather than a new theorem: matching the
four terms of \cref{prop:risk} to their deterministic equivalents, $c_T$
isolates the channel created by the teacher's shrinkage bias, and their
threshold is where filtering $\Sigma-\Gamma_T$ no longer pays for $B_T-2c_T$
(details in \cref{comp:ext:moniri}). Pseudolabel correction and coverage expansion
\citep{lang2024theoretical} and nonlinear feature learning \citep{oh2025linear}
lie outside our linear-learner framework. \citet{mulgund2025relating} also
combine several \emph{strong} models, not weak supervisors.

\paragraph{Multiple weak supervisors and label models.}
Dawid--Skene, data programming and later label models identify supervisor
quality from conditional independence, agreement or higher-order dependence
\citep{dawid1979maximum,ratner2016data}. Our results specify what such
assumptions must control: the part of the supervisors' common error aligned
with the student's task spaces. Empirical multi-supervisor and ensemble W2S
methods combine, weight or route weak models, typically with hand-set or
heuristic weights \citep{sang2024improving,liu2024cosupervised,
agrawal2024ensemw2s,cui2025bayesian}. Routing by teacher--student consistency
\citep{liu2024cosupervised} is close in spirit to a student-relative
criterion, but by \cref{thm:counter} agreement alone cannot identify
student-relative value. Supervisor--student complementarity matters
empirically beyond supervisor accuracy \citep{goel2025great}; our theory
gives a sharper, post-learning notion of complementarity. Panels of LLM
judges with correlated errors carry far fewer effective votes than their
nominal count suggests \citep{kohli2026nine}; with a downstream student, even
the full raw error correlation does not settle value (\cref{thm:fixed-gram}).
We specify which objective to optimize, when it is observable, and how much
clean information it takes.

\paragraph{Stacking, aggregation and forecast combination.}
Given the right Gram and target score, our weight selection is constrained
least squares, stacking or quadratic aggregation
\citep{wolpert1992stacked,breiman1996stacked,tsybakov2003optimal}; with the
Gram estimated on the anchors it \emph{is} stacking (\cref{q4:sec:paired}).
Forecast combination uses the raw forecast-error covariance to minimize the
error of the combined predictor \citep{bates1969combination}. A W2S student
should instead optimize the transmitted error, which in general cannot be
inferred from the raw errors or from unlabeled agreement.

\paragraph{Partial identification, few-label correction and ranking.}
Our sensitivity intervals follow the partial-identification approach
\citep{manski2003partial}, and rankings by LLM judges raise a similar
identifiability question \citep{vossler2026simplex}. Prediction-powered
inference \citep{angelopoulos2023prediction} and semi-supervised estimation
\citep{azriel2022semi,yuval2022semi} use few labels to correct many
predictions; our residualized anchor estimator is a correction of this kind.
MultiPPI \citep{cowenbreen2026multiple} allocates measurements across several
proxies to estimate a population functional, using the joint covariance of
the gold measurement and the proxies and their costs. In W2S the combination
trains a downstream learner instead, so the objective is post-learning
student risk, which in general cannot be recovered from the raw joint
distribution of supervisor predictions. Prediction-powered ranking
\citep{chatzi2024prediction} builds rank-sets from few human and many LLM
comparisons; our certificates play a similar role for student-relative gains,
with simultaneous-sign complexity governed by query geometry and matching
lower bounds.

\paragraph{Local geometry and finite-sample learning.}
Gaussian width, statistical dimension of tangent cones, testing lower bounds,
effective dimension and matrix concentration are standard
\citep{boucheron2013concentration,amelunxen2014living,chatterjee2014new,
tsybakov2009introduction,hsu2012random,tropp2012user}. We connect them to the
two decision tasks of W2S, and show that population spectrum, local decision
geometry and passive coverage are three distinct levels of complexity.

\paragraph{Open problems.}
The optimal dependence on $L>1$ for sign recovery and a matching
high-probability selection lower bound remain open; so does
multi-supervisor identification under general proper losses and with
active anchor selection.

\section{Further Results}\label{app:ext}\label{ext:filters}\label{ext:perturb}\label{ext:moniri}\label{ext:rates}\label{ext:scope}\label{ext:short}

The following results complete the theory but are used neither in the main
text nor in the proof chains of \cref{app:q1,app:q2,app:q3,app:q4}. They are
stated and proved in \cref{comp:ext:filters,comp:ext:perturb,comp:ext:moniri,comp:ext:rates,comp:ext:scope,comp:short}.

\begin{center}\small
\begin{tabularx}{\textwidth}{@{}>{\raggedright}p{0.32\textwidth}>{\raggedright}X>{\raggedright\arraybackslash}p{0.12\textwidth}@{}}
\toprule
Topic & Content & Appendix\tabularnewline
\midrule
Spectral views of a general learner and spectral-filter learners & signed-energy form of the gain; spectral risk formula; ridge, early stopping, gradient flow, cut-off & \cref*{comp:ext:filters}\tabularnewline
Operator perturbation and projection limits & risk, gain and weight stability under learner perturbation; projection limits & \cref*{comp:ext:perturb}\tabularnewline
Interpretation in a proportional ridge-to-ridge model & interpretation of the risk terms in a proportional ridge-to-ridge model & \cref*{comp:ext:moniri}\tabularnewline
Spectral rates and diagnostics for the fitted student & spectral and source-condition rates for the fitted student; Gram diagnostics & \cref*{comp:ext:rates}\tabularnewline
Dependence, kernels, random features and linearization & same-sample dependence; kernels; random features; conditional linearization & \cref*{comp:ext:scope}\tabularnewline
Short consequences & one-line corollaries, examples and remarks used nowhere else & \cref*{comp:short}\tabularnewline
\bottomrule
\end{tabularx}
\end{center}

\section{Spectral views of a general learner and spectral-filter learners}\label{comp:ext:filters}

For a general common linear learner the gain is a signed energy
removal, $G_T=\ip e{(I-T^*T)e}+2\ip{Te}{b_T}-B_T$, with positive and negative
parts and a singular-value form. For a self-adjoint learner with spectral
value $t$ the risk is
$\int t^2\,d\mu_e+\int(1-t)^2\,d\mu_f-2\int t(1-t)\,d\mu_{e,f}$, so each
spectral component transmits a fraction $t^2$ of the error energy and enters
the bias alignment with weight $t(1-t)$; ridge, early-stopped gradient
descent, gradient flow and spectral cut-off are the instances $t=\phi(\mu)$.
The cross term vanishes for all targets and errors if and only if $T$ is an
orthogonal projection.
These results justify treating the learners listed in \cref{sec:setup} as
common linear learners; the main text uses only \cref{prop:risk}.

\subsection{Signed energy removal and spectral views of a general learner}\label{q1:sec:spectral-views}

\begin{proposition}[Contraction criterion and residual correction]\label{q1:prop:Eop}
$\Eop_T=I-T^*T\succeq0$ iff $\norm T_\op\le1$. Moreover
$\Eop_T=(I-T)^*(I-T)+T^*(I-T)+(I-T)^*T$, so
$\ip e{\Eop_Te}=\norm{(I-T)e}^2+2\ip{Te}{(I-T)e}$.
\end{proposition}
\begin{proof}
$\Eop_T\succeq0$ means $\norm{Te}^2\le\norm e^2$ for all $e$. The three
operators on the right sum to $(I-T-T^*+T^*T)+(T^*-T^*T)+(T-T^*T)=I-T^*T$.
\end{proof}

\begin{proposition}[Positive and negative parts]\label{q1:prop:posneg}
Let $\Eop_{T,\pm}$ be the positive and negative parts of $\Eop_T$. Then
$\Eop_T=\Eop_{T,+}-\Eop_{T,-}$, $\Eop_{T,+}\Eop_{T,-}=0$ and
\[
 G_T(e)=\norm{\Eop_{T,+}^{1/2}e}^2-\norm{\Eop_{T,-}^{1/2}e}^2+2\ip{Te}{b_T}-B_T .
\]
For a dictionary, $\Lambda_T=E^*\Eop_TE\succeq0$ iff $\norm{Th}\le\norm h$ for
every $h\in\Ran E$, which can hold although $\Eop_T$ is indefinite on $\Hh$.
\end{proposition}
\begin{proof}
Spectral calculus applied to $\max(x,0)$ and $\max(-x,0)$; and
$w^\top\Lambda_Tw=\norm{Ew}^2-\norm{TEw}^2$, where $Ew$ exhausts $\Ran E$.
\end{proof}

\begin{proposition}[Singular-value form]\label{q1:prop:svd}
In $\R^d$ let $T=U\diag(\sigma_1,\dots,\sigma_d)V^\top$ (zero singular values
included), $\epsilon_i=\ip e{v_i}$ and $\beta_i=\ip{b_T}{u_i}$. Then
\[
 R_T=B_T+\sum_i\sigma_i^2\epsilon_i^2-2\sum_i\sigma_i\epsilon_i\beta_i,\qquad
 G_T=\sum_i(1-\sigma_i^2)\epsilon_i^2+2\sum_i\sigma_i\epsilon_i\beta_i-B_T .
\]
\end{proposition}
\begin{proof}
$e=\sum_i\epsilon_iv_i$ and $Te=\sum_i\sigma_i\epsilon_iu_i$; substitute in
Theorem~\ref{q1:thm:single-general}.
\end{proof}

\begin{example}[Eigenvalues do not control nonnormal amplification]\label{q1:auto:eigenvalues-do-not-control-nonnormal-amp}
For $M>1$, $T=\left(\begin{smallmatrix}0&M\\0&0\end{smallmatrix}\right)$, $f=0$,
$e=(0,1)^\top$: both eigenvalues are $0$, $\norm T_\op=M$ and $G_T=1-M^2<0$.
\end{example}

\begin{theorem}[Spectral risk formula]\label{q1:thm:spectral}
Let $T$ be bounded self-adjoint with spectral measure $\mathsf E_T$, and
$\mu_{h,k}(\mathcal B)=\ip h{\mathsf E_T(\mathcal B)k}$, $\mu_h=\mu_{h,h}$. Then
\begin{align*}
 R_T&=\int t^2\,d\mu_e+\int(1-t)^2\,d\mu_f-2\int t(1-t)\,d\mu_{e,f},\\
 G_T&=\int(1-t^2)\,d\mu_e-\int(1-t)^2\,d\mu_f+2\int t(1-t)\,d\mu_{e,f}.
\end{align*}
For a positive contraction the removal factor $1-t^2\ge0$, but the cross
integral has no prescribed sign.
\end{theorem}
\begin{proof}
Spectral calculus gives $T^2=\int t^2d\mathsf E_T$, $(I-T)^2=\int(1-t)^2d\mathsf E_T$,
$T(I-T)=\int t(1-t)d\mathsf E_T$; insert in Theorem~\ref{q1:thm:single-general}.
For a finite disjoint partition $(\mathcal B_\ell)$,
$\sum_\ell\abs{\ip{\mathsf E_T(\mathcal B_\ell)e}{\mathsf E_T(\mathcal B_\ell)f}}\le\norm e\norm f$
by Cauchy--Schwarz, so the cross measure has finite variation and all
integrals of bounded integrands exist.
\end{proof}

\paragraph{Discrete modes.} If $T$ has an orthonormal eigenbasis $(u_i)$ with
$Tu_i=a_iu_i$, and $f_i,e_i,g_i$ are coordinates, then
\[
 R_T=\sum_i[a_ie_i-(1-a_i)f_i]^2,\qquad
 G_T=\sum_i[(1-a_i^2)e_i^2-(1-a_i)^2f_i^2+2a_i(1-a_i)e_if_i].
\]
For one mode with $0\le a\le1$, $G_i=(1-a)g_i[(1+a)g_i-2f_i]$ and
$R_i(a)=(ag_i-f_i)^2$; for $g_i\ne0$ the risk-minimizing multiplier is
$\operatorname{clip}_{[0,1]}(f_i/g_i)$. More attenuation is not uniformly
beneficial; in ridge or early stopping one parameter links all multipliers.
For a dictionary, with $\epsilon_i=(\ip{e_1}{u_i},\dots,\ip{e_m}{u_i})^\top$,
\[
 \Sigma=\sum_i\epsilon_i\epsilon_i^\top,\quad
 \Gamma_T=\sum_ia_i^2\epsilon_i\epsilon_i^\top,\quad
 c_T=\sum_ia_i(1-a_i)f_i\epsilon_i,\quad
 B_T=\sum_i(1-a_i)^2f_i^2 .
\]
Transmitted error is weighted by $a_i^2$, and bias alignment by
$a_i(1-a_i)$.

\subsection{Spectral-filter learners: ridge, early stopping, flow, cutoff}\label{q1:sec:filters}

\paragraph{Population feature model.} Let $\Ff$ be a real Hilbert parameter
space and $\Psi:\Ff\to\Hh$ a bounded feature-synthesis operator; set
$\mathcal L=\Psi\Psi^*$, $M_{\mathcal L}=\norm{\mathcal L}_\op>0$,
$\Ss=\overline{\Ran\Psi}=(\Ker\mathcal L)^\perp$ and $P=\mathbf 1_{(0,\infty)}(\mathcal L)$.
Indeed $\ip h{\mathcal Lh}=\norm{\Psi^*h}^2$, so $\Ker\mathcal L=\Ker\Psi^*=(\Ran\Psi)^\perp$.
$\mathcal L$ is a positive feature operator, not an error Gram.

\begin{remark}[Population operators versus finite-sample smoothers]\label{q1:auto:population-operators-versus-finite-sampl}
The formulation below is a population least-squares problem. A map that
evaluates an $L^2(P_X)$ equivalence class at finitely many training points
is not automatically well defined or bounded on $L^2(P_X)$, so an empirical
training rule is not automatically a bounded operator satisfying
Assumption~\ref{q1:ass:T}; its domain must be specified separately (\cref{app:q4}).
\end{remark}

For a bounded real Borel filter $a$ with $a(0)=0$ let $T=a(\mathcal L)$, and
$\mu^{\mathcal L}_{h,k}(\mathcal B)=\ip h{\mathsf E_{\mathcal L}(\mathcal B)k}$.
Theorem~\ref{q1:thm:spectral} becomes
\begin{align*}
 B_T&=A(f)+\int_{(0,M_{\mathcal L}]}(1-a(s))^2\,d\mu^{\mathcal L}_f(s),\\
 (\Gamma_T)_{jk}&=\int_{(0,M_{\mathcal L}]}a(s)^2\,d\mu^{\mathcal L}_{e_j,e_k}(s),\qquad
 (c_T)_j=\int_{(0,M_{\mathcal L}]}a(s)(1-a(s))\,d\mu^{\mathcal L}_{e_j,f}(s),
\end{align*}
because at $s=0$ the three multipliers $(1-a)^2,a^2,a(1-a)$ equal $1,0,0$ and
$\mathsf E_{\mathcal L}(\{0\})=Q$.

\begin{theorem}[Ridge]\label{q1:thm:ridge}
For $\lambda>0$ the unique minimizer of $\norm{\Psi u-g}^2+\lambda\norm u^2$
over $u\in\Ff$ is $u_\lambda=(\Psi^*\Psi+\lambda I)^{-1}\Psi^*g$, with
prediction $T_\lambda g$, where
$T_\lambda=\Psi(\Psi^*\Psi+\lambda I)^{-1}\Psi^*=\mathcal L(\mathcal L+\lambda I)^{-1}$.
Thus $a_\lambda(s)=s/(s+\lambda)$, $T_\lambda$ is a positive contraction, and
$1-a_\lambda=\lambda/(s+\lambda)$, $a_\lambda^2=s^2/(s+\lambda)^2$,
$a_\lambda(1-a_\lambda)=s\lambda/(s+\lambda)^2$.
\end{theorem}
\begin{proof}
The directional derivative $2\ip v{\Psi^*(\Psi u-g)+\lambda u}$ vanishes iff
$(\Psi^*\Psi+\lambda I)u=\Psi^*g$, and $\Psi^*\Psi+\lambda I\succeq\lambda I$
is boundedly invertible. At $u_\lambda$ the increment in direction $v$ is
$\norm{\Psi v}^2+\lambda\norm v^2>0$. The identity
$(\mathcal L+\lambda I)\Psi=\Psi(\Psi^*\Psi+\lambda I)$ gives
$\Psi(\Psi^*\Psi+\lambda I)^{-1}=(\mathcal L+\lambda I)^{-1}\Psi$; multiply by
$\Psi^*$.
\end{proof}

\begin{remark}[The penalty must be specified]\label{q1:auto:the-penalty-must-be-specified}
Penalizing $\norm u^2$ yields the filter above. Penalizing the prediction,
$\min_{h\in\Ss}\norm{h-g}^2+\lambda\norm h^2$, yields $T=(1+\lambda)^{-1}P$, a
different algorithm. An unpenalized intercept must be represented
separately: its direction is not shrunk.
\end{remark}

\begin{theorem}[Early-stopped gradient descent and gradient flow]\label{q1:thm:gd}
Gradient descent $u_{t+1}=u_t+\eta\Psi^*(g-\Psi u_t)$ from $u_0=0$ has
prediction operator $T_t=I-(I-\eta\mathcal L)^t$, $a_t(s)=1-(1-\eta s)^t$. For
$0<\eta\le1/M_{\mathcal L}$, $0\preceq T_t\preceq I$ and the multipliers
increase in $t$; for $1/M_{\mathcal L}<\eta<2/M_{\mathcal L}$ the iteration still
converges strongly to $P$, but odd iterates have $a_t(s)>1$ when $\eta s>1$.
Gradient flow $\dot h_t=\mathcal L(g-h_t)$, $h_0=0$, gives
$T_t=I-e^{-t\mathcal L}$, a positive contraction for all $t\ge0$.
\end{theorem}
\begin{proof}
With $h_t=\Psi u_t$, $g-h_{t+1}=(I-\eta\mathcal L)(g-h_t)$. For
$\eta M_{\mathcal L}\le1$, $1-\eta s\in[0,1]$. For $0<\eta<2/M_{\mathcal L}$,
$\abs{1-\eta s}<1$ for $s>0$; a negative base gives $a_t>1$ for odd $t$.
Strong convergence follows by dominated convergence of the squared residual
multiplier, bounded by one on positive spectrum. For flow differentiate
$I-e^{-t\mathcal L}$.
\end{proof}

\begin{theorem}[Spectral cutoff]\label{q1:auto:spectral-cutoff}
For $\tau>0$, $T_\tau=\mathbf 1_{[\tau,\infty)}(\mathcal L)$ is the orthogonal
projection onto $\Ss_\tau\subseteq\Ss$, with
$B_{T_\tau}=A(f)+\mu^{\mathcal L}_f((0,\tau))$, $c_{T_\tau}=0$ and
$(\Gamma_{T_\tau})_{jk}=\mu^{\mathcal L}_{e_j,e_k}([\tau,M_{\mathcal L}])$. The
projection identities hold exactly for the smaller class $\Ss_\tau$, not with
$A(f)$ in place of $B_{T_\tau}$.
\end{theorem}
\begin{proof}
A $\{0,1\}$-valued real multiplier gives $T=T^*=T^2$, with
$a(1-a)\equiv0$; use the filter formulas.
\end{proof}

\begin{proposition}[Effective transmitted mode count]\label{q1:prop:decay}
Let $\mathcal L$ be compact with positive eigenvalues
$c_-i^{-p}\le\mu_i\le c_+i^{-p}$ and $d_{\mathrm{tr}}(T)=\sum_ia(\mu_i)^2$. If
$p>1/2$, then for small $\lambda$ and large $t$:
$d_{\mathrm{tr}}(T_\lambda^{\mathrm{ridge}})\asymp\lambda^{-1/p}$,
$d_{\mathrm{tr}}(T_t^{\mathrm{flow}})\asymp t^{1/p}$ and
$d_{\mathrm{tr}}(T_t^{\mathrm{GD}})\asymp(\eta t)^{1/p}$ ($0<\eta\le1/M_{\mathcal L}$);
for cutoff and any $p>0$, $d_{\mathrm{tr}}(T_\tau)\asymp\tau^{-1/p}$. Constants
depend on $p,c_\pm$.
\end{proposition}
\begin{proof}
Ridge: $a_\lambda\le\min\{1,s/\lambda\}$ and $a_\lambda\ge1/2$ for $s\ge\lambda$;
at least $\lfloor(c_-/\lambda)^{1/p}\rfloor$ modes have $a_\lambda\ge1/2$.
With $J_0=\lceil(c_+/\lambda)^{1/p}\rceil$,
$\sum_ia_\lambda(\mu_i)^2\le J_0+\lambda^{-2}c_+^2\sum_{i>J_0}i^{-2p}
\le J_0+c_+^2J_0^{1-2p}/((2p-1)\lambda^2)\lesssim\lambda^{-1/p}$. Flow:
$1-e^{-ts}\le\min\{1,ts\}$ and $\ge1-e^{-1}$ for $ts\ge1$. GD with $x=\eta s$:
$1-e^{-tx}\le1-(1-x)^t\le\min\{1,tx\}$. Cutoff counts $\mu_i\ge\tau$.
\end{proof}

$d_{\mathrm{tr}}(T)$ is a filter-energy trace; it is neither the number of
supervisors nor the selection dimension $k$ of Appendices~\ref{app:q2}--\ref{app:q3}, and it does not
determine risk, which still depends on target and error coefficients and their
alignment.

\section{Operator perturbation and projection limits}\label{comp:ext:perturb}

Exact perturbation expansions of risk, gain and optimal weights when
the learner changes from $T$ to $T'$; uniform risk, gain and weight stability
in operator norm; the passage from feature-operator to learner perturbation;
and convergence of ridge and other filters to the projection case as the
regularization vanishes, with source-condition rates. \Cref{app:q4} uses only
the finite-sample counterparts proved there.

\subsection{Operator perturbation and projection limits}\label{q1:sec:perturb}

Let $\widetilde T=T+\Xi$ be another bounded linear learner,
$\delta_\Xi=\norm\Xi_\op$; tildes denote replacement of $T$ by $\widetilde T$.

\begin{theorem}[Exact perturbation expansions]\label{q1:thm:pert}
For $w\in\Waff$,
\[
 R_{\widetilde T}(w)-R_T(w)=2\ip{Tg_w-f}{\Xi g_w}+\norm{\Xi g_w}^2=-[G_{\widetilde T}(w)-G_T(w)],
\]
\[
 \widetilde b-b=-\Xi f,\quad \widetilde B-B=-2\ip b{\Xi f}+\norm{\Xi f}^2,\quad
 \widetilde\Gamma-\Gamma=E^*(T^*\Xi+\Xi^*T+\Xi^*\Xi)E,\quad
 \widetilde c-c=E^*[\Xi^*b-\widetilde T^*\Xi f],
\]
so that $\abs{\widetilde B-B}\le2\delta_\Xi\norm b\norm f+\delta_\Xi^2\norm f^2$,
$\varepsilon_\Gamma:=\norm{\widetilde\Gamma-\Gamma}_\op\le\delta_\Xi(\norm T_\op+\norm{\widetilde T}_\op)\norm E_\op^2$
and $\varepsilon_c:=\norm{\widetilde c-c}\le\delta_\Xi\norm E_\op(\norm b+\norm{\widetilde T}_\op\norm f)$.
\end{theorem}
\begin{proof}
$\widetilde Tg_w-f=(Tg_w-f)+\Xi g_w$; weak risk is unchanged.
$(T^*+\Xi^*)(b-\Xi f)-T^*b=\Xi^*b-\widetilde T^*\Xi f$. For $\varepsilon_\Gamma$
use the expansion $T^*\Xi+\Xi^*\widetilde T$ and submultiplicativity.
\end{proof}

\begin{theorem}[Uniform risk and gain stability]\label{q1:thm:unif-pert}
Let $\mathcal W\subseteq\Waff$ with $M_{\mathcal W}=\sup_{\mathcal W}\norm{g_w}<\infty$. Then
\[
 \sup_{\mathcal W}\abs{R_{\widetilde T}-R_T}=\sup_{\mathcal W}\abs{G_{\widetilde T}-G_T}
 \le\varepsilon_R:=2\delta_\Xi M_{\mathcal W}(\norm T_\op M_{\mathcal W}+\norm f)+\delta_\Xi^2M_{\mathcal W}^2 .
\]
Alternatively, with $W_2=\sup_{\mathcal W}\norm w$, the bound
$\abs{\widetilde B-B}+\varepsilon_\Gamma W_2^2+2\varepsilon_cW_2$ holds. For $\WL$
one may take $M_{\mathcal W}\le L\max_j\norm{g_j}$ and $W_2\le L$. If
$\abs{G_T(w)}>\varepsilon_R$ the sign of the gain is unchanged.
\end{theorem}
\begin{proof}
Cauchy--Schwarz on Theorem~\ref{q1:thm:pert} with
$\norm{Tg_w-f}\le\norm T_\op M_{\mathcal W}+\norm f$ and
$\norm{\Xi g_w}\le\delta_\Xi M_{\mathcal W}$; the alternative subtracts the
two quadratic forms of Theorem~\ref{q1:thm:multi}.
\end{proof}

\begin{corollary}[Optimized values and transferred weights]\label{q1:cor:transfer}
If both objectives attain their minima on $\mathcal W$ (e.g.\ $\mathcal W$
compact), then $\abs{\min_{\mathcal W}R_{\widetilde T}-\min_{\mathcal W}R_T}\le\varepsilon_R$,
and any $\eta$-optimal $\widetilde w$ for $R_{\widetilde T}$ has
$R_T(\widetilde w)-\min_{\mathcal W}R_T\le2\varepsilon_R+\eta$. No uniqueness or
positive definiteness is needed.
\end{corollary}
\begin{proof}
$R_T-\varepsilon_R\le R_{\widetilde T}\le R_T+\varepsilon_R$ on $\mathcal W$;
chain $R_T(\widetilde w)\le R_{\widetilde T}(\widetilde w)+\varepsilon_R\le R_{\widetilde T}(w^\star)+\eta+\varepsilon_R\le R_T(w^\star)+2\varepsilon_R+\eta$.
\end{proof}

\begin{assumption}[Restricted curvature]\label{q1:ass:curv}
For $m>1$ there is $\kappa_{\mathrm c}>0$ with $d^\top\Gamma_Td\ge\kappa_{\mathrm c}\norm d^2$
for all $d\in\one^\perp$, i.e.\ $O^\top\Gamma_TO\succeq\kappa_{\mathrm c}I$.
\end{assumption}

\begin{theorem}[Weight stability]\label{q1:thm:weight-stab}
Let $\mathcal W\subseteq\Waff$ be convex with both minima attained at
$w^\star,\widetilde w^\star$, and let Assumption~\ref{q1:ass:curv} hold. With
$a_\Xi=\varepsilon_\Gamma\norm{\widetilde w^\star}+\varepsilon_c$,
\[
 \norm{\widetilde w^\star-w^\star}\le a_\Xi/\kappa_{\mathrm c},\qquad
 R_T(\widetilde w^\star)-R_T(w^\star)\le a_\Xi^2/\kappa_{\mathrm c}.
\]
For unrestricted affine weights, if moreover $\varepsilon_\Gamma<\kappa_{\mathrm c}$,
both optimizers are unique and
$\norm{\widetilde w^\star-w^\star}\le(\varepsilon_c+\varepsilon_\Gamma\norm{w^\star})/(\kappa_{\mathrm c}-\varepsilon_\Gamma)$,
with regret at most $\norm{\Gamma_T}_\op$ times its square.
\end{theorem}
\begin{proof}
Put $d=\widetilde w^\star-w^\star$ and $h=(\Gamma-\widetilde\Gamma)\widetilde w^\star+(\widetilde c-c)$.
The variational inequalities $d^\top(\Gamma w^\star-c)\ge0$ and
$d^\top(\widetilde\Gamma\widetilde w^\star-\widetilde c)\le0$ subtract to
$\kappa_{\mathrm c}\norm d^2\le d^\top\Gamma d\le d^\top h\le a_\Xi\norm d$.
Expanding from $\widetilde w^\star$,
$R_T(\widetilde w^\star)-R_T(w^\star)=2d^\top(\Gamma\widetilde w^\star-c)-d^\top\Gamma d\le2a_\Xi\norm d-\kappa_{\mathrm c}\norm d^2\le a_\Xi^2/\kappa_{\mathrm c}$.
In the affine case use contrast coordinates (Theorem~\ref{q1:thm:affine-general}):
$\widetilde H(\widetilde u^\star-u^\star)=O^\top[(\widetilde c-c)-(\widetilde\Gamma-\Gamma)w^\star]$
with $\widetilde H\succeq(\kappa_{\mathrm c}-\varepsilon_\Gamma)I$; the affine regret is
$d^\top\Gamma_Td$ (Theorem~\ref{q1:thm:kkt}).
\end{proof}

\begin{example}[Close operators, distant affine optima]\label{q1:ex:unstable}
$\Hh=\R^2$, $f=(1,0)^\top$, $g_1=(1,0)^\top$, $g_2=(-1,0)^\top$, $P=\diag(0,1)$,
$T_\epsilon=\diag(\epsilon,1)$: $\norm{T_\epsilon-P}_\op=\abs\epsilon$ and
$R_{T_\epsilon}(t,1-t)=[\epsilon(2t-1)-1]^2$. For $0<\abs\epsilon<1$ the simplex
optimum is $t=1$ if $\epsilon>0$ and $t=0$ if $\epsilon<0$; the affine optimum
$t=(1+\epsilon^{-1})/2$ has risk $0$, whereas at $\epsilon=0$ every affine
risk equals $1$. Without curvature or bounded weights, operator convergence
does not give convergence of optima or optimal values.
\end{example}

\begin{proposition}[From feature-operator to learner perturbation]\label{q1:prop:filter-pert}
For bounded positive self-adjoint $\mathcal L,\widetilde{\mathcal L}$:
$\norm{T_\lambda(\widetilde{\mathcal L})-T_\lambda(\mathcal L)}_\op\le\norm{\widetilde{\mathcal L}-\mathcal L}_\op/\lambda$;
for GD with $\eta\max\{\norm{\mathcal L},\norm{\widetilde{\mathcal L}}\}\le1$,
$\norm{T_t(\widetilde{\mathcal L})-T_t(\mathcal L)}_\op\le t\eta\norm{\widetilde{\mathcal L}-\mathcal L}_\op$;
for flow the bound is $t\norm{\widetilde{\mathcal L}-\mathcal L}_\op$. No
commutation is required. Cutoff has no global Lipschitz bound:
$\mathcal L=(\tau-\epsilon)I$, $\widetilde{\mathcal L}=(\tau+\epsilon)I$ differ by
$2\epsilon$ but their cutoffs at $\tau$ differ by $I$.
\end{proposition}
\begin{proof}
$T_\lambda=I-\lambda(\mathcal L+\lambda I)^{-1}$ and the resolvent identity
$T_\lambda(\widetilde{\mathcal L})-T_\lambda(\mathcal L)=\lambda(\widetilde{\mathcal L}+\lambda)^{-1}(\widetilde{\mathcal L}-\mathcal L)(\mathcal L+\lambda)^{-1}$.
For GD telescope $A^t-B^t=\sum_{j<t}A^{t-1-j}(A-B)B^j$ with
$\norm A,\norm B\le1$. For flow,
$e^{-t\mathcal L}-e^{-t\widetilde{\mathcal L}}=\int_0^te^{-(t-s)\widetilde{\mathcal L}}(\widetilde{\mathcal L}-\mathcal L)e^{-s\mathcal L}\,ds$.
\end{proof}

\begin{theorem}[Projection limits]\label{q1:thm:limits}
(i) $T_\lambda^{\mathrm{ridge}}\to P$ ($\lambda\downarrow0$), $T_t^{\mathrm{flow}}\to P$,
$T_\tau^{\mathrm{cutoff}}\to P$ ($\tau\downarrow0$) and, for
$0<\eta<2/M_{\mathcal L}$, $T_t^{\mathrm{GD}}\to P$ strongly, without a
spectral gap. (ii) If positive spectrum accumulates at $0$, each difference
has operator norm $1$ at every finite parameter; if
$\spec\mathcal L\subseteq\{0\}\cup[\mu_0,M_{\mathcal L}]$, $\mu_0>0$, then
$\norm{T_\lambda-P}\le\lambda/(\mu_0+\lambda)$, $\norm{T_t^{\mathrm{flow}}-P}\le e^{-\mu_0t}$,
$\norm{T_t^{\mathrm{GD}}-P}\le\max_{s\in[\mu_0,M_{\mathcal L}]}\abs{1-\eta s}^t$ and
$T_\tau=P$ for $\tau\le\mu_0$. (iii) If $T_\nu\to P$ strongly, with fixed
$f,g_1,\dots,g_m$, then $B_{T_\nu}\to A$, $c_{T_\nu}\to0$,
$\Gamma_{T_\nu}\to\Gpar$, $\Lambda_{T_\nu}\to\Gperp$, and for every finite $L$,
$\sup_{\WL}\abs{R_{T_\nu}-R_P}\le2L\epsilon_\nu(LM_g+\norm f)+L^2\epsilon_\nu^2\to0$,
where $\epsilon_\nu=\max_j\norm{(T_\nu-P)g_j}$ and $M_g=\max_j\norm{g_j}$.
Optimal values on compact $\mathcal W\subseteq\WL$ converge, cluster points of
minimizers are $P$-minimizers, and a unique $P$-minimizer is the limit.
\end{theorem}
\begin{proof}
(i) Each residual multiplier tends to $0$ on $(0,M_{\mathcal L}]$ and is bounded
by one, and $\norm{(T_\nu-P)h}^2=\int\abs{a_\nu-1}^2d\mu_h^{\mathcal L}$; dominated
convergence. (ii) The norm of a multiplier is its spectral essential supremum.
(iii) Continuity of inner products on finitely many entries; the uniform bound
is Theorem~\ref{q1:thm:pert} with reference $P$; then
Corollary~\ref{q1:cor:transfer} and compactness.
\end{proof}

\begin{remark}[A task-local operator norm]\label{q1:auto:a-task-local-operator-norm}
With $\mathcal K_0=\Span\{f,g_1,\dots,g_m\}$ and an orthonormal basis $(u_\ell)$
of it, strong convergence gives
$\norm{(T_\nu-P)\Pi_{\mathcal K_0}}_\op^2\le\sum_\ell\norm{(T_\nu-P)u_\ell}^2\to0$.
All predictions, biases and Grams are unchanged when $T_\nu,P$ are replaced by
$T_\nu\Pi_{\mathcal K_0},P\Pi_{\mathcal K_0}$, so the perturbation theorems apply
with this task-local norm even when global norm convergence fails.
\end{remark}

\begin{proposition}[Source-condition rates]\label{q1:prop:source}
If $Ph=\mathcal L^ru$ with $r>0$, then
$\norm{(T_\lambda-P)h}\le\lambda^r\norm u$ for $0<r\le1$ and
$\le\lambda M_{\mathcal L}^{r-1}\norm u$ for $r>1$;
$\norm{(T_t^{\mathrm{flow}}-P)h}\le(r/(et))^r\norm u$;
$\norm{(T_\tau-P)h}\le\tau^r\norm u$; GD with $0<\eta\le1/M_{\mathcal L}$
satisfies the flow bound with $t$ replaced by $\eta t$.
\end{proposition}
\begin{proof}
Every difference annihilates $\Ker\mathcal L$. For ridge,
$\lambda s^r/(s+\lambda)=\lambda^r x^r/(1+x)\le\lambda^r$ with $x=s/\lambda$ when
$r\le1$, and $\le\lambda s^{r-1}$ otherwise. For flow, $\max_ss^re^{-ts}=(r/(et))^r$.
Cutoff residuals live on $(0,\tau)$. For GD, $(1-\eta s)^t\le e^{-\eta ts}$.
\end{proof}

Applied to $h=f$ these bound $B_T-A$; applied to all $g_j$ they bound
$\epsilon_\nu$ in Theorem~\ref{q1:thm:limits}. Source conditions are additional
regularity assumptions; strong convergence alone gives no rate.

\section{Interpretation in a proportional ridge-to-ridge model}\label{comp:ext:moniri}

In the proportional ridge-to-ridge model of \citet{moniri2025mechanisms},
the four terms of \cref{prop:risk} have explicit deterministic equivalents
that recover their limiting risk difference; the bias alignment $c_T$
isolates the channel created by the teacher's shrinkage bias and is
proportional to its ridge parameter, and their threshold is where filtering
$\Sigma-\Gamma_T$ no longer pays for $B_T-2c_T$.

\subsection{Example: a proportional ridge-to-ridge model}\label{q1:sec:moniri}

This section evaluates the four terms of Theorem~\ref{q1:thm:single-general} in
the linear model of \citet[Section~2.1]{moniri2025mechanisms}, where a ridge
student is trained on the predictions of a ridge teacher. It shows how their
regularization-compensation threshold decomposes into filtering, the
student's own bias and the bias alignment.

\paragraph{Model.} $X\sim N(0,I_d)$, $f(x)=\beta_\star^\top x$ with
$\norm{\beta_\star}=1$, so that $\Hh$ restricted to linear functions is
$\R^d$ with the Euclidean inner product. The teacher is ridge regression with
parameter $\lambda_t$ on $n_t$ labelled samples with noise variance
$\sigma^2$; the student is ridge regression with parameter $\lambda_s$ on
$n_s$ fresh covariates labelled by the teacher; $d/n_t\to\gamma_t$ and
$d/n_s\to\gamma_s$. Write $\widehat C_p$ for the sample covariances,
$R_p=(\widehat C_p+\lambda_pI)^{-1}$, and let $m_{p,1}$ and $m_{p,2}$ be the
limits of $d^{-1}\tr R_p$ and $d^{-1}\tr R_p^2$ (Marchenko--Pastur), $p\in\{t,s\}$.

\paragraph{The student is a fixed linear learner.} Conditionally on the
student design, the student coefficient is
$S\widehat\beta_t$ with $S=R_s\widehat C_s=I-\lambda_sR_s$, a bounded linear
map of the teacher. Hence Theorem~\ref{q1:thm:single-general} applies exactly,
with $m=1$, $e=\widehat\beta_t-\beta_\star$, $T=S$ and
$b_T=(I-S)\beta_\star=\lambda_sR_s\beta_\star$. (Their weighted-ridge student
of Section~2.2 also has a fixed regularization matrix and is therefore a
fixed linear learner as well, as long as that matrix is not adapted to the
supervisor.)

\paragraph{Deterministic equivalents of the four terms.} The teacher error
is $e=-\lambda_tR_t\beta_\star+R_tX_t^\top\varepsilon/n_t$. Using the
independence of the two designs, isotropy, and the trace factorization of
their proof,
\begin{align*}
 \Sigma=\norm e^2&\to\lambda_t^2m_{t,2}+\sigma^2\gamma_t(m_{t,1}-\lambda_tm_{t,2}),\\
 \Gamma_T=\norm{Se}^2&\to\kappa\,\Sigma,\qquad \kappa=1-2\lambda_sm_{s,1}+\lambda_s^2m_{s,2}\in(0,1),\\
 B_T=\norm{\lambda_sR_s\beta_\star}^2&\to\lambda_s^2m_{s,2},\\
 c_T=\ip{Se}{\lambda_sR_s\beta_\star}&\to-\lambda_t\lambda_s\,m_{t,1}\,(m_{s,1}-\lambda_sm_{s,2})\le0 .
\end{align*}
The first line splits into the teacher's shrinkage bias and the fitted
noise; the second uses $d^{-1}\tr S^2=1-2\lambda_sd^{-1}\tr R_s+\lambda_s^2d^{-1}\tr R_s^2$;
in the last, the noise part of $e$ is independent of $b_T$ and contributes
nothing, while $m_{s,1}-\lambda_sm_{s,2}=\lim d^{-1}\tr(\widehat C_sR_s^2)\ge0$.
These are heuristic term-wise limits obtained with the same asymptotic
factorization as their proof; we do not reproduce a rigorous proof of each.

\paragraph{Recovery of their risk difference.} Inserting the four limits in
$G_T=\Sigma-\Gamma_T-B_T+2c_T$ gives exactly the negative of their limiting
risk difference $\Delta$ \citep[Theorem~4]{moniri2025mechanisms}, as a
polynomial identity in $(\lambda_t,\lambda_s,\sigma^2\gamma_t,m_{t,1},m_{t,2},m_{s,1},m_{s,2})$.
Three readings follow.
\begin{enumerate}
\item The bias alignment is proportional to the teacher's ridge parameter and
 nonpositive: the teacher's shrinkage bias $-\lambda_tR_t\beta_\star$ and the
 student's shrinkage bias point the same way, so the cross term always hurts,
 and it vanishes for an unregularized teacher. $c_T$ is not the only trace of
 $\lambda_t$ in the risk, since $\Sigma$ and $\Gamma_T$ also depend on it; it
 isolates the channel induced by the teacher's shrinkage bias.
\item Their statement that a student cannot beat an over-regularized teacher
 ($\lambda_t\ge\sigma^2\gamma_t$; \citealp[Theorem~5]{moniri2025mechanisms})
 reads $\Sigma-\Gamma_T\le B_T-2c_T$: filtering does not pay for the
 student's own bias and the alignment loss.
\item At the threshold $\lambda_t=\sigma^2\gamma_t$ the risk difference is
 $-G_T=B_T(1-\lambda_tm_{t,1})\ge0$, a nonnegative residual shrinkage term.
\end{enumerate}

\paragraph{Finite-dimensional check.} With $d=400$ and $30$ replicates per
case, the four terms computed directly from simulated teachers and students
agree with the limits above to within finite-$d$ error (largest deviation
about $2.5$ standard errors):

\begin{center}\small
\begin{tabular}{@{}lcccc@{}}
\toprule
$(\lambda_t,\lambda_s,\gamma_t,\gamma_s,\sigma^2)$ & $\Sigma$ & $\Gamma_T$ & $B_T$ & $c_T$\\
\midrule
$(0.1,0.5,0.5,0.5,1)$, under-regularized & $0.604\,/\,0.597$ & $0.229\,/\,0.224$ & $0.207\,/\,0.209$ & $-0.031\,/\,-0.030$\\
$(1,0.5,0.5,0.5,1)$, over-regularized & $0.455\,/\,0.449$ & $0.172\,/\,0.168$ & $0.207\,/\,0.209$ & $-0.116\,/\,-0.115$\\
$(0.2,1,0.8,2,0.5)$, $\gamma_s>1$ & $0.464\,/\,0.454$ & $0.088\,/\,0.087$ & $0.604\,/\,0.603$ & $-0.032\,/\,-0.031$\\
$(0.05,0.3,1.5,0.7,2)$, $\gamma_t>1$ & $2.940\,/\,2.947$ & $1.380\,/\,1.389$ & $0.177\,/\,0.177$ & $-0.068\,/\,-0.068$\\
\bottomrule
\end{tabular}

{\footnotesize Each entry: limit / simulation mean. Script:
\texttt{docs/literature/scripts/moniri\_term\_check.py}.}
\end{center}

\section{Spectral rates and diagnostics for the fitted student}\label{comp:ext:rates}

Explicit learning rates for the fitted student in three spectral
regimes and under a source condition (with saturation), an explicit
projection-learning rate, and empirical diagnostics for rank deficiency and
spectral bias in the candidate Gram. \Cref{thm:e2e} takes the fitting error
$\epsilon_D$ as given; these results instantiate it.

\subsection{Spectral rates for the student}\label{q4:sec:rates}

Let $(\varrho_k)$ be the positive eigenvalues of $\mathcal L$ with orthonormal
eigenfunctions $(u_k)$ spanning $\Ss$; $\deff(\lambda)=\sum_k\varrho_k/(\varrho_k+\lambda)$.

\begin{proposition}[Three spectral regimes]\label{q4:prop:regimes}
(i) If $\rank\mathcal L=d$, $\deff\le d$. (ii) If $\varrho_k\le ck^{-1/p}$ with $0<p<1$,
then $\deff(\lambda)\le1+(c/\lambda)^p/(1-p)$. (iii) If $\varrho_k\le ce^{-bk}$, then
$\deff(\lambda)\le2+b^{-1}\log_+(c/\lambda)+(e^b-1)^{-1}$.
\end{proposition}
\begin{proof}
Each summand is $\le\min\{1,\varrho_k/\lambda\}$. (ii) With $K=\lceil(c/\lambda)^p\rceil$,
$\deff\le K+(c/\lambda)\sum_{k>K}k^{-1/p}\le K+(c/\lambda)\tfrac p{1-p}K^{1-1/p}$ and
$K^{1-1/p}\le(c/\lambda)^{p-1}$. (iii) With $K=\max\{1,\lceil b^{-1}\log(c/\lambda)\rceil\}$,
$\deff\le K+ce^{-b(K+1)}/(\lambda(1-e^{-b}))$ and $ce^{-bK}\le\lambda$.
\end{proof}

\begin{proposition}[Source condition and saturation]\label{q4:prop:source}
If $Ph=\mathcal L^ru$ with $\norm u\le\mathsf R$ and $0<r\le1$, then
$\norm{(P-T_\lambda)h}\le\mathsf R\lambda^r$. For $r>1$ only
$\norm{(P-T_\lambda)h}\le\lambda\norm{\mathcal L}_\op^{r-1}\mathsf R$ holds in general (ridge has
qualification one).
\end{proposition}
\begin{proof}
$T_\lambda Qh=0$ and $(P-T_\lambda)h=\sum_k\frac{\lambda\varrho_k^r}{\varrho_k+\lambda}\ip u{u_k}u_k$, with
$\frac{\lambda\varrho_k^r}{\varrho_k+\lambda}=\lambda^r\frac{s^r}{1+s}\le\lambda^r$ for $s=\varrho_k/\lambda$; Parseval. For
$r>1$, $\lambda\varrho_k^r/(\varrho_k+\lambda)\le\lambda\varrho_k^{r-1}$.
\end{proof}

\begin{theorem}[An explicit projection-learning rate]\label{q4:thm:rate}
Assume, for the fixed functions considered, $Ph=\mathcal L^ru_h$ with
$\norm{u_h}\le\mathsf R$ and $\tfrac12\le r\le1$, $\norm{h-T_\lambda h}_\infty\le R_0$,
$\deff(\lambda)\le C_N\lambda^{-p}$ and $\varkappa_\psi(\lambda)\le C_\varkappa\deff(\lambda)$ with $0<p<1$. On the
simultaneous learning event with $\eta_{N_{\mathrm{fit}}}(\lambda,t)\le\tfrac12$,
\[
 \norm{\widehat T_\lambda h-Ph}^2\le2\mathsf R^2\lambda^{2r}+16R_0^2C_N(1+2C_\varkappa t)\frac{\lambda^{-p}}{N_{\mathrm{fit}}},
\]
and for $\lambda\asymp((1+t)/N_{\mathrm{fit}})^{1/(2r+p)}$,
$\boxed{\norm{\widehat T_\lambda h-Ph}^2\lesssim((1+t)/N_{\mathrm{fit}})^{2r/(2r+p)}}$; the design gate holds
eventually.
\end{theorem}
\begin{proof}
Triangle inequality with Proposition~\ref{q4:prop:source} and
Theorem~\ref{q4:thm:learning}:
$\varepsilon_h\le2R_0\sqrt{C_N\lambda^{-p}/N_{\mathrm{fit}}}(1+\sqrt{2C_\varkappa t})$; square with
$(x+y)^2\le2x^2+2y^2$ twice. The gate is
$\eta\le C_N\lambda^{-p}(\sqrt{C_\varkappa}+C_\varkappa\sqrt{2t})N_{\mathrm{fit}}^{-1/2}$, whose polynomial part at
the chosen $\lambda$ is $N_{\mathrm{fit}}^{(p-2r)/(2(2r+p))}\to0$ since $2r\ge1>p$.
\end{proof}

\begin{remark}[A nonempty residual-envelope regime]\label{q4:auto:a-nonempty-residual-envelope-regime}
If $\sup_k\norm{u_k}_\infty\le M_u$, $\sum_k\varrho_k^{2r}<\infty$ and $\norm{Qh}_\infty<\infty$, then
$\norm{T_\lambda Ph}_\infty,\norm{Ph}_\infty\le\mathsf RM_u(\sum_k\varrho_k^{2r})^{1/2}$ uniformly in $\lambda$, so
$\norm{h-T_\lambda h}_\infty$ is bounded; polynomial decay gives summability when $2r>p$.
\end{remark}

\paragraph{Squared fitting error is not risk regret.} With
$\epsilon_P=\max_j\norm{\widehat s_j-Pg_j}$ and $K_P=\sup_{\WL}\norm{Pg_w-f}$,
$\sup_{\WL}\abs{\norm{\widehat s_w-f}^2-\norm{Pg_w-f}^2}\le2K_PL\epsilon_P+L^2\epsilon_P^2$: an
$N_{\mathrm{fit}}^{-2r/(2r+p)}$ squared fitting rate gives only an $N_{\mathrm{fit}}^{-r/(2r+p)}$
risk perturbation through this bound, because in
$\norm{\widehat s_w-f}^2-\norm{Pg_w-f}^2=2\ip{PEw}{\widehat s_w-Pg_w}+\norm{\widehat s_w-Pg_w}^2$ the
transmitted error $PEw$ need not be orthogonal to the fitting error
(Section~\ref{q4:sec:closure} gives the exact cross term and a counterexample).

\subsection{Rank deficiency, spectral bias and empirical diagnostics}\label{q4:sec:rank}

All observed $x$ lie in $\Ran C_x$ a.s., so $\widehat C_x$ and $\widehat a$ vanish on true null
directions. With $\alpha=0$, if the relative event holds on $\Ran C_x$ then
$\Ker\widehat C_x=\Ker C_x$, and the preceding proofs apply on the quotient with $b_H=0$,
$d_1=d_2=k$, $\varkappa(0)=\esssup x^\top C_x^\dagger x$; outside that event a pseudoinverse is no
substitute for coverage (Example~\ref{q4:ex:coverage}). For a comparator,
$b_H(\beta,\alpha)=\sum_j\alpha^2\beta_j^2/(\mu_j+\alpha)+\alpha\norm{\Pi_{\Ker C_x}\beta}^2\le\alpha\norm\beta^2$ and
$b_P\le b_H(\beta_*,\alpha)$. If $\theta_*=C_x^{1/2}\beta_*=C_x^ru$ with $\norm u\le\mathsf R$ and $0<r\le1$,
then $b_P(\alpha)\le\mathsf R^2\alpha^{2r}$ (as in Proposition~\ref{q4:prop:source}). Small effective
dimension does not mean small bias: for $C_x=\diag(1,\varepsilon)$ and
$\beta_*=(0,\varepsilon^{-1/2})$, $\norm{\theta_*}=1$ but $b_P=\alpha^2/(\varepsilon+\alpha)^2\approx1$ when $\varepsilon\ll\alpha$.

\paragraph{Empirical effective dimension.} $\widehat d_1(\alpha)=\tr[\widehat C_x(\widehat C_x+\alpha I)^{-1}]$
satisfies, on the relative event,
$\abs{d_1-\widehat d_1}\le\eta(m-1-\widehat d_1)$ (from
$(1-\eta)\widehat H_\alpha^{-1}\preceq H_\alpha^{-1}\preceq(1+\eta)\widehat H_\alpha^{-1}$ and $d_1=m-1-\alpha\tr H_\alpha^{-1}$): an
additive certificate, since missed directions defeat multiplicative ones. The
largest observed leverage does not bound $\varkappa$.

\paragraph{A finite regularization grid.} For a deterministic finite grid
$\mathcal A$, replacing $t$ by $\log(6m\abs{\mathcal A}/\delta)$ makes Theorem~\ref{q4:thm:fast} hold
jointly at every grid point satisfying its gate, for a fixed comparator; the
observable gate $\alpha\ge64\kappa_x^2t/N$ certifies membership. Selecting the best
bias--variance trade-off needs an additional validation argument.

\section{Dependence, kernels, random features and linearization}\label{comp:ext:scope}

Transfer for functions chosen adaptively on the fitting sample;
out-of-sample kernel perturbation; bounded random features (exact
conditionally on the draw); and a conditional transfer to nonlinear
fine-tuning under explicit linearization and stability assumptions. These
extend the scope of \cref{app:q4} but are not used by the main chain.

\subsection{Dependence between pseudo-labels and the learner}\label{q4:sec:dependence}

\paragraph{Fixed teachers on the student inputs.} The residual equation uses
$\psi(X_i)(h-T_\lambda h)(X_i)$, which incorporates input--label dependence; the
inverse empirical covariance and the residual average are dependent, and the
proofs intersect events without factoring expectations.

\paragraph{Teachers trained on independent data.} If $\mathcal F_0$ contains
teacher training data, frozen representations and random-feature draws, and
$D,U,A$ are conditionally iid given $\mathcal F_0$, apply every theorem
conditionally: if its assumptions hold on $\mathcal E_0$ with
$\Prob(\mathcal E_0^c)\le\delta_0$ and the conditional failure is $\le\delta_1$, then
$\Prob(\text{failure})\le\Prob(\mathcal E_0^c)+\E[\one_{\mathcal E_0}\Prob(\text{failure}\mid\mathcal F_0)]\le\delta_0+\delta_1$.

\begin{proposition}[Same-sample adaptive functions: transfer]\label{q4:prop:adaptive}
For data-dependent $h_D$ and $h_0$ fixed independently of $D$ with
$\norm{h_D-h_0}_D\le\tau_D$ and $\norm{h_D-h_0}\le\tau_P$,
$\norm{(\widehat T_\lambda-T_\lambda)h_D}\le\norm{(\widehat T_\lambda-T_\lambda)h_0}+\kappa\lambda^{-1/2}\tau_D+\tau_P$. A small
population distance alone is insufficient: on a nonatomic law
$h_D=\one\{x\in\{X_1,\dots,X_{N_{\mathrm{fit}}}\}\}$ has population norm $0$ and empirical
norm $1$.
\end{proposition}
\begin{proof}
Insert $\pm\widehat T_\lambda h_0,\pm T_\lambda h_0$; Proposition~\ref{q4:prop:opt} and $\norm{T_\lambda}_\op\le1$.
\end{proof}

For selection from a predetermined finite collection of $M$ functions, apply
the simultaneous bound to all of them (an extra $\log M$). For $K$-fold
cross-fitting, train each teacher outside its student fold; foldwise events
can be union-bounded, and averaging foldwise students is controlled by the
triangle inequality, but a new full-sample refit is a different estimator. If
$\psi$ is learned on $D$, valid routes are learning it on independent data, a
uniform bound over a predetermined feature class, or an explicit perturbation
comparison with a fixed feature map.

\subsection{Random features, frozen representations and network linearization}\label{q4:sec:rf}

\paragraph{Exact conditional specialization.} For
$k(x,x')=\E_\omega[\psi_\omega(x)\psi_\omega(x')]$ and independent draws $\omega_1,\dots,\omega_P$,
$\psi_P=P^{-1/2}(\psi_{\omega_\ell})_\ell$ makes all preceding results exact with
$\Ff=\R^P$, for the realized random-feature learner (not infinite-feature KRR).

\begin{theorem}[Out-of-sample kernel perturbation]\label{q4:thm:kernel}
On a fixed design let $\mathsf A=N_{\mathrm{fit}}^{-1}[k(X_i,X_j)]$,
$\mathsf b(x)=N_{\mathrm{fit}}^{-1/2}(k(x,X_i))_i$, $s_k(h)(x)=\mathsf b(x)^\top(\mathsf A+\lambda I)^{-1}h_D$, and
similarly $\tilde{\mathsf A},\tilde{\mathsf b}$ for a second PSD kernel. If
$\norm{\tilde{\mathsf A}-\mathsf A}_\op\le\epsilon_A$,
$(\E\norm{\tilde{\mathsf b}(X)-\mathsf b(X)}^2)^{1/2}\le\epsilon_b$ and $(\E\norm{\mathsf b(X)}^2)^{1/2}\le B_b$, then
$\norm{s_{\tilde k}(h)-s_k(h)}\le(\epsilon_b/\lambda+B_b\epsilon_A/\lambda^2)\norm h_D$.
\end{theorem}
\begin{proof}
With resolvents $\mathsf R,\tilde{\mathsf R}$ of norm $\le1/\lambda$,
$\tilde{\mathsf R}-\mathsf R=-\tilde{\mathsf R}(\tilde{\mathsf A}-\mathsf A)\mathsf R$ and
$s_{\tilde k}-s_k=(\tilde{\mathsf b}-\mathsf b)^\top\tilde{\mathsf R}h_D+\mathsf b^\top(\tilde{\mathsf R}-\mathsf R)h_D$.
\end{proof}

\begin{proposition}[Bounded random features]\label{q4:prop:rf}
If $\abs{\psi_\omega}\le\kappa$ and $k_P=P^{-1}\sum_\ell\psi_{\omega_\ell}\otimes\psi_{\omega_\ell}$, then
conditionally on $D$, with probability $\ge1-2e^{-t}$, Theorem~\ref{q4:thm:kernel}
applies with $\epsilon_A,\epsilon_b\le\epsilon_P:=\kappa^2P^{-1/2}(1+\sqrt{2t})$ and $B_b\le\kappa^2$, so
$\norm{s_{k_P}(h)-s_k(h)}\le\epsilon_P(\lambda^{-1}+\kappa^2\lambda^{-2})\norm h_D$.
\end{proposition}
\begin{proof}
Lemma~\ref{q4:lem:hilbert} with $M=\kappa^2$, $v=\kappa^4$ in the Frobenius space for
$a_\omega a_\omega^\top$, $a_\omega=N_{\mathrm{fit}}^{-1/2}(\psi_\omega(X_i))_i$, and in $L^2(P_X;\R^{N_{\mathrm{fit}}})$
for $x\mapsto\psi_\omega(x)a_\omega$; $\abs k\le\kappa^2$.
\end{proof}

The resulting feature budget
$P\ge\kappa^4M^2(1+\sqrt{2t})^2\epsilon^{-2}(\lambda^{-1}+\kappa^2\lambda^{-2})^2$ (for
$\norm h_D\le M$) is sufficient, not optimal. Training Gram closeness alone does
not imply out-of-sample closeness; the cross-kernel term is essential.

\paragraph{Frozen representations and tangent models.} Ridge on a frozen
pretrained representation is an exact instance. For a network $N_\vartheta$ with
reference $\vartheta_0$, $h_0=N_{\vartheta_0}$, $J_0=\nabla_\vartheta N_{\vartheta_0}$, tangent ridge gives
$s_j^{\mathrm{lin}}=h_0+\widehat T_0(g_j-h_0)$; with $f'=f-h_0$, $g_j'=g_j-h_0$ the errors are
unchanged, and for affine $w$, $\sum_jw_js^{\mathrm{lin}}_j=h_0+\widehat T_0(g_w-h_0)$ (the
affine constraint carries the offset).

\begin{theorem}[Conditional nonlinear transfer]\label{q4:thm:ntk}
Suppose network training returns $v_w^{\mathrm{net}}$ with
$\norm{v^{\mathrm{net}}_w}\le R_\vartheta$, $\norm{v^{\mathrm{net}}_w-v^{\mathrm{lin}}_w}\le\epsilon_\vartheta$ uniformly on $\mathcal W$,
$\E\norm{J_0(X)}^2\le\kappa_J^2$, and
$\norm{\nabla^2_\vartheta N_{\vartheta_0+tv^{\mathrm{net}}_w}(x)}_\op\le H(x)$ on $[0,1]$ with
$\norm H_{L^2}\le M_H$. With $\epsilon_{\mathrm{net}}=\kappa_J\epsilon_\vartheta+\tfrac12M_HR_\vartheta^2$,
$\sup_{\mathcal W}\norm{N_{\vartheta_0+v^{\mathrm{net}}_w}-s^{\mathrm{lin}}_w}\le\epsilon_{\mathrm{net}}$, and risks and gains differ by
at most $2K_{\mathrm{lin}}\epsilon_{\mathrm{net}}+\epsilon_{\mathrm{net}}^2$, $K_{\mathrm{lin}}=\sup_{\mathcal W}\norm{s^{\mathrm{lin}}_w-f}$.
\end{theorem}
\begin{proof}
Taylor's formula with integral remainder bounds the remainder by
$H(x)\norm v^2/2$; the tangent difference is $\le\kappa_J\epsilon_\vartheta$ in $L^2$; expand the
squared norm as in Proposition~\ref{q4:prop:risk-transfer}.
\end{proof}

This is a conditional extension, not a proof that training stays in a tangent
regime; nonsmooth activations need a direct linearization bound; and
separately trained nonlinear students averaged afterwards are not the student
trained on aggregated labels.

\section{Short consequences}\label{comp:short}

One-line corollaries, examples and remarks deferred from Appendices B--E.

\subsection{From \cref{app:q1}}

\begin{corollary}[Sign criterion and bounds]\label{q1:cor:sign-bounds}
With $a=\norm{Te}$ and $b=\norm{b_T}$,
\[
 (a-b)^2\le R_T(e)\le(a+b)^2,\qquad
 \norm e^2-(a+b)^2\le G_T(e)\le\norm e^2-(a-b)^2 .
\]
Strict improvement holds iff $\ip e{(I-T^*T)e}+2\ip{Te}{b_T}>B_T$. The lower
risk bound is attained when the nonzero vectors $Te,b_T$ point in the same
direction and the upper one when they point in opposite directions.
\end{corollary}
\begin{proof}
Cauchy--Schwarz bounds $\ip{Te}{b_T}$ between $\pm ab$ in
\eqref{q1:eq:single-risk}, with the stated equality cases; the criterion is
\eqref{q1:eq:single-gain}.
\end{proof}

\begin{corollary}[Algorithmic realizability]\label{q1:cor:alg-real}
If $Tf=f$ and $\norm T_\op\le1$, then $R_T(e)=\norm{Te}^2\le\norm e^2$ and
$G_T(e)=\ip e{(I-T^*T)e}\ge0$. For a positive self-adjoint contraction,
equality holds exactly when $e$ lies in the spectral subspace where $T=I$.
\end{corollary}
\begin{proof}
$Tf=f$ gives $b_T=0$; contractivity gives the inequality. For a positive
contraction, $1-t^2\ge0$ on $\spec T$ and vanishes only at $t=1$; the
integral of $1-t^2$ against the spectral measure of $e$ vanishes exactly
under the stated support condition.
\end{proof}

\begin{corollary}[Realizable and misspecified regimes]\label{q1:cor:regimes}
If $f\in\Ss$, then $R_S(e)=\norm{Pe}^2\le R_W(e)$ and $G(e)=\norm{Qe}^2\ge0$,
strictly positive iff $e\notin\Ss$. If $A>0$, strict improvement, equality and
strict degradation are all possible: an entirely transmitted error $e\in\Ss$
gives $R_S=R_W+A$, an entirely filtered error $e\in\Ss^\perp$ gives $R_S=A$.
\end{corollary}

\begin{remark}[Strength versus realizability]\label{q1:auto:strength-versus-realizability}
$A>0$ does not mean the student is weaker than the supervisor. A natural
operational strength condition is $A<R_W(e)$ (the clean-label student beats
the supervisor). Under it, weak-label training can still improve or degrade
depending on the orientation of $e$ (Section~\ref{q1:sec:orientation}).
\end{remark}

\begin{corollary}[Regimes for aggregates]\label{q1:cor:multi-regimes}
If $A=0$ then $R_S(w)\le R_W(w)$ for all $w\in\Waff$, strictly iff
$Qe_w\ne0$. If $A>0$ then $R_S(w)<R_W(w)\iff w^\top\Gperp w>A$. The sign of
the gain can vary with both the weight and the orientation of the error
family relative to $\Ss$.
\end{corollary}

\begin{proposition}[Residual Gram is not the complement Gram]\label{q1:prop:residual-gram}
With $\Gamma_{I-T}=E^*(I-T)^*(I-T)E\succeq0$ and $C_T=E^*T^*(I-T)E$,
$\Sigma=\Gamma_T+\Gamma_{I-T}+C_T+C_T^\top$. In general
$\Sigma-\Gamma_T\ne\Gamma_{I-T}$.
\end{proposition}
\begin{proof}
Expand the Gram of $E=TE+(I-T)E$; the cross-Grams are $C_T$ and $C_T^\top$.
\end{proof}

\begin{remark}[The comparisons are logically distinct]\label{q1:auto:the-comparisons-are-logically-distinct}
$R_{\mathrm{student}}(w)<R_{\mathrm{agg}}(w)$ does not imply
$R_{\mathrm{student}}(w)<R_{\mathrm{best}}$, because the aggregate may be worse
than the best supervisor; conversely a student can beat the best supervisor
with small same-aggregate gain. Every W2S claim must name its benchmark.
\end{remark}

\begin{proposition}[Rotating errors equals rotating the student]\label{q1:auto:rotating-errors-equals-rotating-the-stud}
$\tr(QUCU^\top)=\tr((U^\top QU)C)$, and $U^\top QU$ ranges over all rank-$q$
orthogonal projections as $U$ ranges over $O(d)$. Rotating every error of a
family by a common $U$ maps $C$ to $UCU^\top$ and preserves its spectrum.
\end{proposition}
\begin{proof}
Cyclicity of trace. Any two rank-$q$ projections are conjugate by an
orthogonal map sending bases of $(\Ran Q',\Ran(I-Q'))$ to bases of
$(\Ran Q,\Ran(I-Q))$. Finally $\sum_a\pi_aUe_a\otimes Ue_a=UCU^\top$.
\end{proof}

\begin{corollary}[No spectrum-only sign diagnostic]\label{q1:cor:no-spectrum}
If $\sum_{i>d_\Ss}\lambda_i<A<\sum_{i\le q}\lambda_i$, there are $C_+,C_-$
with identical spectra and $\overline G(C_+)>0>\overline G(C_-)$; no function
of the eigenvalues alone determines the sign of the average gain.
\end{corollary}

\begin{proposition}[Common rotations preserve raw geometry]\label{q1:prop:common-rot}
For an orthogonal $U$ on $\Hh$ and $e_j^{(U)}=Ue_j$: $\Sigma^{(U)}=\Sigma$, while
$\Gpar^{(U)}=E^*U^*PUE$ and $\Gperp^{(U)}=E^*U^*QUE$, which agree with the
original matrices only under additional alignment such as $UP=PU$.
\end{proposition}
\begin{proof}
$\ip{Ue_j}{Ue_k}=\ip{e_j}{U^*Ue_k}=\ip{e_j}{e_k}$; replace $E$ by $UE$.
\end{proof}

\begin{example}[Ignoring bias alignment selects different weights]\label{q1:ex:bias-align}
$\Hh=\R^2$, $T=\tfrac12I$, $f=(1,0)^\top$, $e_1=(1,0)^\top$, $e_2=(0,1)^\top$:
$\Gamma_T=\tfrac14I$, $c_T=(\tfrac14,0)^\top$, $B_T=\tfrac14$. Minimizing
transmitted error alone selects $(\tfrac12,\tfrac12)$ with student risk
$\tfrac18$; the full objective selects $(1,0)$ with risk $0$. Both are simplex
weights; neither negativity nor singularity explains the discrepancy.
\end{example}

\begin{remark}[Mechanism]\label{q1:auto:mechanism}
Supervisor $1$ has tiny transmitted error $\varepsilon$ and huge filterable
error $M$. Raw weighting treats $M+\varepsilon$ as harmful and nearly
discards it; the student would filter $M$, so student-optimal weighting does
the opposite. The failure is caused by optimizing the wrong geometry, not by
estimation noise.
\end{remark}

\subsection{From \cref{app:q2}}

\begin{proposition}[Additional known linear restrictions]\label{q2:prop:extra}
If $Bf=b$ is also known for a bounded $B:\Hh\to\mathcal Z$ and the combined
system is compatible, every fiber statement holds with $\Nn$ replaced by
$\Nn\cap\Ker B$. A known restriction $f\in\Ss$ is $B=Q$, $b=0$.
\end{proposition}
\begin{proof}
Apply Theorem~\ref{q2:thm:fiber} to $f\mapsto(\mathcal Mf,Bf)$.
\end{proof}

\begin{remark}[Compatibility with the unstructured case]\label{q2:auto:compatibility-with-the-unstructured-case}
For $T=P$, $U\perp V$, but $\Pi_\Nn U$ and $\Pi_\Nn V$ need not be orthogonal:
projecting onto a third subspace can destroy orthogonality. Without structure
($\Nn=\Hh$) the leakage law reduces to Theorem~\ref{q2:thm:tasks}: $\ell_j=Tg_j$,
and selection is identified iff all students coincide.
\end{remark}

\begin{corollary}[Realizable projection student with structured teachers]\label{q2:cor:struct-real}
If $T=P$, $f\in\Ss$ is known and all $g_j\in\Kk$, then all gains and all
contrasts are identified on the restricted class, even if $f$ is not.
\end{corollary}
\begin{proof}
The restricted nullspace is $\Nn'=\Nn\cap\Ss$. For $n\in\Nn'$,
$\ip n{q_j}=\ip{Qn}{g_j}=0$ and $\ip n{s_j}=\ip{Pn}{g_j}=\ip n{g_j}=0$.
Realizability alone (Corollary~\ref{q2:cor:realizable}) does not supply the second
equality; the teacher equations do.
\end{proof}

\begin{corollary}[Invariant information space]\label{q2:cor:invariant}
If Assumption~\ref{q2:ass:range} holds and $T\Kk\subseteq\Kk$ (or just
$T\Span\{g_j\}\subseteq\Kk$), all gains and contrasts are identified. For
$T=P$, $P\Kk\subseteq\Kk$ iff $P\Pi_\Kk=\Pi_\Kk P$.
\end{corollary}
\begin{proof}
Invariance gives $\ell_j=0$. If $P\Kk\subseteq\Kk$ then for $n\in\Nn,k\in\Kk$,
$\ip{Pn}k=\ip n{Pk}=0$, so $P\Nn\subseteq\Nn$ and $P$ commutes with $\Pi_\Kk$;
the converse is immediate.
\end{proof}

\begin{proposition}[Finite dimensions]\label{q2:auto:finite-dimensions}
If $\Hh=\R^d$ and $\mathbf M$ stacks the $M_j$, then $f_\Kk=\mathbf M^\dagger y$,
$\Pi_\Kk=\mathbf M^\dagger \mathbf M$, $\Pi_\Nn=I-\mathbf M^\dagger \mathbf M$, and for $h\in\Kk$ the
minimum-norm dual is $(\mathbf M^\dagger)^\top h$. If the nonzero singular values of $\mathbf M$
are at least $\alpha_{\mathbf M}>0$, then $C_h\le\norm h/\alpha_{\mathbf M}$ and
$\norm{\mathbf M^\dagger\widetilde y-f_\Kk}\le\norm{\widetilde y-y}/\alpha_{\mathbf M}$.
\end{proposition}
\begin{proof}
With a reduced SVD $\mathbf M=U\diag(\sigma_i)V^\top$, $\mathbf M^\dagger=V\diag(\sigma_i^{-1})U^\top$,
$\Kk=\Ran V$, and compatibility gives $y\in\Ran U$. For $h=Vb$ the vector
$U\diag(\sigma_i^{-1})b$ maps to $h$ and is orthogonal to $\Ker \mathbf M^\top$.
\end{proof}

\subsection{From \cref{app:q3}}

\begin{proposition}[Uniform contrast control]\label{q3:prop:contrast}
With $L(w)=\norm{t(w)}^2-2\theta^\top t(w)$ and $\widehat L$ its plug-in,
$\abs{(\widehat L(w)-\widehat L(v))-(L(w)-L(v))}\le2\norm{t(w)-t(v)}\norm\varepsilon$;
on the simplex the supremum is at most $2\max_{j,\ell}\norm{t_j-t_\ell}\norm\varepsilon$.
\end{proposition}
\begin{proof}
The left side is $2\abs{\varepsilon^\top(t(w)-t(v))}$, and
$t(w)-t(v)=\sum_{j,\ell}w_jv_\ell(t_j-t_\ell)$ for simplex weights.
\end{proof}

\begin{proposition}[Exact affine Gaussian minimax risk]\label{q3:auto:exact-affine-gaussian-minimax-risk}
For $\mathcal C=\R^k$ and $Z\sim N(\theta,\varsigma^2I_k)$,
$\inf_{\widehat t}\sup_\theta\E\norm{\widehat t-\theta}^2=k\varsigma^2$.
\end{proposition}
\begin{proof}
$Z$ attains $k\varsigma^2$; under the prior $N(0,b^2I_k)$ the Bayes risk is
$k(b^{-2}+\varsigma^{-2})^{-1}\to k\varsigma^2$ as $b\to\infty$.
\end{proof}

\begin{remark}[Which dictionary is measured]\label{q3:auto:which-dictionary-is-measured}
The relevant vectors are the unresolved, noise-whitened, margin-normalized
queries $2\Omega^{1/2}\Phi_U^*\Pi_\Nn q_{w^{(i)}}/\gamma_i$. Clustering of raw
teacher functions does not imply clustering of these vectors. If
$h_{(i)}=0$ the gain is known without anchors; a zero gain has no separated
sign and is excluded from positive-margin identification.
\end{remark}

\begin{theorem}[Sign cost despite exact structured teachers]\label{q3:thm:struct-sign}
In the structured example of \cref{app:q2} (teachers $M_1=P_{\Span\{u\}}$,
$M_2=P_{\Span\{v\}}$, student $P=P_{\Span\{t_0\}}$, $t_0=(u-v+2z)/\sqrt6$), let
$0<\gamma\le\tfrac14$, $c_\pm=\tfrac74\pm\tfrac32\gamma$ and $f_\pm=u+v+c_\pm z$.
The observations $g_1=u$, $g_2=v$ are unchanged, $G(\mathbf e_1;f_\pm)=\pm\gamma$,
and the clean student beats both teachers. With Gaussian anchor noise, every
sign estimator has worst-world error $\ge\tfrac14e^{-9n\gamma^2/(2\sigma^2)}$, so
error $\delta<\tfrac14$ needs $n\ge\tfrac{2\sigma^2}{9\gamma^2}\log\tfrac1{4\delta}$.
\end{theorem}
\begin{proof}
For $f_c=u+v+cz$: $\norm{f_c}^2=2+c^2$, $\ip{f_c}{t_0}=2c/\sqrt6$,
$R_W(\mathbf e_j)=1+c^2$, $A(f_c)=2+c^2/3$,
$R_S(\mathbf e_1)=2+c^2+\tfrac16-\tfrac{2c}3$, so $G(\mathbf e_1;f_c)=(4c-7)/6$, equal to
$\pm\gamma$ at $c_\pm$. For $\gamma\le\tfrac14$, $c_\pm\ge\tfrac{11}8$, so
$2c_\pm^2/3>1$, i.e.\ $A(f_\pm)<1+c_\pm^2$. The targets differ by $3\gamma z$;
teacher observations contribute no divergence; Theorem~\ref{q3:thm:sign-lower}.
\end{proof}

\subsection{From \cref{app:q4}}

\begin{proposition}[Stochastic pseudo-label noise]\label{q4:prop:pl-noise}
If supervisor $j$'s observed labels are $(g_j)_D+\zeta_j$, let
$A_D=\Psi(C_D+\lambda I)^{-1}S_D^*$, $\zeta_w=\sum_jw_j\zeta_j$,
$\mu_w=\E[\zeta_w\mid D]$, $\Omega_w=\Cov(\zeta_w\mid D)$. Then
$\E[\norm{\widehat s_w+A_D\zeta_w-f}^2\mid D]=\norm{\widehat s_w-f+A_D\mu_w}^2+\tr(A_D^*A_D\Omega_w)$, with no
diagonal or independence assumption on $\Omega_w$. Teacher function errors
$e_j(X_i)$ are not mean-zero noise; they stay in the geometric terms.
\end{proposition}
\begin{proof}
Write $\zeta_w=\mu_w+\varepsilon_w$; the cross term has conditional mean zero.
\end{proof}

\end{document}